%% file: main_arXiv.tex
\documentclass[11pt]{article}

\usepackage{amsmath,amssymb,amsthm,bm}
\usepackage{stmaryrd} 

\usepackage{xcolor}
\usepackage{graphicx}
\usepackage{subcaption}
\usepackage[normalem]{ulem} 
\usepackage{enumerate}

\usepackage[linesnumbered,ruled,vlined]{algorithm2e}
\makeatletter
\newcommand{\nonl}{\renewcommand{\nl}{\let\nl\oldnl}}
\makeatother

\newcommand{\dosemic}{\renewcommand{\@endalgocfline}{\algocf@endline}}
\SetKw{KwInit}{Initialization}

\usepackage{natbib}
\usepackage{url}
\usepackage[colorlinks,linkcolor=red,citecolor=blue]{hyperref}
\usepackage{notation}
\graphicspath{{pics/}}

\newtheorem{remark}{Remark}[section]
\newtheorem{theorem}{Theorem}[section]
\newtheorem{assumption}{Assumption}[section]

\newtheorem{lemma}{Lemma}[section]

\title{\bf Online Tensor Inference}

\author{
    Xin Wen\footnote{New York University. Email: \href{mailto:xin.wen@stern.nyu.edu}{xin.wen@stern.nyu.edu}} 
    \quad 
    Will Wei Sun\footnote{Purdue University. Email: \href{mailto:sun244@purdue.edu}{sun244@purdue.edu}} 
    \quad 
    Yichen Zhang\footnote{Purdue University. Email: \href{mailto:zhang@purdue.edu}{zhang@purdue.edu}}
}
\date{}

\begin{document}
\def\spacingset#1{\renewcommand{\baselinestretch}{#1}\small\normalsize} 
\spacingset{1}

\maketitle

\begin{abstract}
\normalsize
Contemporary applications, such as recommendation systems and mobile health monitoring, require real-time processing and analysis of sequentially arriving high-dimensional tensor data. Traditional offline learning, involving the storage and utilization of all data in each computational iteration, becomes impractical for these tasks. Furthermore, existing low-rank tensor methods lack the capability for online statistical inference, which is essential for real-time predictions and informed decision-making. This paper addresses these challenges by introducing a novel online inference framework for low-rank tensors. Our approach employs Stochastic Gradient Descent (SGD) to enable efficient real-time data processing without extensive memory requirements. We establish a non-asymptotic convergence result for the online low-rank SGD estimator, nearly matches the minimax optimal estimation error rate of offline models. Furthermore, we propose a simple yet powerful online debiasing approach for sequential statistical inference. The entire online procedure, covering both estimation and inference, eliminates the need for data splitting or storing historical data, making it suitable for on-the-fly hypothesis testing. In our analysis, we control the sum of constructed super-martingales to ensure estimates along the entire solution path remain within the benign region. Additionally, a novel spectral representation tool is employed to address statistical dependencies among iterative estimates, establishing the desired asymptotic normality.
\end{abstract}


\spacingset{1.2}

\section{Introduction}\label{sec:Intro}
\input{body/Intro}

\input{body/RelatedWork}

\section{Online Low-Rank Tensor SGD}\label{sec:method}
\input{body/Estimation}

\input{body/Inference}

\section{Numerical Simulations} \label{sec:simulations}
\input{body/Simulation_revision}

\section{Real Data Analysis} \label{sec:real data}
\input{body/RealData}

\bibliographystyle{apalike}
\bibliography{reference/ref1}

\newpage
\bigskip
\begin{center}
{\large\bf Supplementary Material for Online Tensor Inference}
\end{center}
\setcounter{section}{0}
\renewcommand{\thesection}{\Alph{section}}

The supplementary material is organized as follows.
\begin{itemize}
  \item Section \ref{sec:appendix_literature} reviews additional related literature.
  \item Section \ref{sec:appendix_simulation} reports extended simulation results.
  \item Section \ref{sec:appendix_highorder} presents a high-order extension of our algorithm.
  \item Section \ref{sec:notations} summarizes the notation used in the main paper.
  \item Proofs of Theorems \ref{thm:converge}, \ref{thm:entry inference}, and \ref{thm:entry inference consistent}
  are provided in Sections \ref{sec:Proof of Theorem convergence}, \ref{sec:Proof of Theorem entry inference},
  and \ref{sec:Proof of Theorem entry inference consistent}, respectively.
  \item Theorems \ref{thm:U normality} and \ref{thm:Tensor regression} characterize the distribution of the low-rank factors
  \(\mathbf{U}_k^{\star}\in\mathbb{R}^{p_k\times r_k}\); their proofs appear in Sections \ref{sec:Proof of U normality}
  and \ref{sec:Proof of Tensor regression}.
  \item Section \ref{sec:Proof of Lemmas of Theorem converge} collects proofs of auxiliary technical lemmas.
\end{itemize}
Throughout this companion, we set \(\lambda_{\max}=\kappa_0=1\), and define
\(r=\max_{k\in[3]} r_k\) and \(p=\max_{k\in[3]} p_k\).

\section{Additional Related Literature} \label{sec:appendix_literature}
\input{body/Appendix_literature}

\section{Extended Simulation Results}
\label{sec:appendix_simulation}
\input{body/Appendix_simulation}

\section{Extension to General-order Tensor Case}
\label{sec:appendix_highorder}
\input{body/Appendix_highorder}

\section{Notations of the Paper}
\label{sec:notations}
\input{body/Notation}

\input{body/ProofEstimation}

\input{body/ProofEntry}

\input{body/ProofEntryConsis}

\input{body/ProofComponent}

\input{body/Some_lemmas}
\input{body/LemmaEstimation}

\input{body/LemmaEntry}

\input{body/LemmaComponent}

\end{document}

%% file: body/Intro.tex
Digital advertising expenditures in the United States reached 225 billion in 2023, a 7.3\% increase from 2022 \citep{statista2024us}. In response, online retailers are continuously innovating their advertising strategies 
to boost consumer engagement and drive purchases 
\citep{sawhney2005collaborating,kumar2016competitive,shankar2011innovations}. 
For example, companies like Amazon leverage user preference data to introduce targeted features (e.g., ``See Price in Cart," ``Subscribe \& Save," and ``Collects") that not only enhance the shopping experience but also facilitate personalized advertising (for example, alerting customers to price drops on items of interest) \citep{potoo2024amazons}.

Yet significant challenges persist. Consumer behavior in large marketplaces is highly heterogeneous—varying with device, demographics, and geography—complicating the evaluation of such strategies \citep{kamakura1996modeling,keane1997modeling,allenby1998marketing,blake2015consumer}.  Additionally, the need for real-time decision-making across millions of products and daily interactions renders traditional offline methods (e.g., batch-processing historical data) insufficient \citep{grbovic2015ecommerce}.

More importantly, the ultimate objective of collecting customer data on online platforms---whether for personalized recommendations, targeted advertising, or dynamic pricing---is not merely to obtain precise estimates of population-level behavioral metrics, but rather to improve decision making in fast-changing markets \citep{dwivedi2021setting,imbens2022comment,bojinov2022online}. However, even advanced machine learning models struggle to distinguish true effects from random noise, leaving decision-makers with point estimates (e.g., ``Strategy X increases sales by 3\%") that may lack clear statistical significance.

\textbf{Our contributions:}
Motivated by the challenges above, we develop an online low‑rank tensor estimation and inference framework that supplies both point estimates and valid confidence intervals, thereby guiding hypothesis-driven advertising decisions. Our contributions are twofold, spanning both methodological and theoretical aspects.

First, a rich literature shows that customer–item interactions are well‑captured by low‑rank structures \citep{spearman1904general,hotelling1933analysis,funk2006netflix,witten2009penalized,farias2019learning,udell2019why,kallus2020dynamic,xu2021groupsparse,bayati2022speed,zhou2024stochastic}.  
        We cast context‑adaptive advertising as a low‑rank tensor regression problem—e.g., modes for customer segments, time windows, and product categories—and design a streaming Stochastic Gradient Descent (SGD) algorithm that estimates the tensor parameter \( \mathcal{T}^\star \) on the fly.  By processing data sequentially and discarding observations after gradient evaluation, our approach circumvents the memory limitations of conventional offline methods. Moreover, by focusing on a low-rank factorization, we reduce the computational complexity from \( O(p_1p_2p_3) \) to \( O(\max_k p_k \cdot \max_k r_k) \) for a three-mode tensor of dimensions \( p_1 \times p_2 \times p_3 \) with Tucker rank \((r_1, r_2, r_3)\). This reduction facilitates accurate estimates that support effective business decisions on e-commerce platforms, where feature dimensionality is very high \citep{miao2022online, jiang2024highdimensional}.

Second, we propose an online inference procedure that constructs confidence intervals for the tensor parameters of interest. Although standard SGD-based estimates \( \mathcal{T}^{(t)} \) are effective for sequential estimation, their inherent bias—due to low-rank constraints—and high variance from the stochastic updates complicate direct statistical inference. In contrast to existing offline methods that often rely on sample splitting to correct bias, our online debiasing approach yields exact confidence intervals for linear combinations of tensor entries without sacrificing data efficiency. For instance, by constructing confidence intervals for linear forms such as  
\begin{align}\label{equ:hypo test}
\begin{aligned}
        H_0: \mathcal{T}^\star(j_1,j_2,j_3) = \mathcal{T}^\star(j_4,j_5,j_6) \\
    \text{versus} \quad H_1: \mathcal{T}^\star(j_1,j_2,j_3) \neq \mathcal{T}^\star(j_4,j_5,j_6),
\end{aligned}
\end{align}
   advertisers can rigorously test whether different strategies (e.g., varied product exposure or pricing displays) yield statistically significant lift for the same item in the same time band. 
   
Importantly, our work offers a clear practical insight for managers. Our online inference framework enables a dynamic, sequential approach to advertising decisions. In practice, an advertiser begins by formulating a hypothesis about which strategy may be more effective.  As real-time consumer response data is collected, the framework analyzes the results and delivers a statistical decision regarding the hypothesis. Based on this outcome, the advertiser can refine or propose a new hypothesis, with the framework continuously incorporating both previous results and new data. This iterative cycle of hypothesis formulation, data analysis, and decision making supports more effective and adaptive advertising strategies and replaces guesswork with statistically grounded choice.

In addition to the aforementioned methodological contribution, our work yields the following key results from a theoretical perspective:
\begin{enumerate}[(1)]
    \item Non-asymptotic convergence: Theorem \ref{thm:converge} proves that the sequentially updated SGD estimator \( \mathcal{T}^{(t)} \) converges to the true parameter \( \mathcal{T}^\star \) at a nearly minimax optimal rate. That is, with high probability,  
     \[
     \|\mathcal{T}^{(t)} - \mathcal{T}^\star\|_{\mathrm{F}} \leq C \sigma \sqrt{\frac{\mathsf{df}}{t^\alpha}},
     \]
     where \( \sigma \) is the noise level, \( \mathsf{df} = r_1 r_2 r_3 + \sum_{k=1}^3 p_k r_k \) represents the effective degrees of freedom, \( \alpha \in (0, 1) \), and \( C \) is a constant. 
In proving Theorem \ref{thm:converge}, we show that, with the help of a regularizer, a properly initialized low‑rank SGD estimator remains within a local region characterized by benign geometrical properties with high probability. Moreover, a two‑stage step‑size schedule controls the cumulative noise, ensuring the iterates stay in this region throughout the online process.

\item  Online inference for general linear forms of tensors:  We establish the asymptotic normality of the online estimator for a general linear form \( h^\star = \langle \mathcal{T}^\star, \mathcal{H} \rangle \)  in Theorem \ref{thm:entry inference}. Our analysis employs new proof techniques to handle statistical dependencies among sequential estimates, including a spectral representation tool that keeps the singular space of the estimator close to that of the true parameter. We further propose a fully online estimator for the variance of \( \hat{h}^{(t)} \) and prove its consistency in Theorem \ref{thm:entry inference consistent}, thus ensuring the reliability of our constructed confidence intervals.

\item Online inference for low-rank factors: as a byproduct of our online inference framework, we derive the asymptotic distribution of the estimated low-rank factors in Theorem \ref{thm:U normality}, presented in Section \ref{sec:Proof of U normality} of the Supplementary Material. This result allows the construction of confidence regions for these factors and provides deeper insights into the latent structures underlying consumer behavior.
\end{enumerate}

Overall, our contributions offer a practical tool for sequential, context-adaptive advertising decisions as well as a rigorous theoretical framework for online low-rank tensor estimation and inference. We validate the performance of our methods using extensive simulations in Section \ref{sec:simulations} and an online marketplace example in Section \ref{sec:real data}.

%% file: body/RelatedWork.tex
\section{Related Literature}
\label{sec:literature}
This section discusses three lines of related works: online inference based on SGD, estimation of low-rank models, and statistical inference for low-rank models. Our literature review highlights the fundamental differences between these existing approaches and our work, particularly in terms of problem settings and analysis tools.  Additional related literature--on low-rank models in business applications, tensor-based deep learning methods, and human-computer interfaces--is provided in Section \ref{sec:appendix_literature} of the Supplementary Material.

\textbf{Online Inference Based on SGD.}
Our work is related to a growing body of literature on online statistical inference based on SGD.The foundation for this was laid by the establishment of the asymptotic distribution of the averaged version of stochastic approximation, first established in the work by \cite{ruppert1988efficient, polyak1992acceleration}.  Several key contributions in this area include \cite{fang2018online}, who introduced a perturbation-based resampling procedure for inference, and \cite{liang2019statistical}, who developed moment-adjusted stochastic gradient descents for this purpose. \cite{chen2020statistical} proposed online methods to estimate the asymptotic covariance matrix for conducting inference. Recent developments by \cite{chen2021statistical, chen2022online} have focused on SGD-based algorithms in online decision-making contexts, incorporating decision rules into the analysis.
\begin{maroontext}
\cite{han2024online} propose a debiased SGD algorithm for online statistical inference with high‐dimensional data that does not require storing the entire dataset or its sample covariance matrix.
\end{maroontext}
However, none of these works handles the low-rankness and non-convexity in an online tensor learning problem, which demands new analytical tools.

\textbf{Estimation of Low-Rank Models.}
In recent years, there has been notable interest in low-rank tensor models \citep{zhou2013tensor, li2017parsimonious, li2018tucker, zhang2020islet,cai2022nonconvex,cai2023uncertainty, zhen2024nonnegative,zhang2024change}. 
Specifically, gradient descent-based algorithms for tensor parameter estimation have recently gained attention \citep{chen2019nonconvex, han2022optimal, tong2022scaling}. 
However, all these works focus on offline learning.
The work most closely related to ours is the concurrent study by \cite{cai2023online}, which introduced an online Riemannian gradient descent (oRGrad) algorithm for online tensor estimation and studied convergence rates for the estimator with both constant and doubling step sizes. Our work, however, addresses a more general case with a time-decaying step size and focuses on a factor-based SGD algorithm. Importantly, none of these tensor works studies online statistical inference, which is one major step beyond estimation.

\textbf{Statistical Inference for Low-Rank Models.}
Recent studies have begun demystifying statistical inference for low-rank matrix models. 
\cite{xia2019confidence} introduced a debiased estimator for matrix regression under isotropic Gaussian design, establishing the distribution of the \(\sin\Theta\) norm of singular matrices.
\cite{chen2019inference} proposed a debiased estimator for matrix completion. 
\cite{xia2021statistical} focused on matrix linear forms inference, establishing entry-level confidence intervals. \cite{chen2023statistical} explored estimation and inference of low-rank components in high-dimensional matrix-variate factor models. 
\cite{han2024online} extended inference work to the matrix contextual bandit with online decision-making.
All these methods hinge on suitable debiasing of initial estimates. For offline tensor inference,  \cite{huang2022power} studied the statistical inference and power iteration for tensor PCA. \cite{cai2023uncertainty} investigated entrywise statistical inference for noisy low-rank tensor completion in symmetric tensors with low CP-rank. \cite{xia2022inference} considered statistical inference for low-rank tensors with Tucker decomposition, focusing on the entrywise distribution in rank-one tensor PCA models and the confidence regions for the \(\sin\Theta\) norm of low-rank singular spaces in tensor regression. However, current research has yet to conduct online statistical inference on low-rank tensors. Moreover, compared to the offline inference work in tensor methods, our online SGD estimator requires milder conditions for initialization, and our inference procedure is full-online, without the need for extra iteration at each time step.

%% file: body/Estimation.tex
Accurate estimation serves as the foundational basis for all forms of statistical inference. In this section, we first introduce the problem of interest and discuss its treatment in offline data context. Following this, we present our proposed online method for tensor estimation involving online sequential data. Finally, we provide a non-asymptotic convergence analysis of our proposed method.

\subsection{Low-rank Tensor Models}
In the field of tensor learning, low-rank tensor regression is often regarded as one of the most basic settings of supervised learning involving low-rank tensors.
Speciﬁcally, the sample at time \( t \) is denoted as \( \boldsymbol{\zeta}_t=\left(y_t, \mathcal{X}_t\right) \), where the covariate tensor \( \mathcal{X}_t \in \mathbb{R}^{p_1 \times p_2 \times p_3} \) acts as the predictor, and the response variable \( y_t \in \mathbb{R} \) follows a linear model:
\begin{align}\label{equ:regression}
y_t = \left\langle\mathcal{X}_t,\mathcal{T}^{\star}\right\rangle + \xi_t,
\end{align}
where the error terms \( \left\{\xi_t\right\} \) are independently and identically distributed (i.i.d.) mean-zero random variables and the true parameter \( \mathcal{T}^{\star} \in \mathbb{R}^{p_1\times p_2\times p_3} \) is a low-rank tensor of Tucker rank-$\left(r_1, r_2, r_3\right)$. This implies that the parameter can be effectively represented by a limited number of multi-way principal components, a feature that holds significant practical interest \citep{kolda2006multilinear,kolda2009tensor}.

In machine learning, a parameter estimation problem often translates naturally to an optimization problem. Assuming a tensor parameter $\mathcal{T}^{\star}$ satisfies the constraints of the low-rankness model, it minimizes the population risk function $F(\mathcal{T}): \mathbb{R}^{p_1\times p_2\times p_3}\to\mathbb{R}$,  expressed as: 
\begin{align}\label{equ:def of T*}
\underset{\substack{\mathcal{T} \in \mathbb{R}^{p_1 \times p_2 \times p_3},\\ \operatorname{rank}(\mathcal{T}) \leq\left(r_1, r_2, r_3\right)}}{\min } \left( \vphantom{\frac{x}{y}} F\left(\mathcal{T}\right):=\mathbb{E}_{\boldsymbol{\zeta} \sim \Pi} f\left(\mathcal{T}; \boldsymbol{\zeta}\right)\right),
\end{align}
where $ f(\mathcal{T}; \boldsymbol{\zeta}) $ denotes the quadratic loss function $\frac{1}{2} \left( \left\langle\mathcal{X}, \mathcal{T}\right\rangle - y \right)^2$ in tensor regression. 
To estimate \( \mathcal{T}^{\star} \), one might consider performing a rank-constrained minimization of the risk function \( F(\mathcal{T}) \). However, this approach is computationally challenging due to the non-convexity introduced by the low-rank constraint.

Given the low-rank structure of the true parameter \( \mathcal{T}^{\star} \), the Tucker decomposition provides a natural framework for efficient estimation and enhanced interpretability \citep{kolda2009tensor}. Specifically, \( \mathcal{T}^{\star} \) admits the decomposition: $\mathcal{T}^\star = \mathcal{G}^\star \times_{k\in[3]} \mathbf{U}_k^\star,$
which generalizes the matrix singular value decomposition (SVD) to higher-order tensors. Here, the factor matrices \( \{\mathbf{U}_k^\star\}_{k\in[3]} \) capture the principal directions of variation along each mode of \( \mathcal{T}^\star \), analogous to the singular vectors in matrix SVD. The core tensor \( \mathcal{G}^\star \), meanwhile, acts as a compressed representation of \( \mathcal{T}^\star \), with its dimensions \( (r_1, r_2, r_3) \) controlling the degree of dimensionality reduction.  This structural compression directly translates to parameter efficiency: while the full tensor \( \mathcal{T}^\star \) requires \( p_1 p_2 p_3 \) parameters, the Tucker decomposition reduces this to  $\df = r_1 r_2 r_3 + \sum_{k=1}^3 p_k r_k$, dramatically lowering the model complexity when \( r_k \ll p_k \) for all modes \( k \).

In practice, the population risk function \( F \) is inaccessible since the data distribution $\Pi$ is unknown. Instead, researchers often turn to the empirical risk as a reliable approximation. In traditional offline learning setting,  we possess \( n \) i.i.d. samples, represented as \( \left\{\boldsymbol{\zeta}_t\right\}_{t=1}^n \), to facilitate parameter estimation. The prevalent approach in such scenarios is to minimize the empirical risk through deterministic optimization: $\widehat{\mathcal{T}}^{(n)}_{\mathrm{ERM}} = \arg \min\frac{1}{n} \sum_{t=1}^n f\left(\mathcal{T}; \boldsymbol{\zeta}_t\right)$, 
where the empirical risk represents the mean loss computed over the dataset up to size \( n \). This empirical risk minimizer \( \widehat{\mathcal{T}}^{(n)}_{\mathrm{ERM}} \) is widely used in both the statistical and machine learning domains \citep{zhang2020islet, han2022optimal, tong2022scaling}. 
Traditional offline learning methods typically require processing all available samples at once.
However, with the advancement of modern technology enabling data collection at an unprecedented scale, the traditional offline framework can become computationally burdensome. This is primarily due to constraints in memory capacity and processing power. In response to these challenges, there is a growing trend towards adopting online learning approaches, which present a more feasible solution. 
A recent example is the work by \cite{cai2023online}, which introduced an online algorithm oRGrad for tensor estimation. 
Differing from their focus on estimation, our paper aims to introduce a new online algorithm designed to naturally facilitate sequential statistical inference.

\subsection{Online Low-Rank Tensor Estimation}

Online learning permits the sequential processing of data points. In this framework, at every time step \( t \), the model receives and processes only one observation $\boldsymbol{\zeta}_t = \left(y_t, \mathcal{X}_t\right)$. Consequently, in online tensor estimation, the tensor parameter estimate \( {\mathcal{T}}^{(t)} \) is updated in real time, incorporating each newly received data point.
This approach incrementally improves estimation accuracy, leading the model to ultimately converge towards the true tensor parameter \( \mathcal{T}^{\star} \).

 The Robbins-Monro procedure, widely recognized in online learning as SGD \citep{robbins1951stochastic}, offers significant computational and storage benefits compared to conventional deterministic optimization methods. Specifically, SGD's requirement for only a single pass through the data presents a substantial computational advantage over batch methods like traditional gradient descent. Furthermore, SGD has the distinct benefit of being able to discard data points immediately after evaluating the gradient, thereby rendering it naturally online and eliminating the need for huge memory storage. The vanilla tensor SGD algorithm refines parameter estimates by following the gradient of the loss function, formalized as:
\begin{align}\label{equ:vanilla tensor sgd}
    {\mathcal{T}}^{(t)}={\mathcal{T}}^{(t-1)}-\eta_t \nabla_{\mathcal{T}} f\left({\mathcal{T}}^{(t-1)} ; \boldsymbol{\zeta}_t\right),
\end{align}
$t = 1, 2, \cdots$, where $\nabla_{\mathcal{T}} f(\cdot ; \boldsymbol{\zeta})$ represents the gradient of $f(\cdot ; \cdot)$ with respect to $\mathcal{T}$ evaluated at point  $\boldsymbol{\zeta}$.  For notational simplicity, we suppress the argument \(\boldsymbol{\zeta}\) in \(\nabla_{\mathcal{T}}f(\cdot;\boldsymbol{\zeta})\) whenever it is unambiguous. Here, $\left\{\eta_t\right\}_{t=1}^\infty$ denotes a sequence of positive, non-increasing learning rates or step sizes.

Considering the inherent non-convexity of our loss function (\ref{equ:def of T*}) and the high-dimensionality in this problem, traditional methods like vanilla SGD are not suitable.  
Since the true tensor \( \mathcal{T}^{\star} \) has a low-rank structure, we develop a specialized low-rank of tensor SGD that features updating the factor matrices \( \mathbf{U}_k^{(t-1)} \) and the core tensor $\mathcal{G}^{(t-1)}$ instead of directly updating the tensor parameter $\mathcal{T}^{(t-1)}$. Our approach involves adjusting the factor matrices \( \mathbf{U}_k^{(t-1)} \) in the opposite direction of their loss function's gradient, using a decaying step size \( \eta_t \). Similarly, we update the core tensor \( {\mathcal{G}}^{(t-1)} \) by following the gradient specific to the core tensor \( \mathcal{G} \). Specifically, the updates for our tensor factor parameters are given by:
\begin{gather}\label{equ:sgd rule-1}
\begin{aligned}
		\mathbf{U}_k^{(t)}&=\mathbf{U}_k^{(t-1)}-\eta_t \nabla_{\mathbf{U}_k} f\left({\mathcal{T}}^{(t-1)} ; \boldsymbol{\zeta}_t\right), \text{ for } k\in[3],\\
{\mathcal{G}}^{(t)}&={\mathcal{G}}^{(t-1)}-\eta_t \nabla_{\mathcal{G}} f\left({\mathcal{T}}^{(t-1)} ; \boldsymbol{\zeta}_t\right).
\end{aligned}
\end{gather}
Here, \( \nabla_{\mathbf{U}_k} f\left(\mathcal{T}; \boldsymbol{\zeta}\right) \) represents the gradient of \( f\left( \cT ; \cdot\right) \) with respect to \( \mathbf{U}_k \) evaluated at \( \boldsymbol{\zeta} \), and \( \nabla_{\mathcal{G}} f\left(\mathcal{T}; \boldsymbol{\zeta}\right) \) is analogously defined. The tensors \( \mathcal{G}^{(t-1)} \) and \( \mathbf{U}_k^{(t-1)} \) represent the estimates at the prior time step \( t-1 \). 
These gradients can be derived using the chain rule:
$	 \nabla_{\mathbf{U}_k} f(\mathcal{T}; \boldsymbol{\zeta})  = 
  (\left\langle\mathcal{X}, \mathcal{T}\right\rangle-y)\mathcal{M}_k\left(\mathcal X\right)\left(\mathbf{U}_{k+2} \otimes \mathbf{U}_{k+1}\right) \mathcal{M}^{\top}_k(\mathcal{G})\in\mathbb{R}^{p_k\times r_{k+1} r_{k+2}},
  \nabla_{\mathcal{G}} f (\mathcal{T}; \boldsymbol{\zeta}) =   (\left\langle\mathcal{X}, \mathcal{T}\right\rangle-y)\mathcal X \times_{k\in[3]} \mathbf{U}_k^{\top}  \in\mathbb{R}^{r_1\times r_2\times r_3}.$
Building on the factor matrices \( \mathbf{U}_k^{(t)} \in \mathbb{R}^{p_k\times r_k} \) and the core tensor \( {\mathcal{G}}^{(t)} \in \mathbb{R}^{r_1\times r_2\times r_3} \), rather than the full tensor \( {\mathcal{T}}^{(t)} \in \mathbb{R}^{p_1\times p_2\times p_3} \), our low-rank tensor SGD offers lower computational and storage requirements compared to vanilla tensor SGD in Equation (\ref{equ:vanilla tensor sgd}).

Despite its simplicity, this algorithm does not capture a critical identifiability issue in the low-rank tensor modeling.
Specifically, for any set of invertible matrices \( \mathbf{R}_k \in \mathbb{R}^{r_k \times r_k} \) for \( k\in[3] \), an identity $\mathcal{G} \times_{k\in[3]} \mathbf{U}_k = 
\left( \mathcal{G} \times_{k\in[3]} \mathbf{R}_k^{-1} \right)  \times_{k\in[3]} \mathbf{U}_k\mathbf{R}_k$ exists. 
This non-uniqueness may cause the factor matrices to become nearly singular or numerically unstable. 
To address this identifiability issue, a common tactic is to introduce regularization to promote balanced factor matrices:
\begin{align}\label{equ:def of L}
F_1\left(\mathcal{T}\right):= F\left(\mathcal{T}\right) + \frac{1}{2} h\left(\mathbf{U}_1, \mathbf{U}_2, \mathbf{U}_3\right),
\end{align}
where the regularization function $h\left(\mathbf{U}_1, \mathbf{U}_2, \mathbf{U}_3\right) = \frac{1}{2}\sum_{k=1}^3 \left\| \mathbf{U}_k^{\top} \mathbf{U}_k - \mathbf{I}_{r_k} \right\|_{\mathrm{F}}^2.$
This setting is driven by the underlying assumption that the true parameter \(\mathcal{T}^{\star}\) has a Tucker decomposition with ranks $(r_1, r_2, r_3)$ and that each factor matrix \(\mathbf{U}^\star_k\) possesses orthonormal columns for all \( k \in [3] \).
\begin{maroontext}
This assumption ensures that the integrated regularization in our model does not alter the optimal solution of the population risk function.  
In the absence of the regularizer, the factor matrices may become nearly singular.
Furthermore, this regularizer encourages the SGD algorithm to remain within the strong convex region, a crucial aspect for guaranteeing the effectiveness of the first-order optimization procedure in non-convex problems. 
Such regularization has been widely adopted in tackling non-convex challenges in matrix and tensor optimizations, as highlighted in literature \citep{zheng2016convergence, han2022optimal}. Several alternatives exist for ensuring model identifiability. Techniques proposed by \cite{jin2016provable} suggest matrix normalization through SVD at each iteration. Furthermore, \cite{tong2022scaling} introduces a preconditioned gradient descent technique tailored for tensor factor recovery, using preconditioners corresponding to inverse blocks of the Hessian from the population loss. Our approach differs from these strategies by not relying on SVD for matrix renormalization nor requiring the computation of inverse matrix operations to overcome non-uniqueness and hence our approach is computationally more efficient.
\end{maroontext}

Building upon our defined population risk function (\ref{equ:def of L}), we can specify the SGD update rule as:
\begin{align}\label{equ:SGD formula}
\begin{aligned}
	\mathbf{U}_k^{(t)} = & \mathbf{U}_k^{(t-1)} - \eta_t  \nabla_{\mathbf{U}_k} f \left({\mathcal{T}}^{(t-1)} ; \zeta_t\right) \\
    & -  \frac{\eta_t}{2} \nabla_{\mathbf{U}_k} h\left(\mathbf{U}_1^{(t-1)}, \mathbf{U}_2^{(t-1)}, \mathbf{U}_3^{(t-1)}\right), \\
	\mathcal{G}^{(t)} = & \mathcal{G}^{(t-1)} - \eta_t \nabla_{\mathcal{G}} f \left({\mathcal{T}}^{(t-1)} ; \zeta_t\right).
\end{aligned}
\end{align}
Here, $\nabla_{\mathbf{U}_k} h\left(\mathbf{U}_1, \mathbf{U}_2, \mathbf{U}_3\right)$ denotes the partial gradient of function $h$ with respect to $\mathbf{U}_k$, which is defined as: $  \nabla_{\mathbf{U}_k}h(\mathbf{U}_1, \mathbf{U}_2, \mathbf{U}_3) =  \mathbf{U}_k\left(\mathbf{U}_k^{\top} \mathbf{U}_k - \mathbf{I}_{r_k}\right).$ The selection of step size $\{\eta_t\}_{t=1}^\infty$ will be discussed in Remark \ref{rmk:step size}.  We present this specialized stochastic gradient update procedure for the $t$-th step in Algorithm \ref{alg:Low rank tensor SGD}, and refer to it as \emph{Single-step low-rank Tensor SGD}. The indices for \( k+1 \) and \( k+2 \) in \( \mathbf{U}_{k+1} \) and \( \mathbf{U}_{k+2} \) are determined using modulo 3 operations.

\begin{algorithm}[!ht]
\small
\DontPrintSemicolon
\caption{Single-step Low-Rank Tensor SGD for Step $t$}
\label{alg:Low rank tensor SGD}

\KwIn{Previous core tensor estimate \(\mathcal{G}^{(t-1)}\), previous factor matrices \(\{\mathbf{U}_k^{(t-1)}\}_{k=1}^3\), new observation \(\bigl(y_t,\mathcal{X}_t\bigr)\), and step size \(\eta_t\).}

{\color{blue}{\tcc{Update Factor Matrices}}}
\For{$k \in [3]$}{
    $\begin{aligned}
        \mathbf{U}_k^{(t)} = &\mathbf{U}_k^{(t-1)} - \eta_{t} \Big(\left\langle\mathcal{X}_{t}, \mathcal{T}^{(t-1)}\right\rangle - y_{t}\Big)\mathcal{M}_k(\mathcal{X}_{t}) \left(\mathbf{U}_{k+2}^{(t-1)} \otimes \mathbf{U}_{k+ 1}^{(t-1)}\right) \mathcal{M}_k^{\top}(\mathcal{G}^{(t-1)}) \\
        & - \frac{\eta_t }{2}\mathbf{U}_k^{(t-1)} \left(\mathbf{U}_k^{(t-1) \top} \mathbf{U}_k^{(t-1)} - \mathbf{I}_{r_k}\right).
    \end{aligned}$ 
    \;
}

{\color{blue}{\tcc{Update Core Tensor}}}
$\mathcal{G}^{(t)} = \mathcal{G}^{(t-1)} - \eta_{t} \Big(\left\langle\mathcal{X}_{t}, \mathcal{T}^{(t-1)}\right\rangle - y_{t}\Big) \mathcal{X}_{t} \times_{k\in[3]} \mathbf{U}_k^{(t-1)\top}$.\;

\KwOut{
Updated core tensor $\mathcal{G}^{(t)}$, and updated factor matrices $\mathbf{U}_k^{(t)}$ for $k\in[3]$.}
\end{algorithm}

\subsection{Convergence Analysis}
We first introduce the following model assumptions.
\begin{assumption}
\label{cond:1}
\begin{enumerate}[(1)]
  \item The noise \( \{\xi_t\}_{t=1}^\infty \) are i.i.d. sub-Gaussian random variables with parameter \( \sigma \), satisfying \( \mathbb{E}\left[\xi_t \right] = 0 \), \( \mathbb{E}\left[\xi_t^2 \right] = \sigma^2 \), and for all \( x \in \mathbb{R} \), \( \mathbb{E}\left[ \exp\left(x \xi_t\right)\right] \leq \exp\left(\sigma^2x^2/2\right) \). 
\item 
\begin{maroontext}
The design tensor \(\mathcal{X}_t\) consists of i.i.d. sub-Gaussian entries with parameter 1. Each entry satisfies \(\mathbb{E}\left[\mathcal{X}_t\left(j_1, j_2, j_3\right)\right] = 0\), and for all \(x \in \mathbb{R}\), \(\mathbb{E}\left[\exp\left(s\,\mathcal{X}_t\left(j_1, j_2, j_3\right)\right)\right] \leq \exp\left(s^2/2\right)\). Furthermore, assume \( \operatorname{Var}\left(\mathcal{X}_t\left(j_1, j_2, j_3\right)\right)=1\). The tensor \(\mathcal{X}_t\) is independent of \(\xi_t\), and the sequence \(\left\{\mathcal{X}_t\right\}_{t=1}^\infty\) is i.i.d. across \(t\). 
\end{maroontext}
\item The true tensor parameter $\mathcal{T}^{\star}$ is low-rank with Tucker rank $(r_1,r_2,r_3)$, and $r_k \leq \sqrt{p_k}$, for $k\in[3]$, and its condition number is $\kappa_0:=\kappa\left(\mathcal{T}^\star\right)<\kappa$ for a positive constant $\kappa$.
\end{enumerate}
\end{assumption}

Assumption \ref{cond:1} indicates that the observed $y_t$ is affected by a predictable noise $\xi$, the covariate tensor $\mathcal{X}_t$ has i.i.d. sub-Gaussian entries, and the covariate information received at each time is independent from the noise, which are common assumptions in low-rank model literature \citep{raskutti2019convex, tong2022scaling, xia2022inference}. 

In addition, we assume that the true tensor is low-rank and well-conditioned, allowing for efficient estimation and interpretation. The model assumptions outlined in Assumption \ref{cond:1} are standard within the literature, and the low-rank assumption is well justified by real-world examples \citep{zhou2013tensor, chen2019inference, zhang2020islet, han2022optimal}.
\begin{assumption}\label{cond:init}
The initialization $\mathcal{T}^{(0)} = \mathcal{G}^{(0)} \times_{k\in[3]} \mathbf{U}_k^{(0)} $ satisfies $\left\|\mathcal{T}^{(0)}-\mathcal{T}^{\star}\right\|_{\mathrm{F}} \leq C_{\text{init}}\sigma$ for some constant $C_{\text{init}}>0$.
\end{assumption}
Such initialization condition is mild and can be satisfied by spectral methods, such as Higher-Order Singular Value Decomposition (HOSVD) or Higher Order Orthogonal Iteration (HOOI) \citep{delathauwer2000best, zhang2020tensor}. 
\begin{maroontext}
    As shown in \cite[Theorem 1]{zhang2020tensor}, if the signal-to-noise ratio satisfies $\lambda_{\min}/\sigma \gtrsim \sqrt{p^{3/2}/n_0}$, and the initial sample size \(n_0\) is chosen such that $n_0 \gtrsim pr$, HOOI produces an initialization that meets the condition stated in Assumption \ref{cond:init}.
\end{maroontext}

With Assumptions \ref{cond:1} and \ref{cond:init}, we are ready to present the convergence result of our online low-rank estimation obtained through Algorithm \ref{alg:Low rank tensor SGD}. 

\begin{theorem}\label{thm:converge}
For any constant $\alpha \in (0, 1)$, we define the learning rate $\eta_t = \eta_0 \left( \max \left\{t, t^{\star}\right\} \right)^{-\alpha}$ for some constant $\eta_0$, where $t^{\star} = \left(C_{\max} \df\right)^{1/\alpha}$. 
The tensor $\mathcal{T}^{(t)}= \mathcal{G}^{(t)} \times_{k\in[3]} \mathbf{U}_k^{(t)} $ represents the low-rank tensor SGD estimation at time $t$ from Algorithm \ref{alg:Low rank tensor SGD}. Under Assumptions \ref{cond:1}-\ref{cond:init}, if the signal-to-noise ratio (SNR) $\lambda_{\min}/\sigma\geq\widetilde{C}$, where $\widetilde{C}$ is a positive constant, then for any $0 < t \leq n$ and any sufficiently large $\gamma > 0$, with probability at least $1 - 3np^{-\gamma}$, we have
\begin{align}\label{equ:converge thm}
\left\| {\mathcal{T}}^{(t)}- \mathcal{T}^{\star}\right\|_{\mathrm{F}}
\leq C_1 \sigma \left( \frac{\df}{t^\alpha} + \frac{\gamma\log p}{t^\alpha} \sqrt{\frac{\df}{t^\alpha}}\right)^{1/2},
\end{align}
where $\df := r_1 r_2 r_3 + \sum_{k=1}^3 p_k r_k$, and $C_1$ is positive constant.
\end{theorem}
\begin{maroontext}
Under a mild condition \( t^\alpha \geq (\gamma\log p)^2/\df \), the first term on the right side of (\ref{equ:converge thm}) becomes predominant over the second term. Our rate nearly reaches the minimax optimal rate of estimation error \( C\sigma\sqrt{\df/t} \) within the class of \( p_1 \times p_2 \times p_3 \) tensors of Tucker rank-$(r_1 , r_2 , r_3 )$ for offline tensor regression \citep{han2022optimal}.  Moreover, since the constants \(C_1\) and \(\widetilde C\) do not depend on the exponent \(\alpha\), choosing \(\alpha\) arbitrarily close to 1 yields the fastest decay.  Extending to the case \(\alpha=1\) requires delicate handling and refined methods (e.g.\ \citealp{bach2011nonasymptotic}) and is left for future work.
\end{maroontext}
To conclude, we outline several key challenges in our theoretical analysis.
In the offline setting, each iteration at time \( t \) utilizes the entire data set for a new estimate. This process is fundamentally different from our approach, where only a single new data point is available for each update, leading to greater volatility compared to the offline scenario. Moreover, unlike existing gradient-based tensor estimators, which utilize a constant step size, our method employs a time-decaying step size. Given these challenges, we initially establish one-step contraction under expectation. Following this, we construct a super-martingale that includes a time-decaying term. By applying concentration inequalities, we control the sum of these super-martingales up to time \( t \), thereby achieving the desired results.

\begin{remark}[Two-Stage Step-Size Schedule]
\label{rmk:step size}
    The choice of step size $\{\eta_t\}_{t=1}^\infty$ plays a pivotal role in parameter updates. A step size that is excessively large can cause the algorithm to overshoot the minimum, potentially resulting in oscillations around the minimum or even resulting in an explosion due to the initial conditions. Conversely, the algorithm may converge slowly if the step size is too small.  
    In our approach, we introduce a specialized decay step size $\eta_t = \eta_0 \left( \max \left\{t, t^{\star}\right\} \right)^{-\alpha}$, where $t^\star=(C_{\max}\df)^{1/\alpha}$. Equivalently, for $t \le t^\star$ the step size remains constant, $\eta_t = \eta_0\,(C_{\max}\df)^{-1}$, and for $t>t^\star$ it decays at rate $t^{-\alpha}\,. $ To our knowledge, this paper is the first study to apply a decaying step size in an online low-rank tensor regression setting. This specification is different from the constant stepsize specification in the concurrent work of online tensor learning \citep{cai2023online} where they specify the exponent $\alpha=0$ and the constant $\eta_0$ depends on the total sample size. The difference is particularly important since the total sample size is typically unknown at the front in our online setting, and neither can we directly use validation methods to tune the step size in scenarios of streaming data. Further, our specification is indeed different from the stochastic optimization literature (e.g., \citealp{polyak1992acceleration, nemirovski2009robust, bach2011nonasymptotic,  chen2020statistical}), where they use $\eta_t = \eta_0 t^{-\alpha}$ for fixed-dimensional problems. The difference is in the early stage where we choose a less sensitive step size to circumvent excessively large errors and to prevent the estimate from escaping the benign region. This approach
    is essential in high-dimensional problems as $p\rightarrow\infty$, since otherwise the error generated from the first few steps would be irrecoverable in the later stages. Intuitively, if one were using \( \eta_t = \eta_0 t^{-\alpha} \), the first update would lead to a $O(\sqrt{p})$ deviation from the true value due to a constant order of the randomness in the stochastic gradient on each coordinate. In such scenarios, the algorithm would require a much longer trajectory to divert the SGD estimator towards the truth. 
    \begin{maroontext}
    Finally, our SNR requirement coincides with that in \cite{cai2023online}, namely  $\rbr{\lambda_{\min}/\sigma}^2 \geq C\, \df\,\eta_t$. Under our two‑stage schedule, the first stage uses
\(\eta_{t^\star} = (C_{\max}\,\df)^{-1}\) for \(t < t^\star\), which enforces a constant‑order SNR condition: $\rbr{\lambda_{\min}/\sigma}^2 \geq C^\prime$.
If the total learning horizon \(n\) is known in advance, one can instead choose
\(\eta_{t^\star} = O(1/n)\), thereby weakening the SNR requirement to $\rbr{\lambda_{\min}/\sigma}^2 \;\ge\; C^{\prime\prime}\df/n\,$.
\end{maroontext}
\end{remark}

%% file: body/Inference.tex
\section{Online Statistical Inference for Low-Rank Tensors}
\label{sec:inference}
While convergence analysis in optimization informs us about estimation error bounds, accurately determining the outcome distributions of algorithms that tackle complex optimization problems without closed-form solutions remains challenging. 
In this section, we develop inferential procedures for general linear forms \( h^\star = \langle \mathcal{T}^\star, \mathcal{H} \rangle \) of the true parameter tensor \(\mathcal{T}^{\star}\), where \(\mathcal{H}\) is any fixed tensor encoding hypotheses of interest, such as contrasts between specific entries of \(\mathcal{T}^{\star}\). 
\begin{maroontext}
If we are interested in entrywise inference—i.e., making statistical statements about the individual entry \(\mathcal{T}_{j_1,j_2,j_3}^\star\)—we set $\mathcal{H}_1 = e_{j_1}\otimes e_{j_2}\otimes e_{j_3}$,
so that
\begin{align}\label{equ:intro h-1}
    \langle\mathcal{H}_1,\mathcal{T}^{\star}\rangle
    = \langle e_{j_1}\otimes e_{j_2}\otimes e_{j_3},\,\mathcal{T}^\star\rangle
    = \mathcal{T}^\star_{j_1,j_2,j_3}.
\end{align}
To test whether two entries differ, we define
$\mathcal{H}_2 = e_{j_1}\otimes e_{j_2}\otimes e_{j_3}
\;-\;e_{\ell_1}\otimes e_{\ell_2}\otimes e_{\ell_3}$,
which gives
\begin{align}\label{equ:intro h-2}
    \langle\mathcal{H}_2,\mathcal{T}^{\star}\rangle
    = \mathcal{T}^\star_{j_1,j_2,j_3} - \mathcal{T}^\star_{\ell_1,\ell_2,\ell_3}.
\end{align}
\end{maroontext}
Additionally, as a byproduct of our algorithm, we characterize the distribution of the low-rank factors \(\mathbf{U}_k^{\star} \in \mathbb{R}^{p_k \times r_k}\) for \(k \in [3]\) in the Section \ref{sec:Proof of U normality} of the Supplementary Material. This analysis helps us examine the distance between empirical and true singular subspaces through confidence regions for \(\mathbf{U}_k^{\star}\). These two inferential tasks are closely interrelated.  The tensor linear form estimate is computed by projecting the debiased average of the tensor-based SGD estimator onto the space spanned by the low-rank factor matrices. This projection process not only yields the desired estimate but also provides the factor matrices with distributional characteristics.

\subsection{Constructing De-biased Estimators}
Given that our objective function (\ref{equ:def of L}) is an optimization problem with low-rank constraint, the regularization term behaves effectively like shrinkage estimators, indicating that the provided estimates necessarily suffer from non-negligible bias. To enable desired statistical inference, it is important to correct the estimation bias. For low-rank models, it is typical to apply a sample-splitting procedure to achieve an unbiased estimator. However, thanks to the online nature of our approach, we circumvent the need for data splitting, thereby automatically avoiding the potential loss of information. 

Based on the tensor-based SGD estimator \( \mathcal{T}^{(t)} \), we introduce a natural online procedure for bias correction.
Specifically, with the arrival of new data \( \boldsymbol{\zeta}_t \) at time \( t \), we guide $ {\mathcal{T}}^{(t-1)}$ to move a fixed step length in the direction of $\nabla_{\mathcal{T}} f \left(\mathcal{T}^{(t-1)};\boldsymbol{\zeta}_t\right)$ to obtain 
\begin{align}\label{equ:debias}
\begin{aligned}
	\widehat{\mathcal{T}}^{(t)}=&\frac{t-1}{t}\widehat{\mathcal{T}}^{(t-1)}+\frac{1}{t}\left[ \vphantom{\frac{x}{y}} {\mathcal{T}}^{(t-1)}-  \nabla_{\mathcal{T}} f \left(\mathcal{T}^{(t-1)}\right)\right] \\
 =& \frac{1}{t} \sum_{\tau=1}^{t} {\mathcal{T}}^{(\tau-1)}- \frac{1}{t}\sum_{\tau=1}^{t}\nabla_{\mathcal{T}} f \left(\mathcal{T}^{(\tau-1)}\right), 
\end{aligned}
\end{align}
where $ {\mathcal{T}}^{(t-1)}$ denotes the SGD estimator, and $\widehat{\mathcal{T}}^{(t)}$ refers to the debiased estimator.
 The intuition of this online debias procedure is that \( \nabla_{\mathcal{T}} f(\mathcal{T}^{(t-1)}; \boldsymbol{\zeta}_t) \), the gradient of the loss function at the \( (t-1) \)-th estimate \( \mathcal{T}^{(t-1)} \) and the sample at time \( t \), does not enforce a low-rank constraint, thereby pushing \( \mathcal{T}^{(t-1)} \) in the direction of the true parameter \( \mathcal{T}^\star \). 
The term $\frac{1}{t}\sum_{\tau=1}^{t}\nabla_{\mathcal{T}} f \left(\mathcal{T}^{(\tau-1)}; \boldsymbol{\zeta}_{\tau}\right)$ computes the average of all the gradients of the loss function over time steps up to \(t\) and can be viewed as utilizing all previous online data samples $\left\{\boldsymbol{\zeta}_t\right\}_{t=1}^n$ to form an estimator of $\nabla F\left(\mathcal{T}\right)$ at time $t$. This approach effectively ``kills two birds with one stone": it offsets the bias introduced by the low-rankness and, at the same time, reduces the inherent variance of SGD.

 Despite being unbiased, the tensor estimates $\widehat{\mathcal{T}}^{(t)}$ are not necessarily low-rank, with non-negligible energy spread across the entire spectrum, which increases the variability in the estimates. 
 To remedy this issue, we propose to further project $\widehat{\mathcal{T}}^{(t)}$ onto the low-rank space, leading to the following estimator
\begin{align}\label{equ:project}
 \mathcal{P}_{\text{rank-}\mathbf{r}}\left[\widehat{\mathcal{T}}^{(t)}\right] = \widehat{\mathcal{T}}^{(t)} \times_{k\in[3]} \mathcal{P}_{\widehat{\mathbf{U}}_{k}^{(t)}},
\end{align}
where $\mathcal{P}_{\text{rank-}\mathbf{r} }(\widetilde{\mathcal{T}})=\arg \min _{{\mathcal{T}}: \operatorname{rank}(\mathcal{T}) \leq \mathbf{r}}\|\widetilde{\mathcal{T}}-\mathcal{T}\|_{\mathrm{F}}$. This projection step suppresses the variability outside the $\mathbf{r}$-dimensional principal subspace.
The factor matrices \( \widehat{\mathbf{U}}_k^{(t)} \) for $k\in[3]$ in Equation (\ref{equ:project}) can be estimated utilizing HOSVD \citep{delathauwer2000multilinear}. As pointed by \cite{xia2022inference},
solving
$\min_{\operatorname{rank}(\mathcal{T}) \leq\left(r_1, r_2, r_3\right)}\|\widetilde{\mathcal{T}}-{\mathcal{T}}\|_{\mathrm{F}}$ is equivalent to solving $\max _{\mathbf{U}_k \in \mathbb{O}_{p_k, r_k}}\| \widetilde{\mathcal{T}} \times_{k\in[3]}  \mathbf{U}_k^{\top}\|_{\mathrm{F}}$. Furthermore, the Eckart-Young-Mirsky Theorem \citep{eckart1936approximation} implies that when $\widehat{\mathbf{U}}_{k+1}^{(t-1)}$ and $\widehat{\mathbf{U}}_{k+2}^{(t-1)}$ are held constant, the optimal solution for this maximization is attainable via SVD: $
\widehat{\mathbf{U}}_k^{(t)} = \text{SVD}_{r_k}( \mathcal{M}_k(\widehat{\mathcal{T}}^{(t)} \times_{j\neq k} \widehat{\mathbf{U}}_{j}^{(t-1) \top} ))$, for $k\in[3]. $
Intuitively, the HOSVD operates by holding certain tensor modes constant, subsequently identifying the subspace that maximizes the projection value.
  We formalize the procedure for online tensor inference at step \(t\) in Algorithm \ref{alg:Single-step Tensor Linear Form Estimator Update}.

 \begin{algorithm}[!ht]
 \small
\caption{Single-step Tensor Linear Form Estimator Update for Step $t$}\label{alg:Single-step Tensor Linear Form Estimator Update}
\KwIn{
    Linear Form $\mathcal{H}$, online low-rank SGD estimator $\mathcal{T}^{(t-1)}$ and its 
    projected matrices $\widehat{\mathbf{U}}_{k}^{(t-1)}$ \vspace{-0.2cm} for $k \in [3]$, 
    new data $\zeta_{t} = (\mathcal{X}_{t}, y_{t})$.
}

{\color{blue}{\tcc{Update Average SGD Estimator}}}
$\widehat{\mathcal{T}}^{(t)}=\frac{t-1}{t} \widehat{\mathcal{T}}^{(t-1)}+\frac{1}{t}\left( {\mathcal{T}}^{(t-1)}- \left(\left\langle {\mathcal{T}}^{(t-1)}, \mathcal{X}_{t} \right\rangle-y_{t}\right) \mathcal{X}_{t}\right)$.

{\color{blue}{\tcc{Updating Factor Matrices and Corresponding Singular Values}}}
$ \widehat{\mathbf{U}}_k^{(t)}, \widehat{\mathbf{\Lambda}}_k^{(t)} = \text{SVD}_{r_k}\left( \mathcal{M}_k\left(\widehat{\mathcal{T}}^{(t)} \times_{k+1} \widehat{\mathbf{U}}_{k+1}^{(t-1) \top} \times_{k+2} \widehat{\mathbf{U}}_{k+2}^{(t-1) \top}\right)\right)\text{, for }k\in[3]$.

{\color{blue}{\tcc{Update Tensor Linear Form Estimator}}}
$ \hat{h}^{(t)}
 =\left\langle\widehat{\mathcal{T}}^{(t)}\times_1\mathcal{P}_{\widehat{\mathbf{U}}_1^{(t)}}  \times_2 \mathcal{P}_{\widehat{\mathbf{U}}_2^{(t)}}  \times_3 \mathcal{P}_{\widehat{\mathbf{U}}_3^{(t)}} , \mathcal{H}\right\rangle.$
 	
\KwOut{
    Linear form estimate $\hat{h}^{(t)}$, factor matrices $\widehat{\mathbf{U}}_{k}^{(t)}$ and singular values $\widehat{\mathbf{\Lambda}}_k^{(t)}$ for $k \in [3]$.
}
\end{algorithm}

\subsection{Asymptotic Normality of $\hat{h}^{(n)}$} 
Our main result is an asymptotic normality theorem for the estimator
\begin{align}\label{equ:def of mt}
    \hat{h}^{(n)} = \left\langle  \mathcal{P}_{\text{rank-}\mathbf{r}}\left[\widehat{\mathcal{T}}^{(n)}\right], \mathcal{H} \right\rangle,
\end{align}
where \(\mathcal{P}_{\text{rank-}\mathbf{r}}\left[\widehat{\mathcal{T}}^{(n)}\right]\) denotes the low-rank estimator defined in Equation (\ref{equ:project}) at time horizon $n$, and \(\mathcal{H}\) is any fixed tensor encoding hypotheses of interest.  
We begin by introducing an assumption necessary for the theoretical distributional analysis. This assumption addresses the challenge of making inferences about general linear forms of \( \mathcal{T}^{\star} \), which is complicated by the complex dependence among the estimated entries.
\begin{assumption}\label{cond:4}
Let \(n\) denote the total sample size and suppose that there exist a constant \(C>0\) such that $n^\alpha \left( \lambda_{\min}/\sigma \right)^2 \geq C\, \df^2$, and $S_\mathcal{H} \geq C\, \max_{k} \{ (1/p) \|\mathcal{H}\|_{\mathrm{F}}, \; \sqrt{1/p} \|\mathcal{H} \times_k \mathbf{U}_{k}^\star\|_{\mathrm{F}} \}$, where $S_\mathcal{H}^2 = \|\mathcal{H}  \times_{k\in[3]} \UT{k}\|_{\mathrm{F}}^2 +  \sum_{k=1}^3\| \mathcal{P}_{\mathbf{U}_k^\star}^{\perp}\mathbf{H}_k \mathcal{P}_{\left(\mathbf{U}_{k+2}^\star\otimes\mathbf{U}_{k+1}^\star\right)  \mathbf{V}_k^\star}\|_\mathrm{F} ^2$ and $\mathbf{V}_k^\star$ is the right singular space of $\mathcal{M}_k\left(\mathcal{G}\right) \in \mathbb{R}^{r_k \times r_{k+1}r_{k+2}}$. In addition, there exist a positive constant $\gamma$ such that $n = o(p^\gamma)$.
\end{assumption}
\begin{maroontext}
    The lower bound on \(S_{\mathcal H}\) guarantees the variance component arising from the fixed tensor \(\mathcal H\) is sufficiently large to dominate the error terms caused by the dependence between our online estimate (Theorem \ref{thm:converge}) and the debiasing step (Equation~\eqref{equ:debias}). Under this assumption, the estimation error bound presented in Theorem \ref{thm:converge} vanishes as \( n, p \rightarrow \infty \), ensuring that any additional variance introduced by our debiasing procedure becomes negligible. It also precludes the cases where \(\mathcal H\) is nearly orthogonal to the singular spaces of \(\mathcal T^\star\). Unlike prior works \citep{chen2019inference,xia2021statistical} that require an incoherence condition \(\|\mathbf{U}^\star_j\|_{2,\infty}\asymp p^{-1/2}\), our framework relaxes these constraints on the factor matrices, thereby broadening the applicability of low‑rank tensor inference.  Furthermore, Assumption \ref{cond:4} sets an error bound for factor matrices, specifically:
 $\max _{k\in[3]}\|\sin \Theta (\widehat{\mathbf{U}}_k^{(n)}, \mathbf{U}_k^\star ) \|=O_p( (\sigma/\lambda_{\min }) \sqrt{p/n})$. This bound ensures that the asymptotic standard deviation of the main term in the CLT is dominant over other remainder terms within our theoretical framework.
\end{maroontext}

Finally, we are now ready to present the asymptotic normality of the estimator $\hat{h}^{(n)}$ in (\ref{equ:def of mt}).
\begin{theorem}\label{thm:entry inference}
Under the assumptions for Theorem \ref{thm:converge} and Assumption \ref{cond:4}, as $n, p \rightarrow \infty$ we have $$\sqrt{n}\left(\hat{h}^{(n)}-h^{\star}\right)/(\sigma S_\mathcal{H}) \stackrel{\text {d}}{\longrightarrow} \mathcal{N}\left(0, 1\right).$$
\end{theorem}
The detailed proof of this theorem is provided in the Section \ref{sec:Proof of Theorem entry inference}. It is worth mentioning that our online debiasing procedure in Equation \eqref{equ:debias} accelerates the convergence rate. The averaging procedure in Equation \eqref{equ:debias} enhances the convergence rate from \(O(n^{-\alpha})\) for \(\alpha \in (0,1)\) shown in Theorem \ref{thm:converge} to the optimal \(O(n^{-1})\). This improvement is analogous to results in the vector setting (see, e.g., \cite{polyak1992acceleration, bach2011nonasymptotic}). We outline several key challenges faced in our theoretical analysis in Theorem \ref{thm:entry inference}. To the best of our knowledge, there is no existing literature addressing the asymptotic normality of the general linear form in tensor regression model with a Tuck low-rank structure, especially in an online setting. Typically, statistical analysis of low-rank models relies on SVD operations. While substantial progress has been made in methodologies and theories for matrix SVD and matrix regression, literature on tensors of order three or higher is scarce. This scarcity is notable because SVD for high-order tensors presents more complex challenges than for matrices. Furthermore, converting a third-order tensor into a matrix often results in a highly unbalanced matrix, leading to suboptimal sample complexity if matrix theory is applied directly.
To address these issues, \cite{delathauwer2000multilinear,  delathauwer2000best} introduced methods such as the HOSVD and HOOI, targeting efficient spectral and power iteration methods for optimal low-rank approximation. However, HOSVD introduces complex statistical dependencies during the iterative optimization process. Given that our data are collected sequentially, traditional offline sample analysis and sample splitting approaches are not suitable. Instead, we utilize the spectral representation \citep{xia2021statistical,xia2022inference,zhou2023heteroskedastic,agterberg2024estimating} to handle this dependence.

\subsection{Online Parameter Inference of $\hat{h}^{(n)}$}
\label{sec:parameter inference of mt}
The distributional guarantees established in Theorem \ref{thm:entry inference} lay the groundwork for statistical inference concerning \( h^\star \). 
To construct the confidence intervals for model parameters, it is essential to estimate the variance of \( \hat{h}^{(n)} \) in an online manner without the need to store all historical data. A practical approach is using the online plugin estimator, as considered in the works of \cite{chen2020statistical}.
In our setting, the online plugin estimators for $\sigma^2$ and $S^2$ can be constructed by
\begin{align}\label{equ:sigma and S}
\begin{aligned}
    \hat{\sigma}_n^2 
= & \frac{n-1}{n} \hat{\sigma}_{n-1}^2+\frac{1}{n} \left(y_n-\left\langle\mathcal{T}^{(n)}, \mathcal{X}_n\right\rangle\right)^2,\\
\widehat{S}_{\mathcal{H},n}^{2}  = & \sum_{k=1}^3 \left\| \mathcal{P}_{\widehat{\mathbf{U}}_k^{(n)}}^{\perp}\,\mathbf{H}_k\,\mathcal{P}_{\bigl(\widehat{\mathbf{U}}_{k+2}^{(n)}\otimes \widehat{\mathbf{U}}_{k+1}^{(n)}\bigr)\, \widehat{\mathbf{V}}_k^{(n)}} \right\|_{\text{F}}^{2}\\
& + \left\|\mathcal{H} \times_{k\in[3]} {\widehat{\mathbf{U}}_k^{(n)\top }}  \right\|_{\mathrm{F}}^2,
\end{aligned}
\end{align}
where $\widehat{\mathbf{V}}_k^{(n)} = \mathrm{QR}[\mathcal{M}_k(\widehat{\mathcal{G}}^{(n)})^{\top}] = \mathrm{QR}[\mathcal{M}_k(\widehat{\mathcal{T}}^{(n)} \times_{k=1}^3 \widehat{\mathbf{U}}_k^{(n)\top})^{\top}]$ is the estimate of the right singular space of the mode-$j$ matricization of the core tensor $\mathcal{G} \in \mathbb{R}^{r_1 \times r_2 \times r_3}$. Notably, both \(\hat{\sigma}_n^2\) and \(\widehat{S}_{\mathcal{H},n}^2\) can be updated in an online fashion, without retaining all past observations.
To ensure the validity of the online inference procedure, it only remains to prove the consistency of the proposed variance estimator.
\begin{theorem}\label{thm:entry inference consistent}
Under the assumptions of Theorem \ref{thm:entry inference}, as $n, p \rightarrow \infty$, we have
\begin{align*}
\begin{gathered}
\sqrt{n}\left(\hat{h}^{(n)}-h^{\star}\right)/  (\hat{\sigma}_n \widehat{S}_{\mathcal{H},n} )\stackrel{\text {d}}{\longrightarrow} \mathcal{N}(0,1).
\end{gathered}
\end{align*}
\end{theorem}
Given the result of Theorem \ref{thm:entry inference consistent}, we can construct an asymptotic exact confidence interval for the true parameter $h^{\star}$. In particular, at any confidence level $\alpha \in(0,1)$, we can define the $100(1-\alpha) \%$-th confidence interval as 
$
\widehat{\mathrm{CI}}^\alpha_{h,n} = \left(\hat{h}^{(n)}-z_{\alpha / 2} \hat{\sigma}_n \widehat{S}_{\mathcal{H},n} / \sqrt{n}, \hat{h}^{(n)}+z_{\alpha / 2} \hat{\sigma}_n \widehat{S}_{\mathcal{H},n} / \sqrt{n}\right),
$
where $z_{\alpha}=\Phi^{-1}(1-\alpha)$ denotes the score of standard normal distribution for the upper $\alpha$-quantile.  
By Theorem \ref{thm:entry inference consistent}, we have
$ \lim _{n, p \rightarrow \infty} \mathbb{P}\left(h^{\star} \in \widehat{\mathrm{CI}}^\alpha_{h,n}\right)=1-\alpha.$
\begin{maroontext}
Applying Theorems \ref{thm:entry inference} and \ref{thm:entry inference consistent}, we construct the following confidence interval for \(\mathcal{T}_{j_1,j_2,j_3}\) in Equation \eqref{equ:intro h-1}: $\widehat{\mathrm{CI}}^\alpha_{h,n} = \left( \widetilde{\mathcal{T}}^{(n)}_{j_1,j_2,j_3} - z_{\alpha/2}{\hat{\sigma}_n \widehat{S}_{\mathcal{H},n}}/{\sqrt{n}}, \  \widetilde{\mathcal{T}}^{(n)}_{j_1,j_2,j_3} + z_{\alpha/2}{\hat{\sigma}_n \widehat{S}_{\mathcal{H},n}}/{\sqrt{n}} \right),$ where \(\widetilde{\mathcal{T}}^{(n)}\) is the projected debiased estimator defined in Equation \eqref{equ:project}. The variance component is estimated by
$\widehat{S}_{\mathcal{H},n}^{2} = \prod_{k=1}^3 \|\widehat{\mathbf{U}}_k^{(n)\top} e_{j_k}\|_{\mathrm{F}}^2 + \sum_{k=1}^3 \|\mathcal{P}_{\widehat{\mathbf{U}}_k^{(n)}}^\perp e_{j_k}\|^2 \|\widehat{\mathbf{U}}_{k+1}^{(n)\top} e_{j_{k+1}}\|^2 \|\widehat{\mathbf{U}}_{k+2}^{(n)\top} e_{j_{k+2}}\|^2$. The corresponding confidence interval for the difference in Equation \eqref{equ:intro h-2} is constructed as $\widehat{\mathrm{CI}}^\alpha_{h,n} = \left( \widetilde{\mathcal{T}}^{(n)}_{j_1,j_2,j_3} - \widetilde{\mathcal{T}}^{(n)}_{l_1,l_2,l_3} - z_{\alpha/2}{\hat{\sigma}_n \widehat{S}_{\mathcal{H},n}}/{\sqrt{n}}, \ \widetilde{\mathcal{T}}^{(n)}_{j_1,j_2,j_3} - \widetilde{\mathcal{T}}^{(n)}_{l_1,l_2,l_3} + z_{\alpha/2}{\hat{\sigma}_n \widehat{S}_{\mathcal{H},n}}/{\sqrt{n}} \right).$
In this case, the variance component \(\widehat{S}_{\mathcal{H},n}^{2}\) is estimated as $	\widehat{S}_{\mathcal{H},n}^{2} =  \prod_{k=1}^3\|\widehat{\mathbf{U}}_k^{(n)\top}(e_{j_k} - e_{l_k})\|_{\mathrm{F}}^2 
	+  \sum_{k=1}^3 \|\mathcal{P}_{\widehat{\mathbf{U}}_k^{(n)}}^\perp(e_{j_k} - e_{l_k})\|_{\mathrm{F}}^2 \|\widehat{\mathbf{V}}_k^{(n)\top}(\widehat{\mathbf{U}}_{k+2}^{(n)} \otimes \widehat{\mathbf{U}}_{k+1}^{(n)})^\top[(e_{j_{k+1}} - e_{l_{k+1}}) \otimes (e_{j_{k+2}} - e_{l_{k+2}})]\|_{\mathrm{F}}^2.$
\end{maroontext}
The entire procedure for conducting inference for \( h^{\star} \) is summarized in Algorithm \ref{alg:OnlineSeqInference}.

\begin{algorithm}[!ht]
\small
\DontPrintSemicolon  
\SetAlCapHSkip{0em}  
\SetInd{0.5em}{0.5em}  
\DecMargin{1em}  
\caption{Online Sequential Inference Algorithm}\label{alg:OnlineSeqInference}
\KwIn{ 
Initial estimate $\mathcal{T}^{(0)}$, $\widehat{\mathbf{U}}_k^{(t_0)}=\mathbf{U}_k^{(0)}$, for $k\in[3]$, $\hat{\sigma}^2_{0}=0$, 
step size $\{\eta_t\}$, rank $(r_1, r_2, r_3)$, significance level $\alpha$.}

\For{$t = 1, 2, \ldots$}{
 Receive new observation $(\mathcal{X}_t,y_t)$.
 
{\color{blue}{\tcc{ Estimation Task}}}
$\mathbf{U}_1^{(t)}, \mathbf{U}_2^{(t)}, \mathbf{U}_3^{(t)}, \mathcal{G}^{(t)} \leftarrow$ Algorithm \ref{alg:Low rank tensor SGD} $\left(\mathbf{U}_1^{(t-1)}, \mathbf{U}_2^{(t-1)}, \mathbf{U}_3^{(t-1)}, \mathcal{G}^{(t-1)}, \mathcal{X}_t, y_t, \eta_t\right).$

{\color{blue}{\tcc{Inference Task}}}
$\hat{h}^{(t)}, \widehat{\mathbf{U}}_k^{(t)}, \widehat{\mathbf{\Lambda}}_k^{(t)},$
$k\in[3]$ $\leftarrow$ Algorithm \ref{alg:Single-step Tensor Linear Form Estimator Update} $\left(\mathcal{T}^{(t-1)}, \widehat{\mathbf{U}}_1^{(t-1)}, \widehat{\mathbf{U}}_2^{(t-1)}, \widehat{\mathbf{U}}_3^{(t-1)}, \mathcal{X}_t, y_t, \mathcal{M}\right)$.

{\color{blue}{\tcc{Update Plug in Estimate}}}
$\hat{\sigma}_t^2
=\frac{t-1}{t} \hat{\sigma}_{t-1}^2+\frac{1}{t} \left(y_t-\left\langle\mathcal{T}^{(t)}, \mathcal{X}_t\right\rangle\right)^2$. \;
$\widehat{S}_{\mathcal{H},t}^{2}  = \|\mathcal{H} \times_{k\in[3]} {\widehat{\mathbf{U}}_k^{(t)\top }}  \|_{\mathrm{F}}^2 +  \sum_{k=1}^3 \| \mathcal{P}_{\widehat{\mathbf{U}}_k^{(t)}}^{\perp}\,\mathbf{H}_k\,\mathcal{P}_{\bigl(\widehat{\mathbf{U}}_{k+2}^{(t)}\otimes \widehat{\mathbf{U}}_{k+1}^{(t)}\bigr)\, \widehat{\mathbf{V}}_k^{(t)}} \|_{\text{F}}^{2}$.

    {\color{blue}{\tcc{Calculate $(1-\alpha)$-level Confidence Intervals}}}
$\widehat{\mathrm{CI}}^\alpha_{h,t}= \left(\hat{h}^{(t)}-z_{\alpha / 2} \hat{\sigma}_t \widehat{S}_{\mathcal{H},t} / \sqrt{t} ,
\quad \hat{h}^{(t)}+z_{\alpha / 2} \hat{\sigma}_t \widehat{S}_{\mathcal{H},t} / \sqrt{t}\right)$.

{\color{blue}{\tcc{Update SGD Tensor Estimator}}}
$\mathcal{T}^{(t)} = \mathcal{G}^{(t)} \times_1 \mathbf{U}_1^{(t)} \times_2 \mathbf{U}_2^{(t)} \times_3 \mathbf{U}_3^{(t)}$.\;
}
\KwOut{$\left\{\hat{h}^{(t)}\right\}$, $\left\{ \widehat{\mathrm{CI}}^\alpha_{h, t}\right\}$. }
 	\end{algorithm}

%% file: body/Simulation_revision.tex
In this section, we evaluate the empirical performance of our proposed online tensor estimation and inference procedures through numerical simulations. We first describe the data‑generating process, then evaluate our estimation algorithm across varying tensor dimensions, true Tucker ranks, and noise levels. Next, we compare its performance against the recently proposed oRGrad method \citep{cai2023online}. In the second part, we examine the accuracy of our inference procedure for different linear form tests, tensor dimensions, true ranks, and noise intensities. We also investigate the sensitivity of Algorithm \ref{alg:Low rank tensor SGD} to its hyperparameters and extend our experiments beyond Gaussian designs and batch settings in Section \ref{sec:appendix_simulation} of the Supplementary Material.

We first generate a core tensor \(\widetilde{\mathcal{G}} \in \mathbb{R}^{r_1 \times r_2 \times r_3}\) with independent standard Gaussian entries. To control the signal strength, we rescale it as  \( \mathcal{G}^\star = \widetilde{\mathcal{G}} \cdot \lambda / \min_{k\in[3]} \sigma_{r_k}(\mathcal{M}_k(\widetilde{\mathcal{G}})) \) 
   where \(\lambda = 2\) represents the signal level, ensuring \(\min_{k \in [3]} \sigma_{r_k}(\mathcal{M}_k(\mathcal{T}^\star)) = \lambda\).   
   For each mode \(k \in [3]\), we generate \(\widetilde{\mathbf{U}}_k \in \mathbb{R}^{p_k \times r_k}\) with independent standard uniform entries. These are orthonormalized via QR decomposition to obtain \(\mathbf{U}_k^\star = \mathrm{QR}(\widetilde{\mathbf{U}}_k)\), ensuring \(\mathbf{U}_k^\star\) is uniformly distributed over the Stiefel manifold \(\mathbb{O}_{p_k, r_k}\).   
   The ground-truth tensor is constructed as  $\mathcal{T}^\star = \mathcal{G}^\star \times_{k \in [3]} \mathbf{U}_k^\star$.
   Covariate tensors \(\{\mathcal{X}_t\}\) are generated with independent standard Gaussian entries. Observations \(\{y_t\}\) follow the regression model in Equation (\ref{equ:regression}), with additive Gaussian noise \(\xi_t \sim \mathcal{N}(0, \sigma^2)\) and noise level \(\sigma = 1\).  
 The step size \(\eta_t\) follows Theorem \ref{thm:converge} with decay rate \(\alpha = 0.999\).     
 We set \(n_0 = 30 \sqrt{\lambda/\sigma}\df\) initial samples and the initial estimate \(\mathcal{T}^{(0)}\) is computed via a two-step spectral method, first constructing the averaged weighted tensor \(\widetilde{\mathcal{T}} =  \sum_{i=1}^{n_0} y_i \mathcal{X}_i/n_0\) and then applying the HOOI to \(\widetilde{\mathcal{T}}\), yielding the factorization \( \mathcal{G}^{(0)}\times_{k\in[3]} \mathbf{U}_k^{(0)} = \operatorname{HOOI}(\widetilde{\mathcal{T}}, (r_1, r_2, r_3))\).

For the learning task, we adopt relative error as the evaluation metric, defined as $\text{Relative Error} = \|\mathcal{T}^{(t)} - \mathcal{T}^\star\|_{\mathrm{F}}/\|\mathcal{T}^\star\|_{\mathrm{F}}$, where \(\mathcal{T}^{(t)}\) is the tensor output by the Algorithm \ref{alg:Low rank tensor SGD} and \(\mathcal{T}^\star\) is the true tensor in Equation (\ref{equ:regression}). For our baseline configuration, we set \(p=20\), \(r=2\), \(\sigma = 1\), \(\eta_0 = 5 \times 10^{-5}\), \(\alpha = 0.999\), \(t^\star= 10{,}000\), and $T = 20,000$. 
Each experimental setting is repeated over 100 Monte Carlo replicates, and we report the median values in our plots. For the inference task, we present the results using blue histograms (derived from 1,000 simulation runs) with an overlaid red standard normal density curve, demonstrating the accuracy of our inference procedure.

\begin{figure}[!ht]
    \centering
    \includegraphics[width=1\linewidth]{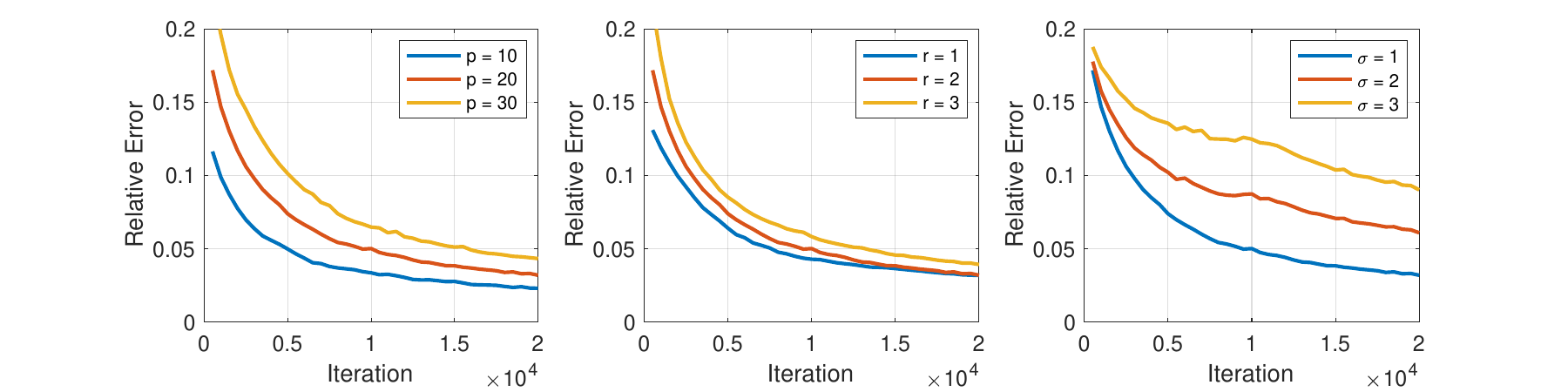}
    \caption{Error analysis for our online tensor estimation across different dimensions, ranks, and noise levels.}
    \label{fig:learning-1}
\end{figure}

We examine the effect of varying tensor dimensions, true rank, and noise level in Figure \ref{fig:learning-1}.  The left subplot presents experiments with tensor dimensions \(p = 10, 20,\) and \(30\), while keeping the rank fixed at \(r = 2\) and the noise level at \(\sigma = 1\). All curves exhibit a monotonic decrease in relative error, converging to low error values. 
 The central subplot compares performance for \(r = 1, 2, 3\) (with fixed \(p = 20\) and \(\sigma = 1\)). As expected, higher ranks require more iterations to stabilize, reflecting the increased complexity of estimating higher-dimensional core tensors. 
  The right subplot examines noise resilience by varying \(\sigma\) (1, 2, 3) for \(p = 20\) and \(r = 2\). final error levels scale approximately linearly with noise intensity (e.g., 0.03 for \(\sigma = 1\) versus 0.09 for \(\sigma = 3\)), consistent with the theoretical relationship \(\|\mathcal{T}^{(t)}-\mathcal{T}^\star\|_\mathrm{F} \propto \sigma\).

\begin{figure}[!ht]
    \centering
    \includegraphics[width=0.75\linewidth]{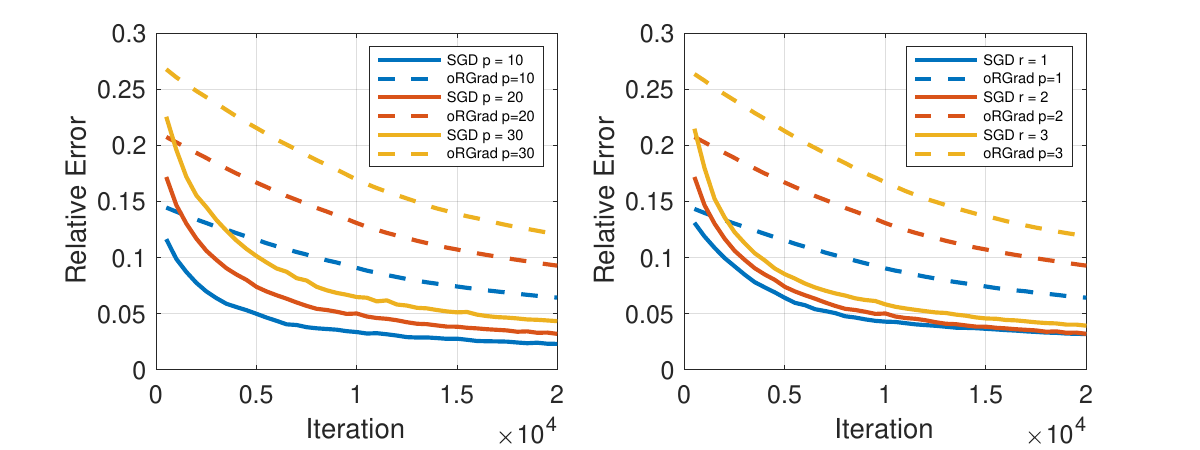}
    \caption{Error analysis for our online tensor estimation across different dimensions $p$ and ranks $r$.}
    \label{fig:learning-4}
\end{figure}

We also compare our method with the oRGrad algorithm \citep{cai2023online} across varying dimensions and ranks in Figure \ref{fig:learning-4}. 
For \(p = 10\), our method achieves a final relative error of approximately 0.02, while oRGrad stagnates near 0.06. As \(p\) increases to 20 and 30, our method maintains stable convergence (around 0.04 error), whereas oRGrad’s error increases (up to approximately 0.12). For \(r = 1\), our method converges to a relative error of about 0.04 compared to 0.06 for oRGrad. For higher ranks (\(r = 2, 3\)), our method consistently preserves a low error floor (around 0.04), while oRGrad’s error increases to around 0.13. Additionally, our method runs approximately three times faster than oRGrad, partly due to avoiding the need to construct a low-rank gradient at every step.

\begin{figure}[!ht]
    \centering
\includegraphics[width=1\linewidth]{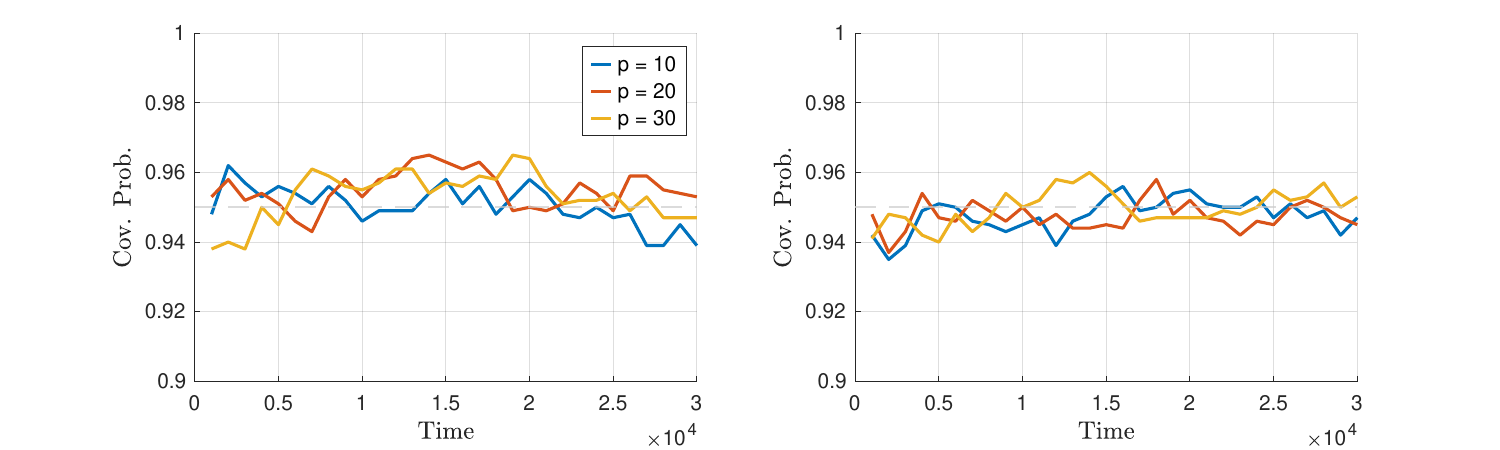}
\caption{Coverage probabilities for hypothesis tests in \eqref{equ:example-1} (left plot) and \eqref{equ:example-2} (right plot).
}
    \label{fig:clt-2}
\end{figure}

    Next, we present the coverage probabilities of our proposed confidence intervals in Figure \ref{fig:clt-2}. Specifically, we evaluate our method under varying dimensions (\(p = 10\), \(20\), and \(30\)) and perform two distinct hypothesis tests. The left subfigure corresponds to the single-entry test:
\begin{align}\label{equ:example-1}
    H_0: \mathcal{T}^\star(1,1,1) = 0 \quad \text{vs} \quad H_1: \mathcal{T}^\star(1,1,1) \neq 0 .
\end{align}
The right subfigure corresponds to the linear combination test:
\begin{align}\label{equ:example-2}
\small
\begin{aligned}
	 & H_0: \mathcal{T}^\star(1,1,1) + 2\,\mathcal{T}^\star(2,2,2) - 3\,\mathcal{T}^\star(3,3,3) = 0 \\
	   \text{vs.} \quad & H_1: \mathcal{T}^\star(1,1,1) + 2\,\mathcal{T}^\star(2,2,2) - 3\,\mathcal{T}^\star(3,3,3) \neq 0.
\end{aligned}
\end{align}
For both tests, the observed coverage probabilities are around 95\%,  which aligns well with our pre-specified confidence level. 
\begin{figure}[!ht]
    \centering
\includegraphics[width=0.75\linewidth]{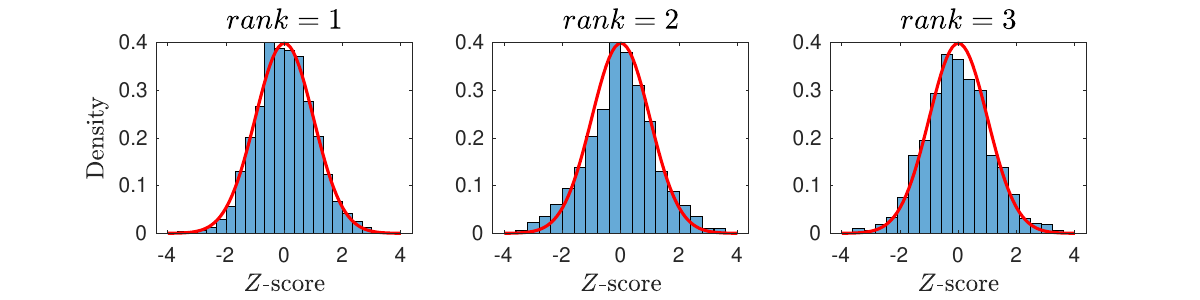}
\caption{Histogram of normal approximation over 1000 independent trails for different ranks.}
    \label{fig:clt-3}
\end{figure}
\begin{figure}[!ht]
    \centering
\includegraphics[width=0.75\linewidth]{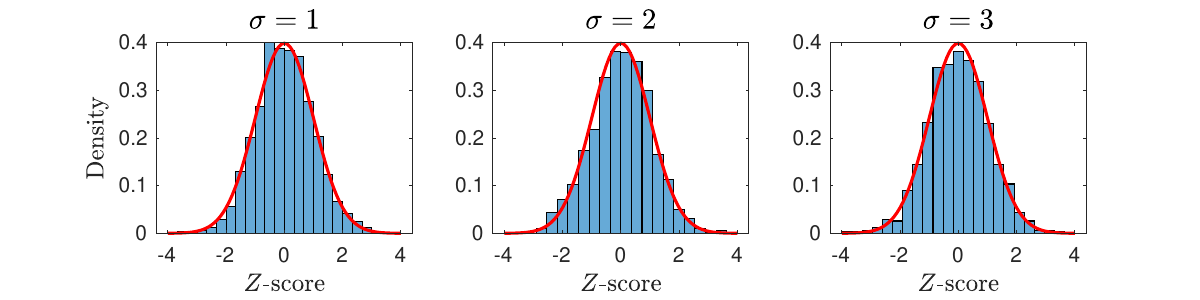}
\caption{Histogram of normal approximation over 1000 independent trails for different noise levels.}
    \label{fig:clt-4}
\end{figure}
We further evaluate the robustness of our method by varying additional parameters under two distinct regimes. Figure \ref{fig:clt-3} examines stability over different ranks: for fixed \(d = 20\) and \(\sigma = 1\), we vary the rank \(r\) over \(\{1, 2, 3\}\). Figure \ref{fig:clt-4} explores the effect of noise magnitude: for fixed \(d = 20\) and \(r = 1\), the noise level \(\sigma\) is varied over \(\{1, 2, 3\}\). In all cases, the histogram closely matches the \(\mathcal{N}(0,1)\) density.

%% file: body/RealData.tex
\begin{maroontext}
We evaluate our framework using a dataset from Alibaba’s Taobao platform, China’s largest e-commerce marketplace (available at \url{https://tianchi.aliyun.com/dataset/649}). The dataset spans from November 25 to December 3, 2017, and records user-item interactions (including user ID, item ID, category ID, behavior type, and timestamp). The behavior types include clicks (page views), purchases, add-to-cart actions, and item-favoring. To focus on peak shopping activity, we restrict our analysis to interactions occurring between 06:00 and 24:00 each day. Moreover, each day is segmented into three six-hour intervals (06:00--12:00, 12:00--18:00, and 18:00--24:00) to capture temporal variations in consumer behavior. The data is divided into two phases: an initialization phase (November 25--29) and a parameter estimation phase (November 30--December 3).

Given the dataset’s scale (\(10^4\) item categories and \(10^7\) users) and its inherent sparsity, we narrow our focus to the top 50 categories by sales volume and the top 10\% of active users in order to enhance the signal-to-noise ratio. In addition, extreme outliers in daily purchase counts (i.e., those exceeding the 0.999 quantile) are truncated to mitigate undue influence. After these preprocessing steps, the model initialization and training phases comprise 84,111 and 86,419 users, respectively.  

\begin{figure}[!htbp]
   \centering
   \includegraphics[width=0.9\linewidth]{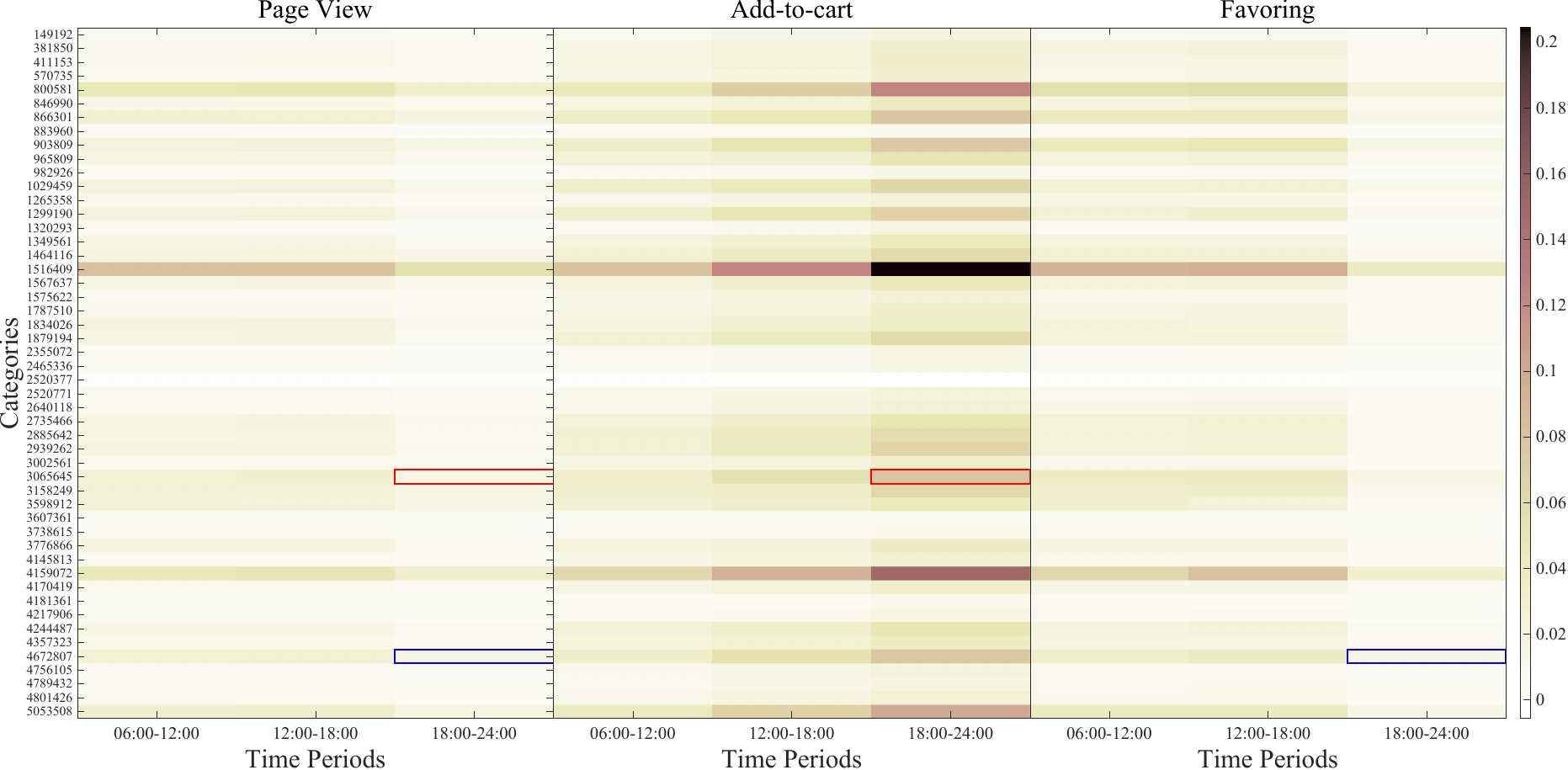}
   \caption{Estimated low-rank tensor coefficients. Each matrix corresponds to a different user behavior mode over the three time segments, with columns representing the top 50 best-selling categories. Blue and red boxes indicate entries used in the hypothesis tests in Equations \eqref{equ:test1} and \eqref{equ:test2}, respectively.}
   \label{fig:Real data result}
\end{figure}

 User interactions are then structured as a third-order count-valued tensor \(\mathcal{T} \in \mathbb{R}^{I \times J \times K}\), where each entry \((i, j, k)\) aggregates the count of behavior type \(j\) (e.g., add-to-cart) for category \(i\) during time segment \(k\). The response variable---daily purchases---is modeled as a function of these tensor covariates.
    Our objective is to analyze and test the effects of different customer behaviors on purchases. In doing so, advertisers can tailor their promotional strategies for various items based on the insights derived from customer behavior data.
In our analysis, each user–day observation is treated as independent and identically distributed, with the assumption of homogeneity within user segments. We initialize the tensor decomposition using the ISLET method \citep{zhang2020islet}, with rank \((1, 2, 2)\) determined via cross-scheme \citep{zhang2019cross}. All covariates are standardized (mean 0, variance 1) to ensure scale comparability. Algorithm \ref{alg:Low rank tensor SGD} is performed with an initial step size of \(\eta_0 = 1 \times 10^{-5}\) and a decay rate of \(\alpha = 0.999\).

Figure \ref{fig:Real data result} displays the estimated low-rank tensor coefficients for the top 50 categories. For ease of interpretation, we decompose the tensor along its behavior dimension into three matrices---each corresponding to a distinct user behavior. Within these matrices, columns represent the top 50 best-selling categories, and rows correspond to the three daily time segments. Each matrix entry indicates the estimated coefficient from our low-rank tensor trace model. Notably, the coefficients associated with the add-to-cart behavior exhibit a larger positive effect on subsequent purchases relative to those for page views and item-favoring. This finding suggests that strategies aimed at encouraging customers to add items to their shopping carts may significantly boost purchase likelihood.

\begin{figure}[!ht]
   \centering
   \includegraphics[width=0.75\linewidth]{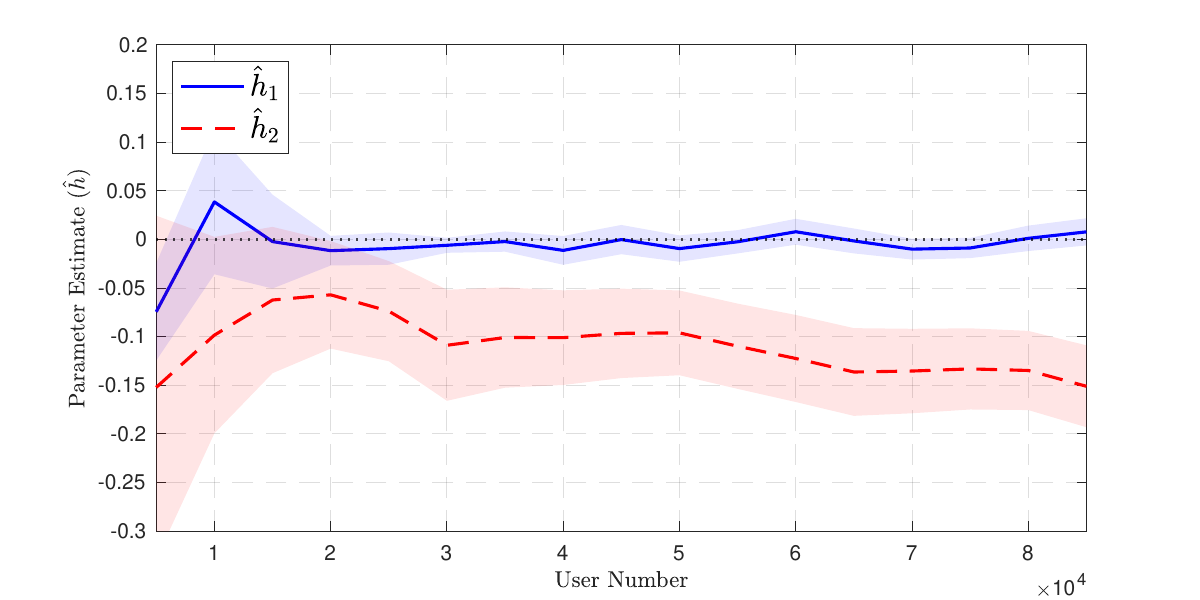}
      \caption{Estimated hypothesis test statistics \(\hat{h}_1\) and \(\hat{h}_2\) for Equations~\eqref{equ:test1} and~\eqref{equ:test2}, respectively, across increasing user sample sizes. Solid blue and dashed red lines represent the estimated values \(\hat{h}_1\) and \(\hat{h}_2\), with shaded areas indicating the corresponding 95\% confidence intervals.}
   \label{fig:real data test}
\end{figure}

To illustrate the practical utility of our approach, we consider two advertising decision-making scenarios. In the first scenario, a seller has secured an advertising slot during the 18:00--24:00 period and must choose between two promotional strategies for category ``4672807." The seller can either enhance exposure by increasing page views or encourage customer engagement by promoting item collections. We formalize this decision as a hypothesis test
\begin{align}\label{equ:test1}
\begin{aligned}
		H_0: \mathcal{T}^\star(\text{``4672807''}, \text{Page View}, \text{18:00--24:00}) = \mathcal{T}^\star(\text{``4672807''}, \text{Collects}, \text{18:00--24:00}),\\
H_1: \mathcal{T}^\star(\text{``4672807''}, \text{Page View}, \text{18:00--24:00}) \neq \mathcal{T}^\star(\text{``4672807''}, \text{Collects}, \text{18:00--24:00}),
\end{aligned}
\end{align}
to determine whether the two promotional strategies exhibit a statistically significant difference.
In Figure~\ref{fig:Real data result}, the tensor entries corresponding to these strategies are highlighted in blue.

 In the second scenario, the seller aims to promote category ``3065645" but is uncertain whether to feature it prominently on the homepage or to adopt a pricing strategy in which the price is omitted from the item page and revealed only in the shopping cart.  This strategic choice is captured through the following hypothesis test:
\begin{align}\label{equ:test2}
	\begin{aligned}
		H_0: \mathcal{T}^\star(\text{``3065645''}, \text{Page View}, \text{18:00--24:00}) = \mathcal{T}^\star(\text{``3065645''}, \text{Add-to-Cart}, \text{18:00--24:00}),\\
H_1: \mathcal{T}^\star(\text{``3065645''}, \text{Page View}, \text{18:00--24:00}) \neq \mathcal{T}^\star(\text{``3065645''}, \text{Add-to-Cart}, \text{18:00--24:00}).
	\end{aligned}
\end{align}
The tensor entries relevant to this comparison are indicated by the red boxes in Figure \ref{fig:Real data result}.

Our proposed method provides a confidence interval for testing these hypotheses. With a confidence level set to 0.95, we expect that if the experiment were repeated 100 times, the estimated parameter would fall within this interval at least 95 times.
Figure \ref{fig:real data test} presents the hypothesis testing results for both scenarios. For the hypothesis in Equation \eqref{equ:test1}, the confidence interval for the parameter \(\hat{h}_1\) includes zero at the end of the experiment. Consequently, we cannot reject the null hypothesis, implying that there is no statistically significant difference between increasing page view exposure and encouraging customers to add the item to their collections for category ``4672807." In contrast, for the hypothesis in Equation \eqref{equ:test2}, the confidence interval for \(\hat{h}_2\) does not include zero at the end of the experiment. This indicates a statistically significant difference between the two advertising strategies, with the data suggesting that encouraging customers to add items to their carts is a more effective strategy for category ``3065645" than relying solely on increased page views. These findings provide actionable insights for managers by clarifying which advertising strategies yield significant differences in consumer behavior under specific conditions.
\end{maroontext}

%% file: body/Appendix_literature.tex
In this section, we discuss three additional strands of related work: low‑rank models in business settings, tensor‑based deep learning methods, and human–computer interface applications. We highlight the key distinctions between these approaches and our framework.

\textbf{Low-Rank Models in Business Settings:} Low-rank models are increasingly recognized for their ability to capture complex, multi-dimensional interactions in business applications. For instance, \cite{farias2019learning} formalize the task of learning customer preferences as the recovery of a three-dimensional tensor from noisy observations, proposing an efficient algorithm to tackle this challenge. Similarly, \cite{kallus2020dynamic} explore dynamic assortment personalization by leveraging low-rank structures to optimize product offerings over time. In textual analytics, \cite{xu2021groupsparse} employ transfer learning of word embeddings within a low-rank matrix trace regression framework to boost analytical performance. Other studies, such as \cite{bayati2022speed}, address two-sided product problems by modeling rewards using matrices, while \cite{farias2024fixing} and \cite{tang2024match} further demonstrate the effectiveness of low-rank models in detecting anomalies and accelerating reward learning in online settings. In contrast to these contributions, our work focuses on statistical inference following the online low-rank tensor learning.

\textbf{Tensor-based Deep Learning Methods:} 
Deep learning has become a dominant approach in areas such as recommender systems, yet it often requires highly complex models and large datasets. To address these challenges, many researchers have incorporated tensor methods to reduce the number of parameters and improve computational efficiency \citep{frolov2017tensor,bi2018multilayer,song2019tensor,zhang2021dynamic,entezari2021tensorbased}. For example, low-rank approximations are used to compress convolutional layers in CNNs by reducing the dimensionality of activation tensors \citep{denton2014exploiting,lebedev2015speedingup,tai2016convolutional,kim2015compression,hayashi2019exploring,kossaifi2020factorized} and to decrease parameters in fully connected layers \citep{novikov2015tensorizing,ye2020blockterm,kossaifi2020tensor}. These methods eliminate the need for the flattening operations typical of traditional architectures, thereby streamlining model training and inference.

\textbf{Human–Computer Interface Example:}
Beyond online advertising, our framework is also well suited to brain–computer interface applications.  The brain operates as a complex dynamical system, with spatially distributed neural regions interacting to generate multivariate temporal signals that convey both functional and structural information \citep{bassett2011understanding}. Neuroimaging techniques--such as electroencephalography (EEG), magnetoencephalography (MEG), functional magnetic resonance imaging (fMRI), and near-infrared spectroscopy (NIRS)--naturally produce multidimensional datasets best represented as tensors \citep{cichocki2008noninvasive}. Given the continuously evolving nature of brain activity, online tensor analysis has emerged as a powerful tool for modeling dynamic brain processes. Recent advances in tensor decomposition have particularly enhanced our understanding of dynamic functional connectivity networks (FCNs), which capture transient synchronization patterns among neural populations \citep{prabhakaran2006eventrelated,goebel2006analysis,boveroux2010breakdown,chang2010timefrequency}. FCNs often exhibit low-rank structural properties—a feature leveraged by tensor-based models to reduce dimensionality while preserving network topology \citep{ozdemir2017recursive,yeung2004neural,mahyari2017tensor,al-sharoa2019tensor,xu2023temporal,gabrielson2024mode}. In this context, the neuroimaging data collected over time serves as the covariate, while various signal treatments or behavioral outcomes are modeled as responses. The large scale of neuroimaging data and the necessity for real-time analysis make online tensor methods particularly relevant in clinical settings, where timely treatment decisions are critical.

%% file: body/Appendix_simulation.tex
\begin{enumerate}
\item \textbf{Hyperparameters of the our online tensor estimation algorithm:}
We analyze the impact of our online tensor estimation hyperparameters, namely the initial learning rate \(\eta_0\), decay rate \(\alpha\), and constant period \(t^\star\) in Figure \ref{fig:learning-2}.

\begin{figure}[!ht]
    \centering
    \includegraphics[width=1\linewidth]{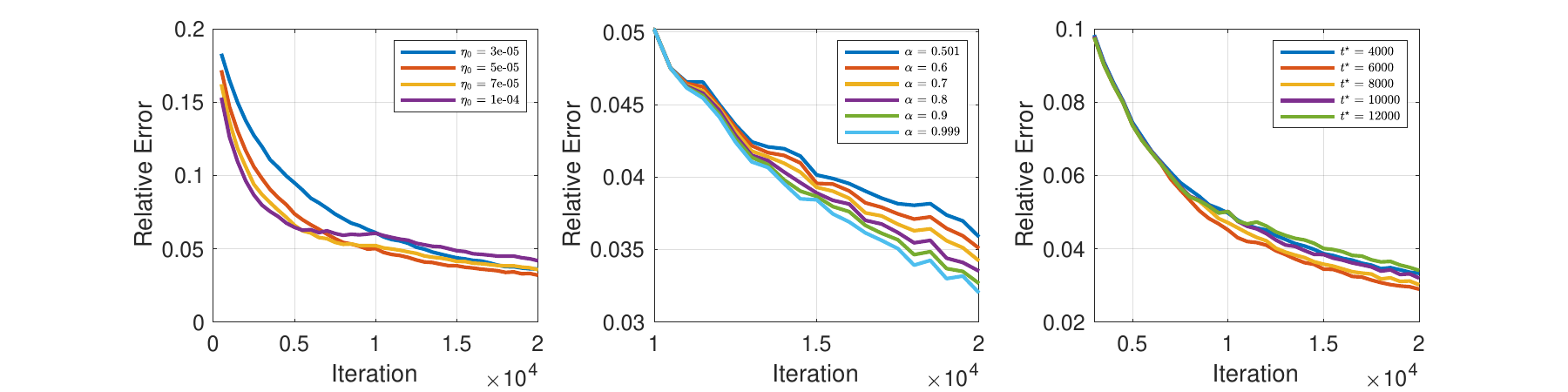}
    \caption{Error analysis for our online tensor estimation across different hyperparameters.}
    \label{fig:learning-2}
\end{figure}
The left panel presents experiments that vary the initial learning rate (\(\eta_0\)). We observe that smaller values (e.g., \(3 \times 10^{-5}\)) yield slower initial error reduction but lead to stable convergence. In contrast, larger values (e.g., \(7 \times 10^{-5}\) or \(1 \times 10^{-4}\)) accelerate early convergence, albeit with a risk of oscillatory behavior. In particular, for relatively large step sizes such as \(\eta_0 = 1 \times 10^{-4}\), the error exhibits initial oscillations; however, due to our two-stage step-size strategy, the error decreases again in the second phase.

	 The middle panel illustrates the impact of the decay rate (\(\alpha\)) on convergence performance. Our findings show that a decay rate close to 1 (e.g., \(\alpha = 0.999\)) produces both a faster error decay and a lower final error (approximately 0.033), compared to lower values such as \(\alpha = 0.6\), which yield a final error of around 0.035.  These experimental results are in line with our theoretical predictions, which state that as \(\alpha\) approaches 1, the convergence error decreases.

The right panel compares different constant periods ($t^\star$), which dictates the duration of the constant learning rate phase before decay commences. The results indicate that all values of \(t^\star\) yield very similar convergence errors. This demonstrates that our method is robust with respect to the choice of this parameter.

\paragraph{Practical Recommendations}  
Based on our experiments, we observe that the parameter \(t^\star\) is quite robust---its specific choice generally does not have a significant impact on convergence. For the decay parameter \(\alpha\), our results and theoretical analysis both suggest selecting a value as close to 1 as possible. Regarding the initial step size \(\eta_0\), we recommend starting with a small value and gradually increasing it until an optimal performance is reached. Notably, our two-stage step size strategy is designed to mitigate issues that can arise with a slightly large initial step size. While a larger \(\eta_0\) (\(\eta_0 = 1 \times 10^{-4}\)) might cause oscillations during the constant step size period, the subsequent decaying step size effectively dampens these oscillations, as demonstrated in Figure \ref{fig:learning-2}.

\item\textbf{Beyond Gaussian Designs}

To assess the robustness of our method, we extend our simulations beyond the standard sub-Gaussian setting by comparing different distributions for both noise $\xi$ and the covariate $\cX$ in Figure \ref{fig:learning-3-1}. 

  \begin{figure}[!ht]
    \centering
    \includegraphics[width=0.75\linewidth]{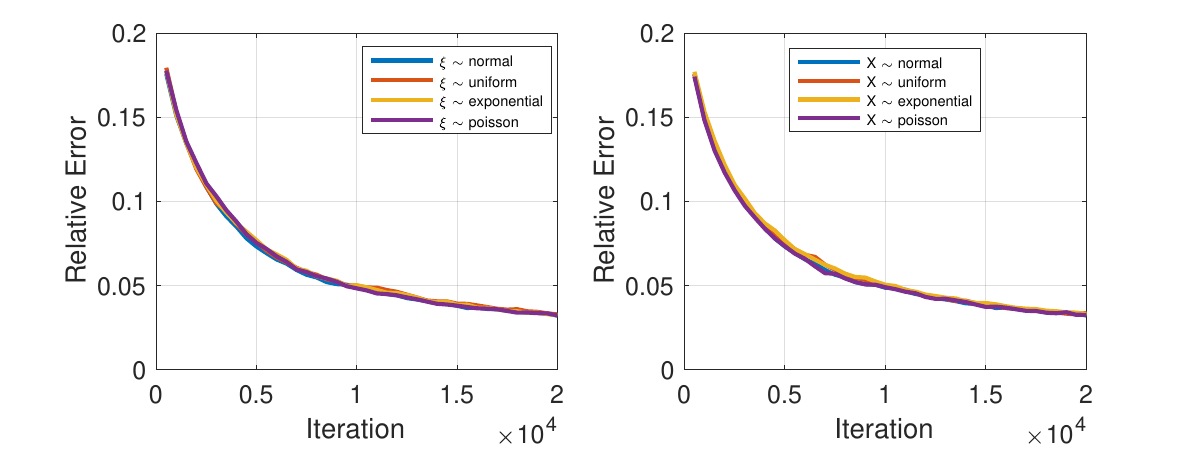}
    \caption{Error analysis for our online tensor estimation across different distributions for noise $\xi$ and the covariate $\cX$.}
    \label{fig:learning-3-1}
\end{figure}

We compare noise drawn from normal, uniform (scaled to \([-\sqrt{3}, \sqrt{3}]\)), exponential (shifted to have zero mean and normalized), and Poisson (centered and variance-scaled) distributions. All noise distributions in the left subfigure exhibit monotonic error decay with iterations, converging to similar error levels.  The right subfigure shows nearly identical performance across different design distributions, confirming that the sub-Gaussian condition is sufficient for optimal performance. These observations indicate that our framework is robust and generalizes well beyond its theoretical assumptions. Figure \ref{fig:clt-5} investigates the inference results of non-subgaussian noise: for fixed \(d = 20\), \(r = 1\), and \(\sigma = 1\), we compare results under different noise distributions, including uniform, exponential, and Poisson. Finally, Figure \ref{fig:clt-6} considers non-subgaussian design: for fixed \(d = 20\), \(r = 1\), and \(\sigma = 1\), we assess performance when the design tensors are drawn from uniform, exponential, and Poisson distributions.  In all cases, the histogram closely matches the \(\mathcal{N}(0,1)\) density. 
\begin{figure}[!ht]
    \centering
\includegraphics[width=0.75\linewidth]{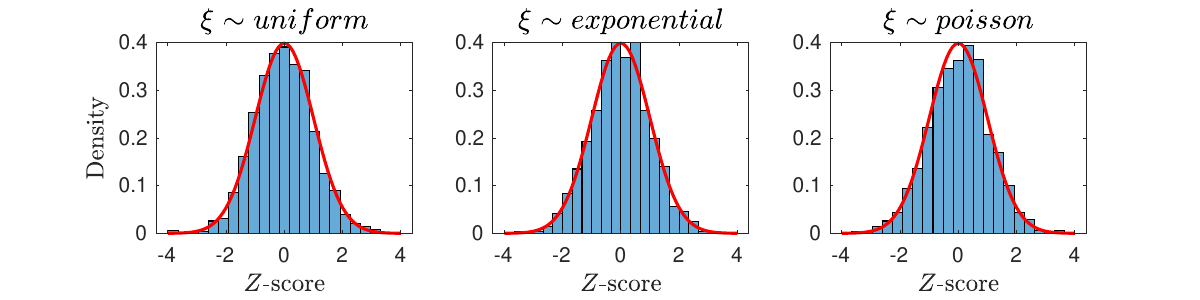}
\caption{Histogram of normal approximation over 1000 independent trails for different noise distributions.}
    \label{fig:clt-5}
\end{figure}

\begin{figure}[!ht]
    \centering
\includegraphics[width=0.75\linewidth]{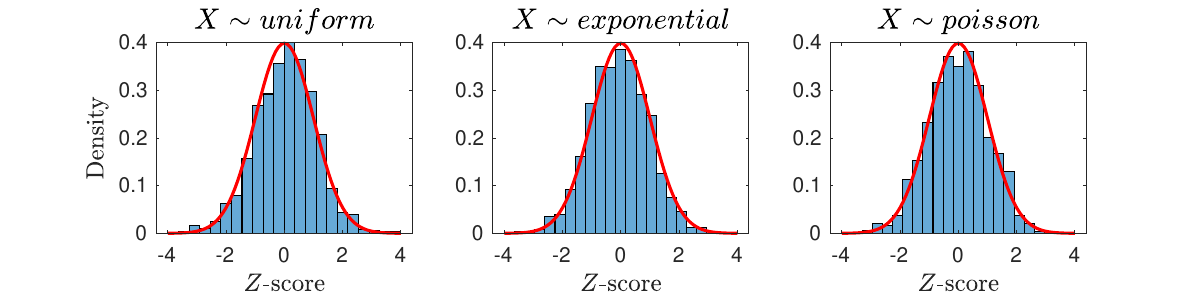}
\caption{Histogram of normal approximation over 1000 independent trails for different design distributions.}
    \label{fig:clt-6}
\end{figure}

\item  \textbf{Batch Settings}

We extend our model to allow data to arrive sequentially in batches. Specifically, we introduce a batch size parameter \(b\) and adjust the step size schedule to $\eta_t = \eta_0 \left(\max\{b\cdot t, t^{\star}\}\right)^{-\alpha}$, with $t^{\star} = \bigl(C_{\max}\, \df\bigr)^{1/\alpha}$. 
        We detail the resulting Online Batch Low-Rank Tensor Inference algorithm in Algorithm \ref{alg:Online_Batch_Inference} with supporting subroutines shown in Algorithm \ref{alg:Low rank tensor batch SGD}-\ref{alg:Single-step batch Tensor Linear Form Estimator Update} and provide simulation results for the batch algorithm. We investigate the effect of batch size in Figure \ref{fig:learning-5}.

\begin{figure}[!ht]
    \centering
    \includegraphics[width=0.5\linewidth]{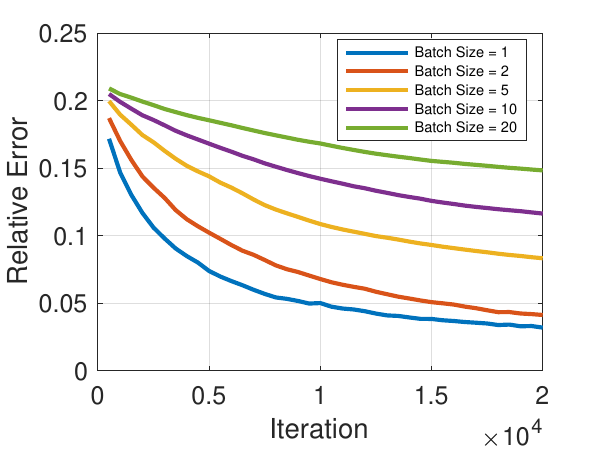}
    \caption{Error analysis of SGD across different batch sizes.}
    \label{fig:learning-5}
\end{figure}

A batch size of 1 (i.e., no mini-batch averaging) performs best, yielding the smallest relative error (approximately 0.05) with stable convergence. At this learning rate, the small gradient updates are less susceptible to stochastic noise, and larger batch sizes may unnecessarily smooth the updates, thereby slowing convergence.

\begin{algorithm}[!ht]
\small
\DontPrintSemicolon
\caption{Single-step Low-Rank Tensor Batch SGD for Step $t$}
\label{alg:Low rank tensor batch SGD}

\KwIn{$\mathcal{G}^{(t-1)}$, $\mathbf{U}_{k}^{(t-1)}$ for $k\in[3]$, new batch data pair $\left\{ \rbr{y_i, \mathcal{X}_i} \right\}_{i\in[b]}$, and the step size $\eta_{t}$.}

{\color{blue}{\tcc{Update Factor Matrices}}}

\For{$k \in [3]$}{
    $\begin{aligned}
        \mathbf{U}_k^{(t)} = &\mathbf{U}_k^{(t-1)} - \frac{\eta_t}{b}\sum_{i\in[b]} \Big(\left\langle\mathcal{X}_{i}, \mathcal{T}^{(t-1)}\right\rangle - y_{i}\Big)\mathcal{M}_k(\mathcal{X}_{i}) \left(\mathbf{U}_{k+2}^{(t-1)} \otimes \mathbf{U}_{k+ 1}^{(t-1)}\right) \mathcal{M}_k^{\top}(\mathcal{G}^{(t-1)}) \\
        & - \frac{\eta_t }{2}\mathbf{U}_k^{(t-1)} \left(\mathbf{U}_k^{(t-1) \top} \mathbf{U}_k^{(t-1)} - \mathbf{I}_{r_k}\right).
    \end{aligned}$ 
    \;
}

{\color{blue}{\tcc{Update Core Tensor}}}

$\mathcal{G}^{(t)} = \mathcal{G}^{(t-1)} - \frac{\eta_t}{b}\sum_{i\in[b]} \Big(\left\langle\mathcal{X}_{i}, \mathcal{T}^{(t-1)}\right\rangle - y_{i}\Big) \mathcal{X}_{i} \bigtimes_{k\in[3]} \mathbf{U}_k^{(t-1)\top}$.\;

\KwOut{
Updated core tensor $\mathcal{G}^{(t)}$, and updated factor matrices $\mathbf{U}_k^{(t)}$ for $k\in[3]$.}
\end{algorithm}
        
\begin{algorithm}[!ht]
 \small
\caption{Single-step Tensor Linear Form Estimator Batch Update for Step $t$}\label{alg:Single-step batch Tensor Linear Form Estimator Update}
\KwIn{
    Linear Form $\mathcal{H}$, online low-rank SGD estimator $\mathcal{T}^{(t-1)}$ and its 
    projected matrices $\widehat{\mathbf{U}}_{k}^{(t-1)}$ for $k \in [3]$, 
    new batch data pair $\left\{ \rbr{y_i, \mathcal{X}_i} \right\}_{i\in[b]}$.
}
{\color{blue}{\tcc{Update Average SGD Estimator}}}
$\widehat{\mathcal{T}}^{(t)}=\frac{t-1}{t} \widehat{\mathcal{T}}^{(t-1)}+\frac{1}{bt}\sum_{i\in[b]}\left( {\mathcal{T}}^{(t-1)}- \left(\left\langle {\mathcal{T}}^{(t-1)}, \mathcal{X}_{i} \right\rangle - y_{i}\right) \mathcal{X}_{i}\right)$. 

{\color{blue}{\tcc{Updating Factor Matrices and Corresponding Singular Values}}}
$ \widehat{\mathbf{U}}_k^{(t)}, \widehat{\mathbf{\Lambda}}_k^{(t)} = \text{SVD}_{r_k}\left( \mathcal{M}_k\left(\widehat{\mathcal{T}}^{(t)} \times_{k+1} \widehat{\mathbf{U}}_{k+1}^{(t-1) \top} \times_{k+2} \widehat{\mathbf{U}}_{k+2}^{(t-1) \top}\right)\right)\text{, for }k\in[3]$.

{\color{blue}{\tcc{Update Tensor Linear Form Estimator}}}
$ \hat{h}^{(t)}
 =\left\langle\widehat{\mathcal{T}}^{(t)}\times_1\mathcal{P}_{\widehat{\mathbf{U}}_1^{(t)}}  \times_2 \mathcal{P}_{\widehat{\mathbf{U}}_2^{(t)}}  \times_3 \mathcal{P}_{\widehat{\mathbf{U}}_3^{(t)}} , \mathcal{H}\right\rangle.$
 	
\KwOut{
    Linear form estimate $\hat{h}^{(t)}$, factor matrices $\widehat{\mathbf{U}}_{k}^{(t)}$ and singular values $\widehat{\mathbf{\Lambda}}_k^{(t)}$ for $k \in [3]$.
}
\end{algorithm}

\begin{algorithm}[!ht]
\small
\DontPrintSemicolon  
\SetAlCapHSkip{0em}  
\SetInd{0.5em}{0.5em}  
\DecMargin{1em}  
\caption{Online Batch Low-Rank Tensor Inference}\label{alg:Online_Batch_Inference}
\KwIn{ 
Initial estimate $\mathcal{T}^{(0)}$, $\widehat{\mathbf{U}}_k^{(0)}=\mathbf{U}_k^{(0)}$, for $k\in[3]$, $\hat{\sigma}^2_{0} = 0$,  step size $\{\eta_t\}$, rank $(r_1, r_2, r_3)$, significance level $\alpha$.}
\For{$t = 1, 2, \ldots$}{
 Receive new batch data pair $\left\{ \rbr{y_i, \mathcal{X}_i} \right\}_{i\in[b]}$.\;
 
{\color{blue}{\tcc{ Estimation Task}}}
$\mathbf{U}_1^{(t)}, \mathbf{U}_2^{(t)}, \mathbf{U}_3^{(t)}, \mathcal{G}^{(t)} \leftarrow$ Algorithm \ref{alg:Low rank tensor batch SGD} $\left(\mathbf{U}_1^{(t-1)}, \mathbf{U}_2^{(t-1)}, \mathbf{U}_3^{(t-1)}, \mathcal{G}^{(t-1)}, \left\{ \rbr{y_i, \mathcal{X}_i} \right\}_{i\in[b]}, \eta_t\right).$

{\color{blue}{\tcc{Inference Task}}}
$\hat{h}^{(t)}, \widehat{\mathbf{U}}_k^{(t)}, \widehat{\mathbf{\Lambda}}_k^{(t)},$
$k\in[3]$ $\leftarrow$ Algorithm \ref{alg:Single-step batch Tensor Linear Form Estimator Update} $\left(\mathcal{T}^{(t-1)}, \widehat{\mathbf{U}}_1^{(t-1)}, \widehat{\mathbf{U}}_2^{(t-1)}, \widehat{\mathbf{U}}_3^{(t-1)}, \left\{ \rbr{y_i, \mathcal{X}_i} \right\}_{i\in[b]}, \mathcal{H}\right)$.

 	    {\color{blue}{\tcc{Update Plug in Estimate}}}
 	$\hat{\sigma}_t^2
=\frac{t-1}{t} \hat{\sigma}_{t-1}^2+\frac{1}{bt}\sum_{i\in[b]} \left(y_i-\left\langle\mathcal{T}^{(t)}, \mathcal{X}_i\right\rangle\right)^2$. \;
$\widehat{S}_{\mathcal{H},t}^{2}  = \|\mathcal{H} \bigtimes_{k\in[3]} {\widehat{\mathbf{U}}_k^{(t)\top }}  \|_{\mathrm{F}}^2 +  \sum_{k=1}^3 \| \mathcal{P}_{\widehat{\mathbf{U}}_k^{(t)}}^{\perp}\,\mathbf{H}_k\,\mathcal{P}_{\bigl(\widehat{\mathbf{U}}_{k+2}^{(t)}\otimes \widehat{\mathbf{U}}_{k+1}^{(t)}\bigr)\, \widehat{\mathbf{V}}_k^{(t)}} \|_{\text{F}}^{2}$.\;

    {\color{blue}{\tcc{Calculate $(1-\alpha)$-level Confidence Intervals}}}
$\widehat{\mathrm{CI}}^\alpha_{h,t}= \left(\hat{h}^{(t)}-z_{\alpha / 2} \hat{\sigma}_t \widehat{S}_{\mathcal{H},t} / \sqrt{t} ,
\quad \hat{h}^{(t)}+z_{\alpha / 2} \hat{\sigma}_t \widehat{S}_{\mathcal{H},t} / \sqrt{t}\right)$.\;

 	{\color{blue}{\tcc{Update SGD Tensor Estimator}}}
$\mathcal{T}^{(t)} = \mathcal{G}^{(t)} \times_1 \mathbf{U}_1^{(t)} \times_2 \mathbf{U}_2^{(t)} \times_3 \mathbf{U}_3^{(t)}$.\;
}
\KwOut{$\left\{\hat{h}^{(t)}\right\}$, $\left\{ \widehat{\mathrm{CI}}^\alpha_{h,t}\right\}$. }
 	\end{algorithm}

\end{enumerate}

%% file: body/Appendix_highorder.tex
We would like to clarify that our method is not restricted to third-order tensors but is indeed applicable to tensors of general order \(m\). The variable \(m\) denotes the number of modes. Importantly, the theoretical results presented in Theorems 1 and 2 extend naturally to higher-order tensors without alteration. We provide Algorithm \ref{alg:Low rank High-order tensor SGD} for the learning component and  Algorithm \ref{alg:Single-step High-order Tensor Linear Form Estimator Update} for inference. The full procedure is summarized in Algorithm \ref{alg:Online_high_order_Inference}.

\begin{algorithm}[!ht]
\small
\DontPrintSemicolon
\caption{Single-step Low-Rank High-order Tensor SGD for Step $t$}
\label{alg:Low rank High-order tensor SGD}

\KwIn{$\mathcal{G}^{(t-1)}$, $\mathbf{U}_{k}^{(t-1)}$ for $k\in[m]$, new data pair $\left(y_t, \mathcal{X}_t\right)$, and the step size $\eta_{t}$.}

{\color{blue}{\tcc{Update Factor Matrices}}}

\For{$k \in [m]$}{
    $\begin{aligned}
        \mathbf{U}_k^{(t)} = &\mathbf{U}_k^{(t-1)} - \eta_{t} \Big(\left\langle\mathcal{X}_{t}, \mathcal{T}^{(t-1)}\right\rangle - y_{t}\Big)\mathcal{M}_k(\mathcal{X}_{t}) \left(\otimes_{j\neq k}  \mathbf{U}_{j}^{(t-1)}\right) \mathcal{M}_k^{\top}(\mathcal{G}^{(t-1)}) \\
        & - \frac{\eta_t }{2}\mathbf{U}_k^{(t-1)} \left(\mathbf{U}_k^{(t-1) \top} \mathbf{U}_k^{(t-1)} - \mathbf{I}_{r_k}\right).
    \end{aligned}$ 
    \;
}

{\color{blue}{\tcc{Update Core Tensor}}}

$\mathcal{G}^{(t)} = \mathcal{G}^{(t-1)} - \eta_{t} \Big(\left\langle\mathcal{X}_{t}, \mathcal{T}^{(t-1)}\right\rangle - y_{t}\Big) \mathcal{X}_{t} \times_{k\in[m]} \mathbf{U}_k^{(t-1)\top}$.\;

\KwOut{
Updated core tensor $\mathcal{G}^{(t)}$, and updated factor matrices $\mathbf{U}_k^{(t)}$ for $k\in[m]$.}
\end{algorithm}

 \begin{algorithm}[!ht]
 \small
\caption{Single-step  High-order Tensor Linear Form Estimator Update for Step $t$}
\label{alg:Single-step High-order Tensor Linear Form Estimator Update}
\KwIn{
    Linear Form $\mathcal{H}$, online low-rank SGD estimator $\mathcal{T}^{(t-1)}$ and its 
    projected matrices $\widehat{\mathbf{U}}_{k}^{(t-1)}$  for $k \in [m]$, 
    new data $\zeta_{t} = (\mathcal{X}_{t}, y_{t})$.
}

{\color{blue}{\tcc{Update Average SGD Estimator}}}

$\widehat{\mathcal{T}}^{(t)}=\frac{t-1}{t} \widehat{\mathcal{T}}^{(t-1)}+\frac{1}{t}\left( {\mathcal{T}}^{(t-1)}- \left(\left\langle {\mathcal{T}}^{(t-1)}, \mathcal{X}_{t} \right\rangle-y_{t}\right) \mathcal{X}_{t}\right)$.

{\color{blue}{\tcc{Updating Factor Matrices and Corresponding Singular Values}}}

$ \widehat{\mathbf{U}}_k^{(t)}, \widehat{\mathbf{\Lambda}}_k^{(t)} = \text{SVD}_{r_k}\left( \mathcal{M}_k\left(\widehat{\mathcal{T}}^{(t)} \times_{\substack{j\in[m] \\ j\neq k}}  \widehat{\mathbf{U}}_{j}^{(t-1) \top}\right)\right)\text{, for }k\in[m]$.

{\color{blue}{\tcc{Update Tensor Linear Form Estimator}}}

$ \hat{h}^{(t)}
 =\left\langle\widehat{\mathcal{T}}^{(t)}\times_{k\in[m]}\mathcal{P}_{\widehat{\mathbf{U}}_k^{(t)}} , \mathcal{H}\right\rangle.$
 	
\KwOut{
    Linear form estimate $\hat{h}^{(t)}$, factor matrices $\widehat{\mathbf{U}}_{k}^{(t)}$ and singular values $\widehat{\mathbf{\Lambda}}_k^{(t)}$ for $k \in [m]$.
}
\end{algorithm}

\begin{algorithm}[!ht]
\small
\DontPrintSemicolon    
\SetAlCapHSkip{0em}  
\SetInd{0.5em}{0.5em}  
\DecMargin{1em}  
\caption{Online High-order Tensor Inference}\label{alg:Online_high_order_Inference}
\KwIn{ 
Initial estimate $\mathcal{T}^{(0)}$, $\widehat{\mathbf{U}}_k^{(0)}=\mathbf{U}_k^{(0)}$, for $k\in[m]$, $\hat{\sigma}^2_{t_0} = 0$, step size $\{\eta_t\}$, rank $(r_1, \cdots, r_m)$, significance level $\alpha$.}

\For{$t = 1, 2, \ldots$}{
 Receive new observation $(\mathcal{X}_t,y_t)$.\;

{\color{blue}{\tcc{ Estimation Task}}}

$\mathbf{U}_k^{(t)}, \mathcal{G}^{(t)},k\in[m] \leftarrow$ Algorithm \ref{alg:Low rank High-order tensor SGD} $\left(\mathbf{U}_1^{(t-1)}, \cdots, \mathbf{U}_m^{(t-1)}, \mathcal{G}^{(t-1)}, \mathcal{X}_t, y_t, \eta_t\right).$

{\color{blue}{\tcc{Inference Task}}}

$\hat{h}^{(t)}, \widehat{\mathbf{U}}_k^{(t)}, \widehat{\mathbf{\Lambda}}_k^{(t)},$
$k\in[m]$ $\leftarrow$ Algorithm \ref{alg:Single-step High-order Tensor Linear Form Estimator Update} $\left(\mathcal{T}^{(t-1)}, \widehat{\mathbf{U}}_1^{(t-1)},\cdots, \widehat{\mathbf{U}}_m^{(t-1)}, \mathcal{X}_t, y_t, \mathcal{H}\right)$.

 	    {\color{blue}{\tcc{Update Plug in Estimate}}}
      
 	$\hat{\sigma}_t^2
=\frac{t-1}{t} \hat{\sigma}_{t-1}^2+\frac{1}{t} \left(y_t-\left\langle\mathcal{T}^{(t)}, \mathcal{X}_t\right\rangle\right)^2$. \;
$\widehat{S}_{\mathcal{H},t}^{2}  = \|\mathcal{H} \times_{k\in[m]} {\widehat{\mathbf{U}}_k^{(t)\top }}  \|_{\mathrm{F}}^2 +  \sum_{k=1}^3 \| \mathcal{P}_{\widehat{\mathbf{U}}_k^{(t)}}^{\perp}\,\mathbf{H}_k\,\mathcal{P}_{\bigl(\otimes_{j\neq k} \widehat{\mathbf{U}}_{j}^{(t)}\bigr)\, \widehat{\mathbf{V}}_k^{(t)}} \|_{\text{F}}^{2}$.\;

    {\color{blue}{\tcc{Calculate $(1-\alpha)$-level Confidence Intervals}}}
    
$\widehat{\mathrm{CI}}^\alpha_{h,t}= \left(\hat{h}^{(t)}-z_{\alpha / 2} \hat{\sigma}_t \widehat{S}_{\mathcal{H},t} / \sqrt{t} ,
\quad \hat{h}^{(t)}+z_{\alpha / 2} \hat{\sigma}_t \widehat{S}_{\mathcal{H},t} / \sqrt{t}\right)$.\;

 	{\color{blue}{\tcc{Update SGD Tensor Estimator}}}
  
$\mathcal{T}^{(t)} = \mathcal{G}^{(t)} \times_{k\in[m]} \mathbf{U}_k^{(t)}$.\;
}
\KwOut{$\left\{\hat{h}^{(t)}\right\}$, $\left\{ \widehat{\mathrm{CI}}^\alpha_{h,t}\right\}$. }
\end{algorithm}

%% file: body/Notation.tex
We introduce notational conventions and provide some preliminaries on tensor algebra.

\subsubsection{Basic Notations}
Let $|\cdot|$ denote the cardinality of a set and $[k] = \{1,2,\ldots,k\}$ for any integer $k \geq 1$.
Scalars are represented by lowercase letters such as $a, \lambda$. Vectors use bold lowercase letters like $\mathbf{x}, \mathbf{y}$, matrices by bold uppercase $\mathbf{U}$, and tensors by calligraphic letters, e.g., $\mathcal{T}, \mathcal{M}$.
For any matrix $\mathbf{U}$, the notations $\mathbf{U}_{ij}$, $\mathbf{U}_{i\cdot}$, and $\mathbf{U}_{\cdot j}$ indicate the entry at the $i$th row and $j$-th column, the $i$th row, and the $j$-th column, respectively.  
The transpose of a matrix is given by $\mathbf{U}^{\top}$, while $\|\mathbf{U}\|_\mathrm{F}$ denotes its Frobenius norm.  
The matrix inner product between $\mathbf{U}_1$ and $\mathbf{U}_2$ is defined as $\langle \mathbf{U}_1,\mathbf{U}_2\rangle = \operatorname{tr}\left(\mathbf{U}_1^{\top}\mathbf{U}_2\right)$.  
The symbols \(\|\cdot\|\) are used to represent the Euclidean norm for vectors and the matrix spectral norm for matrices.  
We let $\|\mathbf{U}\|_{2, \infty}$ be the $\ell_{2, \infty}$ norm of a matrix, defined as $\|\mathbf{U}\|_{2, \infty}=\max _i\left\|\mathbf{U}_i.\right\|$.
We use $\mathbf{e}_k$ for the standard basis vector, and the identity is represented as either $\mathbf{I}$ or $\mathbf{I}_k$, where \(k\) specifies the dimension. 
Let \(\mathbb{O}_{p, r}\) be the set of all \(p \times r\) matrices with orthonormal columns, defined as: \(\mathbb{O}_{p, r} = \{\mathbf{U} \in \mathbb{R}^{p \times r}: \mathbf{U}^{\top} \mathbf{U} = \mathbf{I}_r\}\) where \(\mathbf{I}_r\) is the \(r \times r\) identity matrix.
Let $\sigma_r(\cdot)$ be the $r$-th largest singular value of a matrix.  In particular, we use $\sigma_{\min }(\cdot)$, and $\sigma_{\max }(\cdot)$ as the smallest and largest nontrivial singular values of a matrix.  
Let $\mathbf{U}_1, \mathbf{U}_2\in\mathbb{O}_{p, r}$ be two matrices of the same dimension having orthonormal columns. We define their (spectral) \(\sin \Theta\) distance, denoted by \(\left\|\sin \Theta\left(\mathbf{U}_1, \mathbf{U}_2\right)\right\|\), as: \(\left\|\sin \Theta\left(\mathbf{U}_1, \mathbf{U}_2\right)\right\|=\sqrt{1-\sigma_{\min }^2\left(\mathbf{U}_1^{\top} \mathbf{U}_2\right)} = \left\|\mathbf{U}_{1\perp}^{\top} \mathbf{U}_2\right\|\)  and \(\left\|\sin \Theta\left(\mathbf{U}_1, \mathbf{U}_2\right)\right\|_\mathrm{F}=\sqrt{r-\left\|\mathbf{U}_1^{\top} \mathbf{U}_2\right\|_\mathrm{F}^2} = \left\|\mathbf{U}_{1\perp}^{\top} \mathbf{U}_2\right\|_\mathrm{F}\).
For a matrix $\mathbf{U}$ with orthonormal columns, let \(\mathcal{P}_\mathbf{U} = \mathbf{U}\mathbf{U}^\top\) represent the projection onto the subspace spanned by \(\mathbf{U}\). The matrix \(\operatorname{SVD}_r(\mathbf{U}) \in \mathbb{O}_{p, r}\) is defined as the matrix comprising the top \(r\) left singular vectors of \(\mathbf{U}\). 
Given any matrix \(\mathbf{U}=\left[\mathbf{u}_1, \ldots, \mathbf{u}_J\right] \in \mathbb{R}^{I \times J}\) and \(\mathbf{V} \in \mathbb{R}^{K \times L}\), the Kronecker product is represented as: \(\mathbf{U} \otimes \mathbf{V} = \left[\mathbf{u}_1 \otimes \mathbf{V}, \ldots, \mathbf{u}_J \otimes \mathbf{V}\right],\) yielding a \((I K) \times (J L)\) matrix.
Furthermore, the notation \( \stackrel{d}{\longrightarrow} \) is used to denote convergence in distribution, while \( \stackrel{p}{\longrightarrow} \) indicates convergence in probability.

\subsubsection{Tensor Notations}
A tensor is described as a multi-dimensional array. Its order, or the number of its dimensions, is referred to as its mode.
  For tensors $\mathcal{X}$ and $\mathcal{Y}$ in $\mathbb{R}^{p_1 \times \cdots \times p_d}$, the tensor inner product is defined as $\langle\mathcal{X}, \mathcal{Y}\rangle = \sum_{i_1 \in [p_1], \ldots, i_d \in [p_d]} \mathcal{X}_{i_1, \ldots, i_d} \mathcal{Y}_{i_1, \ldots, i_d}$, and its Frobenius norm as $\|\mathcal{X}\|_\text{F} = \sqrt{\langle\mathcal{X}, \mathcal{X}\rangle}$.
The mode-$k$ matricization of an order-$d$ tensor $\mathcal{T}$ is denoted as $\mathcal{M}_k(\mathcal{T})$ and reshapes the tensor into a matrix by aligning the $k$-th mode along the rows while consolidating all other modes as columns.
  For instance, for an order-3 tensor $\mathcal{T}  \in \mathbb{R}^{p_1 \times p_2 \times p_3}$, its mode-1 matricization $\mathcal{M}_1(\mathcal{T} ) \in \mathbb{R}^{p_1 \times\left(p_2 p_3\right)}$ is defined as, for $i \in\left[p_1\right], j \in\left[p_2\right], k \in\left[p_3\right]$, $\left[\mathcal{M}_1(\mathcal{T} )\right]_{i,(j-1) p_3+k}=\mathcal{T} _{i, j, k}.$
  Given a tensor $\mathcal{T} \in \mathbb{R}^{p_1 \times \cdots \times p_d}$ and a matrix $\mathbf{U} \in \mathbb{R}^{r_1 \times p_1}$, the marginal multiplication is defined as
$
\mathcal{T} \times_1 \mathbf{U} =  \sum_{i_1'=1}^{p_1} \mathcal{T}_{i_1', i_2, \ldots, i_d} \mathbf{U}_{i_1, i_1'} .
$
Marginal multiplications for other modes, $\times_2, \ldots, \times_d$, are analogously defined.
An essential identity that connects matrix-tensor products with matricization is
\begin{align*}
\mathcal{M}_k(\mathcal{G} \times_{k\in[d]} \mathbf{U}_k) = \mathbf{U}_k \mathcal{M}_k(\mathcal{G})\left( \mathbf{U}_d\otimes\cdots\otimes\mathbf{U}_{k+1}\otimes\mathbf{U}_{k-1}\cdots\otimes\mathbf{U}_1\right)^\top.
\end{align*}
For matrices $\mathbf{U}_k\in \mathbb{O}_{p_k, r_k}$ for $k\in[d]$, representing the left singular vectors of $\mathcal{M}_k(\mathcal{T})$, a tensor $\mathcal{T}$ has a Tucker decomposition of rank $ (r_1, r_2, \cdots, r_d)$ if there exists a core tensor $\mathcal{G} \in \mathbb{R}^{r_1 \times \cdots \times r_d}$ such that
\begin{align}\label{equ:def of Tucker}
    \mathcal{T} = \mathcal{G} \times_1 \mathbf{U}_1 \times_2 \ldots \times_d \mathbf{U}_d =\mathcal{G} \times_{k\in[d]} \mathbf{U}_k  .
\end{align}
If $\mathcal{T}$ has Tucker ranks $\left(r_1, \cdots, r_d\right)$, the signal strength of $\mathcal{T}$ is defined by
$ \lambda_{\text {min }}(\mathcal{T})=\min_{i\in[d]}\sigma_{r_i}\left(\mathcal{M}_1(\mathcal{T})\right), $
i.e., the smallest positive singular value of all matricizations. Similarly, define $\lambda_{\max }(\mathcal{T})=\max _k \sigma_1\left(\mathcal{M}_k(\mathcal{T})\right)$. To ease notation, we use $\lambda_{\min }$ and $\lambda_{\max }$ to refer to $\lambda_{\min }(\mathcal{T}^\star)$ and $\lambda_{\max }(\mathcal{T}^\star)$ of the true tensor throughout the paper.  The condition number of $\mathcal{T}$ is defined by $\kappa(\mathcal{T}):=$ $\lambda_{\max }(\mathcal{T}) \lambda_{\text {min }}^{-1}(\mathcal{T})$. We let $\mathbf{\Lambda}_k$ be the $r_k \times r_k$ diagonal matrix containing the singular values of $\mathcal{M}_k(\mathcal{G})$ (or equivalently the singular values of $\mathcal{M}_k(\mathcal{T})$ ). 
Readers seeking comprehensive discussions on tensor algebra are referred to \cite{kolda2009tensor}. In this study, we focus on third-order tensors, i.e., \( d = 3 \). Given this setting, the indices for \( k+1 \) and \( k+2 \) in \( \mathbf{U}_{k+1} \) and \( \mathbf{U}_{k+2} \) are determined using modulo 3 operations. This choice streamlines our notation and facilitates a clearer presentation of the core methodology.

%% file: body/ProofEstimation.tex
\section{Proof of Theorem \ref{thm:converge}}
\label{sec:Proof of Theorem convergence}

In this section, we present the proof of technical results concerning error contraction, organized into six steps. Step 1, detailed in Section \ref{sec:step 1 of convergence}, lays the groundwork by introducing essential notations and conditions necessary for developing the theoretical framework. Following this, Steps 2 through 4, detailed in Sections \ref{sec:step 2 of convergence} to \ref{sec:step 4 of convergence}, are dedicated to establishing the one-step error contraction. Step 5, found in Section \ref{sec:step 5 of convergence}, involves the construction of a super-martingale and the derivation of a high probability bound for the cumulative sum of this super-martingale. Finally, in Step 6, detailed in Section \ref{sec:step 6 of convergence}, we apply the union bound and provide the convergence analysis.

\subsection{Step 0: notations and conditions}
\label{sec:step 1 of convergence}
To quantify the difference between \( \mathcal{T}^{(t)} \) and \( \mathcal{T}^\star \), we utilize the error measurement \( J^{(t)} \), which is commonly utilized in factor-based gradient descent methods \citep{han2022optimal}:
\begin{align}\label{equ:def of Jt}
\begin{aligned}
	J^{(t)}&=\sum_{k=1}^3\left\|\mathbf{U}_k^{(t)}-\mathbf{U}_k^{\star} \mathbf{R}_k^{(t)}\right\|_{\mathrm{F}}^2+\left\| \mathcal{G}^{(t)} -  \mathcal{G}^\star \times_{k\in[3]} \mathbf{R}_k^{(t) \top}\right\|_{\mathrm{F}}^2,\\
J^{\prime(t)}&=\sum_{k=1}^3\left\|{\mathbf{U}}_k^{(t)}-\mathbf{U}_k^\star \mathbf{R}_k^{(t-1)}\right\|_{\mathrm{F}}^2+\left\|{\mathcal{G}}^{(t)}-  \mathcal{G}^\star \times_{k\in[3]}  \mathbf{R}_k^{(t-1) \top} \right\|_{\mathrm{F}}^2,
\end{aligned}
\end{align}
where 
\begin{align}\label{equ:def of Rt}
  \left(\mathbf{R}_1^{(t)}, \mathbf{R}_2^{(t)}, \mathbf{R}_3^{(t)}\right)=\underset{\substack{\mathbf{R}_k \in \mathbb{O}_{p_k, r_k} \\
k\in[3]}}{\arg \min }\left\{\sum_{k=1}^3\left\|\mathbf{U}_k^{(t)}-\mathbf{U}_k^{\star} \mathbf{R}_k\right\|_{\mathrm{F}}^2 + 
\left\|\mathcal{G}^{(t)}- \mathcal{G}^\star \times_{k\in[3]} \mathbf{R}_k^{\top}\right\|_{\mathrm{F}}^2\right\}.
\end{align}
It is crucial to recognize that \(J^{(t)}\) does not equate to \(\left\|\mathcal{T}^{(t)}-\mathcal{T}^\star\right\|_{\mathrm{F}}^2\) in the absence of a similar singular structure between \(\mathbf{U}_k^{(t)}\) and \(\mathbf{U}_k^\star\).
Intuitively, \( J^{(t)} \) quantifies the difference between the tensor components \( \mathcal{G}^\star \times_{k\in[3]} \mathbf{U}_k^\star \) and \( \mathcal{G}^{(t)} \times_{k\in[3]}  \mathbf{U}_k^{(t)} \) under rotation.  Based on the defined rotation matrices \( \big(\mathbf{R}_1^{(t)}, \mathbf{R}_2^{(t)}, \mathbf{R}_3^{(t)}\big) \), it becomes evident that \(J^{(t)}\) is bounded above by \(J^{\prime(t)}\).  To facilitate our analysis, \( J^{\prime(t)} \) will be utilized to establish an upper bound for \( J^{(t)} \).

Recalling Assumption \ref{cond:init} regarding the initial estimation \( \mathcal{T}^{(0)} \), and based on the equivalence between \( J^{(t)} \) and \( \left\|\mathcal{T}^{(t)}-\mathcal{T}^\star\right\|_{\mathrm{F}}^2 \) as explicated in Lemma E.2 of \cite{han2022optimal}, we can have the initial condition for \( J^{(0)} \):
\begin{align}\label{equ:init of J0}
  J^{(0)}\leq 480\lambda_{\min}^{-2}\|\mathcal{T}^{(0)}-\mathcal{T}^\star\|_{\mathrm{F}}^2\leq C_\text{init}^\prime.
\end{align} 
\xin{Here we use the assumption that the bound of initilizaiton is lambda min nor the sigma.}

For the low-rank tensor SGD algorithm, defining a benign region is crucial for analyzing convergence properties. The benign region, formally denoted as
\begin{align}\label{cond:def of benign region}
\mathcal{D}=\left\{\mathcal{G}^{} \times_{k\in[3]} \mathbf{U}_k^{} \mid J \leq c_d\lambda_{\min}^{2}\right\},
\end{align}
is a subset of the parameter space where the error measurement \( J \) is sufficiently small specifically, not exceeding \( c_d \lambda_{\min}^{2} \), where \( c_d \) is a constant. This region provides a controlled environment where the tensors are well-behaved, as described in the subsequent lemma:
\begin{lemma}\label{lem:property in region D}
   For $\mathcal{G}^{} \times_{k\in[3]} \mathbf{U}_k^{} \in \mathcal{D}$, then we have   
\begin{align}\label{equ:benign region}
   \max _{k\in[3]}\left\|\mathbf{U}_k\right\| \leq 1.01  \quad \text{ and } \quad 
  \max _{k\in[3]}\left\|\mathcal{M}_k(\mathcal{G})\right\| \leq 1.01 \lambda_{\max}.
\end{align} 
\end{lemma}
Proof in Section \ref{sec:Proof of Lemma 1}. To establish the one-step error contraction for \( J^{(t)} \), it is essential to define an event as a quantitative measure of desirable progression:
\begin{align}\label{equ:def of event e}
\mathcal{E}_t=\left\{\forall \tau \leq t: J^{(\tau)} \leq J^{(0)}\prod_{i=1}^\tau\left(1-\frac{\eta_i}{\phi}\right)^{-1} + C_\mathcal{E} \sigma^2 \left( \gamma\log(p)\sqrt{\df\eta_\tau^3} + \df \eta_\tau  \right) \right\},
\end{align}
where $\phi=\left(2c_0\lambda_{\min}^2\right)^{-1}$ for some constant $c_0$\xin{check why phi}. By definition $\mathbb{P}\left[\mathcal{E}_0\right] = 1$. This event \( \mathcal{E}_t \) represents a sequence of states where the error measurement \( J^{(\tau)} \) remains controlled throughout the online process. It is worth pointing out that by the definition of $t^{\star}$ and $\eta_t$ in Theorem \ref{thm:converge}, we have
\begin{align}\label{equ:bound of J under E}
C_\mathcal{E} \sigma^2 \left( \gamma\log(p)\sqrt{\df\eta_\tau^3} + \df \eta_\tau  \right) \leq  C_\mathcal{E}  \sigma^2 \df \eta_{t^{\star}} \leq C_\mathcal{E} \sigma^2,
\end{align}
which means that event $\mathcal{E}_t$ implies $J^{(t)}\leq C_\mathcal{E} \sigma^2$ for some constant $C_\mathcal{E}>0$. Recall the definition of the region $\mathcal{D}$, one can see that for large enough signal-to-noise ratio \xin{(SNR Assumption)}, i.e., $\left(\frac{\lambda_{\min}}{\sigma}\right)^2 \geq  \frac{C_\text{init}}{c_d} $, and when the event $\mathcal{E}_{t}$ happens, we have $\mathcal{G}^{(t)}\times_{k\in[3]} \mathbf{U}_k^{(t)} \in \mathcal{D}$. This observation ensures that our online process remains within a region where the tensor maintains desirable properties for convergence. The following lemma is a crucial component in the theoretical analysis.
 We first define
\begin{align}\label{equ:def of Xi}
\begin{aligned}
	\epsilon_{1}^{(t)}:
=& \sup _{\substack{\mathcal{T}^\prime \in \mathbb{R}^{p_1 \times p_2 \times p_3},\\\|\mathcal{T}^\prime\|_\mathrm{F} \leq 1, \\ \operatorname{rank}(\mathcal{T}^\prime) \leq\left(r_1, r_2, r_3\right)}}\left|\left\langle\nabla_\mathcal{T}f\left(\mathcal{T}^{\star};\boldsymbol{\zeta}_{t+1}\right), \mathcal{T}^\prime\right\rangle\right|, \ \text{ and } \epsilon_{2}^{(t)}:
=& \fbr{\nabla_\mathcal{T}f\left(\mathcal{T}^{(t)} - \mathcal{T}^{\star};\boldsymbol{\zeta}_{t+1}\right)}.
\end{aligned}
\end{align}
\begin{lemma}\label{lem:def of Xi}
  Under the Assumption \ref{cond:1}, there exist some universal constants $C_2, c_2$ and event $\mathcal{E}^\epsilon_t$, such that $\mathbb{P}\left[\left(\mathcal{E}^\epsilon_t\right)^c\right] \leq 2\exp \left(-c_2 \sqrt{\df}\right)$, where $\mathcal{E}^\epsilon_t = \left\{\epsilon_{1}^{(t)} + \epsilon_{2}^{(t)}I\{\cE_t\}\leq C_2 \sigma \sqrt{\df}\right\}$, and $\df = r_1 r_2 r_3+\sum_{k=1}^3 p_k r_k$. 
\end{lemma}
Proof in Section \ref{sec:proof of lemma def of Xi}. 
Intuitively speaking, $\epsilon_{1}^{(t)}$ and $\epsilon_{2}^{(t)}$ measure the fluctuation of the gradient of the loss function \( f \) at the true parameters \( \mathcal{T}^\star \) and 
 the difference between the gradient at the current estimate \( \mathcal{T}^{(t)} \) and the gradient at the true parameters \( \mathcal{T}^\star \) projected onto the manifold of low-rank tensors.

Under the conditions outlined in Assumption \ref{cond:1}, the tensor $\mathcal{X}_t$ is composed of i.i.d. sub-Gaussian entries with variance 1. This statistical structure  imparts a  characteristic to  $\mathcal{X}_t$: specifically, for any fixed tensor $\Delta$
with the same shape of $\mathcal{X}_t$, we have
\begin{align}\label{equ:X inner product}
\mathbb{E}\left[\left\langle \mathcal{X}_t, \Delta\right\rangle\mathcal{X}_t\right] = \Delta.
\end{align}
When involved in an inner-outer product operation, the tensor $\mathcal{X}_t$ behaves as a kind of  ``identity operator" for the fixed tensor $\Delta$.  Before we start, we recall the definition of function $\psi_p$: it is defined as $\psi_p(u)=\exp \left(u^p\right)-1,$ when $u>u_0$, and $\psi_p(u)$ is linear for $u \leq u_0$ to preserve the convexity of function $\psi$.  Subsequently, the Orlicz norm of a random variable $y$ with respect to $\psi_p$ is established as $\|y\|_{\psi_p}=\inf\{v > 0: \mathbb{E}[\psi_p(|y| / v)] \leq 1\}$.

The following analysis decomposes the error measurement into two distinct components: the factor matrices and the core tensor. To address these errors, we proceed by establishing a one-step contraction for the factor matrices $\mathbf{U}_k^{(t+1)}$ in Step 2. Following that, in Step 3, we focus on constructing a one-step contraction for the core tensor ${\mathcal{G}}^{(t+1)}$. Finally, Step 4 involves formulating a one-step contraction for the entire error term $J^{(t+1)}$.

\subsection{Step 1: One-step Contraction}
\label{sec:step 2 of convergence}
We have the following decomposition by plugging in the gradient in Algorithm \ref{alg:Low rank tensor SGD}.

\begin{lemma}\label{lem:contraction of U}
For $k\in[3]$, we have
\begin{align}\label{equ:one step U}
&\left\|{\mathbf{U}}_k^{(t+1)}-\mathbf{U}_k^\star \mathbf{R}_k^{(t)}\right\|_{\mathrm{F}}^2 I\left\{\mathcal{E}_{t}\right\} \leq \rbr{\left\|\mathbf{U}_k^{(t)}-\mathbf{U}_k^\star \mathbf{R}_k^{(t)}\right\|_{\mathrm{F}}^2 -2 \eta_{t+1}J_{k, 1}^{(t)} +\eta_{t+1}^2J_{k, 2}^{(t)}}I\left\{\mathcal{E}_{t}\right\} ,
\end{align}
where
\begin{align*}
  J_{k, 1}^{(t)} = &\left\langle\mathcal{T}^{(t)}-\mathcal{T}_k^{(t)}, \nabla f\left(\mathcal{T}^{(t)}\right)\right\rangle + \frac{1}{8} \left( \left\|\mathbf{U}_k^{(t) \top} \mathbf{U}_k^{(t)}-\mathbf{U}_k^{\star \top} \mathbf{U}_k^{\star}\right\|_{\mathrm{F}}^2-c_d\lambda_{\min}^{2} \left\|\mathbf{U}_k^{(t)}-\mathbf{U}_k^{\star} \mathbf{R}_k^{(t)}\right\|_{\mathrm{F}}^2\right), \\
  J_{k, 2}^{(t)} = &6 \lambda_{\max}^2  \left( \left(\epsilon_{1}^{(t)}\right)^2+\left(\epsilon_{2}^{(t)}\right)^2\right)+\frac{5}{8} \left\|\mathbf{U}_k^{(t) \top} \mathbf{U}_k^{(t)}-\mathbf{U}_k^{\star \top} \mathbf{U}_k^{\star}\right\|_{\mathrm{F}}^2. 
\end{align*}
\end{lemma}
Proof in Section \ref{sec:Proof of Lemma contraction of U}.

\begin{lemma}\label{lem:contraction of G}
	\begin{align}\label{equ:one step G}
\begin{aligned}
& \left\|\mathcal{G}^{(t+1)}-\mathcal{G}  \times_{k\in[3]} \mathbf{R}_k^{(t) \top} \right\|_{\mathrm{F}}^2I\left\{\mathcal{E}_{t}\right\}  
\leq  \rbr{\left\|\mathcal{G}^{(t)}-\mathcal{G}  \times_{k\in[3]} \mathbf{R}_k^{(t) \top}\right\|_{\mathrm{F}}^2 -2 \eta_{t+1}J_{\mathcal{G}, 1}^{(t)}+\eta_{t+1}^2J_{\mathcal{G}, 2}^{(t)}}I\left\{\mathcal{E}_{t}\right\},
\end{aligned}
\end{align}
where $J_{\mathcal{G}, 1}^{(t)} = \left\langle\mathcal{T}^{(t)}-\mathcal{T}_{\mathcal{G}}^{(t)}, \nabla_\mathcal{T}f\left(\mathcal{T}^{(t)}\right)\right\rangle$ and $J_{\mathcal{G}, 2}^{(t)} =  3 \left( \left(\epsilon_{1}^{(t)}\right)^2+\left(\epsilon_{2}^{(t)}\right)^2\right)$.
\end{lemma}
Proof in Section \ref{sec:Proof of Lemma contraction of G}.
\label{sec:step 4 of convergence}
Incorporating the one-step contraction of $\mathbf{U}_1^{(t+1)}$ as described in Equation (\ref{equ:one step U}) and the one-step contraction of $\mathcal{G}^{(t+1)}$ as described in Equation (\ref{equ:one step G}) into the established definition of $J^{(t+1)}$ given in Equation (\ref{equ:def of Jt}), we derive the following expression:
  \begin{align}\label{equ:D.8-1}
  J^{(t+1)} \leq  J^{\prime(t+1)} \leq J^{(t)}-2 \eta_{t+1}\left(J_{\mathcal{G}, 1}^{(t)}+\sum_{k=1}^3 J_{k, 1}^{(t)}\right)+\eta_{t+1}^2\left(J_{\mathcal{G}, 2}^{(t)}+\sum_{k=1}^3 J_{k, 2}^{(t)}\right).
\end{align}
Next, we will further control the two terms on the right side of the formula above.
Let $\mathcal{F}_t$ denote the filtration generated by all the historical randomness up to time $t$, i.e., $\mathcal{F}_t=\sigma\left(\cX_1, y_1, \ldots, \cX_t, y_t\right)$. 
\begin{lemma} \label{lem:bound of JG1}
	In this step, we provide a sharper lower bound for $J_{\mathcal{G}, 1}^{(t)}+\sum_{k=1}^3 J_{k, 1}^{(t)}$:
\begin{align}\label{equ:def of Q1 bound}
  \mathbb{E}\left[J_{\mathcal{G}, 1}^{(t)}+\sum_{k=1}^3 J_{k, 1}^{(t)} \mid\mathcal{F}_t \right]I\left\{\mathcal{E}_{t}\cap\mathcal{E}_t^\epsilon\right\} \geq & \left(c_0 \lambda_{\min}^{2} J^{(t)} + \frac{1}{24} \sum_{k=1}^3\left\|\mathbf{U}_k^{(t) \top} \mathbf{U}_k^{(t)}-\mathbf{U}_k^{\star \top} \mathbf{U}_k^{\star}\right\|_{\mathrm{F}}^2 \right) I\left\{\mathcal{E}_{t}\right\}.
\end{align}
\end{lemma}
Proof in Section \ref{sec:Proof of Lemma bound of JG1}.
For $J_{\mathcal{G}, 2}^{(t)}+\sum_{k=1}^3 J_{k, 2}^{(t)}$, based on the definitions of \(J_{k, 2}^{(t)}\) in Equation (\ref{def of Q12}) and \(J_{\mathcal{G}, 2}^{(t)}\) in Equation (\ref{def of Gg2}), we can conclude:
\begin{align}\label{equ:def of Q2 bound}
\begin{aligned}
 \left(J_{\mathcal{G}, 2}^{(t)} +\sum_{k=1}^3 J_{k, 2}^{(t)}\right) I\left\{\mathcal{E}_{t}\right\} 
\leq & \left(21 \left( \left(\epsilon_{1}^{(t)}\right)^2+\left(\epsilon_{2}^{(t)}\right)^2\right) +2 \sum_{k=1}^3\left\|\mathbf{U}_k^{(t) \top} \mathbf{U}_k^{(t)}-\mathbf{U}_k^{\star \top} \mathbf{U}_k^{\star}\right\|_{\mathrm{F}}^2\right)I\left\{\mathcal{E}_{t}\right\}\\
\leq & \left( 21 C_2 \sigma^2 \df+2 \sum_{k=1}^3\left\|\mathbf{U}_k^{(t) \top} \mathbf{U}_k^{(t)}-\mathbf{U}_k^{\star \top} \mathbf{U}_k^{\star}\right\|_{\mathrm{F}}^2\right)I\left\{\mathcal{E}_{t}\right\}.
\end{aligned}
\end{align}
The last line is based on Lemma \ref{lem:def of Xi}.

\subsubsection{Error contraction of $J^{(t+1)}$}
When $\eta_t\leq\frac{1}{24}$, we have $ \left(-\frac{1}{12}\eta_t + 2 \eta_t^2 \right) \sum_{k=1}^3\left\|\mathbf{U}_k^{(t) \top} \mathbf{U}_k^{(t)}-\mathbf{U}_k^{\star \top} \mathbf{U}_k^{\star}\right\|_{\mathrm{F}}^2 \leq 0 $.
Substituting Equations \eqref{equ:def of Q1 bound} and (\ref{equ:def of Q2 bound}) into (\ref{equ:D.8-1}), we obtain:
\begin{align}\label{equ:one contra of EJ}
\begin{aligned}
 \mathbb{E}\left[J^{(t+1)}\mid\mathcal{F}_t\right]I\left\{\mathcal{E}_{t}\cap\mathcal{E}_t^\epsilon\right\} \leq &\mathbb{E}\left[J^{\prime(t+1)}\mid\mathcal{F}_t\right]I\left\{\mathcal{E}_{t}\cap\mathcal{E}_t^\epsilon\right\} \\ 
\leq & \left(1-2c_0\lambda_{\min}^2\eta_{t+1}\right) J^{(t)} I\left\{\mathcal{E}_{t}\right\} + 21 C_2 \sigma^2 \df\eta_{t+1}^2 .
\end{aligned}
\end{align}
We have now obtained a one-step error contraction of $J^{(t+1)}$ under the conditional expectation of $\mathcal{F}_t$. Next, we will employ a sup-martingale to establish an upper bound for $J^{(t+1)}$ without relying on conditional expectation.

\subsection{Step 2: Construct a super-martingale}
\label{sec:step 5 of convergence}
\begin{lemma}\label{lem:super-martingale}
  If we define $ J_{M, t}= 21 C_2 \sigma^2 \df\phi\eta_t$, 
where $\phi=\left(2c_0\lambda_{\min}^2\right)^{-1}$ and some constant $C_2>0$ does not depend on $t$, and we define
\begin{align*}
\mathcal{J}_t= \prod_{\tau=1}^t\left(1-\frac{\eta_\tau}{\phi}\right)^{-1}\left(J^{\prime(t)}I\left\{\mathcal{E}_{t-1}\cap\mathcal{E}_{t-1}^\epsilon\right\} -J_{M, t}\right).
\end{align*}
Then $\mathcal{J}_t$ is a super-martingale, i.e., $\mathbb{E}\left[\mathcal{J}_{t} \mid \mathcal{F}_{t-1}\right] \leq \mathcal{J}_{t-1}$.
\end{lemma}
Proof in Section \ref{section:proof of super-martingale}. \xin{Check this lemma.}
Given $\mathcal{J}_t$ is a super-martingale, the following holds
\begin{align}\label{equ:Mt-M0}
\begin{aligned}
	\mathcal{J}_t-\mathcal{J}_0=&\sum_{\tau=1}^{t}\left(\mathcal{J}_\tau-\mathcal{J}_{\tau-1}\right) \\
	\leq &  \sum_{\tau=1}^{t}\left(\mathcal{J}_\tau-\mathbb{E}\left[\mathcal{J}_\tau \mid \mathcal{F}_{\tau-1}\right]\right)\\
= &\sum_{\tau=1}^{t}\prod_{s=1}^\tau\left(1-\frac{\eta_s}{\phi}\right)^{-1}\left(J^{\prime(\tau)} -\mathbb{E}\left[J^{\prime(\tau)}\mid\mathcal{F}_{\tau-1}\right]\right)I\left\{\mathcal{E}_{t-1}\cap\mathcal{E}_{t-1}^\epsilon\right\}.
\end{aligned}
\end{align}
By the definition of $J^{\prime(t+1)}$, we have
\begin{align*}
\begin{aligned}
	  J^{\prime(t+1)}  =&\sum_{k=1}^3\left\|\mathbf{U}_k^{(t+1)}-\mathbf{U}_k^\star \mathbf{R}_k^{(t)}\right\|_{\mathrm{F}}^2+\left\|\mathcal{G}^{(t+1)}- \mathcal{G}^\star \times_{k\in[3]} \mathbf{R}_k^{(t) \top} \right\|_{\mathrm{F}}^2\\
= & \sum_{k=1}^3\left\|\mathbf{U}_k^{(t)}-\mathbf{U}_k^\star \mathbf{R}_k^{(t)}-\eta_{t+1}\left[\mathcal{M}_k\left( \nabla_\mathcal{T}f\left(\mathcal{T}^{(t)}\right)\right) \breve{\mathbf{U}}^{(t)}_k+\frac{1}{2} \mathbf{U}_k^{(t)}\left(\mathbf{U}_k^{(t) \top} \mathbf{U}_k^{(t)} - \mathbf{U}_k^{\star \top} \mathbf{U}_k^{\star}\right)\right]\right\|_{\mathrm{F}}^2 \\
&+\left\|\mathcal{G}^{(t)}- \mathcal{G}^\star \times_{k\in[3]} \mathbf{R}_k^{(t) \top}
 -\eta_{t+1} \nabla_\mathcal{T}f\left(\mathcal{T}^{(t)}\right) \times_{k\in[3]} \mathbf{U}_k^{(t) \top} \right\|_{\mathrm{F}}^2\\
= & \sum_{k=1}^3\left\|\mathbf{U}_k^{(t)}-\mathbf{U}_k^\star \mathbf{R}_k^{(t)}\right\|_{\mathrm{F}}^2 +\eta_{t+1}^2\sum_{k=1}^3\left\|\mathcal{M}_k\left( \nabla_\mathcal{T}f\left(\mathcal{T}^{(t)}\right)\right) \breve{\mathbf{U}}^{(t)}_k\right\|_{\mathrm{F}}^2\\
& +\eta_{t+1}^2\sum_{k=1}^3\left\|\frac{1}{2} \mathbf{U}_k^{(t)}\left(\mathbf{U}_k^{(t) \top} \mathbf{U}_k^{(t)} - \mathbf{U}_k^{\star \top} \mathbf{U}_k^{\star}\right)\right\|_{\mathrm{F}}^2 \\
& +  \eta_{t+1}^2\sum_{k=1}^3\left\langle\mathcal{M}_k\left( \nabla_\mathcal{T}f\left(\mathcal{T}^{(t)}\right)\right) \breve{\mathbf{U}}^{(t)}_k,  \mathbf{U}_k^{(t)}\left(\mathbf{U}_k^{(t) \top} \mathbf{U}_k^{(t)} - \mathbf{U}_k^{\star \top} \mathbf{U}_k^{\star}\right)\right\rangle\\
& -2 \eta_{t+1}\sum_{k=1}^3\left\langle\mathbf{U}_k^{(t)}-\mathbf{U}_k^\star \mathbf{R}_k^{(t)}, \mathcal{M}_k\left( \nabla_\mathcal{T}f\left(\mathcal{T}^{(t)}\right)\right) \breve{\mathbf{U}}^{(t)}_k\right\rangle \\
& - \eta_{t+1} \sum_{k=1}^3\left\langle\mathbf{U}_k^{(t)}-\mathbf{U}_k^\star \mathbf{R}_k^{(t)}, \mathbf{U}_k^{(t)}\left(\mathbf{U}_k^{(t) \top} \mathbf{U}_k^{(t)} - \mathbf{U}_k^{\star \top} \mathbf{U}_k^{\star}\right)\right\rangle\\
&+  \left\|\mathcal{G}^{(t)}- \mathcal{G}^\star \times_{k\in[3]} \mathbf{R}_k^{(t) \top} \right\|_{\mathrm{F}}^2
+\eta_{t+1}^2\left\| \nabla_\mathcal{T}f\left(\mathcal{T}^{(t)}\right) \times_{k\in[3]} \mathbf{U}_k^{(t) \top} \right\|_{\mathrm{F}}^2 \\
& -2 \eta_{t+1}\left\langle\mathcal{G}^{(t)}- \mathcal{G}^\star \times_{k\in[3]} \mathbf{R}_k^{(t) \top} ,  \nabla_\mathcal{T}f\left(\mathcal{T}^{(t)}\right) \times_{k\in[3]} \mathbf{U}_k^{(t) \top} \right\rangle .
\end{aligned}
\end{align*}
We subtract the conditional expectation from $J^{\prime(t+1)}$, yielding the following result:
\begin{align*}
  &J^{\prime(t+1)}-\mathbb{E}\left[J^{\prime(t+1)}\mid \mathcal{F}_{t}\right]\\
  = & J_1^{(t+1)} + J_2^{(t+1)} + J_3^{(t+1)}- \mathbb{E}\left[J_1^{(t+1)}\mid \mathcal{F}_{t}\right]  -\mathbb{E}\left[J_2^{(t+1)}\mid \mathcal{F}_{t}\right] -\mathbb{E}\left[J_3^{(t+1)}\mid \mathcal{F}_{t}\right],
\end{align*}
where
\begin{align*}
   J_1^{(t+1)}
   =& -2 \eta_{t+1}\sum_{k=1}^3\left\langle\mathbf{U}_k^{(t)}-\mathbf{U}_k^\star \mathbf{R}_k^{(t) \top}, \mathcal{M}_k\left( \nabla_\mathcal{T}f\left(\mathcal{T}^{(t)}\right)\right) \breve{\mathbf{U}}^{(t)}_k\right\rangle\\
  & -2 \eta_{t+1}\left\langle\mathcal{G}^{(t)}- \mathcal{G}^\star\times_{k\in[3]} \mathbf{R}_k^{(t)  \top},  \nabla_\mathcal{T}f\left(\mathcal{T}^{(t)}\right) \times_{k\in[3]} \mathbf{U}_k^{(t) \top} \right\rangle,\\
    J_2^{(t+1)}=&\eta_{t+1}^2\left(\sum_{k=1}^3\left\|\mathcal{M}_k\left( \nabla_\mathcal{T}f\left(\mathcal{T}^{(t)}\right)\right) \breve{\mathbf{U}}^{(t)}_k\right\|_{\mathrm{F}}^2 + \left\| \nabla_\mathcal{T}f\left(\mathcal{T}^{(t)}\right) \times_{k\in[3]} \mathbf{U}_k^{(t) \top} \right\|_{\mathrm{F}}^2\right), \\
      J_3^{(t+1)}= & \eta_{t+1}^2\sum_{k=1}^3\left\langle\mathcal{M}_k\left( \nabla_\mathcal{T}f\left(\mathcal{T}^{(t)}\right)\right) \breve{\mathbf{U}}^{(t)}_k, \mathbf{U}_k^{(t)}\left(\mathbf{U}_k^{(t) \top} \mathbf{U}_k^{(t)} - \mathbf{U}_k^{\star \top} \mathbf{U}_k^{\star}\right)\right\rangle.
\end{align*}
Using the notations \(J_1^{(t+1)}\), \(J_2^{(t+1)}\), and \(J_3^{(t+1)}\), we can express Equation (\ref{equ:Mt-M0}) as follows:
\begin{align*}
\begin{aligned}
 \mathcal{J}_t-\mathcal{J}_0 \leq &\sum_{\tau=1}^{t}\left(\mathcal{J}_\tau-\mathbb{E}\left[\mathcal{J}_\tau \mid \mathcal{F}_{\tau-1}\right]\right) \\
\leq & \left|\sum_{\tau=1}^{t} \prod_{s=1}^\tau\left(1-\frac{\eta_s}{\phi}\right)^{-1}\rbr{J_1^{(\tau)}-\mathbb{E}\left[J_1^{(\tau)}\mid \mathcal{F}_{\tau-1}\right]}\right|I\left\{\mathcal{E}_{t-1}\cap\mathcal{E}_{t-1}^\epsilon\right\}\\
&+\left|\sum_{\tau=1}^{t} \prod_{s=1}^\tau\left(1-\frac{\eta_s}{\phi}\right)^{-1}\rbr{J_2^{(\tau)}-\mathbb{E}\left[J_2^{(\tau)}\mid \mathcal{F}_{\tau-1}\right]}\right|I\left\{\mathcal{E}_{t-1}\cap\mathcal{E}_{t-1}^\epsilon\right\}\\
&+\left|\sum_{\tau=1}^{t} \prod_{s=1}^\tau\left(1-\frac{\eta_s}{\phi}\right)^{-1}\rbr{J_3^{(\tau)}-\mathbb{E}\left[J_3^{(\tau)}\mid \mathcal{F}_{\tau-1}\right]}\right|I\left\{\mathcal{E}_{t-1}\cap\mathcal{E}_{t-1}^\epsilon\right\}.
\end{aligned}
\end{align*}
then recall the definition of the super-martingale $\mathcal{J}_t$, the following relationship holds true,
\begin{align}\label{equ:upper bound of J of J1J2J3}
\begin{aligned}
J^{(t)}I\left\{\mathcal{E}_{t-1}\cap\mathcal{E}_{t-1}^\epsilon\right\}
 \leq & \underbrace{\prod_{s=1}^t\left(1-\frac{\eta_s}{\phi}\right)\left(J^{\prime(0)} -J_{M,0}\right)}_{\mathcal{J}_0}\\
 & +\left|\sum_{\tau=1}^{t} \prod_{s=\tau+1}^t\left(1-\frac{\eta_s}{\phi}\right)\rbr{J_1^{(\tau)}-\mathbb{E}\left[J_1^{(\tau)}\mid \mathcal{F}_{\tau-1}\right]}\right|I\left\{\mathcal{E}_{t-1}\cap\mathcal{E}_{t-1}^\epsilon\right\} \\
& +\left|\sum_{\tau=1}^{t} \prod_{s=\tau+1}^t\left(1-\frac{\eta_s}{\phi}\right)\rbr{J_2^{(\tau)}-\mathbb{E}\left[J_2^{(\tau)}\mid \mathcal{F}_{\tau-1}\right]}\right|I\left\{\mathcal{E}_{t-1}\cap\mathcal{E}_{t-1}^\epsilon\right\}\\
& +\left|\sum_{\tau=1}^{t} \prod_{s=\tau+1}^t\left(1-\frac{\eta_s}{\phi}\right)\rbr{J_3^{(\tau)}-\mathbb{E}\left[J_3^{(\tau)}\mid \mathcal{F}_{\tau-1}\right]}\right|I\left\{\mathcal{E}_{t-1}\cap\mathcal{E}_{t-1}^\epsilon\right\}\\
& +J_{M, t} .
\end{aligned}
\end{align}
Then, to bound the term $J^{(t)}$ with high probability, it remains to show the right-hand side of the above expression can be upper bounded with high probability. 

\subsection{Step 3: convergence Analysis}
\label{sec:step 6 of convergence}
For the second term on the right-hand side of Equation (\ref{equ:upper bound of J of J1J2J3}), we can establish the following lemma:
\begin{lemma}\label{lem:bound of J1}
For any large enough constant $\gamma>0$, there exists an absolute constant $C_2$ such that with probability at least $1-2 p^{-\gamma}$, we have
\begin{align*}
\begin{aligned}
&\left|\sum_{\tau=1}^{t} \prod_{s=\tau+1}^t\left(1-\frac{\eta_s}{\phi}\right) \left(J_1^{(\tau)}-\mathbb{E}\left[J_1^{(\tau)}\mid \mathcal{F}_{\tau-1}\right]\right)\right|I\left\{\mathcal{E}_{t-1}\right\} \leq C_2\gamma\sigma^2\log p\sqrt{\df\eta_t^3}.
\end{aligned}
\end{align*}
\end{lemma}
Proof in Section \ref{section:proof of bound of J1}.
For the final two terms on the right-hand side of Equation (\ref{equ:upper bound of J of J1J2J3}), we can establish the following lemma:
\begin{lemma}\label{lem:bound of J2}
There exists an absolute constant $C_3$ and $C_4$ such that we can  have
\begin{align*}
\begin{aligned}
&\left|\sum_{\tau=1}^{t} \prod_{s=\tau+1}^t\left(1-\frac{\eta_s}{\phi}\right) \left(J_2^{(\tau)}-\mathbb{E}\left[J_2^{(\tau)}\mid \mathcal{F}_{\tau-1}\right]\right)\right|I\left\{\mathcal{E}_{t-1}\cap\mathcal{E}_{t-1}^\epsilon\right\} \leq C_3 \sigma^2 \df \eta_t,\\
&\left|\sum_{\tau=1}^{t} \prod_{s=\tau+1}^t\left(1-\frac{\eta_s}{\phi}\right) \left(J_3^{(\tau)}-\mathbb{E}\left[J_3^{(\tau)}\mid \mathcal{F}_{\tau-1}\right]\right)\right|I\left\{\mathcal{E}_{t-1}\cap\mathcal{E}_{t-1}^\epsilon\right\} \leq C_4 \sigma^2 \df\sqrt{\eta_t^3}.
\end{aligned}
\end{align*}
\end{lemma}
Proof in Section \ref{section:proof of bound of J2}.
By combining the results from Lemmas \ref{lem:bound of J1} and Lemma \ref{lem:bound of J2}, along with Equation (\ref{equ:upper bound of J of J1J2J3}), we can assert with a probability of at least $1-2p^{-\gamma}$ that
\begin{align*}
\begin{aligned}
J^{(t)}I\left\{\mathcal{E}_{t-1}\cap\mathcal{E}_{t-1}^\epsilon\right\} \leq & \prod_{s=1}^t\left(1-\frac{\eta_s}{\phi}\right)\left(J^{(0)} -J_{M,0}\right) + C_2\gamma\sigma^2\log p\sqrt{\df\eta_t^3}\\
&+C_3 \sigma^2 \df \eta_t+ C_4 \sigma^2 \df\sqrt{\eta_t^3} + 21 C_2 \sigma^2 \df\phi\eta_t.
\end{aligned}
\end{align*}
Since we know that by Assumption \ref{cond:init} and Equation (\ref{equ:init of J0}), we have $J^{(0)} \leq C_\text{init}$ for some constant $C_\text{init}>$ 0 . Due to the fact that $J_{M, 0}= 21C_2 \sigma^2 \df\phi\eta_{t^\star}$, together with the definition of $\eta_{t^{\star}}$, we can have $J^{(0)}-J_{M, 0} \leq 0$ as long as $C_\text{init}$ is small enough in Assumption \ref{cond:init}. Finally, by the definition of event $\mathcal{E}_{t-1}^\epsilon$ in Lemma \ref{lem:def of Xi}, we can conclude that with probability $1-3p^{-\gamma}$,
\begin{align*}
\begin{aligned}
J^{(t)}I\left\{\mathcal{E}_{t-1}\right\} \leq & C_2^\prime \sigma^2\left(\gamma\log p \sqrt{\df \eta_t^3}+\df\eta_t\right),
\end{aligned}
\end{align*}
for some large enough $C_2$. Therefore, what we have shown is,
\begin{align*}
\mathbb{P}\left(\mathcal{E}_{t-1} \cap \mathcal{E}_t^c\right) \leq \frac{3}{p^\gamma},
\end{align*}
where $\mathcal{E}_t^c$ denotes the complementary event of $\mathcal{E}_t$. We have the probability of the event $\mathcal{E}_t$ as
\begin{align*}
\begin{aligned}
\mathbb{P}\left(\mathcal{E}_t\right) = 1-\mathbb{P}\left(\mathcal{E}_t^c\right) \geq 1-\sum_{\tau=1}^t \mathbb{P}\left(\mathcal{E}_{\tau-1} \cap \mathcal{E}_\tau^c\right) \geq 1-3tp^{-\gamma}.
\end{aligned}
\end{align*}
Thus, we conclude the proof of Theorem \ref{thm:converge}.

%% file: body/ProofEntry.tex
\section{Proof of Theorem \ref{thm:entry inference}}
\label{sec:Proof of Theorem entry inference}
In this proof, we use a generic index $t$ to prove the distribution of $\hat{h}^{(t)}$:
\begin{align}\label{equ:mt in proof}
\hat{h}^{(t)}-h^{\star}=\left\langle\widehat{\mathcal{T}}^{(t)} \times_{k \in[3]} \mathcal{P}_{\widehat{\mathbf{U}}_k^{(t)}}-\mathcal{T}^{\star}, \mathcal{H}\right\rangle,
\end{align}
where $\widehat{\mathcal{T}}^{(t)}$ is the average estimator defined in \eqref{equ:debias},
and $\widehat{\mathbf{U}}_k^{(t)}$ is the singular vectors from HOSVD. Then the result of Theorem holds when we set $t=n$.
Using the notation \(\Delta_{\tau-1} = \mathcal{T}^\star - {\mathcal{T}}^{(\tau-1)}\), the  explicit representation formulas for $\widehat{\mathcal{T}}^{(t)}$ can be:
\begin{align}\label{equ:t=t+z}
\begin{aligned}
	\widehat{\mathcal{T}}^{(t)} - \mathcal{T}^\star &= \underbrace{\frac{1}{t} \sum_{\tau=1}^{t}  \xi_{\tau} \mathcal{X}_{\tau}}_{ \mathcal{Z}_1^{(t)}} + \underbrace{\frac{1}{t} \sum_{\tau=1}^{t}\left(\left\langle\Delta_{\tau-1}, \mathcal{X}_{\tau}\right\rangle \mathcal{X}_{\tau} - \Delta_{\tau-1}\right)}_{\mathcal{Z}_2^{(t)}}=:\mathcal{Z}^{(t)}.
\end{aligned}
\end{align}
Based on Theorem \ref{thm:converge}, we have established that \( \left\|\Delta_t\right\|_{\mathrm{F}} = o_{\bP}\left(\sigma\right) \) and that the influence of \( {\mathcal{Z}_2^{(t)}} \) is predominantly determined by that of \( {\mathcal{Z}_1^{(t)}} \). It is worth noting that the perturbation induced by \( {\mathcal{Z}_1^{(t)}} \) resembles or closely approximates a random perturbation with i.i.d. entry-wise noise. 

We begin by providing an upper bound for the singular vector \( \widehat{\mathbf{U}}_k^{(t)} \). Since online inference tasks rely on SGD estimators \( \mathcal{T}^{(t)} \) and \( \mathbf{U}^{(t)} \), it is necessary for the time \( t \) to exceed a certain threshold \( t_0 \) to achieve satisfactory accuracy. After reaching the threshold \( t_0 \), we use \( \mathbf{U}_k^{(t_0)} \) as the initial estimate for \( \widehat{\mathbf{U}}_k^{(t_0)} \). 

For subsequent steps, starting from \(t_0 +1\), we establish the error bound under the following conditions for all $k\in[3]$: \(\widehat{\mathbf{U}}_k^{(t_0)} = \mathbf{U}_k^{(t_0)}\) and \(\widehat{\mathbf{U}}_k^{(t_0+1)}\) is the leading \(r_k\) left singular vectors of \(\mathcal{M}_k\left(\widehat{\mathcal{T}}^{(t_0+1)} \times_{j\neq k} \widehat{\mathbf{U}}_{j}^{(t_0) \top} \right)\). The following lemma states that for sufficiently large values of \( t \), the corresponding $\sin\Theta$ norm of the matrix \( \mathbf{U}_k^{(t)} \) is subject to an upper bound.
\begin{lemma}\label{lem:bound for inference}
Let \(L_t\) denote the spectral \(\sin \Theta\) norm error for \(\widehat{\mathbf{U}}_k^{(t)}\) at time $t$. There exist absolute constants \(C_1 > 0\), for all 
$t^\alpha\rbr{\lambda_{\min}/\sigma}^2 \geq C_1 \df$,
we have
\[
L_t = \max_{k\in[3]} \left\|\sin \Theta\left(\widehat{\mathbf{U}}_k^{(t)}, \mathbf{U}_k^\star\right)\right\| = \Op{\frac{\sigma}{\lambda_{\min}}\sqrt{\frac{p}{t}}}.
\]
\end{lemma}
\xin{Check that the bound in Lemma A.1 is without $\kappa_0$ term while Lemma A.2 has. And is that correct that there is no log term in the bound. I found that we can remove the probability event used in the proof of Theorem 1 and replace directly with Op language without adding log term. I am not sure about that.}
Proof in Section \ref{sec:Proof of Lemma bound for inference}.
To establish a more precise bound for the leading term in \( \hat{h}^{(t)}-h^\star \), we will demonstrate the first-order expansion of the tensor singular vectors \( \left(\mathcal{P}_{\widehat{\mathbf{U}}_1^{(t)}}-\mathcal{P}_{\mathbf{U}_1^\star}\right) \).
Following Algorithm \ref{alg:OnlineSeqInference}, \(\widehat{\mathbf{U}}_1^{(t)}\) comprises the top-\(r_1\) eigenvectors of  $\widehat{\mathbf{T}}_1^{(t)}\left(\mathcal{P}_{\widehat{\mathbf{U}}_3^{(t-1)}} \otimes \mathcal{P}_{\widehat{\mathbf{U}}_2^{(t-1)}}\right) \widehat{\mathbf{T}}_1^{(t)\top}$. Consequently, \(\widehat{\mathbf{U}}_1^{(t)}\widehat{\mathbf{U}}_1^{(t)\top}\) serves as the spectral projector and is decomposed as:
\begin{align}\label{equ:def of E1}
\begin{aligned}
&\mathcal{M}_1(\widehat{\mathcal{T}}^{(t)})\left(\widehat{\mathbf{U}}_3^{(t-1)} \widehat{\mathbf{U}}_3^{(t-1)\top} \otimes \widehat{\mathbf{U}}_2^{(t-1)} \widehat{\mathbf{U}}_2^{(t-1)\top}\right) \mathcal{M}_1^{\top}(\widehat{\mathcal{T}}^{(t)})\\
=& \mathbf{T}_1^\star\left(\mathcal{P}_{\mathbf{U}_3^\star} \otimes \mathcal{P}_{\mathbf{U}_2^\star}\right) \mathbf{T}_1^{\star\top}+\mathfrak{J}_1^{(t)}+\mathfrak{J}_2^{(t)}+\mathfrak{J}_3^{(t)}+\mathfrak{J}_4^{(t)}
=: \mathbf{U}_1^\star \mathbf{G}_1^\star \mathbf{G}_1^{\star\top} \mathbf{U}_1^{\star\top}+\mathfrak{E}_1^{(t)},
\end{aligned}
\end{align}
where we use the notation $ \mathcal{M}_1\left({\mathcal{T}}^{\star}\right) = \mathbf{T}_1^\star = \mathbf{U}_1^\star \mathbf{G}_1^\star \left(\mathbf{U}_3^\star\otimes\mathbf{U}_2^\star \right)^\top$.
Here, the terms \(\mathfrak{J}_1^{(t)}\), \(\mathfrak{J}_2^{(t)}\), \(\mathfrak{J}_3^{(t)}\), and \(\mathfrak{J}_4^{(t)}\) are defined as follows: $\mathfrak{J}_1^{(t)} = \mathbf{T}_1^\star\left(\mathcal{P}_{\widehat{\mathbf{U}}_3^{(t-1)}} \otimes \mathcal{P}_{\widehat{\mathbf{U}}_2^{(t-1)}}\right) \mathbf{Z}_1^{(t)\top},
\mathfrak{J}_2^{(t)} = \rbr{\mathfrak{J}_1^{(t)}}^\top,
\mathfrak{J}_3^{(t)} = \mathbf{Z}_1^{(t)}\left(\mathcal{P}_{\widehat{\mathbf{U}}_3^{(t-1)}} \otimes \mathcal{P}_{\widehat{\mathbf{U}}_2^{(t-1)}}\right) \mathbf{Z}_1^{(t)\top},
\mathfrak{J}_4^{(t)} = \mathbf{T}_1^\star \left(\left(\mathcal{P}_{\widehat{\mathbf{U}}_3^{(t-1)}} - \mathcal{P}_{\mathbf{U}_3^\star}\right) \otimes \mathcal{P}_{\widehat{\mathbf{U}}_2^{(t-1)}}\right) \mathbf{T}_1^{\star\top} + \mathbf{T}_1^\star\left(\mathcal{P}_{\mathbf{U}_3^\star} \otimes \left(\mathcal{P}_{\widehat{\mathbf{U}}_2^{(t-1)}} - \mathcal{P}_{\mathbf{U}_2^\star}\right)\right) \mathbf{T}_1^{\star\top}.$
By definition, \(\mathbf{\Lambda}_k^2\) is a diagonal matrix consisting of the eigenvalues of \(\mathbf{G}_k^\star \mathbf{G}_k^{\star\top}\). Assuming without loss of generality that \(\mathbf{G}_k^\star \mathbf{G}_k^{\star\top} = \mathbf{\Lambda}_k^2\), which is diagonal, we deduce:
\begin{align}\label{B.2}
	\left\|\mathbf{\Lambda}_k^{-1} \mathbf{G}_k^\star\right\| = \nbr{\mathbf{\Lambda}_k^{-1} \mathbf{\Lambda}_k\mathbf{V}_k^{\star\top}} = 1, \quad \forall k \in [3]. 
\end{align}
where $\mathbf{V}_k^\star$ is the right singular space of $ \mathbf{G}_k^\star \in \mathbb{R}^{r_k \times r_{-k}}$.
\begin{lemma}\label{lem:bound for first round}
There exist absolute constants \(C_1 > 0\), for all 
$t^\alpha\rbr{\lambda_{\min}/\sigma}^2 \geq C_1 \df$,  we have $\left\|\mathfrak{J}_1^{(t)}\right\| = \left\|\mathfrak{J}_2^{(t)}\right\| = \Op{\kappa_0 \lambda_{\min} \sigma \sqrt{p/t}}, \left\|\mathfrak{J}_3^{(t)}\right\| = \Op{\sigma^2 p/t}, \left\|\mathfrak{J}_4^{(t)}\right\| =\Op{\kappa_0^2 \sigma^2 p/(t-1)}, \left\|\mathfrak{E}_1^{(t)}\right\| = \Op{\kappa_0 \lambda_{\min} \sigma \sqrt{\frac{p}{t}}}$,
and
	\begin{align*}
		\left\|\mathfrak{E}_1^{(t)} - \mathbf{T}_1^{\star}\left(\mathcal{P}_{\mathbf{U}_2^{\star}} \otimes \mathcal{P}_{\mathbf{U}_3^{\star}}\right) \cM_1^\top\rbr{\mathcal{Z}_1^{(t)}}  -\cM_1\rbr{\mathcal{Z}_1^{(t)}} \left(\mathcal{P}_{\mathbf{U}_2^{\star}} \otimes \mathcal{P}_{\mathbf{U}_3^{\star}}\right) \mathbf{T}_1^{\star \top}\right\| = \Op{\kappa_0^2\sigma^2\frac{p}{t} + \kappa_0\lambda_{\min}\sigma\sqrt{\frac{p\df}{t^{1+\alpha}}}}.
	\end{align*}
\end{lemma}
Proof in Section \ref{sec:Proof of Lemma bound for first round}.
For a positive integer \( n \), define $\mathfrak{P}_k^{-n} = \mathbf{U}_k^\star \mathbf{\Lambda}_k^{-2 n} \mathbf{U}_k^{\star\top}.$
By a slight abuse of notation, let
$ \mathfrak{P}_k^0 = \mathfrak{P}_k^{\perp} = \mathcal{P}_{\mathbf{U}_k^\star}^{\perp}. $
Note that, when $t^\alpha\rbr{\lambda_{\min}/\sigma}^2 > C_1\kappa_0^2\df$ for some constant $C_1$, we have $\left\|\mathfrak{E}_1^{(t)}\right\| < \frac{\lambda_{\text{min}}^2}{2}$ with high probability, which implies that the condition of Theorem 1 in \cite{xia2021normal} is satisfied.
\begin{lemma}[Theorem 1 in \cite{xia2021normal}] \label{lem:theorem 1 in xia2021normal}
If $\left\|\mathfrak{E}_1^{(t)}\right\| \leq \frac{\lambda_{\min }^2}{2}$, the following equation holds
\begin{align}\label{B.3}
\mathcal{P}_{\widehat{\mathbf{U}}_1^{(t)}}-\mathcal{P}_{\mathbf{U}_1^\star}=\sum_{n \geq 1} \mathcal{S}_{\mathbf{G}_1, n}\left(\mathfrak{E}_1^{(t)}\right),
\end{align}
where for each positive integer $n$, $\mathcal{S}_{\mathbf{G}_1, n}\left(\mathfrak{E}_1^{(t)} \right)=\sum_{s_1+\cdots+s_{n+1}=n}(-1)^{1+\tau(\mathbf{s})} \cdot \mathfrak{P}_1^{-s_1} \mathfrak{E}_1^{(t)} \mathfrak{P}_1^{-s_2} \ldots \mathfrak{P}_1^{-s_n} \mathfrak{E}_1^{(t)} \mathfrak{P}_1^{-s_{n+1}},$
where $s_1, \cdots, s_{n+1}$ are non-negative integers and $\tau(\mathbf{s})=\sum_{j=1}^{n+1} \mathbf{I}\left(s_j>0\right)$.
\end{lemma}
Having presented all the basic lemmas relevant to the proof of Theorem \ref{thm:entry inference}, the following subsection will begin with the decomposition of \( \hat{h}^{(t)}-h^\star \).

\subsection{Step 1: decomposing $\hat{h}^{(t)}-h^\star$}
\label{sec:proof of theorem entry step1}
By Equation (\ref{equ:mt in proof}), we have
\begin{align}\label{equ:decomposition of h}
\begin{aligned}
    \hat{h}^{(t)} - h^\star
=&\left\langle\widehat{\mathcal{T}}^{(t)}\times_{k\in[3]}\mathcal{P}_{\widehat{\mathbf{U}}_k^{(t)}} - \mathcal{T}^\star, \mathcal{H}\right\rangle\\
\stackrel{(\ref{equ:t=t+z})}{=} &\left\langle \rbr{{\mathcal{T}^\star} + {\mathcal{Z}}^{(t)}  } \times_{k\in[3]}\mathcal{P}_{\widehat{\mathbf{U}}_k^{(t)}} - \mathcal{T}^\star , \mathcal{H}\right\rangle\\
= & \left\langle{\mathcal{T}^\star}\times_{k\in[3]}\mathcal{P}_{\widehat{\mathbf{U}}_k^{(t)}} -\mathcal{T}^\star + {\mathcal{Z}}^{(t)}\times_{k\in[3]}\mathcal{P}_{\widehat{\mathbf{U}}_k^{(t)}},    \mathcal{H}\right\rangle 
\end{aligned}
\end{align}
For the first term on the right-hand side of Equation (\ref{equ:decomposition of h}), we can proceed to expand it further:
\begin{align*}
\mathcal{T}^\star \times_{k\in[3]}\mathcal{P}_{\widehat{\mathbf{U}}_k^{(t)}} \;-\; \mathcal{T}^\star
             = &\; \mathcal{T}^\star 
        \times_1\Bigl(\mathcal{P}_{\widehat{\mathbf{U}}_1^{(t)}}
                         - \mathcal{P}_{\mathbf{U}_1^\star}\Bigr) 
        \times_{2}\Bigl(\mathcal{P}_{\widehat{\mathbf{U}}_{2}^{(t)}}
                         - \mathcal{P}_{\mathbf{U}_{2}^\star}\Bigr) 
        \times_{3}\Bigl(\mathcal{P}_{\widehat{\mathbf{U}}_{3}^{(t)}}
                         - \mathcal{P}_{\mathbf{U}_{3}^\star}\Bigr) \\
& + \sum_{j=1}^3 \mathcal{T}^\star 
        \times_j\Bigl(\mathcal{P}_{\widehat{\mathbf{U}}_j^{(t)}}
                         - \mathcal{P}_{\mathbf{U}_j^\star}\Bigr) 
        \times_{j+1}\Bigl(\mathcal{P}_{\widehat{\mathbf{U}}_{j+1}^{(t)}}
                         - \mathcal{P}_{\mathbf{U}_{j+1}^\star}\Bigr) 
        \times_{j+2}\mathcal{P}_{\mathbf{U}_{j+2}^\star} \\
        & + \underbrace{\sum_{j=1}^3 \mathcal{T}^\star 
                \times_j \Bigl(\mathcal{P}_{\widehat{\mathbf{U}}_j^{(t)}}
                         - \mathcal{P}_{\mathbf{U}_j^\star}\Bigr) 
        \times_{j+1} \mathcal{P}_{\mathbf{U}_{j+1}^\star} 
        \times_{j+2} \mathcal{P}_{\mathbf{U}_{j+2}^\star}}_{\mathcal{O}_1}.
\end{align*}
and for the second term on the right-hand side of Equation (\ref{equ:decomposition of h}), we have:
\begin{align*}
\mathcal{Z}^{(t)}
        \times_{k\in[3]}\mathcal{P}_{\widehat{\mathbf{U}}_k^{(t)}}
         = & {\mathcal{Z}}^{(t)} 
        \times_1\Bigl(\mathcal{P}_{\widehat{\mathbf{U}}_1^{(t)}}
                         - \mathcal{P}_{\mathbf{U}_1^\star}\Bigr)
        \times_{2}\Bigl(\mathcal{P}_{\widehat{\mathbf{U}}_{2}^{(t)}}
                         - \mathcal{P}_{\mathbf{U}_{2}^\star}\Bigr)
        \times_{3}\Bigl(\mathcal{P}_{\widehat{\mathbf{U}}_{3}^{(t)}}
                         - \mathcal{P}_{\mathbf{U}_{3}^\star}\Bigr) \\
& + \sum_{j=1}^3 {\mathcal{Z}}^{(t)} 
        \times_j\Bigl(\mathcal{P}_{\widehat{\mathbf{U}}_j^{(t)}}
                         - \mathcal{P}_{\mathbf{U}_j^\star}\Bigr)
        \times_{j+1}\Bigl(\mathcal{P}_{\widehat{\mathbf{U}}_{j+1}^{(t)}}
                         - \mathcal{P}_{\mathbf{U}_{j+1}^\star}\Bigr)
        \times_{j+2}\,\mathcal{P}_{\mathbf{U}_{j+2}^\star} \\
& + \sum_{j=1}^3 {\mathcal{Z}}^{(t)} 
        \times_j\Bigl(\mathcal{P}_{\widehat{\mathbf{U}}_j^{(t)}}
                         - \mathcal{P}_{\mathbf{U}_j^\star}\Bigr)
        \times_{j+1}\,\mathcal{P}_{\mathbf{U}_{j+1}^\star}
        \times_{j+2}\,\mathcal{P}_{\mathbf{U}_{j+2}^\star} \\
& + \mathcal{Z}^{(t)} 
        \times_{k\in[3]}\,\mathcal{P}_{\mathbf{U}_k^\star}.
\end{align*}
We present the following lemma to provide an upper bound for these two terms:
\begin{lemma}\label{lem:bound of O1} Under Lemma \ref{lem:bound for inference},\ref{lem:bound for first round},\ref{lem:theorem 1 in xia2021normal}, when $t^\alpha \rbr{\lambda_{\min}/\sigma}^2\geq Cp^2$, we have
   \begin{align*}
& \inner{\mathcal{T}^\star \times_{k\in[3]}\mathcal{P}_{\widehat{\mathbf{U}}_k^{(t)}} \;-\; \mathcal{T}^\star - \cO_1, \cH} =    \Op{ \kappa_0 \frac{\sigma^2}{\lambda_{\min}^{}} \sqrt{\frac{r\log p}{t^2}} \rbr{\fbr{\cH\times  \Pu{3}}+ \frac{\sigma}{\lambda_{\min}}\sqrt{\frac{1}{t}}\fbr{\cH } } } \\
& \inner{\mathcal{Z}^{(t)}\times_{k\in[3]}\mathcal{P}_{\widehat{\mathbf{U}}_k^{(t)}} - \mathcal{Z}^{(t)} \times_{k\in[3]}\,\mathcal{P}_{\mathbf{U}_k^\star},    \mathcal{H}} =  \Op{\frac{\sigma^2}{\lambda_{\min}}\sqrt{\frac{p^2r}{t^2}}\left(  L_t^2\left\|\mathcal{H}\right\|_\mathrm{F} + L_t\sum_{k=1}^3\left\|\mathcal{H} \times_k \mathbf{U}_{k}^\star\right\|_{\mathrm{F}} +  \sum_{k=1}^3\left\|\mathcal{H} \times_{j\neq k} \mathbf{U}_{j}^\star \right\|_\mathrm{F} \right)}.
   \end{align*}
\end{lemma}
Proof in Section \ref{sec:proof of lemma bound of O1}.  

\begin{lemma}\label{lem:bound of O3}
	\begin{align*}
\begin{aligned}
        \cO_1
    =& \sum_{k\in[3]}\left\langle \mathbf{Z}_k^{(t)}, \mathcal{P}_{\mathbf{U}_k^\star}^{\perp}\mathbf{H}_k \mathcal{P}_{\left(\mathbf{U}_{k+2}^\star\otimes\mathbf{U}_{k+1}^\star\right)  \mathbf{V}_{k}^\star} \right\rangle  + \Op{\kappa_0^3 \frac{\sigma^2}{\lambda_{\min }}\sqrt{\frac{p^2r}{t^2}} \left\|\mathbf{H}_1 
   \left(\mathbf{U}_3^\star \otimes\mathbf{U}_2^\star \right)\right\|_\mathrm{F}}.
\end{aligned}
\end{align*}
where $\mathcal{P}_{\left(\mathbf{U}_{k+2}^\star\otimes\mathbf{U}_{k+1}^\star\right)  \mathbf{V}_k^\star} = \left(\mathbf{U}_{k+2}^\star\otimes\mathbf{U}_{k+1}^\star\right)  \mathbf{V}_k^\star\mathbf{V}_k^{\star\top} \left(\mathbf{U}_{k+1}^\star \otimes \mathbf{U}_{k+2}^\star\right)^\top$ and  $\mathbf{V}_k^\star$ is the right singular space of $\mathcal{M}_k\left(\mathcal{G}\right) \in \mathbb{R}^{r_k \times r_{-k}}$.
\end{lemma}
Proof in Section \ref{sec:proof of lemma bound of O3}.
For Equation (\ref{equ:decomposition of h}), with above two lemmas, we can establish the following:
\begin{align}\label{equ:m-1}
\begin{aligned}
      	\hat{h}^{(t)}-h^\star  = & \left\langle {\mathcal{Z}}^{(t)} \times_{k\in[3]} \mathcal{P}_{\mathbf{U}_k^\star},
        \mathcal{H} \right\rangle + \sum_{k=1}^3   \left\langle \mathbf{Z}_k^{(t)}, \mathcal{P}_{\mathbf{U}_k^\star}^{\perp}\mathbf{H}_k \mathcal{P}_{\left(\mathbf{U}_{k+2}^\star\otimes\mathbf{U}_{k+1}^\star\right)  \mathbf{V}_k^\star} \right\rangle  \\
             & + O_\bP\left( \frac{\sigma^2}{\lambda_{\min}}\sqrt{\frac{p^2r}{t^2}}\left(  L_t^2\left\|\mathcal{H}\right\|_\mathrm{F} + L_t\sum_{k=1}^3\left\|\mathcal{H} \times_k \mathbf{U}_{k}^\star\right\|_{\mathrm{F}} + \kappa_0^3 \sum_{k=1}^3\left\|\mathcal{H} \times_{j\neq k} \mathbf{U}_{j}^\star \right\|_\mathrm{F} \right) \right.
             \\ &\quad\quad\quad\quad + \left.\kappa_0 \frac{\sigma^2}{\lambda_{\min}^{}} \sqrt{\frac{r\log p}{t^2}} \rbr{\sum_{k=1}^3\left\|\mathcal{H} \times_k \mathbf{U}_{k}^\star\right\|_{\mathrm{F}}+ \frac{\sigma}{\lambda_{\min}}\sqrt{\frac{1}{t}}\fbr{\cH } } \right).
\end{aligned}
\end{align}

\subsection{Step 3: characterizing the distribution of leading terms}
\label{sec:proof of theorem entry step3}
Our objective is to characterize the distribution of the first term on the right-hand side of Equation (\ref{equ:m-1}). Subsequently, we will demonstrate that the variance of this leading term exceeds that of the second term on the right-hand side of Equation (\ref{equ:m-1}).
\begin{lemma}\label{lem:clt}
 Under the Assumptions of Theorem \ref{thm:entry inference}, as $t, p \rightarrow \infty$, we have
\begin{align*}
\frac{\sqrt{t}}{\sigma S_\mathcal{H}} \left( \left\langle {\mathcal{Z}}^{(t)} 
        \times_{k\in[3]} \mathcal{P}_{\mathbf{U}_k^\star}, \mathcal{H} \right\rangle + \sum_{k=1}^3   \left\langle \mathbf{Z}_k^{(t)}, \mathcal{P}_{\mathbf{U}_k^\star}^{\perp}\mathbf{H}_k \mathcal{P}_{\left(\mathbf{U}_{k+2}^\star\otimes\mathbf{U}_{k+1}^\star\right)  \mathbf{V}_k^\star} \right\rangle \right) \stackrel{d}{\longrightarrow} \mathcal{N}\left(0, 1\right),
\end{align*}
where $ S_\mathcal{H}^2 = \left\|\mathcal{H} \times_{k\in[3]} \mathbf{U}_k^\star \right\|_{\mathrm{F}}^2 
+  \sum_{k=1}^3 \left\| \mathcal{P}_{\mathbf{U}_k^\star}^{\perp}\mathbf{H}_k \mathcal{P}_{\left(\mathbf{U}_{k+1}^\star\otimes\mathbf{U}_{k+2}^\star\right)  \mathbf{V}_k^\star}  \right\|_\mathrm{F} ^2 $.
\end{lemma}
Proof in Section \ref{sec:proof of lemma clt}.
In the following, we will show that the negligible terms are bounded and converge to 0 asymptotically.
Meanwhile, due to Equation (\ref{equ:m-1}), and Assumption \ref{thm:entry inference}, one can see that as $t, p \rightarrow \infty$
\begin{align}\label{equ:S and Stlide}
\begin{aligned}
    &\frac{\sqrt{t}}{\sigma S_\mathcal{H}}\left|	\hat{h}^{(t)}-h^\star - \left\langle {\mathcal{Z}}^{(t)} 
       \times_{k\in[3]} \mathcal{P}_{\mathbf{U}_k^\star}, \mathcal{H} \right\rangle - \sum_{k=1}^3   \left\langle \mathbf{Z}_k^{(t)}, \mathcal{P}_{\mathbf{U}_k^\star}^{\perp}\mathbf{H}_k \mathcal{P}_{\left(\mathbf{U}_{k+2}^\star\otimes\mathbf{U}_{k+1}^\star\right)  \mathbf{V}_k^\star} \right\rangle \right| \\
= &  O_\bP\left( \frac{\sigma}{\lambda_{\min} S_\mathcal{H} } \sqrt{\frac{p^2r}{t}}\left(  L_t^2\left\|\mathcal{H}\right\|_\mathrm{F} + L_t\sum_{k=1}^3\left\|\mathcal{H} \times_k \mathbf{U}_{k}^\star\right\|_{\mathrm{F}} + \kappa_0^3 \sum_{k=1}^3\left\|\mathcal{H} \times_{j\neq k} \mathbf{U}_{j}^\star \right\|_\mathrm{F} \right) \right.
             \\ &\quad\quad\quad\quad + \left.\kappa_0 \frac{\sigma}{\lambda_{\min}S_\mathcal{H}} \sqrt{\frac{r\log p}{t}} \rbr{\sum_{k=1}^3\left\|\mathcal{H} \times_k \mathbf{U}_{k}^\star\right\|_{\mathrm{F}}+ \frac{\sigma}{\lambda_{\min}}\sqrt{\frac{1}{t}}\fbr{\cH } } \right)\\
             = & o_\bP\rbr{1}.
\end{aligned}
\end{align}
\xin{Under Assumption \ref{cond:4}, we have $\frac{\fbr{\cH}}{S_\mathcal{H}}\leq p, \frac{\fbr{\cH \times_k \mathbf{U}_{k}^\star}}{S_\mathcal{H}}\leq\sqrt{p}$.}
Finally, if we combine the above equation and  Equation (\ref{equ:m-1}), we show that the variance of the main CLT term dominates the other terms. Now we have proved that,  when $t,p\rightarrow\infty$,
\begin{align*}
\frac{\sqrt{t}\left(\hat{h}^{(t)}-h^\star\right)}{\sigma S} \stackrel{d}{\longrightarrow} \mathcal{N}(0,1).
\end{align*}
Thus, we conclude the proof of Theorem \ref{thm:entry inference}.

%% file: body/ProofEntryConsis.tex
\section{Proof of Theorem \ref{thm:entry inference consistent}}
\label{sec:Proof of Theorem entry inference consistent}
Our idea is to prove the consistency of the plug-in estimator by showing the $\widehat{S}^2_t$ and $\widehat{\sigma}^2_t$ converge in probability to $S^2$ and $\sigma^2$ respectively. Again, in this proof, we use a generic index $t$ to prove the consistency. Theorem holds when we set $t=n$. We start by showing the consistency of the $\widehat{S}_t^2$.

\subsection{consistency of $\widehat{S}^2_t$}
\begin{align*}
S_\mathcal{H}^2 = \left\|\mathcal{H} \bigtimes_{k\in[3]} \UT{k} \right\|_{\mathrm{F}}^2 +  \sum_{k=1}^3 \left\| \mathcal{P}_{\mathbf{U}_k^\star}^{\perp}\mathbf{H}_k \mathcal{P}_{\left(\mathbf{U}_{k+2}^\star\otimes\mathbf{U}_{k+1}^\star\right)  \mathbf{V}_k^\star}  \right\|_\mathrm{F} ^2 .
\end{align*}
and
\begin{align*}
\widehat{S}_{\mathcal{H},t}^{2} = 
\left\|
\mathcal{H} \bigtimes_{k\in[3]} {\widehat{\mathbf{U}}_k^{(t)\top}} 
\right\|_{\mathrm{F}}^2 +  \sum_{k=1}^3 \left\| \mathcal{P}_{\widehat{\mathbf{U}}_k^{(t)}}^{\perp}\,\mathbf{H}_k\,
    \mathcal{P}_{\bigl(\widehat{\mathbf{U}}_{k+2}^{(t)}\otimes \widehat{\mathbf{U}}_{k+1}^{(t)}\bigr)\,\widehat{\mathbf{V}}_k^{(t)}}
  \right\|_{\text{F}}^{2} .
\end{align*}

Since we have
\begin{align*}
\begin{aligned}
& \mathcal{H} \bigtimes_{k\in[3]} {\widehat{\mathbf{U}}_k^{(t)\top}}  
- \mathcal{H} \bigtimes_{k\in[3]} \UT{k}  \\
= & \mathcal{H} \times{ }_1\left({\widehat{\mathbf{U}}_1^{(t)}}-{\mathbf{U}_1^\star}\right) \times_2\left({\widehat{\mathbf{U}}_2^{(t)}}-{\mathbf{U}_{2}^\star}\right) \times_3\left({\widehat{\mathbf{U}}_3^{(t)}}-{\mathbf{U}_3^\star}\right) \\
& +\sum_{k=1}^3 \mathcal{H} \times_k\left({\widehat{\mathbf{U}}_k^{(t)}}-{\mathbf{U}_k^\star}\right) \times_{k+1}\left({\widehat{\mathbf{U}}_{k+1}^{(t)}} - {\mathbf{U}_{k+1}^\star}\right) \times_{k+2} {\mathbf{U}_{k+2}^\star} \\
& +\sum_{k=1}^3 \mathcal{H} \times_k\left({\widehat{\mathbf{U}}_k^{(t)}}-{\mathbf{U}_k^\star}\right) \times_{k+1} {\mathbf{U}_{k+1}^\star} \times_{k+2} {\mathbf{U}_{k+2}^\star},
\end{aligned}
\end{align*}
thus, for the first part of $\widehat{S}_\mathcal{H}^2$, we have
\begin{align*}
   & \left|\left\|\mathcal{H} \bigtimes_{k\in[3]} {\widehat{\mathbf{U}}_k^{(t)\top}} \right\|_{\text{F}}^2 - \left\|\mathcal{H} \bigtimes_{k\in[3]} \UT{k} \right\|_{\text{F}}^2\right|\\
  = & \Op{L_t^6 \left\|\mathcal{H} \right\|_{\text{F}} ^2  + L_t^4 \sum_{k=1}^3 \left\|\mathcal{H}\times_k  {\mathbf{U}_k^\star}\right\|_{\text{F}}^2  + L_t^2 \sum_{k=1}^3 \left\|\mathcal{H} \bigtimes_{j\neq k}  {\mathbf{U}_{j}^\star} \right\|_{\text{F}}^2}.
\end{align*}
The above inequality is due to  triangle inequality.

For the second term:
\begin{align*}
& \left|\,
  \bigl\| \mathcal{P}_{\widehat{\mathbf{U}}_k^{(t)}}^{\perp}\,\mathbf{H}_k\,
    \mathcal{P}_{\bigl(\widehat{\mathbf{U}}_{k+2}^{(t)} \otimes \widehat{\mathbf{U}}_{k+1}^{(t)}\bigr)\,\widehat{\mathbf{V}}_k^{(t)}}
  \bigr\|_{\text{F}}^{2}
  \;-\;
  \bigl\|\mathcal{P}_{\mathbf{U}_k^\star}^{\perp}\,\mathbf{H}_k\,
    \mathcal{P}_{\bigl(\mathbf{U}_{k+2}^\star\otimes \mathbf{U}_{k+1}^\star\bigr)\,\mathbf{V}_k^\star}
  \bigr\|_{\text{F}}^{2}\right|
\\[4pt]
\le\;&
  \bigl\|\mathcal{P}_{\mathbf{U}_k^\star}^{\perp}\,\mathbf{H}_k\,
    \bigl(\,\mathcal{P}_{\bigl(\widehat{\mathbf{U}}_{k+1}^{(t)}\otimes \widehat{\mathbf{U}}_{k+2}^{(t)}\bigr)\,\widehat{\mathbf{V}}_k^{(t)}} 
    \;-\;
    \mathcal{P}_{\bigl(\mathbf{U}_{k+2}^\star\otimes \mathbf{U}_{k+1}^\star\bigr)\,\mathbf{V}_k^\star}
    \bigr)
  \bigr\|_{\text{F}}^{2}
\\[4pt]
&\;+\;
  \bigl\|\bigl(\mathcal{P}_{\widehat{\mathbf{U}}_k^{(t)}}^{\perp} 
    \;-\; 
    \mathcal{P}_{\mathbf{U}_k^\star}^{\perp}\bigr)\,\mathbf{H}_k\,
    \mathcal{P}_{\bigl(\mathbf{U}_{k+2}^\star\otimes \mathbf{U}_{k+1}^\star\bigr)\,\mathbf{V}_k^\star}
  \bigr\|_{\text{F}}^{2}
\\[4pt]
&\;+\;
  \bigl\|\bigl(\mathcal{P}_{\widehat{\mathbf{U}}_k^{(t)}}^{\perp} 
    \;-\; 
    \mathcal{P}_{\mathbf{U}_k^\star}^{\perp}\bigr)\,\mathbf{H}_k\,
    \bigl(\,\mathcal{P}_{\bigl(\widehat{\mathbf{U}}_{k+1}^{(t)}\otimes \widehat{\mathbf{U}}_{k+2}^{(t)}\bigr)\,\widehat{\mathbf{V}}_k^{(t)}}
    \;-\;
    \mathcal{P}_{\bigl(\mathbf{U}_{k+2}^\star\otimes \mathbf{U}_{k+1}^\star\bigr)\,\mathbf{V}_k^\star}\bigr)
  \bigr\|_{\text{F}}^{2}.
\end{align*}
For the first term, we have
\begin{align*}
& \Bigl\| 
     \mathcal{P}_{\Bigl(\widehat{\mathbf{U}}_{k+2}^{(t)}\otimes\widehat{\mathbf{U}}_{k+1}^{(t)}\Bigr)\,\widehat{\mathbf{V}}_k^{(t)}} 
    \;-\;
    \mathcal{P}_{\bigl(\mathbf{U}_{k+2}^\star\otimes \mathbf{U}_{k+1}^\star\bigr)\,\mathbf{V}_k^\star}
    \Bigr\| \\
  =\; & \Bigl\|  
        \bigl(\widehat{\mathbf{U}}_{k+2}^{(t)} \otimes \widehat{\mathbf{U}}_{k+1}^{(t)}\bigr) 
         \,\widehat{\mathbf{V}}_k^{(t)}\widehat{\mathbf{V}}_k^{(t)\top} 
         \bigl(\widehat{\mathbf{U}}_{k+2}^{(t)} \otimes \widehat{\mathbf{U}}_{k+1}^{(t)}\bigr)^\top 
       \;-\; 
       \Bigl(\mathbf{U}_{k+1}^\star\otimes\mathbf{U}_{k+2}^\star\Bigr) 
         \,\mathbf{V}_k^\star\mathbf{V}_k^{\star\top} 
         \Bigl(\mathbf{U}_{k+1}^\star \otimes \mathbf{U}_{k+2}^\star\Bigr)^\top 
      \Bigr\| \\
  \leq\; & 
      \frac{2 \Bigl\|\Bigl(\mathcal{P}_{\widehat{\mathbf{U}}_{k+1}^{(t)}} \otimes 
             \mathcal{P}_{\widehat{\mathbf{U}}_{k+2}^{(t)}}\Bigr) 
             \widehat{\mathbf{T}}_k^{(t)\top} 
             \mathcal{P}_{\widehat{\mathbf{U}}_k^{(t)}}  
             \;-\; 
             \mathbf{T}_k^\top\Bigr\|}
           {\lambda_{\min}}.
\end{align*}

Since
\begin{align*}
& \Bigl(\mathcal{P}_{\widehat{\mathbf{U}}_{k+2}^{(t)}} \otimes \mathcal{P}_{\widehat{\mathbf{U}}_{k+1}^{(t)}}\Bigr) 
             \widehat{\mathbf{T}}_k^{(t)\top} 
             \mathcal{P}_{\widehat{\mathbf{U}}_k^{(t)}}  
             \;-\; 
             \mathbf{T}_k^{\star\top} \\
=  &\; \widehat{\mathcal{T}}^{(t)} \times_k \mathcal{P}_{\widehat{\mathbf{U}}_k^{(t)}}
                     \times_{k+1} \mathcal{P}_{\widehat{\mathbf{U}}_{k+1}^{(t)}}
                     \times_{k+2} \mathcal{P}_{\widehat{\mathbf{U}}_{k+2}^{(t)}}
             \;-\; 
             \mathcal{T}^\star \\
=  &\; \Bigl(\mathcal{T}^\star + \widehat{\mathcal{Z}}^{(t)}\Bigr) 
        \times_k \mathcal{P}_{\widehat{\mathbf{U}}_k^{(t)}}
        \times_{k+1} \mathcal{P}_{\widehat{\mathbf{U}}_{k+1}^{(t)}}
        \times_{k+2} \mathcal{P}_{\widehat{\mathbf{U}}_{k+2}^{(t)}}
             \;-\; 
             \mathcal{T}^\star \\
= &\; \mathcal{T}^\star 
        \times_k \Bigl(\mathcal{P}_{\widehat{\mathbf{U}}_k^{(t)}}
                         - \mathcal{P}_{\mathbf{U}_k^\star}\Bigr) 
        \times_{k+1}\Bigl(\mathcal{P}_{\widehat{\mathbf{U}}_{k+1}^{(t)}}
                         - \mathcal{P}_{\mathbf{U}_{k+1}^\star}\Bigr) 
        \times_{k+2}\Bigl(\mathcal{P}_{\widehat{\mathbf{U}}_{k+2}^{(t)}}
                         - \mathcal{P}_{\mathbf{U}_{k+2}^\star}\Bigr) \\
& + \sum_{j=1}^3 \mathcal{T}^\star 
        \times_j\Bigl(\mathcal{P}_{\widehat{\mathbf{U}}_j^{(t)}}
                         - \mathcal{P}_{\mathbf{U}_j^\star}\Bigr) 
        \times_{j+1}\Bigl(\mathcal{P}_{\widehat{\mathbf{U}}_{j+1}^{(t)}}
                         - \mathcal{P}_{\mathbf{U}_{j+1}^\star}\Bigr) 
        \times_{j+2}\mathcal{P}_{\mathbf{U}_{j+2}^\star} \\
& + \sum_{j=1}^3 \mathcal{T}^\star 
        \times_j\Bigl(\mathcal{P}_{\widehat{\mathbf{U}}_j^{(t)}}
                         - \mathcal{P}_{\mathbf{U}_j^\star}\Bigr) 
        \times_{j+1} \mathcal{P}_{\mathbf{U}_{j+1}^\star} 
        \times_{j+2} \mathcal{P}_{\mathbf{U}_{j+2}^\star} \\
& +  {\mathcal{Z}}^{(t)} 
        \times_k\Bigl(\mathcal{P}_{\widehat{\mathbf{U}}_k^{(t)}}
                         - \mathcal{P}_{\mathbf{U}_k^\star}\Bigr)
        \times_{k+1}\Bigl(\mathcal{P}_{\widehat{\mathbf{U}}_{k+1}^{(t)}}
                         - \mathcal{P}_{\mathbf{U}_{k+1}^\star}\Bigr)
        \times_{k+2}\Bigl(\mathcal{P}_{\widehat{\mathbf{U}}_{k+2}^{(t)}}
                         - \mathcal{P}_{\mathbf{U}_{k+2}^\star}\Bigr) \\
& + \sum_{j=1}^3 {\mathcal{Z}}^{(t)} 
        \times_j\Bigl(\mathcal{P}_{\widehat{\mathbf{U}}_j^{(t)}}
                         - \mathcal{P}_{\mathbf{U}_j^\star}\Bigr)
        \times_{j+1}\Bigl(\mathcal{P}_{\widehat{\mathbf{U}}_{j+1}^{(t)}}
                         - \mathcal{P}_{\mathbf{U}_{j+1}^\star}\Bigr)
        \times_{j+2}\,\mathcal{P}_{\mathbf{U}_{j+2}^\star} \\
& + \sum_{j=1}^3 {\mathcal{Z}}^{(t)} 
        \times_j\Bigl(\mathcal{P}_{\widehat{\mathbf{U}}_j^{(t)}}
                         - \mathcal{P}_{\mathbf{U}_j^\star}\Bigr)
        \times_{j+1}\,\mathcal{P}_{\mathbf{U}_{j+1}^\star}
        \times_{j+2}\,\mathcal{P}_{\mathbf{U}_{j+2}^\star} \\
& + {\mathcal{Z}}^{(t)} 
        \times_1\,\mathcal{P}_{\mathbf{U}_1^\star}
        \times_2\,\mathcal{P}_{\mathbf{U}_2^\star}
        \times_3\,\mathcal{P}_{\mathbf{U}_3^\star},
\end{align*}
thus
\begin{align*}
  &  \Bigl\|\Bigl(\mathcal{P}_{\widehat{\mathbf{U}}_{k+1}^{(t)}} \otimes 
             \mathcal{P}_{\widehat{\mathbf{U}}_{k+2}^{(t)}}\Bigr) 
             \widehat{\mathbf{T}}_k^{(t)\top} 
             \mathcal{P}_{\widehat{\mathbf{U}}_k^{(t)}}  
             \;-\; 
             \mathbf{T}_k^\top\Bigr\| = \Op{\kappa_0\sigma\sqrt{\frac{p}{t}}  + \frac{\sigma^2}{\lambda_{\min}}\sqrt{\frac{p^2r}{t^2}} + \sigma\sqrt{\frac{r^2}{t}}}.   
\end{align*}
and
\begin{align*}
	\Bigl\| 
     \mathcal{P}_{\Bigl(\widehat{\mathbf{U}}_{k+2}^{(t)}\otimes\widehat{\mathbf{U}}_{k+1}^{(t)}\Bigr)\,\widehat{\mathbf{V}}_k^{(t)}} 
    \;-\;
    \mathcal{P}_{\bigl(\mathbf{U}_{k+2}^\star\otimes \mathbf{U}_{k+1}^\star\bigr)\,\mathbf{V}_k^\star}
    \Bigr\|
    = \Op{\kappa_0 \frac{\sigma}{\lambda_{\min}}\sqrt{\frac{p}{t}}  + \frac{\sigma^2}{\lambda_{\min}^2}\sqrt{\frac{p^2r}{t^2}}}
\end{align*}
In addition, for the terms involving \(\mathcal{P}_{\mathbf{U}_k^\star}^{\perp}\), we can establish the following property:
\begin{align*}
\begin{aligned}
&\left\| \mathcal{P}_{\widehat{\mathbf{U}}_k^{(t)}}^{\perp}- \mathcal{P}_{\mathbf{U}_k^\star}^{\perp}\right\| = \left\| \widehat{\mathbf{U}}_{k\perp}^{(t)}\widehat{\mathbf{U}}_{k\perp}^{(t)\top} -\mathbf{U}_{k\perp}^\star \mathbf{U}_{k\perp}^{\star\top}\right\|  \\
= & \left\|\mathbf{U}_{k\perp}^\star \mathbf{U}_{k\perp}^{\star\top} + \mathbf{U}_{k}^\star \mathbf{U}_{k}^{\star\top} +\widehat{\mathbf{U}}_{k}^{(t)}\widehat{\mathbf{U}}_{k}^{(t)\top} - \mathbf{U}_{k}^\star \mathbf{U}_{k}^{\star\top} - \left(\widehat{\mathbf{U}}_{k}^{(t)}\widehat{\mathbf{U}}_{k}^{(t)\top}  + \widehat{\mathbf{U}}_{k\perp}^{(t)}\widehat{\mathbf{U}}_{k\perp}^{(t)\top} \right)\right\|\\
= & \left\|\mathbf{I}_{r_k}+\widehat{\mathbf{U}}_{k}^{(t)}\widehat{\mathbf{U}}_{k}^{(t)\top} - \mathbf{U}_{k}^\star \mathbf{U}_{k}^{\star\top} - \mathbf{I}_{r_k}\right\|
=\left\| \widehat{\mathbf{U}}_{k}^{(t)}\widehat{\mathbf{U}}_{k}^{(t)\top} -{\mathbf{U}}_{k}^\star {\mathbf{U}}_{k}^{\star\top}\right\|\\
=& \left\| \mathcal{P}_{\widehat{\mathbf{U}}_k^{(t)}}- \mathcal{P}_{\mathbf{U}_k^\star}\right\| .
\end{aligned}
\end{align*}

Combine previous three, we can get:
    \begin{align*}
        & \left|\,
  \bigl\|\mathcal{P}_{\widehat{\mathbf{U}}_k^{(t)}}^{\perp}\,\mathbf{H}_k\,
    \mathcal{P}_{\bigl(\widehat{\mathbf{U}}_{k+1}^{(t)}\otimes \widehat{\mathbf{U}}_{k+2}^{(t)}\bigr)\,\widehat{\mathbf{V}}_k^{(t)}}
  \bigr\|_{\text{F}}^{2}
  \;-\;
  \bigl\|\mathcal{P}_{\mathbf{U}_k^\star}^{\perp}\,\mathbf{H}_k\,
    \mathcal{P}_{\bigl(\mathbf{U}_{k+2}^\star\otimes \mathbf{U}_{k+1}^\star\bigr)\,\mathbf{V}_k^\star}
  \bigr\|_{\text{F}}^{2}\right|
\\[4pt]
= & O_\bP \left[ \left(\kappa_0^2\frac{\sigma^2}{\lambda_{\min}^2}\frac{p}{t} + \frac{\sigma^4}{\lambda_{\min}^4}\frac{p^2r}{t^2}\right) \left\|\mathcal{H}\right\|_{\text{F}}^2
+ \frac{\sigma^2}{\lambda_{\min}^2}\frac{p}{t}\left\|\mathbf{H}_k\left( \mathbf{U}_{k+1} ^{\star} \otimes\mathbf{U}_{k+2} ^{\star}\right)  \right\|_\mathrm{F}^2 \right].
    \end{align*}
    
    Thus we have
    \begin{align*}
    	\left|\widehat{S}_{\mathcal{H},t}^{2} - S_\mathcal{H}^2 \right| = & O_\bP \left( 
    	 L_t^6 \left\|\mathcal{H} \right\|_{\text{F}} ^2  + L_t^4 \sum_{k=1}^3 \left\|\mathcal{H}\times_k  {\mathbf{U}_k^\star}\right\|_{\text{F}}^2  + L_t^2 \sum_{k=1}^3 \left\|\mathcal{H} \bigtimes_{j\neq k}  {\mathbf{U}_{j}^\star} \right\|_{\text{F}}^2\right.
    	\\
    	& +\left. \left(\kappa_0^2\frac{\sigma^2}{\lambda_{\min}^2}\frac{p}{t} + \frac{\sigma^4}{\lambda_{\min}^4}\frac{p^2r}{t^2}\right) \left\|\mathcal{H}\right\|_{\text{F}}^2
+ \frac{\sigma^2}{\lambda_{\min}^2}\frac{p}{t}\sum_{k\in[3]}\left\|\mathcal{H} \bigtimes_{j\neq k}  {\mathbf{U}_{j}^\star}  \right\|_\mathrm{F}^2 \right) \\
= & \Op{\kappa_0^2\frac{\sigma^2}{\lambda_{\min}^2}\frac{p}{t} \left\|\mathcal{H}\right\|_{\text{F}}^2
 + \frac{\sigma^4}{\lambda_{\min}^4} \frac{p^2}{t^2}\sum_{k=1}^3 \left\|\mathcal{H}\times_k  {\mathbf{U}_k^\star}\right\|_{\text{F}}^2 \frac{\sigma^2}{\lambda_{\min}^2}\frac{p}{t}\sum_{k\in[3]}\left\|\mathcal{H} \bigtimes_{j\neq k}  {\mathbf{U}_{j}^\star}  \right\|_\mathrm{F}^2}\\
 = & o_\bP(1).
     \end{align*}
\subsection{consistency of $\hat{\sigma}_t^2$} 
\label{sec:consistency of sigma}
We then need to consider the consistency of the estimator $\widehat{\sigma}_t^2$, where we have
\begin{align*}
\begin{aligned}
\hat{\sigma}_t^2 = & \frac{1}{t} \sum_{\tau=1}^t\left(y_\tau-\left\langle\mathcal{T}^{(\tau-1)}, \mathcal{X}_\tau\right\rangle\right)^2 \\
= & \frac{1}{t} \sum_{\tau=1}^t\left(\left\langle\mathcal{T}^{\star}, \mathcal{X}_\tau\right\rangle + \xi_\tau-\left\langle\mathcal{T}^{(\tau-1)}, \mathcal{X}_\tau\right\rangle\right)^2 \\
= & \underbrace{\frac{1}{t} \sum_{\tau=1}^t\left\langle \Delta_{\tau-1}, \mathcal{X}_\tau\right\rangle^2}_\mathrm{I}+\underbrace{\frac{2}{t} \sum_{\tau=1}^t\left\langle \Delta_{\tau-1}, \mathcal{X}_\tau\right\rangle \xi_\tau}_{\mathrm{II}}  +\underbrace{\frac{1}{t} \sum_{\tau=1}^t \xi_\tau^2}_{\mathrm{III}} .
\end{aligned}
\end{align*}
For term $\mathrm{I}$, we can see that by Theorem \ref{thm:converge} and Assumption \ref{cond:4}, we have
\begin{align*}
\begin{aligned}
\|\mathrm{I}\|_{\psi_1} & \leq  \frac{1}{t} \sum_{\tau=1}^t\left\|\Delta_{\tau-1}\right\|_{\mathrm{F}}^2  \leq  \frac{1}{t} \int_1^t \sigma^2 \frac{\df }{x^\alpha} d x \leq C \sigma^2 \frac{\df }{t^\alpha} \longrightarrow 0,
\end{aligned}
\end{align*}
where $C$ denotes some positive constant. Similarly, for the term $\mathrm{II}$, we have
\begin{align*}
\|\mathrm{II}\|_{\psi_1} \leq  \frac{2}{t} \sum_{\tau=1}^t \sigma\left\|\Delta_{\tau-1}\right\|_{\mathrm{F}} \leq C\sigma^2  \sqrt{\frac{d f}{t^\alpha}} \longrightarrow 0 .
\end{align*}
For the term $\mathrm{III}$, by the Assumption \ref{cond:1}, we have
\begin{align*}
\mathbb{E}\left[ \xi_\tau^2\right] = \sigma^2 < \infty.
\end{align*}
By the weak law of large numbers \cite[Theorem 4]{ferguson2017course},  we have
\begin{align*}
\mathrm{III}=\frac{1}{t} \sum_{\tau=1}^t \xi_\tau^2 \stackrel{p}{\longrightarrow} \sigma^2 .
\end{align*}
Then combine the results of $\mathrm{I} \stackrel{p}{\longrightarrow} 0, \mathrm{II} \stackrel{p}{\longrightarrow} 0$, and $\mathrm{III} \stackrel{p}{\longrightarrow} \sigma^2$, we conclude the proof of the consistency of $\widehat{\sigma}_t^2$. Finally, since we have shown that $\widehat{S}^2_t \stackrel{p}{\longrightarrow} S^2$, and $\widehat{\sigma}_t^2 \stackrel{p}{\longrightarrow} \sigma^2$, we then apply Slutsky's theorem and the result of Theorem \ref{thm:entry inference}, we conclude that
\begin{align*}
\frac{\hat{h}^{(t)} -h^\star}{\hat{\sigma}_t \widehat{S}_{\mathcal{H},t} / \sqrt{t}} \stackrel{d}{\longrightarrow} \mathcal{N}(0,1),
\end{align*}
and we thus finish proving Theorem \ref{thm:entry inference consistent}.

%% file: body/ProofComponent.tex
\section{Asymptotic Normality and Online Inference of Factors}\label{sec:Proof of U normality}
In the estimation process of the tensor linear form estimator, detailed in Algorithm \ref{alg:Single-step Tensor Linear Form Estimator Update}, we implement a projection of \( \widehat{\mathcal{T}}^{(t)} \) into a low-rank space spanned by factor matrices $\widehat{\mathbf{U}}_k^{(t)}$ for $k\in[3]$. A key aspect of this process is identifying the spectral differences between the true low-rank tensor \( \mathcal{T}^\star \) and the averaged tensor-based SGD \( \widehat{\mathcal{T}}^{(t)} \). This projection serves a dual purpose: it not only reduces variability but also provides the factor matrices \( \widehat{\mathbf{U}}_k^{(t)} \) with favorable distributional characteristics. It is noteworthy that the subspace \( \cP_{\widehat{\mathbf{U}}_k^{(t)}} \) is unique and remains invariant under rotation, as \( \widehat{\mathbf{U}}_k^{(t)}\widehat{\mathbf{U}}_k^{(t)\top} = \widehat{\mathbf{U}}_k^{(t)} \mathbf{R} \mathbf{R}^{\top} \widehat{\mathbf{U}}_k^{(t)\top} \) for any rotation matrix \( \mathbf{R} \in \mathbb{O}_{r_k, r_k} \). 
 The rotational invariance motivates us to characterize the distribution of the distance between the estimated low-rank space $\cP_{\widehat{\mathbf{U}}_k^{(t)}}$ and the true low-rank space $\Pu{k}$, as expressed by:
\begin{align*}
\left\|\sin \Theta\left(\widehat{\mathbf{U}}_k^{(t)}, \mathbf{U}_k^\star\right)\right\|_{\mathrm{F}}^2 = \frac{1}{2}\left\|\widehat{\mathbf{U}}_k^{(t)}\widehat{\mathbf{U}}_k^{(t)\top}-\mathbf{U}_k^{\star} \mathbf{U}_k^{\star \top}\right\|_{\mathrm{F}}^2.
\end{align*} 
\begin{assumption}
\label{cond:6}
As $t,p\rightarrow\infty$, $\max\left\{\frac{pr^3 \vee p^{3/2}r^{1/2}}{t\left(\lambda_{\min}/\sigma\right)^2}, \frac{\df^{3/2}}{t^\alpha} ,  \frac{r^3}{p}\right\}\rightarrow 0 $.
\end{assumption}
\begin{theorem}\label{thm:U normality}
Under Assumptions \ref{cond:1}, \ref{cond:init} and \ref{cond:6}, we further assumption that he design tensor \(\mathcal{X}_t\) consists of i.i.d. standard normal distribution entries. Then, as $t, p \rightarrow \infty$, we have
\begin{align*}
\frac{\left\|\sin \Theta\left(\widehat{\mathbf{U}}_k^{(t)}, \mathbf{U}_k^\star\right)\right\|_{\mathrm{F}}^2-p_k t^{-1} \sigma^2\left\|\mathbf{\Lambda}_k^{-1}\right\|_{\mathrm{F}}^2}{\sqrt{2 p_k} t^{-1} \sigma^2\left\|\mathbf{\Lambda}_k^{-2}\right\|_{\mathrm{F}}} \stackrel{\text {d}}{\longrightarrow}\mathcal{N}(0,1),
\end{align*}
where $\mathbf{\Lambda}_k$ is the $r_k \times r_k$ diagonal matrix containing the singular values of $\mathcal{M}_k(\mathcal{T}^\star)$, $k\in[3]$.
\end{theorem}
\xin{The reason we need to assume Gaussianity is that, in \eqref{equ:mean and var of clt}, we require the fourth moment of the design entries.}
From Theorem \ref{thm:U normality}, an asymptotic distribution of $\big\|\sin \Theta\big(\widehat{\mathbf{U}}_k^{(t)}, \mathbf{U}_k^\star\big)\big\|_{\mathrm{F}}^2$
involves two parameters,  \(\sigma^2\) and \(\mathbf{\Lambda}_k\).
The first parameter \(\sigma^2\) has already been estimated in Section \ref{sec:parameter inference of mt}. We next estimate the remaining parameter \(\mathbf{\Lambda}_k\) using online methods. Following a similar approach as in the previous section, an online plugin estimator for \(\mathbf{\Lambda}_k\) is as follows:
\begin{equation*}
\widehat{\mathbf{\Lambda}}_k^{(t)} =\text { diagonal matrix with the top } r_k \text { singular values of } \mathcal{M}_k\big(\widehat{\mathcal{T}}^{(t)} \times_{k+1} \widehat{\mathbf{U}}_{k+1}^{(t-1) \top} \times_{k+2} \widehat{\mathbf{U}}_{k+2}^{(t-1) \top}\big).
\end{equation*}
The estimate \( \widehat{\mathbf{\Lambda}}_k^{(t)} \) can be directly obtained as a byproduct during the estimation of \( \widehat{\mathbf{U}}_k^{(t)} \), eliminating the need for an additional, separate estimation procedure. The subsequent theorem addresses the consistency of our proposed variance estimator:
\begin{theorem}\label{thm:Tensor regression}
Under Assumptions of Theorem \ref{thm:U normality}, as $t, p \rightarrow \infty$, we have
\begin{align*}
\begin{gathered}
\frac{\left\|\sin \Theta\left(\widehat{\mathbf{U}}_k, \mathbf{U}_k^\star\right)\right\|_{\mathrm{F}}^2-p_k t^{-1} \widehat{\sigma}_t^2\left\|\big(\widehat{\mathbf{\Lambda}}_k^{(t)}\big)^{-1}\right\|_{\mathrm{F}}^2}{\sqrt{2 p_k} t^{-1} \widehat{\sigma}_t^2\left\|\big(\widehat{\mathbf{\Lambda}}_k^{(t)}\big)^{-2}\right\|_{\mathrm{F}}} \stackrel{\text {d}}{\longrightarrow} \mathcal{N}(0,1). 
\end{gathered}
\end{align*}
\end{theorem}
We detail the proof of Theorem \ref{thm:Tensor regression} in Appendix \ref{sec:Proof of Tensor regression}.
Based on Theorem \ref{thm:Tensor regression}, we are able to construct a confidence region for the true parameter \( \mathbf{U}_k^\star \). Specifically, for any given confidence level \( \alpha \in  (0,1) \), a \( 100(1-\alpha) \% \) confidence region can be constructed as follows:
\begin{align}\label{equ:confidence region}
\widehat{\mathrm{CR}}^\alpha_{U,t}:=\left\{\mathbf{U} \in \mathbb{O}_{p_k, r_k}:
\left\|\sin \Theta\left(\widehat{\mathbf{U}}_k, \mathbf{U}\right)\right\|_{\mathrm{F}}^2
\leq \frac{p_k \widehat{\sigma}_t^2}{t}\left\|\big(\widehat{\mathbf{\Lambda}}_k^{(t)}\big)^{-1}\right\|_{\mathrm{F}}^2+z_\alpha \frac{\sqrt{2 p_k} \widehat{\sigma}_t^2}{t}\left\|\big(\widehat{\mathbf{\Lambda}}_k^{(t)}\big)^{-2}\right\|_{\mathrm{F}}\right\} .
\end{align}
Theorem \ref{thm:Tensor regression} indicates that $\lim _{t, p \rightarrow \infty} \mathbb{P}\left(\mathbf{U}_k^\star \in \widehat{\mathrm{CR}}^\alpha_{U,t}\right)=1-\alpha$.

\section{Proof of Theorem \ref{thm:U normality}}\label{sec:Proof of Tensor regression}
In this section, our objective is to prove Theorem \ref{thm:U normality}, which is structured into three steps. Step 1 focuses on representing the spectral projector and is detailed in Section \ref{sec:proof of theorem U step 1}. Step 2 involves characterizing the distribution of the leading terms in the expansion of \(\big\|\sin \Theta\big(\widehat{\mathbf{U}}_k^{(t)}, \mathbf{U}_k^{\star}\big)\big\|_{\mathrm{F}}^2\), as elaborated in Section \ref{sec:proof of theorem U step 2}. Finally, Step 3 is dedicated to characterizing the distribution of the leading terms, which is discussed in Section \ref{sec:proof of theorem U step 3}.
\subsection{Step 1: representation of Spectral Projector}
\label{sec:proof of theorem U step 1}
Without loss of generality, we focus on the case of  $k = 1$. Our focus now turns to the distribution of $\left\| \mathcal{P}_{\widehat{\mathbf{U}}_1^{(t)}}-\mathcal{P}_{\mathbf{U}_1^\star} \right\|_{\mathrm{F}}^2$. 
Expressing this, we find
\begin{align*}
\left\|\mathcal{P}_{\widehat{\mathbf{U}}_1^{(t)}}-\mathcal{P}_{\mathbf{U}_1^\star}\right\|_{\mathrm{F}}^2=&2 r_1-2\left\langle\widehat{\mathbf{U}}_1^{(t)} \widehat{\mathbf{U}}_1^{(t)\top}, \mathbf{U}_1^\star \mathbf{U}_1^{\star\top}\right\rangle\\
=&-2\left\langle\mathcal{P}_{\widehat{\mathbf{U}}_1^{(t)}}-\mathcal{P}_{\mathbf{U}_1^\star}, \mathbf{U}_1^\star \mathbf{U}_1^{\star\top}\right\rangle .
\end{align*}
Utilizing the spectral representation formula detailed in Lemma \ref{lem:theorem 1 in xia2021normal} and Theorem 1 from \cite{xia2021normal}, we proceed with the expansion as follows. From Equation (\ref{B.3}), it is established that
\begin{align}\label{equ:repres of spectral proj}
\begin{aligned}
      	& \left\langle \mathbf{U}_1^\star \mathbf{U}_1^{\star\top}, \mathcal{P}_{\widehat{\mathbf{U}}_1^{(t)}}-\mathcal{P}_{\mathbf{U}_1^\star} \right\rangle\\
       = &  	\left\langle \mathbf{U}_1^\star \mathbf{U}_1^{\star\top}, \sum_{n \geq 1} \mathcal{S}_{\mathbf{G}_1, n}\left(\mathfrak{E}_1^{(t)}\right)\right\rangle\\
  	=& \left\langle \mathbf{U}_1^\star \mathbf{U}_1^{\star\top}, \mathcal{S}_{\mathbf{G}_1, 1}\left(\mathfrak{E}_1^{(t)}\right) \right\rangle + \left\langle \mathbf{U}_1^\star \mathbf{U}_1^{\star\top}, \mathcal{S}_{\mathbf{G}_1, 2}\left(\mathfrak{E}_1^{(t)}\right) \right\rangle \\
  	& + \left\langle \mathbf{U}_1^\star \mathbf{U}_1^{\star\top}, \mathcal{S}_{\mathbf{G}_1, 3}\left(\mathfrak{E}_1^{(t)}\right) \right\rangle +\left\langle \mathbf{U}_1^\star \mathbf{U}_1^{\star\top}, \sum_{n \geq 4} \mathcal{S}_{\mathbf{G}_1, n}\left(\mathfrak{E}_1^{(t)}\right)\right\rangle
\end{aligned}
\end{align}

\subsection{Step 2: quantification of Spectral Projector Terms}
\label{sec:proof of theorem U step 2}
Now we examine the first term on the right-hand of Equation (\ref{equ:repres of spectral proj}). Given that \(\mathfrak{P}_k^0 \mathbf{U}_k^\star \mathbf{U}_k^{\star\top} = \mathbf{U}_k^\star \mathbf{U}_k^{\star\top} \mathfrak{P}_k^0 = 0\), it follows that:
\begin{align}\label{equ:sg1-1}
    \left\langle\mathcal{S}_{\mathbf{G}_1, 1}\left(\mathfrak{E}_1^{(t)} \right), \mathbf{U}_1^\star \mathbf{U}_1^{\star\top}\right\rangle=\left\langle\mathfrak{P}_1^{-1} \mathfrak{E}_1^{(t)} \mathfrak{P}_1^{\perp}+\mathfrak{P}_1^{\perp} \mathfrak{E}_1^{(t)} \mathfrak{P}_1^{-1}, \mathbf{U}_1^\star \mathbf{U}_1^{\star\top}\right\rangle=0 .
\end{align}
Our next step is to analyze the second and third terms on the right-hand side of the Equation (\ref{equ:repres of spectral proj}).
\begin{lemma}\label{lem:bound of S2}
Under the assumption for Theorem \ref{thm:U normality}, we have
  \begin{align*}
\begin{gathered}
\left|\left\langle\mathcal{S}_{\mathbf{G}_1, 2}\left(\mathfrak{E}_1^{(t)}\right), \mathbf{U}_1^\star \mathbf{U}_1^{\star\top}\right\rangle + \operatorname{tr}\left(\mathbf{\Lambda}_1^{-4} \mathbf{G}_1^\star \left(\mathbf{U}_2^{\star\top} \otimes \mathbf{U}_3^{\star\top}\right) \mathbf{Z}_1^{(t)\top} \mathbf{U}_{1 \perp}^\star \mathbf{U}_{1 \perp}^{\star\top} \mathbf{Z}_1^{(t)}\left(\mathbf{U}_2^\star \otimes \mathbf{U}_3^\star\right) \mathbf{G}_1^{\star\top}\right)\right| \\
= \Op{\frac{ \sigma^3 p r^2}{\lambda_{\min}^{3}t^{3/2}}
+ \frac{ \kappa_0^2\sigma^4p^2r_1}{\lambda_{\min }^4t^2}},
\end{gathered}
\end{align*}
and
  \begin{align}\label{equ:lemma d1-2}
\begin{aligned}
	\left|\left\langle\mathcal{S}_{\mathbf{G}_1, 3}\left(\mathfrak{E}_1^{(t)} \right), \mathbf{U}_1^\star \mathbf{U}_1^{\star\top}\right\rangle \right|= \Op{ \frac{ \sigma^3 p r^2}{\lambda_{\min}^{3}t^{3/2}} 
	+ \frac{ \kappa_0^2\sigma^4p^2r_1}{\lambda_{\min }^4t^2}}.
\end{aligned}
\end{align}
\end{lemma}
Proof in Section \ref{sec:Proof of Lemma S2}.
For the fourth term of Equation (\ref{equ:repres of spectral proj}), from Equation (\ref{equ:bound of Sg gep 2}), it is inferred that:
\begin{align}\label{equ:sg1-4}
\left|\sum_{n \geq 4}\left\langle\mathcal{S}_{\mathbf{G}_1, n}\left(\mathfrak{E}_1^{(t)} \right), \mathbf{U}_1^\star \mathbf{U}_1^{\star\top}\right\rangle\right| \leq r_1 \sum_{n \geq 4}\left(\frac{4\left\|\mathfrak{E}_1^{(t)} \right\|}{\lambda_{\min }^2}\right)^n  = \Op{ r_1 \kappa_0^4 \frac{\sigma^4p^2}{\lambda_{\min }^4t^2}},
\end{align}
The first inequality is derived from the Cauchy-Schwarz inequality, and the latter is established by Lemma \ref{lem:bound for first round}.
Considering Equation (\ref{equ:repres of spectral proj}), (\ref{equ:sg1-1}), (\ref{equ:sg1-4}), and with support from Lemma \ref{lem:bound of S2}, we have: 
\begin{align}\label{B.26}
\begin{gathered}
\left|\left\|\mathcal{P}_{\widehat{\mathbf{U}}_1^{(t)}}-\mathcal{P}_{\mathbf{U}_1^\star}\right\|_{\mathrm{F}}^2-2 \operatorname{tr}\left(\mathbf{\Lambda}_1^{-4} \mathbf{G}_1^\star\left(\mathbf{U}_2^{\star\top} \otimes \mathbf{U}_3^{\star\top}\right) \mathbf{Z}_1^{(t)\top} \mathbf{U}_{1 \perp}^\star \mathbf{U}_{1 \perp}^{\star\top} \mathbf{Z}_1^{(t)}\left(\mathbf{U}_2^\star \otimes \mathbf{U}_3^\star\right) \mathbf{G}_1^{\star\top}\right)\right| \\
= \Op{\frac{ \sigma^3 p r^2}{\lambda_{\min}^{3}t^{3/2}}
+ \frac{ \kappa_0^2\sigma^4p^2r_1}{\lambda_{\min }^4t^2}}.
\end{gathered}
\end{align}
\subsection{Step 3: characterizing the distribution of the leading terms}
\label{sec:proof of theorem U step 3}
By Equation (\ref{B.26}), it suffices to prove the distribution of $\operatorname{tr}\left(\mathbf{\Lambda}_1^{-4} \mathbf{G}_1^\star \left(\mathbf{U}_2^{\star\top} \otimes \mathbf{U}_3^{\star\top}\right) \mathbf{Z}_1^{(t)\top} \mathbf{U}_{1 \perp}^\star \mathbf{U}_{1 \perp}^{\star\top} \mathbf{Z}_1^{(t)}\left(\mathbf{U}_2^\star \otimes \mathbf{U}_3^\star\right) \mathbf{G}_1^{\star\top}\right)$.
We can write
\begin{align}\label{equ:lead-1}
\begin{aligned}
    &\operatorname{tr}\left(\mathbf{\Lambda}_1^{-4} \mathbf{G}_1^\star \left(\mathbf{U}_2^{\star\top} \otimes \mathbf{U}_3^{\star\top}\right) \mathbf{Z}_1^{(t)\top} \mathbf{U}_{1 \perp}^\star \mathbf{U}_{1 \perp}^{\star\top} \mathbf{Z}_1^{(t)}\left(\mathbf{U}_2^\star \otimes \mathbf{U}_3^\star\right) \mathbf{G}_1^{\star\top}\right)\\
=&\left\|\mathbf{\Lambda}_1^{-2} \mathbf{G}_1^\star \left(\mathbf{U}_2^{\star\top} \otimes \mathbf{U}_3^{\star\top}\right) \mathbf{Z}_1^{(t)\top} \mathbf{U}_{1 \perp}^\star \right\|_\mathrm{F}^2\\
=&\left\|\mathbf{\Lambda}_1^{-2} \mathbf{G}_1^\star \left(\mathbf{U}_2^{\star\top} \otimes \mathbf{U}_3^{\star\top}\right) \mathcal{M}_1^{\top}\left(\mathcal{Z}^{(t)}\right) \mathbf{U}_{1 \perp}^\star \right\|_\mathrm{F}^2\\
=&\left\|\mathbf{\Lambda}_1^{-2} \mathbf{G}_1^\star \left(\mathbf{U}_2^{\star\top} \otimes \mathbf{U}_3^{\star\top}\right) \left(\mathcal{M}_1^{\top}\left(\mathcal{Z}_1^{(t)}\right)+\mathcal{M}_1^{\top}\left(\mathcal{Z}_2^{(t)}\right) \right)\mathbf{U}_{1 \perp}^\star \right\|_\mathrm{F}^2\\
\leq &\left\|\mathbf{\Lambda}_1^{-2} \mathbf{G}_1^\star \left(\mathbf{U}_2^{\star\top} \otimes \mathbf{U}_3^{\star\top}\right) \mathcal{M}_1^{\top}\left(\mathcal{Z}_1^{(t)}\right)\mathbf{U}_{1 \perp}^\star \right\|_\mathrm{F}^2 \\
& + \left\|\mathbf{\Lambda}_1^{-2} \mathbf{G}_1^\star \left(\mathbf{U}_2^{\star\top} \otimes \mathbf{U}_3^{\star\top}\right) \mathcal{M}_1^{\top}\left(\mathcal{Z}_2^{(t)}\right)\mathbf{U}_{1 \perp}^\star \right\|_\mathrm{F}^2.
\end{aligned}
\end{align}
Breaking down the above equation, we focus on the first term on the right-hand side:
\begin{align*}
\left\|\mathbf{\Lambda}_1^{-2} \mathbf{G}_1^\star \left(\mathbf{U}_2^{\star\top} \otimes \mathbf{U}_3^{\star\top}\right)\mathcal{M}_1^{\top}\left(\mathcal{Z}_1^{(t)}\right) \mathbf{U}_{1 \perp}^\star \right\|_\mathrm{F}^2
=\frac{1}{t^2}\left\|\sum_{i=1}^t\xi_i\mathbf{\Lambda}_1^{-2} \mathbf{G}_1^\star \left(\mathbf{U}_2^{\star\top} \otimes \mathbf{U}_3^{\star\top}\right) \mathcal{M}_1^{\top}(\mathcal{X}_i) \mathbf{U}_{1 \perp}^\star \right\|_\mathrm{F}^2.
\end{align*}
Recall that \(\mathbf{G}_1^\star \mathbf{G}_1^{\star\top} = \mathbf{\Lambda}_1^2\). For any fixed \(\mathbf{U}_k \in \mathbb{O}_{p_k, r_k}\), Assumption \ref{cond:1} implies that each entry of \(\mathbf{U}_{1 \perp}^{\top} \mathcal{M}_1\left(\mathcal{X}_j\right)\left(\mathbf{U}_2 \otimes \mathbf{U}_3\right) \in \mathbb{R}^{(p_1 - r_1) \times r_2 r_3}\) follows a Gaussian distribution with mean zero and variance 1.
Then, 
\begin{align*}
\begin{aligned}
\text{Var} \sbr{\mathbf{\Lambda}_1^{-2} \mathbf{G}_1^\star \left(\mathbf{U}_2^{\star\top} \otimes \mathbf{U}_3^{\star\top}\right) \mathcal{M}_1(\mathcal{X}_i)^{\top} \mathbf{U}_{1 \perp}^\star} = \mathbf{\Lambda}_1^{-2} \mathbf{G}_1^\star\left[\mathbf{\Lambda}_1^{-2} \mathbf{G}_1^\star\right]^{\top} = \mathbf{\Lambda}_1^{-2} \mathbf{G}_1^\star \mathbf{G}_1^{\star\top} \mathbf{\Lambda}_1^{-2}=  \mathbf{\Lambda}_1^{-2}.
\end{aligned}
\end{align*}
Therefore,
\begin{align}\label{equ:final term}
\begin{aligned}
    & \left\|\mathbf{\Lambda}_1^{-2} \mathbf{G}_1^\star \left(\mathbf{U}_2^{\star\top} \otimes \mathbf{U}_3^{\star\top}\right)\mathcal{M}_1^{\top}\left(\mathcal{Z}_1^{(t)}\right) \mathbf{U}_{1 \perp}^\star \right\|_\mathrm{F}^2\\
= & \frac{1}{t^2}\left\|\sum_{i=1}^t\xi_i\mathbf{\Lambda}_1^{-2} \mathbf{G}_1^\star \left(\mathbf{U}_2^{\star\top} \otimes \mathbf{U}_3^{\star\top}\right) \mathcal{M}_1(\mathcal{X}_i)^{\top} \mathbf{U}_{1 \perp}^\star \right\|_\mathrm{F}^2
\stackrel{\mathrm{d}}{=} \frac{1}{t^2}\left\|\sum_{i=1}^t \xi_i \widetilde{\mathbf{Z}}_i \mathbf{\Lambda}_1^{-1}\right\|_{\mathrm{F}}^2,
\end{aligned}
\end{align}
where each entry of $\widetilde{\mathbf{Z}}_i \in\mathbb{R}^{(p_1-r_1)\times r_1}$ follows a Gaussian distribution with mean zero and variance 1.
For the second term in Equation (\ref{equ:lead-1}), similar to Lemma \ref{lem:compare z1 and z2}, we have
\begin{align}\label{equ:lead-2}
\begin{aligned}
     &\left\|\mathbf{\Lambda}_1^{-2} \mathbf{G}_1^\star \left(\mathbf{U}_2^{\star\top} \otimes \mathbf{U}_3^{\star\top}\right) \mathcal{M}_1^{\top}\left(\mathcal{Z}_2^{(t)}\right) \mathbf{U}_{1 \perp}^\star \right\|_\mathrm{F}^2 \\
      \stackrel{(a)}{\leq} & r  \left\|\mathbf{\Lambda}_1^{-2} \mathbf{G}_1^\star\right\|^2 \left\|  \left(\mathbf{U}_2^{\star\top} \otimes \mathbf{U}_3^{\star\top}\right)\mathcal{M}_1^{\top}\left(\mathcal{Z}_2^{(t)}\right) \mathbf{U}_{1 \perp}^\star \right\|^2
=\Op{r \frac{\sigma^2}{\lambda_{\min}^2} \frac{p\df}{t^{1+\alpha}}}.
\end{aligned}
 \end{align}
Here, (a) arises from the elementary bounds $\|\mathbf{A B}\|_{\mathrm{F}} \leq\|\mathbf{A}\|_{\mathrm{F}}\|\mathbf{B}\|$.
By Equation (\ref{B.26}), (\ref{equ:final term}) and (\ref{equ:lead-2}), 
\begin{align*}
\left|\left\|\mathcal{P}_{\widehat{\mathbf{U}}_1^{(t)}}-\mathcal{P}_{\mathbf{U}_1^\star}\right\|_{\mathrm{F}}^2-\frac{2}{t^2}\left\|\sum_{i=1}^t \xi_i \widetilde{\mathbf{Z}}_i \mathbf{\Lambda}_1^{-1}\right\|_{\mathrm{F}}^2\right|
 = \Op{\frac{ \sigma^3 p r^2}{\lambda_{\min}^{3}t^{3/2}}
+ \frac{ \kappa_0^2\sigma^4p^2r_1}{\lambda_{\min }^4t^2}
 +  \frac{\sigma^2}{\lambda_{\min}^2} \frac{pr\df}{t^{1+\alpha}}}.
\end{align*}

For any integer \( j \) such that \( 1 \leq j \leq p_1-r_1 \):
\begin{align}\label{equ:mean and var of clt}
\begin{aligned}
	\mathbb{E}\left\|\left(\sum_{i=1}^t\xi_i\left(\widetilde{\mathbf{Z}}_i\right)_{[j,:]}\right) \mathbf{\Lambda}_1^{-1}\right\|_2^2 =& t\sigma^2\left\|\mathbf{\Lambda}_1^{-1}\right\|_{\mathrm{F}}^2, \\
\operatorname{Var}\left(\left\|\left(\sum_{i=1}^t\xi_i\left(\widetilde{\mathbf{Z}}_i\right)_{[j,:]}\right) \mathbf{\Lambda}_1^{-1}\right\|_2^2\right) =& 2 t^2\sigma^4\left\|\mathbf{\Lambda}_1^{-2}\right\|_{\mathrm{F}}^2.
\end{aligned}
\end{align}
Here, we use the fact that the fourth moment of the standard normal distribution is 3.
Thus, \( \left\|\left(\sum_{i=1}^t \xi_i \widetilde{\mathbf{Z}}_i\right) \mathbf{\Lambda}_1^{-1}\right\|_{\mathrm{F}}^2 \) is expressed as the sum of \( p_1-r_1 \) random variables, with a mean value of \( t\sigma^2\left\|\mathbf{\Lambda}_1^{-1}\right\|_{\mathrm{F}}^2 \) and a standard deviation of \( \sqrt{2}t\sigma^2\left\|\mathbf{\Lambda}_1^{-2}\right\|_{\mathrm{F}} \).  
Drawing from the Central Limit Theorem as presented in \cite[Theorem 5]{ferguson2017course}, we deduce:
\begin{align*}
	\frac{2t^{-2}\left\|\left(\sum_{i=1}^t \xi_i \widetilde{\mathbf{Z}}_i\right) \mathbf{\Lambda}_1^{-1}\right\|_{\mathrm{F}}^2-2\left(p_1-r_1\right)t^{-1}\sigma^2\left\|\mathbf{\Lambda}_1^{-1}\right\|_{\mathrm{F}}^2}{2\sqrt{2\left(p_1-r_1\right)}t^{-1}\sigma^2\left\|\mathbf{\Lambda}_1^{-2}\right\|_{\mathrm{F}}}  \stackrel{d}{\longrightarrow} \mathcal{N}(0,1).
\end{align*}
Next, we need to show that the reminder terms are less than standard deviation. Note that the standard deviation term 
\begin{align*}
	2\sqrt{2\left(p_1-r_1\right)}t^{-1}\sigma^2\left\|\mathbf{\Lambda}_1^{-2}\right\|_{\mathrm{F}}t^{-1}
	\geq
	\sqrt{2 p_1 r_1}t^{-1}\sigma^2 \kappa_0^{-2} \lambda_{\min }^{-2}.
\end{align*}
Then,
\begin{align*}
	 & \frac{1}{2\sqrt{2\left(p_1-r_1\right)}t^{-1}\sigma^2\left\|\mathbf{\Lambda}_1^{-2}\right\|_{\mathrm{F}}t^{-1}}\sbr{\frac{ \sigma^3 p r^2}{\lambda_{\min}^{3}t^{3/2}}
+ \frac{ \kappa_0^2\sigma^4p^2r_1}{\lambda_{\min }^4t^2}
 +  \frac{\sigma^2}{\lambda_{\min}^2} \frac{pr\df}{t^{1+\alpha}}}\\
 \leq & \frac{1}{\sqrt{2 p_1 r_1}t^{-1}\sigma^2 \kappa_0^{-2} \lambda_{\min }^{-2}}
 \sbr{\frac{ \sigma^3 p r^2}{\lambda_{\min}^{3}t^{3/2}}
+ \frac{ \kappa_0^2\sigma^4p^2r_1}{\lambda_{\min }^4t^2}
+  \frac{\sigma^2}{\lambda_{\min}^2} \frac{pr\df}{t^{1+\alpha}}}\\
 = & \Op{\kappa_0^2\sqrt{\frac{\sigma^2pr^3}{\lambda_{\min}^{2}t}}
+ \frac{\kappa_0^4\sigma^2\sqrt{p^3r}}{\lambda_{\min }^2t}  
 + \frac{\kappa_0^2\df^{3/2}}{t^\alpha}}\\
 = & o_p(1).
\end{align*}
The last equality is due to Assumption \ref{cond:6}. Combining the two inequalities above, we know that
\begin{align*}
	\frac{\left\|\mathcal{P}_{\widehat{\mathbf{U}}_1^{(t)}}-\mathcal{P}_{\mathbf{U}_1^\star}\right\|_{\mathrm{F}}^2-2\left(p_1-r_1\right)t^{-1}\sigma^2\left\|\mathbf{\Lambda}_1^{-1}\right\|_{\mathrm{F}}^2}{2\sqrt{2\left(p_1-r_1\right)}t^{-1}\sigma^2\left\|\mathbf{\Lambda}_1^{-2}\right\|_{\mathrm{F}}}  \stackrel{d}{\longrightarrow} \mathcal{N}(0,1).
\end{align*}
Given Lipschitz property of both $\Phi(\cdot)$ and $|x| e^{-x^2 / 2}<1$ for all $x \in \mathbb{R}$, this replacement is justified for any $x \in \mathbb{R}$.
 \begin{align*}
\begin{aligned}
&\left|\Phi\left(\sqrt{\frac{p_1}{p_1-r_1}} x+\frac{r_1\left\|\mathbf{\Lambda}_1^{-1}\right\|_{\mathrm{F}}^2}{\sqrt{2\left(p_1-r_1\right)}\left\|\mathbf{\Lambda}_1^{-2}\right\|_{\mathrm{F}}}\right)-\Phi(x)\right| \\
\leq & \left|\Phi\left(\sqrt{\frac{p_1}{p_1-r_1}} x\right)-\Phi(x)\right|+C_3 \frac{r_1\left\|\mathbf{\Lambda}_1^{-1}\right\|_{\mathrm{F}}^2}{\sqrt{2\left(p_1-r_1\right)}\left\|\mathbf{\Lambda}_1^{-2}\right\|_{\mathrm{F}}} \\
\leq & \left(\sqrt{\frac{p_1}{p_1-r_1}}-1\right)|x| e^{-x^2 / 2} +C_3 \frac{r_1\left\|\mathbf{\Lambda}_1^{-1}\right\|_{\mathrm{F}}^2}{\sqrt{2\left(p_1-r_1\right)}\left\|\mathbf{\Lambda}_1^{-2}\right\|_{\mathrm{F}}}
\stackrel{(a)}{\leq} C_3 \frac{r^{3 / 2}}{\sqrt{p_1-r_1}} .
\end{aligned}
\end{align*}
In (a), we use the inequality \(\left\|\mathbf{\Lambda}_1^{-1}\right\|_{\mathrm{F}}^2 \leq \sqrt{r}\left\|\mathbf{\Lambda}_1^{-2}\right\|_{\mathrm{F}}\) which is derived from the Cauchy-Schwarz Inequality.
Combining the above two inequalities, we will have:
\begin{align*}
	\frac{\left\|\mathcal{P}_{\widehat{\mathbf{U}}_1^{(t)}}-\mathcal{P}_{\mathbf{U}_1^\star}\right\|_{\mathrm{F}}^2-2p_1t^{-1}\sigma^2\left\|\mathbf{\Lambda}_1^{-1}\right\|_{\mathrm{F}}^2}{2\sqrt{2 p_1}t^{-1}\sigma^2\left\|\mathbf{\Lambda}_1^{-2}\right\|_{\mathrm{F}}}  \stackrel{d}{\longrightarrow} \mathcal{N}(0,1).
\end{align*}
Owing to the equivalence of the sin$\Theta$ distance, we derive the following relationship:
\begin{align*}
\left\|\mathcal{P}_{\widehat{\mathbf{U}}_1^{(t)}}-\mathcal{P}_{\mathbf{U}_1^\star}\right\|_{\mathrm{F}}^2 = 2 \left\|\sin \Theta(\widehat{\mathbf{U}}_1^{(t)}, \mathbf{U}_1^\star)\right\|_{\mathrm{F}}^2.
\end{align*}
We conclude
\begin{align*}
\frac{\left\|\sin \Theta\left(\widehat{\mathbf{U}}_k^{(t)}, \mathbf{U}_k^\star\right)\right\|_{\mathrm{F}}^2-p_k t^{-1} \sigma^2\left\|\mathbf{\Lambda}_k^{-1}\right\|_{\mathrm{F}}^2}{\sqrt{2 p_k} t^{-1} \sigma^2\left\|\mathbf{\Lambda}_k^{-2}\right\|_{\mathrm{F}}} \stackrel{d}{\longrightarrow} \mathcal{N}(0,1).
\end{align*}
Now we conclude the proof of Theorem \ref{thm:U normality}.

\section{Proof of Theorem \ref{thm:Tensor regression}}
\label{sec:Proof of Tensor regression}
We denote $\widehat{\mathbf{U}}_1^{(t)} \in \mathbb{O}_{p_1, r_1}$ the top-$r_1$ left singular vectors of $\mathcal{M}_1\left(\widehat{\mathcal{T}}^{(t)} \times_2 \widehat{\mathbf{U}}_2^{(t-1)} \times_3 \widehat{\mathbf{U}}_3^{(t-1)}\right)$. By Lemma \ref{lem:bound for inference}, it is easy to show that under the event $\mathcal{A}_t\cap\mathcal{C}_t\cap\mathcal{D}_t$, we have 
$$ \left\|\mathcal{P}_{\widehat{\mathbf{U}}_1^{(t)}}-\mathcal{P}_{\mathbf{U}_1^\star}\right\| = \Op{ \frac{\sigma}{\lambda_{\min}}\sqrt{\frac{p}{t}}}.$$
By definition, we know that $\left(\widehat{\mathbf{\Lambda}}_1^{(t)}\right)^2=\operatorname{diag}\left(\hat{\lambda}_1^2, \cdots, \hat{\lambda}_{r_1}^2\right)$ contains the eigenvalues of 
$$\mathcal{M}_1\left(\widehat{\mathcal{T}}^{(t)}\right)\left(\widehat{\mathbf{U}}_2^{(t-1)} \widehat{\mathbf{U}}_2^{(t-1) \top} \otimes \widehat{\mathbf{U}}_3^{(t-1)} \widehat{\mathbf{U}}_3^{(t-1) \top}\right) \mathcal{M}_1^{\top}\left(\widehat{\mathcal{T}}^{(t)}\right).$$
Then, by spectral decompositiona and Weyl’s inequality, we have
\begin{align}\label{equ:lam-1}
\begin{aligned}
& \sup _{1 \leq k \leq r_1}\left|\lambda_k^2-\hat{\lambda}_k^2\right| \\
\leq & \inf _{\mathbf{R} \in \mathbb{O}_{r_1, r_1}}\left\|\widehat{\mathbf{U}}_1^{(t)\top} \mathcal{M}_1\left(\widehat{\mathcal{T}}^{(t)}\right)\left(\mathcal{P}_{\widehat{\mathbf{U}}_2^{(t-1)}} \otimes \mathcal{P}_{\widehat{\mathbf{U}}_3^{(t-1)}}\right) \mathcal{M}_1\left(\widehat{\mathcal{T}}^{(t)}\right)^{\top} \widehat{\mathbf{U}}_1^{(t)}-\mathbf{R}  \mathbf{G}_1^\star \mathbf{G}_1^{\star\top} \mathbf{R} \right\| \\
\leq & \inf _{\mathbf{R}  \in \mathbb{O}_{r_1, r_1}}\left\|\widehat{\mathbf{U}}_1^{(t)\top} \mathbf{T}^{\star}_1\left(\mathcal{P}_{\widehat{\mathbf{U}}_2^{(t-1)}} \otimes \mathcal{P}_{\widehat{\mathbf{U}}_3^{(t-1)}}\right) \mathbf{T}^{\star\top}_1 \widehat{\mathbf{U}}_1^{(t)}-\mathbf{R}  \mathbf{\Lambda}_1^2 \mathbf{R} \right\| \\
& +2\left\|\widehat{\mathbf{U}}_1^{(t)\top} \mathbf{T}^{\star}_1\left(\mathcal{P}_{\widehat{\mathbf{U}}_2^{(t-1)}} \otimes \mathcal{P}_{\widehat{\mathbf{U}}_3^{(t-1)}}\right) \mathbf{Z}_1^{(t)\top} \widehat{\mathbf{U}}_1^{(t)}\right\|+\left\|\widehat{\mathbf{U}}_1^{(t)\top} \mathbf{Z}_1^{(t)}\left(\mathcal{P}_{\widehat{\mathbf{U}}_2^{(t-1)}} \otimes \mathcal{P}_{\widehat{\mathbf{U}}_3^{(t-1)}}\right) \mathbf{Z}_1^{(t)\top} \widehat{\mathbf{U}}_1^{(t)}\right\| \\
 \leq &\left\|\widehat{\mathbf{U}}_1^{(t)\top} \mathbf{U}_1^{\star} \mathbf{G}_1^\star\left(\left(\mathbf{U}_2^{\star\top} \mathcal{P}_{\widehat{\mathbf{U}}_2^{(t-1)}} \mathbf{U}_2^\star\right) \otimes\left(\mathbf{U}_3^{\star\top} \mathcal{P}_{\widehat{\mathbf{U}}_3^{(t-1)}} \mathbf{U}_3^\star\right)\right) \mathbf{G}_1^{\star\top} \mathbf{U}_1^{\star\top} \widehat{\mathbf{U}}_1^{(t)}-\widehat{\mathbf{U}}_1^{(t)\top} \mathbf{U}_1^{\star} \mathbf{G}_1^\star \mathbf{G}_1^{\star\top} \mathbf{U}_1^{\star\top} \widehat{\mathbf{U}}_1^{(t)}\right\| \\
& +\inf _{\mathbf{R}  \in \mathbb{O}_{r_1, r_1}}\left\|\widehat{\mathbf{U}}_1^{(t)\top} \mathbf{U}_1^{\star} \mathbf{\Lambda}_1^2 \mathbf{U}_1^{\star\top} \widehat{\mathbf{U}}_1^{(t)}-\mathbf{R}  \mathbf{\Lambda}_1^2 \mathbf{R} ^{\top}\right\|+2 \kappa_0 \lambda_{\min }\left\|\widehat{\mathbf{U}}_1^{(t)\top} \mathbf{Z}_1^{(t)}\left(\widehat{\mathbf{U}}_2^{(t-1)} \otimes \widehat{\mathbf{U}}_3^{(t-1)}\right)\right\|\\
&+\left\|\widehat{\mathbf{U}}_1^{(t)\top} \mathbf{Z}_1^{(t)}\left(\widehat{\mathbf{U}}_2^{(t-1)} \otimes \widehat{\mathbf{U}}_3^{(t-1)}\right)\right\|^2 .
\end{aligned}
\end{align}
For the last term in Equation (\ref{equ:lam-1}), by Lemma \ref{lem:compare z1 and z2}, Lemma \ref{lem:bound for inference}, we have
\begin{align*}
\left\|\widehat{\mathbf{U}}_1^{(t)\top} \mathbf{Z}_1^{(t)}\left(\widehat{\mathbf{U}}_2^{(t-1)} \otimes \widehat{\mathbf{U}}_3^{(t-1)}\right)\right\| = \Op{\sigma\sqrt{\frac{r^2}{t}} + \frac{\sigma^2}{\lambda_{\min}} \sqrt{\frac{p^2r}{t^2}} }.
\end{align*}
For the first term in Equation (\ref{equ:lam-1}), by Lemma \ref{lem:bound for inference}, we have
\begin{align*}
\begin{aligned}
&\left\|\widehat{\mathbf{U}}_1^{(t)\top} \mathbf{U}_1^{\star} \mathbf{G}_1^{\star}\left(\left(\mathbf{U}_2^{\star\top} \mathcal{P}_{\widehat{\mathbf{U}}_2^{(t-1)}} \mathbf{U}_2^{\star}\right) \otimes\left(\mathbf{U}_3^{\star\top} \mathcal{P}_{\widehat{\mathbf{U}}_3^{(t-1)}} \mathbf{U}_3^{\star}\right)\right) \mathbf{G}_1^{\star\top} \mathbf{U}_1^{\star\top} \widehat{\mathbf{U}}_1^{(t)}-\widehat{\mathbf{U}}_1^{(t)\top} \mathbf{U}_1^{\star} \mathbf{G}_1^\star \mathbf{G}_1^{\star\top} \mathbf{U}_1^{\star\top} \widehat{\mathbf{U}}_1^{(t)}\right\| \\
\leq & \left\|\mathbf{G}_1^\star\left(\left(\mathbf{U}_2^{\star\top} \mathcal{P}_{\widehat{\mathbf{U}}_2^{(t-1)}} \mathbf{U}_2^{\star}\right) \otimes\left(\mathbf{U}_3^{\star\top} \mathcal{P}_{\widehat{\mathbf{U}}_3^{(t-1)}} \mathbf{U}_3^{\star}\right)\right) \mathbf{G}_1^{\star\top}-\mathbf{G}_1^\star \left(\mathbf{I}_{r_2}\otimes\mathbf{I}_{r_3}\right)\mathbf{G}_1^{\star\top}\right\| \\
\stackrel{(a)}{\leq} &\left\|\mathbf{G}_1^\star\left(\left(\mathbf{U}_2^{\star\top} \mathcal{P}_{\widehat{\mathbf{U}}_2^{(t-1)}}^{\perp} \mathbf{U}_2^{\star}\right) \otimes\left(\mathbf{U}_3^{\star\top} \mathcal{P}_{\widehat{\mathbf{U}}_3^{(t-1)}} \mathbf{U}_3^{\star}\right)\right) \mathbf{G}_1^{\star\top}\right\|+\left\|\mathbf{G}_1^\star\left(\mathbf{I}_{r_2} \otimes\left(\mathbf{U}_3^{\star\top} \mathcal{P}_{\widehat{\mathbf{U}}_3^{(t-1)}}^{\perp} \mathbf{U}_3\right)\right) \mathbf{G}_1^{\star\top}\right\| \\
\leq & \kappa_0^2 \lambda_{\min }^2\left(\left\|\mathbf{U}_2^{\star\top} \mathcal{P}_{\widehat{\mathbf{U}}_2^{(t-1)}}^{\perp} \mathbf{U}_2^{\star}\right\|+\left\|\mathbf{U}_3^{\star\top} \mathcal{P}_{\widehat{\mathbf{U}}_3^{(t-1)}}^{\perp} \mathbf{U}_3^{\star}\right\|\right) 
\leq \kappa_0^2 \lambda_{\text {min }}^2\left(\left\|\mathbf{U}_2^{\star\top} \widehat{\mathbf{U}}_{2 \perp}^{(t-1)}\right\|^2+\left\|\mathbf{U}_3^{\star\top} \widehat{\mathbf{U}}_{3 \perp}^{(t-1)}\right\|^2\right) \\
=  & \Op{\kappa_0^2 \sigma^2\frac{p}{t}} .
\end{aligned}
\end{align*}
Here, (a) arises from $\mathcal{P}_{\widehat{\mathbf{U}}_2^{(t-1)}}^{\perp} + \mathcal{P}_{\widehat{\mathbf{U}}_2^{(t-1)}} = \mathbf{I}_{r_2}$.
To deal with $\inf _{\mathbf{R} \in \mathbb{O}_{r_1, r_1}}\left\|\widehat{\mathbf{U}}_1^{(t)\top} \mathbf{U}_1^{\star} \mathbf{\Lambda}_1^2 \mathbf{U}_1^{\star\top} \widehat{\mathbf{U}}_1^{(t)}-\mathbf{R} \mathbf{\Lambda}_1^2 \mathbf{R}^{\top}\right\|$, by the Lemma 6 in \cite{xia2022inference} and Lemma \ref{lem:bound for inference}, we have 
\begin{align*}
\begin{aligned}
& \inf _{\mathbf{R} \in \mathbb{O}_{r_1, r_1}}\left\|\widehat{\mathbf{U}}_1^{(t)\top} \mathbf{U}_1^{\star} \mathbf{\Lambda}_1^2 \mathbf{U}_1^{\star\top} \widehat{\mathbf{U}}_1^{(t)}-\mathbf{R} \mathbf{\Lambda}_1^2 \mathbf{R}^{\top}\right\| & \\
\leq & \inf _{\mathbf{R} \in \mathbb{O}_{r_1, r_1}}\left\{\left\|\left(\widehat{\mathbf{U}}_1^{(t)\top} \mathbf{U}_1^{\star}-\mathbf{R}\right) \mathbf{\Lambda}_1^2 \mathbf{U}_1^{\star\top} \widehat{\mathbf{U}}_1^{(t)}\right\|+\left\|\mathbf{R} \mathbf{\Lambda}_1^2\left(\widehat{\mathbf{U}}_1^{(t)\top} \mathbf{U}_1^{\star}-\mathbf{R}\right)^{\top}\right\|\right\} \\
\leq & 2 \inf _{\mathbf{R} \in \mathbb{O}_{r_1, r_1}}\left\|\widehat{\mathbf{U}}_1^{(t)\top} \mathbf{U}_1^{\star}-\mathbf{R}\right\|\left\|\mathbf{\Lambda}_1^2\right\| \stackrel{(a)}{\leq} 2 \left\|\mathbf{U}_{1\perp}^{\star\top}\widehat{\mathbf{U}}_1^{(t)}\right\|^2\left\|\mathbf{\Lambda}_1^2\right\|\\
=  & \Op{\left(\sqrt{\frac{p}{t}}\sigma \lambda_{\text {min }}^{-1}\right)^2 \cdot \kappa_0^2 \lambda_{\text {min }}^2 } = \Op{\kappa_0^2\sigma^2 \frac{p}{t}}.
\end{aligned}
\end{align*}
Here, (a) is due to Equation (\ref{equ:bound of UU-R}).
Combining together the inequalities above, we have
 \begin{align*}
\sup _{1 \leq k \leq r_1}\left|\lambda_k^2-\hat{\lambda}_k^2\right| =\Op{ \kappa_0 \sigma \left(\lambda_{\min }\sqrt{\frac{r^2}{t}} + \kappa_0 \sigma \sqrt{\frac{p^2r}{t^2}}\right)} .
\end{align*}
Therefore, we have
\begin{align*}
\left|\left\|\mathbf{\Lambda}_1^{-1}\right\|_{\mathrm{F}}^2-\left\|\left(\widehat{\mathbf{\Lambda}}_1^{(t)}\right)^{-1}\right\|_{\mathrm{F}}^2\right| \leq r_1 \sup _{1 \leq k \leq r_1} \frac{\left|\lambda_k^2-\hat{\lambda}_k^2\right|}{\lambda_k^2 \hat{\lambda}_k^2}=\Op{\kappa_0 \sigma\lambda_{\min }^{-3} \left(r\sqrt{\frac{r^2}{t} } + \kappa_0\lambda_{\min }^{-1}\sigma\frac{pr^{3/2}}{t}\right)}, 
\end{align*}
and as a result
\begin{align*}
\begin{aligned}
\left|\left\|\mathbf{\Lambda}_1^{-2}\right\|_{\mathrm{F}}-\left\|\left(\widehat{\mathbf{\Lambda}}_1^{(t)}\right)^{-2}\right\|_{\mathrm{F}}\right| \leq & \left\|\mathbf{\Lambda}_1^{-2}-\left(\widehat{\mathbf{\Lambda}}_1^{(t)}\right)^{-2}\right\|_{\mathrm{F}} 
\leq r_1 \sup _{1 \leq k \leq r_1} \frac{\left|\lambda_k^2-\hat{\lambda}_k^2\right|}{\lambda_k^2 \hat{\lambda}_k^2} \\
= &  \Op{ \kappa_0 \sigma\lambda_{\min }^{-3} \left(r\sqrt{\frac{r^2}{t}} + \kappa_0\lambda_{\min }^{-1}\sigma\frac{pr^{3/2}}{t}\right)}.
\end{aligned}
\end{align*}
Under Assumptions of Theorem \ref{thm:U normality}, as $t, p \rightarrow \infty$,  we have demonstrated $\big\|\big(\widehat{\mathbf{\Lambda}}_1^{(t)}\big)^{-1}\big\|_{\mathrm{F}}^2 \stackrel{p}{\longrightarrow} \left\|\mathbf{\Lambda}_1^{-1}\right\|_{\mathrm{F}}^2$, and $\big\|\big(\widehat{\mathbf{\Lambda}}_1^{(t)}\big)^{-2}\big\|_{\mathrm{F}} \stackrel{p}{\longrightarrow}\left\|\mathbf{\Lambda}_1^{-2}\right\|_{\mathrm{F}}$. Then, we apply Slutsky's theorem in conjunction with the findings of Theorem \ref{thm:U normality} and Theorem \ref{thm:entry inference consistent}. Consequently, we conclude that:
\begin{align*}
\begin{gathered}
\frac{\left\|\sin \Theta\left(\widehat{\mathbf{U}}_k, \mathbf{U}_k^\star\right)\right\|_{\mathrm{F}}^2-p_k t^{-1} \widehat{\sigma}_t^2\left\|\left(\widehat{\mathbf{\Lambda}}_k^{(t)}\right)^{-1}\right\|_{\mathrm{F}}^2}{\sqrt{2 p_k} t^{-1} \widehat{\sigma}_t^2\left\|\left(\widehat{\mathbf{\Lambda}}_k^{(t)}\right)^{-2}\right\|_{\mathrm{F}}} \stackrel{d}{\longrightarrow} \mathcal{N}(0,1), 
\end{gathered}
\end{align*}
and we thus finish proving Theorem \ref{thm:Tensor regression}.

%% file: body/Some_lemmas.tex
\section{Some Lemmas}
\begin{lemma}\label{lem:compare z1 and z2}
	Under the assumptions of Theorem \ref{thm:entry inference},  
	we have,
\begin{align*}
\|\mathbf{U}_1^{\top} \mathcal{M}_1\left(\mathcal{Z}^{(t)}\right) \left(\mathbf{U}_2 \otimes \mathbf{U}_3\right)\|=\Op{\sigma \sqrt{\frac{r^2}{t}}}. 
\end{align*}
\end{lemma}
Proof in Section \ref{sec:Proof of Lemma compare z1 and z2}.
\begin{lemma}\label{lem:Op bound}
	Let \(X\) be a real-valued random variable with finite second moment, i.e., \(\mathbb{E}[X^2] < \infty\). Then
\[ X \;=\; O_p\!\bigl(\sqrt{\mathbb{E}[X^2]}\bigr). \]
\end{lemma}
The following lemma addresses the relation between the error bound of tensor \( \mathcal{T}^{(t)} \) and the matrix \( \mathbf{U}_k^{(t)} \).
\begin{lemma}\label{lem:Norm T and Norm U}
Let both tensors \(\mathcal{T}^{(t)}\) and \(\mathcal{T}^\star\) have the tucker rank-\((r_1, r_2, r_3)\) and $\mathbf{U}_k^{(t)}$ and $ \mathbf{U}_k^\star$ are their factor matricies. For any \(\delta \in [0,1]\), if 
\[
\left\| \mathcal{M}_k\left(\mathcal{T}^{(t)} - \mathcal{T}^\star\right) \right\| \leq \frac{\delta \lambda_{\text{min}}}{2},
\] 
then 
\[
\left\|\sin \Theta\left(\mathbf{U}_k^{(t)}, \mathbf{U}_k^\star\right)\right\| \leq \delta.
\]
\end{lemma}
Proof in Section \ref{sec:proof of Lemma B3}.

%% file: body/LemmaEstimation.tex
\section{Proof of Technical Lemmas}
\label{sec:Proof of Lemmas of Theorem converge}

\subsection{Proof of Lemma \ref{lem:property in region D}}
\label{sec:Proof of Lemma 1}
   For $\left( \mathcal{G}^{}, \mathbf{U}_1^{}, \mathbf{U}_2^{}, \mathbf{U}_3^{} \right) \in \mathcal{D}$, we have $ J \leq c_d \lambda_{\min}^{2}$. For any $k\in[3]$, we just need to assume $c_d\leq \frac{1}{10^2}$, then we will have
\begin{align*}
\begin{aligned}
\left\|\mathbf{U}_k^{}\right\| \leq & \left\|\mathbf{U}_k^\star \mathbf{R}_k^{}\right\|+\left\|\mathbf{U}_k^{}-\mathbf{U}_k^\star \mathbf{R}_k^{}\right\| 
=  1 + \left\|\mathbf{U}_k^{}-\mathbf{U}_k^\star \mathbf{R}_k\right\|_{\mathrm{F}} \stackrel{(a)}{\leq} 1.01, 
\end{aligned}
\end{align*}
and
\begin{align*}
  \left\|\mathcal{M}_k\left(\mathcal{G}^{}\right)\right\| \leq & \left\|\mathbf{R}_k^{ \top} \mathcal{M}_k\left(\mathcal{G}^\star\right)\left(\mathbf{R}_{k+1} \otimes \mathbf{R}_{k+2}\right)\right\|  +\left\|\mathbf{R}_k^{ \top} \mathcal{M}_k\left(\mathcal{G}^\star\right)\left(\mathbf{R}_{k+1}^{} \otimes \mathbf{R}_{k+2}^{}\right)-\mathcal{M}_k\left(\mathcal{G}^{}\right)\right\| \\
 \leq& \lambda_{\max} + \left\|\mathcal{G}^{}- \mathcal{G}^\star \times_1 \mathbf{R}_1 \times_2 \mathbf{R}_2 \times_3 \mathbf{R}_3 \right\|_{\mathrm{F}} 
\stackrel{(b)}{\leq} 1.01 \lambda_{\max}.
\end{align*}
Here, (a) and (b) are due to $ J \leq c_d \lambda_{\min}^{2}$ and the matrix $\mathbf{R}_k$ for $k\in[3]$ is defined in Equation (\ref{equ:def of Rt}).
Thus, we conclude the proof for Lemma \ref{lem:property in region D}.

\subsection{Proof of Lemma \ref{lem:def of Xi}}
\label{sec:proof of lemma def of Xi}

First, by the distribution of $\xi$ and $\cX$ in Assumption \ref{cond:1}, we can find that
\begin{align*}
\begin{aligned}
\left\|\left(\nabla_\mathcal{T}f\left(\mathcal{T}^{\star};\boldsymbol{\zeta}_{t+1}\right)\right)_{ijk}\right\|_{\psi_2} = \left\|\left(\xi_{t+1}\mathcal{X}_{t+1}\right)_{ijk}\right\|_{\psi_2}\leq \sigma,
\end{aligned}
\end{align*}
and for the second term, we have
\begin{align*}
\begin{aligned}
&\left\|\left(\nabla_\mathcal{T}f\left(\mathcal{T}^{(t)} -\mathcal{T}^{\star} ;\boldsymbol{\zeta}_{t+1}\right)\right)_{ijk}\right\|_{\psi_1}I\left\{\mathcal{E}_{t}\right\}\\
=& \left\|\left(\left\langle\mathcal{X}_{t+1}, \mathcal{T}^{(t)}-\mathcal{T}^{\star}\right\rangle\mathcal{X}_{t+1}\right)_{ijk}\right\|_{\psi_1}I\left\{\mathcal{E}_{t}\right\}\\
\stackrel{(a)}{\leq} & \left\|\rbr{\mathcal{T}^{(t)}-\mathcal{T}^{\star}}_{ijk}\right\|_\mathrm{F}I\left\{\mathcal{E}_{t}\right\}
\stackrel{(b)}{\leq}  C_\mathcal{E}\sigma.
\end{aligned}
\end{align*}
Here, (a) is due to the distribution of $\cX$ and under the event $\cE_t$, $\mathcal{T}^{(t)}-\mathcal{T}^{\star}$ is a constant. (b) comes from Equation (\ref{equ:bound of J under E}) and  Lemma E.2 in \cite{han2022optimal}. 
Then, by Lemma E.5 in \cite{han2022optimal} for $\epsilon_1^{(t)}$ and Bernstein-type inequality for $\epsilon_{2}^{(t)}I\left\{\cE_t\right\}$, we can get that there exist some universal constants $C_2, c_2$ and event $\mathcal{E}^\epsilon_t$, such that $\mathbb{P}\left[\left(\mathcal{E}^\epsilon_t\right)^c\right] \leq 2\exp \left(-c_2\sqrt{\df}\right)$, where $  \mathcal{E}^\epsilon_t = \left\{ \epsilon_{1}^{(t)} + \epsilon_{2}^{(t)} I\left\{\cE_t\right\} \leq C_2 \sigma \sqrt{\df}\right\}$, and $\df = r_1 r_2 r_3+\sum_{k=1}^3 p_k r_k$. Thus, we conclude the proof for Lemma \ref{lem:def of Xi}.

\subsection{Proof of Lemma \ref{lem:contraction of U}}\label{sec:Proof of Lemma contraction of U}

We use the notation of 
\begin{align}\label{equ:U deri}
\begin{aligned}
	\breve{\mathbf{U}}^{(t)}_1 & :=\left(\mathbf{U}^{(t)}_3 \otimes \mathbf{U}^{(t)}_2\right) \mathcal{M}_1^{\top}\left(\mathcal{G}^{(t)}\right), \\
\breve{\mathbf{U}}^{(t)}_2 & :=\left(\mathbf{U}^{(t)}_3 \otimes \mathbf{U}^{(t)}_1\right) \mathcal{M}_2^{\top}\left(\mathcal{G}^{(t)}\right), \\
\breve{\mathbf{U}}^{(t)}_3 & :=\left(\mathbf{U}^{(t)}_2 \otimes \mathbf{U}^{(t)}_1\right) \mathcal{M}_3^{\top}\left(\mathcal{G}^{(t)}\right).
\end{aligned}
\end{align}
	\begin{align}\label{equ:U-UR}
\begin{aligned}
& \left\|\mathbf{U}_1^{(t+1)}-\mathbf{U}_1^\star \mathbf{R}_1^{(t)}\right\|_{\mathrm{F}}^2 \\
= & \left\|\mathbf{U}_1^{(t)}-\mathbf{U}_1^\star \mathbf{R}_1^{(t)}-\eta_{t+1}\left[\mathcal{M}_1\left( \nabla_\mathcal{T}f\left(\mathcal{T}^{(t)}\right)\right) \breve{\mathbf{U}}^{(t)}_1+\frac{1}{2} \mathbf{U}_1^{(t)}\left(\mathbf{U}_1^{(t) \top} \mathbf{U}_1^{(t)} - \mathbf{U}_1^{\star \top} \mathbf{U}_1^{\star}\right)\right]\right\|_{\mathrm{F}}^2 \\
= & \left\|\mathbf{U}_1^{(t)}-\mathbf{U}_1^\star \mathbf{R}_1^{(t)}\right\|_{\mathrm{F}}^2+\eta_{t+1}^2\left\|\mathcal{M}_1\left( \nabla_\mathcal{T}f\left(\mathcal{T}^{(t)}\right)\right) \breve{\mathbf{U}}^{(t)}_1+\frac{1}{2} \mathbf{U}_1^{(t)}\left(\mathbf{U}_1^{(t) \top} \mathbf{U}_1^{(t)} - \mathbf{U}_1^{\star \top} \mathbf{U}_1^{\star}\right)\right\|_{\mathrm{F}}^2 \\
& -2 \eta_{t+1}\left\langle\mathbf{U}_1^{(t)}-\mathbf{U}_1^\star \mathbf{R}_1^{(t)}, \mathcal{M}_1\left( \nabla_\mathcal{T}f\left(\mathcal{T}^{(t)}\right)\right) \breve{\mathbf{U}}^{(t)}_1\right\rangle \\
& -2 \eta_{t+1}\frac{1}{2}\left\langle\mathbf{U}_1^{(t)}-\mathbf{U}_1^\star \mathbf{R}_1^{(t)}, \mathbf{U}_1^{(t)}\left(\mathbf{U}_1^{(t) \top} \mathbf{U}_1^{(t)} - \mathbf{U}_1^{\star \top} \mathbf{U}_1^{\star}\right)\right\rangle.
\end{aligned}
\end{align}
We bound the last three terms separately. First, due to $(a+b)^2\leq 2(a^2+b^2)$, we have
\begin{align*}
\begin{aligned}
& \left\|\mathcal{M}_1\left( \nabla_\mathcal{T}f\left(\mathcal{T}^{(t)}\right)\right) \breve{\mathbf{U}}^{(t)}_1+\frac{1}{2} \mathbf{U}_1^{(t)}\left(\mathbf{U}_1^{(t) \top} \mathbf{U}_1^{(t)} - \mathbf{U}_1^{\star \top} \mathbf{U}_1^{\star}\right)\right\|_{\mathrm{F}}^2 \\
\leq & 2\left(\left\|\mathcal{M}_1\left( \nabla_\mathcal{T}f\left(\mathcal{T}^{(t)}\right)\right) \breve{\mathbf{U}}^{(t)}_1\right\|_{\mathrm{F}}^2+\frac{1}{4}\left\|\mathbf{U}_1^{(t)}\left(\mathbf{U}_1^{(t) \top} \mathbf{U}_1^{(t)} - \mathbf{U}_1^{\star \top} \mathbf{U}_1^{\star}\right)\right\|_{\mathrm{F}}^2\right) .
\end{aligned}
\end{align*}
The first term can be bounded as
\begin{align*}
\begin{aligned}
& \left\|\mathcal{M}_1\left( \nabla_\mathcal{T}f\left(\mathcal{T}^{(t)}\right)\right) \breve{\mathbf{U}}^{(t)}_1\right\|_{\mathrm{F}}^2 \\
\leq & 2\left(\left\|\mathcal{M}_1\left( \nabla_\mathcal{T}f\left(\mathcal{T}^{\star}\right)\right) \breve{\mathbf{U}}^{(t)}_1\right\|_{\mathrm{F}}^2+\left\|\mathcal{M}_1\left( \nabla_\mathcal{T}f\left(\mathcal{T}^{(t)}\right)- \nabla_\mathcal{T}f\left(\mathcal{T}^{\star}\right)\right) \breve{\mathbf{U}}^{(t)}_1\right\|_{\mathrm{F}}^2\right) \\
\stackrel{(\ref{equ:U deri})}{=} & 2\left\|\mathcal{M}_1\left( \nabla_\mathcal{T}f\left(\mathcal{T}^{\star}\right)\right)\left(\mathbf{U}_3^{(t)} \otimes \mathbf{U}_2^{(t)}\right) \mathcal{M}_1\left(\mathcal{G}^{(t)}\right)^{\top}\right\|_{\mathrm{F}}^2\\
&+2\left\|\mathcal{M}_1\left( \nabla_\mathcal{T}f\left(\mathcal{T}^{(t)}\right)- \nabla_\mathcal{T}f\left(\mathcal{T}^{\star}\right)\right) \breve{\mathbf{U}}^{(t)}_1\right\|_{\mathrm{F}}^2.
\end{aligned}
\end{align*}
Notice that by the duality of Frobenius norm, we have
\begin{align*}
\begin{aligned}
& \left\|\mathcal{M}_1\left( \nabla_\mathcal{T}f\left(\mathcal{T}^{\star}\right)\right)\left(\mathbf{U}_3^{(t)} \otimes \mathbf{U}_2^{(t)}\right) \mathcal{M}_1\left(\mathcal{G}^{(t)}\right)^{\top}\right\|_{\mathrm{F}} \\
= &\sup_{\substack{\mathbf{W}_1 \in \mathbb{R}^{p_1 \times r_1},\\\left\|\mathbf{W}_1\right\|_{\mathrm{F}} \leq 1}}\left\langle\mathcal{M}_1\left( \nabla_\mathcal{T}f\left(\mathcal{T}^{\star}\right)\right)\left(\mathbf{U}_3^{(t)} \otimes \mathbf{U}_2^{(t)}\right) \mathcal{M}_1\left(\mathcal{G}^{(t)}\right)^{\top}, \mathbf{W}_1\right\rangle \\
= &\sup _{\substack{\mathbf{W}_1 \in \mathbb{R}^{p_1 \times r_1},\\\left\|\mathbf{W}_1\right\|_{\mathrm{F}} \leq 1}}\left\langle\mathcal{M}_1\left( \nabla_\mathcal{T}f\left(\mathcal{T}^{\star}\right)\right), \mathbf{W}_1 \mathcal{M}_1\left(\mathcal{G}^{(t)}\right)\left(\mathbf{U}_3^{(t)} \otimes \mathbf{U}_2^{(t)}\right)^{\top}\right\rangle \\
= & \sup_{\substack{\mathbf{W}_1 \in \mathbb{R}^{p_1 \times r_1},\\\left\|\mathbf{W}_1\right\|_{\mathrm{F}} \leq 1}}\left\langle \nabla_\mathcal{T}f\left(\mathcal{T}^{\star}\right), \mathcal{G}^{(t)} \times \mathbf{W}_1 \times{ }_2 \mathbf{U}_2^{(t)} \times_3 \mathbf{U}_3^{(t)}\right\rangle \\
\stackrel{(a)}{\leq} & \left\|\mathbf{W}_1\right\|_{\mathrm{F}} \left\|\mathcal{M}_1\left(\mathcal{G}^{(t)}\right)\right\| \left\|\mathbf{U}_3^{(t)} \otimes \mathbf{U}_2^{(t)}\right\|  \epsilon_{1}^{(t)}
\leq   \left\|\mathcal{M}_1\left(\mathcal{G}^{(t)}\right)\right\| \left\|\mathbf{U}_3^{(t)} \otimes \mathbf{U}_2^{(t)}\right\|  \epsilon_{1}^{(t)}.
\end{aligned}
\end{align*}
Here, (a) is due to Equation (\ref{equ:def of Xi}) in Lemma \ref{lem:def of Xi}.
By Cauchy-Schwarz inequality, we have
\begin{align*}
  &\left\|\mathcal{M}_1\left( \nabla_\mathcal{T}f\left(\mathcal{T}^{(t)}\right)- \nabla_\mathcal{T}f\left(\mathcal{T}^{\star}\right)\right) \breve{\mathbf{U}}^{(t)}_1\right\|_{\mathrm{F}}\\
  =&\left\|\mathcal{M}_1\left( \nabla_\mathcal{T}f\left(\mathcal{T}^{(t)}\right)- \nabla_\mathcal{T}f\left(\mathcal{T}^{\star}\right)\right) \left(\mathbf{U}_3^{(t)} \otimes \mathbf{U}_2^{(t)}\right) \mathcal{M}_1\left(\mathcal{G}^{(t)}\right)^{\top}\right\|_{\mathrm{F}}\\
\stackrel{(\ref{equ:def of Xi})}{\leq}&  \left\|\mathbf{U}_3^{(t)}\otimes\mathbf{U}_2^{(t)}\right\|\left\|\mathcal{M}_1\left(\mathcal{G}^{(t)}\right)\right\|\epsilon_{2}^{(t)}.
\end{align*}
Then, combing the above two inequalities, we have that
\begin{align}\label{equ:J2-1}
\begin{aligned}
& \left\|\mathcal{M}_1\left( \nabla_\mathcal{T}f\left(\mathcal{T}^{(t)}\right)\right) \breve{\mathbf{U}}^{(t)}_1\right\|_{\mathrm{F}}^2I\left\{\mathcal{E}_{t}\right\} \\
 \leq &  2 \left( \left(\epsilon_{1}^{(t)}\right)^2+\left(\epsilon_{2}^{(t)}\right)^2\right) \left\|\mathcal{M}_1\left(\mathcal{G}^{(t)}\right)\right\|^2 \left\|\mathbf{U}_3^{(t)} \otimes \mathbf{U}_2^{(t)}\right\|^2 I\left\{\mathcal{E}_{t}\right\}  \\
 \stackrel{(\ref{equ:benign region})}{\leq}& 2 (1.01 )^4 (1.01 \lambda_{\max})^2 \left( \left(\epsilon_{1}^{(t)}\right)^2+\left(\epsilon_{2}^{(t)}\right)^2\right) I\left\{\mathcal{E}_{t}\right\} \\
\leq &  3 \lambda_{\max}^2 \left( \left(\epsilon_{1}^{(t)}\right)^2+\left(\epsilon_{2}^{(t)}\right)^2\right) I\left\{\mathcal{E}_{t}\right\} .
\end{aligned}
\end{align}
In addition,
\begin{align}\label{equ:J2-2}
\begin{aligned}
 \left\|\mathbf{U}_1^{(t)}\left(\mathbf{U}_1^{(t) \top} \mathbf{U}_1^{(t)}-\mathbf{U}_1^{\star \top} \mathbf{U}_1^{\star}\right)\right\|_{\mathrm{F}}^2 I\left\{\mathcal{E}_{t}\right\} \stackrel{(a)}{\leq} & \left\|\mathbf{U}_1^{(t)}\right\|^2 \left\|\mathbf{U}_1^{(t) \top} \mathbf{U}_1^{(t)}-\mathbf{U}_1^{\star \top} \mathbf{U}_1^{\star}\right\|_{\mathrm{F}}^2 I\left\{\mathcal{E}_{t}\right\} \\
 \stackrel{(\ref{equ:benign region})}{\leq}& (1.01 )^2\left\|\mathbf{U}_1^{(t) \top} \mathbf{U}_1^{(t)}-\mathbf{U}_1^{\star \top} \mathbf{U}_1^{\star}\right\|_{\mathrm{F}}^2 I\left\{\mathcal{E}_{t}\right\} \\
\leq &  \frac{5}{4} \left\|\mathbf{U}_1^{(t) \top} \mathbf{U}_1^{(t)}-\mathbf{U}_1^{\star \top} \mathbf{U}_1^{\star}\right\|_{\mathrm{F}}^2  I\left\{\mathcal{E}_{t}\right\}.
\end{aligned}
\end{align}
Here, (a) arises from the elementary bounds $\|\mathbf{A} \mathbf{B}\|_{\mathrm{F}} \leq\|\mathbf{A}\|_{\mathrm{F}}\|\mathbf{B}\|$.
Combining the two inequalities above, we have
\begin{align*}
\begin{aligned}
& \left\|\mathcal{M}_1\left( \nabla_\mathcal{T}f\left(\mathcal{T}^{(t)}\right)\right) \breve{\mathbf{U}}^{(t)}_1+\frac{1}{2} \mathbf{U}_1^{(t)}\left(\mathbf{U}_1^{(t) \top} \mathbf{U}_1^{(t)}-\mathbf{U}_1^{\star \top} \mathbf{U}_1^{\star}\right)\right\|_{\mathrm{F}}^2 I\left\{\mathcal{E}_{t}\right\} \\
\leq & \left(6 \lambda_{\max}^2  \left( 
\left(\epsilon_{1}^{(t)}\right)^2 + \left(\epsilon_{2}^{(t)}\right)^2\right)+ \frac{5}{8} \left\|\mathbf{U}_1^{(t) \top} \mathbf{U}_1^{(t)}-\mathbf{U}_1^{\star \top} \mathbf{U}_1^{\star}\right\|_{\mathrm{F}}^2\right)I\left\{\mathcal{E}_{t}\right\} .
\end{aligned}
\end{align*}
We define the right term in the above inequality as $J_{1,2}^{(t)}$:
\begin{align}\label{def of Q12}
\begin{aligned}
	J_{1,2}^{(t)} :=& 6 \lambda_{\max}^2  \left( \left(\epsilon_{1}^{(t)}\right)^2+\left(\epsilon_{2}^{(t)}\right)^2\right) +\frac{5}{8} \left\|\mathbf{U}_1^{(t) \top} \mathbf{U}_1^{(t)}-\mathbf{U}_1^{\star \top} \mathbf{U}_1^{\star}\right\|_{\mathrm{F}}^2.
\end{aligned}
\end{align}
For the third term on the right-hand side of Equation (\ref{equ:U-UR}), we have
\begin{align}\label{equ:first term in U eta}
\begin{aligned}
& \left\langle\mathbf{U}_1^{(t)}-\mathbf{U}_1^\star \mathbf{R}_1^{(t)}, \mathcal{M}_1\left( \nabla_\mathcal{T}f\left(\mathcal{T}^{(t)}\right)\right) \breve{\mathbf{U}}^{(t)}_1\right\rangle \\
\stackrel{(a)}{=}& \left\langle\mathbf{U}_1^{(t)} \breve{\mathbf{U}}^{(t) \top}_1-\mathbf{U}_1^\star \mathbf{R}_1^{(t)} \breve{\mathbf{U}}^{(t) \top}_1, \mathcal{M}_1\left( \nabla_\mathcal{T}f\left(\mathcal{T}^{(t)}\right)\right)\right\rangle\\
\stackrel{(\ref{equ:U deri})}{=}&\left\langle\mathcal{T}^{(t)}-\mathcal{G}^{(t)} \times{ }_1 \mathbf{U}_1^\star \mathbf{R}_1^{(t)} \times{ }_2 \mathbf{U}_2^{(t)} \times_3 \mathbf{U}_3^{(t)},  \nabla_\mathcal{T}f\left(\mathcal{T}^{(t)}\right)\right\rangle\\
\stackrel{(\ref{equ:T notation})}{=}&\left\langle\mathcal{T}^{(t)}-\mathcal{T}_1^{(t)},  \nabla_\mathcal{T}f\left(\mathcal{T}^{(t)}\right)\right\rangle.
\end{aligned}
\end{align}
Here, (a) is due to $\left\langle \mathbf{A}, \mathbf{BC} \right\rangle = \left\langle \mathbf{A}\mathbf{C}^\top , \mathbf{B} \right\rangle$.
For the last term on the right-hand side of Equation (\ref{equ:U-UR}), we have
\begin{align*}
\begin{aligned}
& \left\langle\mathbf{U}_1^{(t)}-\mathbf{U}_1^\star \mathbf{R}_1^{(t)}, \mathbf{U}_1^{(t)}\left(\mathbf{U}_1^{(t) \top} \mathbf{U}_1^{(t)}-\mathbf{U}_1^{\star \top} \mathbf{U}_1^{\star}\right)\right\rangle \\
= & \left\langle\mathbf{U}_1^{(t) \top} \mathbf{U}_1^{(t)}-\mathbf{U}_1^{(t) \top} \mathbf{U}_1^\star \mathbf{R}_1^{(t)}, \mathbf{U}_1^{(t) \top} \mathbf{U}_1^{(t)}-\mathbf{U}_1^{\star \top} \mathbf{U}_1^{\star}\right\rangle \\
= & \frac{1}{2}\left\langle\mathbf{U}_1^{(t) \top} \mathbf{U}_1^{(t)}-\mathbf{U}_1^{\star \top} \mathbf{U}_1^\star, \mathbf{U}_1^{(t) \top} \mathbf{U}_1^{(t)}-\mathbf{U}_1^{\star \top} \mathbf{U}_1^{\star}\right\rangle \\
& +\frac{1}{2}\left\langle\mathbf{U}_1^{\star \top} \mathbf{U}_1^\star-2 \mathbf{U}_1^{(t) \top} \mathbf{U}_1^\star \mathbf{R}_1^{(t)}+\mathbf{U}_1^{(t) \top} \mathbf{U}_1^{(t)}, \mathbf{U}_1^{(t) \top} \mathbf{U}_1^{(t)}-\mathbf{U}_1^{\star \top} \mathbf{U}_1^{\star}\right\rangle \\
= & \frac{1}{2}\left\|\mathbf{U}_1^{(t) \top} \mathbf{U}_1^{(t)}-\mathbf{U}_1^{\star \top} \mathbf{U}_1^{\star}\right\|_{\mathrm{F}}^2+\frac{1}{2}\left\langle\mathbf{U}_1^{(t) \top}\left(\mathbf{U}_1^{(t)}-\mathbf{U}_1^\star \mathbf{R}_1^{(t)}\right), \mathbf{U}_1^{(t) \top} \mathbf{U}_1^{(t)}-\mathbf{U}_1^{\star \top} \mathbf{U}_1^{\star}\right\rangle \\
& +\frac{1}{2}\left\langle\mathbf{U}_1^{\star \top} \mathbf{U}_1^\star-\mathbf{U}_1^{(t) \top} \mathbf{U}_1^\star \mathbf{R}_1^{(t)}, \mathbf{U}_1^{(t) \top} \mathbf{U}_1^{(t)}-\mathbf{U}_1^{\star \top} \mathbf{U}_1^{\star}\right\rangle .
\end{aligned}
\end{align*}
For the last term in the above equation, we have
\begin{align*}
\begin{aligned}
&\left\langle\mathbf{U}_1^{\star \top} \mathbf{U}_1^\star-\mathbf{U}_1^{(t) \top} \mathbf{U}_1^\star \mathbf{R}_1^{(t)}, \mathbf{U}_1^{(t) \top} \mathbf{U}_1^{(t)}-\mathbf{U}_1^{\star \top} \mathbf{U}_1^{\star}\right\rangle \\
\stackrel{(a)}{=} & \left\langle\mathbf{U}_1^{\star \top} \mathbf{U}_1^\star-\mathbf{R}_1^{(t) \top} \mathbf{U}_1^{\star \top} \mathbf{U}_1^{(t)}, \mathbf{U}_1^{(t) \top} \mathbf{U}_1^{(t)}-\mathbf{U}_1^{\star \top} \mathbf{U}_1^{\star}\right\rangle \\
\stackrel{(b)}{=} & \left\langle\mathbf{R}_1^{(t) \top} \mathbf{U}_1^{\star \top} \mathbf{U}_1^\star \mathbf{R}_1^{(t)}-\mathbf{R}_1^{(t) \top} \mathbf{U}_1^{\star \top} \mathbf{U}_1^{(t)}, \mathbf{U}_1^{(t) \top} \mathbf{U}_1^{(t)}-\mathbf{U}_1^{\star \top} \mathbf{U}_1^{\star}\right\rangle \\
=& \left\langle\left(\mathbf{U}_1^\star \mathbf{R}_1^{(t)}\right)^{\top}\left(\mathbf{U}_1^\star \mathbf{R}_1^{(t)}-\mathbf{U}_1^{(t)}\right), \mathbf{U}_1^{(t) \top} \mathbf{U}_1^{(t)}-\mathbf{U}_1^{\star \top} \mathbf{U}_1^{\star}\right\rangle,
\end{aligned}
\end{align*}
where (a) is due to the fact that $\langle\mathbf{A}, \mathbf{B}\rangle=\left\langle\mathbf{A}^{\top}, \mathbf{B}\right\rangle$ for symmetric matrix $\mathbf{B}$ and (b) holds because $\mathbf{U}_1^{\star \top} \mathbf{U}_1^\star= \mathbf{I}_{r_1}$ and $\mathbf{R}_1^{(t)\top} \mathbf{R}_1^{(t)}=\mathbf{I}_{r_1}$.
Combing the above two equations, we further have
\begin{align}\label{equ:second term in U eta}
\begin{aligned}
& \left\langle\mathbf{U}_1^{(t)}-\mathbf{U}_1^\star \mathbf{R}_1^{(t)}, \mathbf{U}_1^{(t)}\left(\mathbf{U}_1^{(t) \top} \mathbf{U}_1^{(t)}-\mathbf{U}_1^{\star \top} \mathbf{U}_1^{\star}\right)\right\rangle I\left\{\mathcal{E}_{t}\right\} \\
= & \frac{1}{2}\left\|\mathbf{U}_1^{(t) \top} \mathbf{U}_1^{(t)}-\mathbf{U}_1^{\star \top} \mathbf{U}_1^{\star}\right\|_{\mathrm{F}}^2 I\left\{\mathcal{E}_{t}\right\} \\
& +\frac{1}{2}\left\langle\left(\mathbf{U}_1^\star \mathbf{R}_1^{(t)}-\mathbf{U}_1^{(t)}\right)^{\top}\left(\mathbf{U}_1^\star \mathbf{R}_1^{(t)}-\mathbf{U}_1^{(t)}\right), \mathbf{U}_1^{(t) \top} \mathbf{U}_1^{(t)}-\mathbf{U}_1^{\star \top} \mathbf{U}_1^{\star}\right\rangle I\left\{\mathcal{E}_{t}\right\}  \\
\geq & \frac{1}{2}\left( \left\|\mathbf{U}_1^{(t) \top} \mathbf{U}_1^{(t)}-\mathbf{U}_1^{\star \top} \mathbf{U}_1^{\star}\right\|_{\mathrm{F}}^2 -\left\|\mathbf{U}_1^\star \mathbf{R}_1^{(t)}-\mathbf{U}_1^{(t)}\right\|_{\mathrm{F}}^2 \cdot\left\|\mathbf{U}_1^{(t) \top} \mathbf{U}_1^{(t)}-\mathbf{U}_1^{\star \top} \mathbf{U}_1^{\star}\right\|_{\mathrm{F}}\right) I\left\{\mathcal{E}_{t}\right\} \\
\stackrel{(a)}{\geq}  & \frac{1}{2}\left(\left\|\mathbf{U}_1^{(t) \top} \mathbf{U}_1^{(t)}-\mathbf{U}_1^{\star \top} \mathbf{U}_1^{\star}\right\|_{\mathrm{F}}^2 -\frac{1}{2}\left\|\mathbf{U}_1^{(t) \top} \mathbf{U}_1^{(t)}-\mathbf{U}_1^{\star \top} \mathbf{U}_1^{\star}\right\|_{\mathrm{F}}^2-\frac{1}{2}\left\|\mathbf{U}_1^{(t)}-\mathbf{U}_1^\star \mathbf{R}_1^{(t)}\right\|_{\mathrm{F}}^4\right)I\left\{\mathcal{E}_{t}\right\}  \\
\stackrel{(b)}{\geq} & \frac{1}{4}\left\|\mathbf{U}_1^{(t) \top} \mathbf{U}_1^{(t)}-\mathbf{U}_1^{\star \top} \mathbf{U}_1^{\star}\right\|_{\mathrm{F}}^2I\left\{\mathcal{E}_{t}\right\} -\frac{1}{4} J^{(t)}\left\|\mathbf{U}_1^{(t)}-\mathbf{U}_1^\star \mathbf{R}_1^{(t)}\right\|_{\mathrm{F}}^2I\left\{\mathcal{E}_{t}\right\}  \\
\stackrel{(c)}{\geq} & \frac{1}{4}\left\|\mathbf{U}_1^{(t) \top} \mathbf{U}_1^{(t)}-\mathbf{U}_1^{\star \top} \mathbf{U}_1^{\star}\right\|_{\mathrm{F}}^2I\left\{\mathcal{E}_{t}\right\} -\frac{c_d\lambda_{\min}^{2}}{4} \left\|\mathbf{U}_1^{(t)}-\mathbf{U}_1^\star \mathbf{R}_1^{(t)}\right\|_{\mathrm{F}}^2 I\left\{\mathcal{E}_{t}\right\} ,
\end{aligned}
\end{align}
where (a) is due to the fact that $ab\leq\frac{a^2+b^2}{2},$  (b) comes from the definition of $J^{(t)}$ in Equation (\ref{cond:def of benign region}), and (c) holds because of 
  $\left\|\mathbf{U}_1^{(t)}-\mathbf{U}_1^\star \mathbf{R}_1^{(t)}\right\|_{\mathrm{F}}^2I\left\{\mathcal{E}_{t}\right\}  \leq J^{(t)}I\left\{\mathcal{E}_{t}\right\}  \leq c_d\lambda_{\min}^2.$
Combining previous Equation (\ref{equ:first term in U eta}) and (\ref{equ:second term in U eta}), we define $J_{1, 1}^{(t)}$:
\begin{align}\label{equ:def of Q11}
\begin{aligned}
      J_{1, 1}^{(t)} = \left\langle\mathcal{T}^{(t)}-\mathcal{T}_1^{(t)}, \nabla_\mathcal{T}f\left(\mathcal{T}^{(t)}\right)\right\rangle + \frac{1}{8} \left(\left\|\mathbf{U}_1^{(t) \top} \mathbf{U}_1^{(t)}-\mathbf{U}_1^{\star \top} \mathbf{U}_1^{\star}\right\|_{\mathrm{F}}^2 - c_d\lambda_{\min}^{2}\left\|\mathbf{U}_1^{(t)}-\mathbf{U}_1^{\star} \mathbf{R}_1^{(t)}\right\|_{\mathrm{F}}^2 \right).
\end{aligned}
\end{align}
Therefore, combining previous Equation (\ref{equ:U-UR}), (\ref{def of Q12}), and (\ref{equ:def of Q11}), we obtain
\begin{align*}
\left\|{\mathbf{U}}_1^{(t+1)}-\mathbf{U}_1^\star \mathbf{R}_1^{(t)}\right\|_{\mathrm{F}}^2 I\left\{\mathcal{E}_{t}\right\} \leq\left\|\mathbf{U}_1^{(t)}-\mathbf{U}_1^\star \mathbf{R}_1^{(t)}\right\|_{\mathrm{F}}^2I\left\{\mathcal{E}_{t}\right\} -2 \eta_{t+1}J_{1, 1}^{(t)}I\left\{\mathcal{E}_{t}\right\} +\eta_{t+1}^2J_{1,2}^{(t)}I\left\{\mathcal{E}_{t}\right\}  .
\end{align*}
 Then more generally, for $k\in[3]$, we have
\begin{align*}
&\left\|{\mathbf{U}}_k^{(t+1)}-\mathbf{U}_k^\star \mathbf{R}_k^{(t)}\right\|_{\mathrm{F}}^2 I\left\{\mathcal{E}_{t}\right\} \leq\left\|\mathbf{U}_k^{(t)}-\mathbf{U}_k^\star \mathbf{R}_k^{(t)}\right\|_{\mathrm{F}}^2I\left\{\mathcal{E}_{t}\right\} -2 \eta_{t+1}J_{k, 1}^{(t)}I\left\{\mathcal{E}_{t}\right\} +\eta_{t+1}^2J_{k, 2}^{(t)}I\left\{\mathcal{E}_{t}\right\} ,
\end{align*}
where
\begin{align*}
  J_{k, 1}^{(t)} = &\left\langle\mathcal{T}^{(t)}-\mathcal{T}_k^{(t)}, \nabla f\left(\mathcal{T}^{(t)}\right)\right\rangle + \frac{1}{8} \left( \left\|\mathbf{U}_k^{(t) \top} \mathbf{U}_k^{(t)}-\mathbf{U}_k^{\star \top} \mathbf{U}_k^{\star}\right\|_{\mathrm{F}}^2-c_d\lambda_{\min}^{2} \left\|\mathbf{U}_k^{(t)}-\mathbf{U}_k^{\star} \mathbf{R}_k^{(t)}\right\|_{\mathrm{F}}^2\right), \\
  J_{k, 2}^{(t)} = &6 \lambda_{\max}^2  \left( \left(\epsilon_{1}^{(t)}\right)^2+\left(\epsilon_{2}^{(t)}\right)^2\right)+\frac{5}{8} \left\|\mathbf{U}_k^{(t) \top} \mathbf{U}_k^{(t)}-\mathbf{U}_k^{\star \top} \mathbf{U}_k^{\star}\right\|_{\mathrm{F}}^2. 
\end{align*}

\subsection{Proof of Lemma \ref{lem:contraction of G}}\label{sec:Proof of Lemma contraction of G}

\begin{proof}
To streamline our analysis, we introduce the following notations:
\begin{align}\label{equ:T notation}
\begin{aligned}
& \mathcal{T}^{(t)}= \mathcal{G}^{(t)} \times_{k\in[3]} \mathbf{U}_k^{(t)}, \\
& \mathcal{T}_{\mathcal{G}}^{(t)}= \mathcal{G}^\star  \times_{k\in[3]} \mathbf{U}_k^{(t)} \mathbf{R}_k^{(t) \top},  \\
& \mathcal{T}_k^{(t)}=\mathcal{G}^{(t)} \times_k \mathbf{U}_k^{\star} \mathbf{R}_k^{(t)} \times_{j\neq k} \mathbf{U}_{j}^{(t)}, \quad k\in[3],
\end{aligned}
\end{align}
	We have the following decomposition by plugging in the gradient in Algorithm \ref{alg:Low rank tensor SGD}.
\begin{align}\label{equ:G sgd expansion}
\begin{aligned}
& \left\|{\mathcal{G}}^{(t+1)}- \mathcal{G}^\star  \times_{k\in[3]} \mathbf{R}_k^{(t) \top}\right\|_{\mathrm{F}}^2 \\
= & \left\|\mathcal{G}^{(t)}- \mathcal{G}^\star  \times_{k\in[3]} \mathbf{R}_k^{(t) \top} -\eta_{t+1} \nabla_\mathcal{T}f\left(\mathcal{T}^{(t)}\right) \times_{k\in[3]} \mathbf{U}_k^{(t) \top} \right\|_{\mathrm{F}}^2 \\
= & \left\|\mathcal{G}^{(t)}- \mathcal{G}^\star  \times_{k\in[3]} \mathbf{R}_k^{(t) \top} \right\|_{\mathrm{F}}^2+\eta_{t+1}^2\left\| \nabla_\mathcal{T}f\left(\mathcal{T}^{(t)}\right) \times_{k\in[3]} \mathbf{U}_k^{(t) \top} \right\|_{\mathrm{F}}^2 \\
& -2 \eta_{t+1}\left\langle\mathcal{G}^{(t)}- \mathcal{G}^\star  \times_{k\in[3]} \mathbf{R}_k^{(t) \top} ,  \nabla_\mathcal{T}f\left(\mathcal{T}^{(t)}\right) \times_{k\in[3]} \mathbf{U}_k^{(t) \top} \right\rangle .
\end{aligned}
\end{align}
For the last term in the inequality above, we have
\begin{align*}
\begin{aligned}
& \left\langle\mathcal{G}^{(t)}- \mathcal{G}^\star  \times_{k\in[3]} \mathbf{R}_k^{(t) \top} ,  \nabla_\mathcal{T}f\left(\mathcal{T}^{(t)}\right) \times_{k\in[3]} \mathbf{U}_k^{(t) \top} \right\rangle \\
= & \left\langle\mathcal{M}_1\left(\mathcal{G}^{(t)}\right)-\mathbf{R}_1^{(t) \top} \mathcal{M}_1\left(\mathcal{G}^\star\right)\left(\mathbf{R}_3^{(t)} \otimes \mathbf{R}_2^{(t)}\right), \mathbf{U}_1^{(t) \top} \mathcal{M}_1\left( \nabla_\mathcal{T}f\left(\mathcal{T}^{(t)}\right)\right)\left(\mathbf{U}_3^{(t)} \otimes \mathbf{U}_2^{(t)}\right)\right\rangle \\
= & \left\langle\mathbf{U}_1^{(t)} \mathcal{M}_1\left(\mathcal{G}^{(t)}\right)\left(\mathbf{U}_3^{(t)} \otimes \mathbf{U}_2^{(t)}\right)^{\top}-\mathbf{U}_1^{(t)} \mathbf{R}_1^{(t) \top} \mathcal{M}_1\left(\mathcal{G}^\star\right)\left(\mathbf{R}_3^{(t)} \mathbf{U}_3^{(t) \top} \otimes \mathbf{R}_2^{(t)} \mathbf{U}_2^{(t) \top}\right),\mathcal{M}_1\left( \nabla_\mathcal{T}f\left(\mathcal{T}^{(t)}\right)\right)\right\rangle \\
= & \left\langle\mathcal{T}^{(t)}-\mathcal{G}^\star  \times_{k\in[3]} \mathbf{U}_k^{(t)} \mathbf{R}_k^{(t) \top},  \nabla_\mathcal{T}f\left(\mathcal{T}^{(t)}\right)\right\rangle \\
\stackrel{(\ref{equ:T notation})}{=} & \left\langle\mathcal{T}^{(t)}-\mathcal{T}_{\mathcal{G}}^{(t)},  \nabla_\mathcal{T}f\left(\mathcal{T}^{(t)}\right)\right\rangle.
\end{aligned}
\end{align*}
We define $J_{\mathcal{G}, 1}^{(t)}$ as:
\begin{align}\label{def of Gg1}
 J_{\mathcal{G}, 1}^{(t)}:= \left\langle\mathcal{T}^{(t)}-\mathcal{T}_{\mathcal{G}}^{(t)},  \nabla_\mathcal{T}f\left(\mathcal{T}^{(t)}\right)\right\rangle.
\end{align}
For the second term in Equation (\ref{equ:G sgd expansion}), by the duality of Frobenius norm,  we can apply Lemma \ref{lem:def of Xi} to obtain the following result:
\begin{align*}
\begin{aligned}
& \left\| \nabla_\mathcal{T}f\left(\mathcal{T}^{\star}\right) \times_{k\in[3]}\mathbf{U}_k^{(t) \top} \right\|_{\mathrm{F}}\\
=& \sup _{\substack{\mathcal{G} \in \mathbb{R}^{r_1 \times r_2 \times r_3} \\
\|\mathcal{G}\|_{\mathrm{F}} \leq 1}}\left\langle \nabla_\mathcal{T}f\left(\mathcal{T}^{\star}\right) \times_{k\in[3]}\mathbf{U}_k^{(t) \top} , \mathcal{G}\right\rangle \\
= & \sup _{\substack{\mathcal{G} \in \mathbb{R}^{r_1 \times r_2 \times r_3} \\
\|\mathcal{G}\|_{\mathrm{F}} \leq 1}}\left\langle \nabla_\mathcal{T}f\left(\mathcal{T}^{\star}\right), \mathcal{G} \times_{k\in[3]}\mathbf{U}_k^{(t) }\right\rangle \\
\stackrel{(\ref{equ:def of Xi})}{\leq} & \left\|\mathbf{U}_1^{(t)}\right\| \cdot\left\|\mathbf{U}_2^{(t)}\right\| \cdot\left\|\mathbf{U}_3^{(t)}\right\| \cdot \epsilon_1^{(t)}.
\end{aligned}
\end{align*}
By Cauchy-Schwarz inequality, we can have
\begin{align*}
  &\left\| \rbr{\nabla_\mathcal{T}f\left(\mathcal{T}^{(t)}\right)- \nabla_\mathcal{T}f\left(\mathcal{T}^{\star}\right)}\times_{k\in[3]} \mathbf{U}_k^{(t) \top} \right\|_{\mathrm{F}} \leq \left\|\mathbf{U}_1^{(t)}\right\| \cdot\left\|\mathbf{U}_2^{(t)}\right\| \cdot\left\|\mathbf{U}_3^{(t)}\right\|\epsilon_{2}^{(t)}.
\end{align*}
we also have
\begin{align}\label{equ:J2-3}
\begin{aligned}
& \left\| \nabla_\mathcal{T}f\left(\mathcal{T}^{(t)}\right) \times_{k\in[3]} \mathbf{U}_k^{(t) \top} \right\|_{\mathrm{F}}^2I\left\{\mathcal{E}_{t}\right\}  \\
\leq & 2 \left\| \nabla_\mathcal{T}f\left(\mathcal{T}^{\star}\right) \times_{k\in[3]}\mathbf{U}_k^{(t) \top}  \right\|_{\mathrm{F}}^2I\left\{\mathcal{E}_{t}\right\}  \\
&+2 \left\|\left[\nabla_\mathcal{T}f\left(\mathcal{T}^{(t)}\right)- \nabla_\mathcal{T}f\left(\mathcal{T}^{\star}\right)\right] \times_{k\in[3]}\mathbf{U}_k^{(t) \top} \right\|_{\mathrm{F}}^2I\left\{\mathcal{E}_{t}\right\} \\
\leq & 2\left\|\mathbf{U}_1^{(t)}\right\|^2 \left\|\mathbf{U}_2^{(t)}\right\|^2 \left\|\mathbf{U}_3^{(t)}\right\|^2 \left(\epsilon_1^{(t)}\right)^2 I\left\{\mathcal{E}_{t}\right\}  + 2 \left\|\mathbf{U}_1^{(t)}\right\|^2\left\|\mathbf{U}_2^{(t)}\right\|^2\left\|\mathbf{U}_3^{(t)}\right\|^2\left(\epsilon_{2}^{(t)}\right)^2I\left\{\mathcal{E}_{t}\right\} \\
\stackrel{(\ref{equ:benign region})}{\leq} & 3 \left( \left(\epsilon_{1}^{(t)}\right)^2+\left(\epsilon_{2}^{(t)}\right)^2\right)I\left\{\mathcal{E}_{t}\right\} .
\end{aligned}
\end{align}
We define $J_{\mathcal{G}, 2}^{(t)}$ as:
\begin{align}\label{def of Gg2}
 J_{\mathcal{G}, 2}^{(t)}:=  3 \left( \left(\epsilon_{1}^{(t)}\right)^2+\left(\epsilon_{2}^{(t)}\right)^2\right) .
\end{align}
Therefore, combining the Equation (\ref{equ:G sgd expansion}), (\ref{def of Gg1}), and (\ref{def of Gg2}), we have
\begin{align}\label{equ:one step G}
\begin{aligned}
& \left\|\mathcal{G}^{(t+1)}-\mathcal{G}  \times_{k\in[3]} \mathbf{R}_k^{(t) \top} \right\|_{\mathrm{F}}^2I\left\{\mathcal{E}_{t}\right\}  \\
\leq & \left\|\mathcal{G}^{(t)}-\mathcal{G}  \times_{k\in[3]} \mathbf{R}_k^{(t) \top}\right\|_{\mathrm{F}}^2I\left\{\mathcal{E}_{t}\right\} -2 \eta_{t+1}J_{\mathcal{G}, 1}^{(t)}I\left\{\mathcal{E}_{t}\right\} +\eta_{t+1}^2J_{\mathcal{G}, 2}^{(t)}I\left\{\mathcal{E}_{t}\right\}.
\end{aligned}
\end{align}
\end{proof}

\subsection{Proof of Lemma \ref{lem:bound of JG1}}\label{sec:Proof of Lemma bound of JG1}

\begin{proof}
	 By definitions of $J_{\mathcal{G}, 1}^{(t)}$ in Equation (\ref{def of Gg1}) and $J_{k, 1}^{(t)}$ in Equation (\ref{equ:def of Q11}),  we have
\begin{align}\label{equ:C.15}
\begin{aligned}
 J_{\mathcal{G}, 1}^{(t)}+\sum_{k=1}^3 J_{k, 1}^{(t)} 
= & \left\langle 4 \mathcal{T}^{(t)}-\mathcal{T}_\mathcal{G}^{(t)}-\sum_{k=1}^3 \mathcal{T}_k^{(t)},  \nabla_\mathcal{T}f\left(\mathcal{T}^{(t)}\right)\right\rangle \\
&+ \frac{1}{8} \sum_{k=1}^3\left( \left\|\mathbf{U}_k^{(t) \top} \mathbf{U}_k^{(t)}-\mathbf{U}_k^{\star \top} \mathbf{U}_k^{\star}\right\|_{\mathrm{F}}^2 - c_d\lambda_{\min}^2 \left\|\mathbf{U}_k^{(t)}-\mathbf{U}_k^{\star} \mathbf{R}_k^{(t)}\right\|_{\mathrm{F}}^2\right). 
\end{aligned}
\end{align}
Utilizing Lemma E.3 from \cite{han2022optimal}, we can represent the first term on the right-hand side of Equation (\ref{equ:C.15}) as follows:
\begin{align*}
4 \mathcal{T}^{(t)}-\mathcal{T}_{\mathcal{G}}^{(t)}-\sum_{k=1}^3 \mathcal{T}_k^{(t)}=\mathcal{T}^{(t)}-\mathcal{T}^\star+\mathcal{H}_{\varepsilon}^{(t)},
\end{align*}
where
\begin{align}\label{equ:D.10}
\begin{aligned}
\mathcal{H}_{\varepsilon}^{(t)}= & \mathcal{G}^{\star} \times_{k\in[3]} \mathbf{H}_k^{(t)}  +
 \sum_{k=1}^3 \mathcal{G}^{\star}  \times_k \mathbf{U}_k^{(t)} \mathbf{R}_k^{(t)\top} \times_{j\neq k} \mathbf{H}_{j}^{(t)} \\
& +\sum_{k=1}^3 \mathcal{H}_{\mathcal{G}}^{(t)} \times_k \mathbf{H}_k^{(t)} 
\times_{j\neq k} \mathbf{U}_{j}^{(t)} \mathbf{R}_{j}^{(t)\top},\\
\mathbf{H}_k^{(t)}=&\mathbf{U}_k^\star-\mathbf{U}_k^{(t)} \mathbf{R}_k^{(t)\top}, \quad k\in[3], \\
\mathcal{H}_{\mathcal{G}}^{(t)}=&\mathcal{G}^\star- \mathcal{G}^{(t)} \times_{k\in[3]} \mathbf{R}_k^{(t) \top}.
\end{aligned}
\end{align}
Then it follows that
\begin{align}\label{equ:D.11}
\begin{aligned}
& \mathbb{E}\left[\left\langle 4 \mathcal{T}^{(t)}-\mathcal{T}_{\mathcal{G}}^{(t)}-\sum_{k=1}^3 \mathcal{T}_k^{(t)}, \nabla_\mathcal{T}f\left(\mathcal{T}^{(t)}\right)\right\rangle \mid\mathcal{F}_t\right]\\
=& \mathbb{E}\left[\left\langle\mathcal{T}^{(t)}-\mathcal{T}^\star, \nabla_\mathcal{T}f\left(\mathcal{T}^{(t)}\right)-\nabla_\mathcal{T}f\left(\mathcal{T}^{\star}\right)\right\rangle \mid\mathcal{F}_t\right]
+ \mathbb{E}\left[\left\langle\mathcal{H}_{\varepsilon}, \nabla_\mathcal{T}f\left(\mathcal{T}^{(t)}\right)-\nabla_\mathcal{T}f\left(\mathcal{T}^{\star}\right)\right\rangle \mid\mathcal{F}_t\right]\\
&+ \mathbb{E}\left[ \left\langle 4 \mathcal{T}^{(t)}-\mathcal{T}_{\mathcal{G}}^{(t)}-\sum_{k=1}^3 \mathcal{T}_k^{(t)}, \nabla_\mathcal{T}f\left(\mathcal{T}^{\star}\right)\right\rangle \mid\mathcal{F}_t\right]\\
\stackrel{(a)}{\geq} & \mathbb{E}\left[ \left\langle\mathcal{T}^{(t)}-\mathcal{T}^\star, \nabla_\mathcal{T}f\left(\mathcal{T}^{(t)}\right)-\nabla_\mathcal{T}f\left(\mathcal{T}^{\star}\right)\right\rangle \mid\mathcal{F}_t\right] - \left|\mathbb{E}\left[\left\langle\mathcal{H}_{\varepsilon}, \nabla_\mathcal{T}f\left(\mathcal{T}^{(t)}\right)-\nabla_\mathcal{T}f\left(\mathcal{T}^{\star}\right)\right\rangle \mid\mathcal{F}_t \right]\right|.
\end{aligned}
\end{align}
Here, (a) arise from $ \mathbb{E} \left[\nabla_\mathcal{T}f\left(\mathcal{T}^{\star}\right) \right] = \mathbb{E} \left[\xi_{t+1}\mathcal{X}_{t+1}\right] = 0$.
For the first term in Equation (\ref{equ:D.11}), since $\nabla_\mathcal{T}f\left(\mathcal{T}^{(t)}-\mathcal{T}^{\star}\right)
=\left\langle \mathcal{X}_{t+1},\mathcal{T}^{(t)}-\mathcal{T}^\star\right\rangle\mathcal{X}_{t+1}$, we firstly have
\begin{align}\label{equ:D.12}
\begin{aligned}
& \mathbb{E}\left[\left\langle\mathcal{T}^{(t)}-\mathcal{T}^\star, \nabla_\mathcal{T}f\left(\mathcal{T}^{(t)}\right)-\nabla_\mathcal{T}f\left(\mathcal{T}^{\star}\right)\right\rangle\mid\mathcal{F}_t\right]I\left\{\mathcal{E}_{t}\right\}\\
=&\mathbb{E}\left[\left\langle\mathcal{T}^{(t)}-\mathcal{T}^\star, \left\langle \mathcal{X}_{t+1},\mathcal{T}^{(t)}-\mathcal{T}^\star\right\rangle\mathcal{X}_{t+1}\right\rangle\mid\mathcal{F}_{t}\right]I\left\{\mathcal{E}_{t}\right\} \\
 \stackrel{(\ref{equ:X inner product})}{=} & \left\langle\mathcal{T}^{(t)}-\mathcal{T}^\star,\mathcal{T}^{(t)}-\mathcal{T}^\star\right\rangle I\left\{\mathcal{E}_{t}\right\}
 =\left\|\mathcal{T}^{(t)}-\mathcal{T}^\star\right\|_{\mathrm{F}}^2I\left\{\mathcal{E}_{t}\right\} .
\end{aligned}
\end{align}
For the second term in Equation (\ref{equ:D.11}), we need to use the fact that $\mathcal{H}_{\varepsilon}$ is a summation of rank- $\left(r_1, r_2, r_3\right)$ tensors:
\begin{align*}
\begin{aligned}
&\mathbb{E}\left[\left\langle\mathcal{H}_{\varepsilon}^{(t)}, \nabla_\mathcal{T}f\left(\mathcal{T}^{(t)}\right)-\nabla_\mathcal{T}f\left(\mathcal{T}^{\star}\right)\right\rangle \mid\mathcal{F}_{t}\right]I\left\{\mathcal{E}_{t}\right\}\\
 = & \mathbb{E}\left[\left\langle\mathcal{H}_{\varepsilon}^{(t)}, \left\langle \mathcal{X}_{t+1},\mathcal{T}^{(t)}-\mathcal{T}^\star\right\rangle\mathcal{X}_{t+1}\right\rangle \mid\mathcal{F}_{t}\right]I\left\{\mathcal{E}_{t}\right\} \\
\stackrel{(\ref{equ:X inner product})}{=} &\left\langle\mathcal{H}_{\varepsilon}^{(t)}, \mathcal{T}^{(t)}-\mathcal{T}^\star\right\rangle I\left\{\mathcal{E}_{t}\right\}
\stackrel{(a)}{\leq} \left\|\mathcal{H}_{\varepsilon}^{(t)}\right\|_{\mathrm{F}}\left\|\mathcal{T}^{(t)}-\mathcal{T}^\star\right\|_{\mathrm{F}}I\left\{\mathcal{E}_{t}\right\} .
\end{aligned}
\end{align*}
where (a) is due to Cauchy-Schwarz inequality and $\left\|\mathcal{H}_{\varepsilon}^{(t)}\right\|_\mathrm{F}$ is derived from Equation (\ref{equ:D.10}):
\begin{align*}
\begin{aligned}
\left\|\mathcal{H}_{\varepsilon}^{(t)}\right\|
= & \left\|\mathcal{G}^{\star} \times_{k\in[3]} \mathbf{R}_k^{(t) \top}\times_1 \mathbf{H}_1^{(t)}\times_2 \mathbf{H}_2^{(t)}\times_3 \mathbf{H}_3^{(t)}\right\| +
 \sum_{k=1}^3\left\| \mathcal{G}^{\star}  \times_k \mathbf{U}_k^{(t)} \mathbf{R}_k^{(t)\top} \times_{k+1} \mathbf{H}_{k+1}^{(t)} \times_{k+2} \mathbf{H}_{k+2}^{(t)}\right\| \\
& +\sum_{k=1}^3 \left\|\mathcal{H}_{\mathcal{G}}^{(t)} \times_k \mathbf{H}_k^{(t)} \times_{k+1} \mathbf{U}_{k+1}^{(t)} \mathbf{R}_{k+1}^{(t)\top} \times_{k+2} \mathbf{U}_{k+2}^{(t)} \mathbf{R}_{k+2}^{(t)\top}\right\|.
\end{aligned}
\end{align*}
Due to Equation (\ref{equ:benign region}) and Lemma E.3 in \cite{han2022optimal}, we have
\begin{align}\label{equ:bound of H eps}
\begin{aligned}
\left\|\mathcal{H}_{\varepsilon}^{(t)}\right\|I\left\{\mathcal{E}_{t}\right\}  \stackrel{(\ref{equ:benign region})}{\leq}
& \left(1.01 \lambda_{\max} \left(J^{(t)}\right)^{3 / 2} + 3(1.01 )^2 J^{(t)}
+3 \left( 1.01 \lambda_{\max} \right)(1.01 ) J^{(t)}\right)I\left\{\mathcal{E}_{t}\right\}  \\
 \stackrel{(\ref{cond:def of benign region})}{\leq}&\left(1.01 \lambda_{\max} \sqrt{c_d}\lambda_{\min} + 3 \cdot 1.01^2 + 3 \cdot 1.01^2\lambda_{\max} \right) J^{(t)} I\left\{\mathcal{E}_{t}\right\} \\
\leq & 6.25 J^{(t)} I\left\{\mathcal{E}_{t}\right\} .
\end{aligned}
\end{align}
The inequality in the last line is because we assume $\lambda_{\max} = 1$. Combining all of the above, we obtain:
\begin{align}\label{equ:D.13}
\begin{aligned}
& \left|\mathbb{E}\left[\left\langle\mathcal{H}_{\varepsilon}^{(t)}, \nabla_\mathcal{T}f\left(\mathcal{T}^{(t)}\right)-\nabla_\mathcal{T}f\left(\mathcal{T}^{\star}\right)\right\rangle\mid\mathcal{F}_{t}\right]\right| I\left\{\mathcal{E}_{t}\right\}\\ 
\leq &  \left\|\mathcal{H}_{\varepsilon}^{(t)}\right\|\left\|\mathcal{T}^{(t)}-\mathcal{T}^\star\right\|_{\mathrm{F}} I\left\{\mathcal{E}_{t}\right\} 
\leq  6.25\left\|\mathcal{T}^{(t)}-\mathcal{T}^\star\right\|_{\mathrm{F}}  J^{(t)} I\left\{\mathcal{E}_{t}\right\} \\
\stackrel{(a)}{\leq} & \left(\frac{1}{2}\left\|\mathcal{T}^{(t)}-\mathcal{T}^\star\right\|_{\mathrm{F}}^2+ \frac{625}{32} \left(J^{(t)}\right)^2\right) I\left\{\mathcal{E}_{t}\right\} \\
\stackrel{(\ref{equ:benign region})}{\leq} & \left(\frac{1}{2}\left\|\mathcal{T}^{(t)}-\mathcal{T}^\star\right\|_{\mathrm{F}}^2 + \frac{625}{32} c_d\lambda_{\min}^2 J^{(t)}\right)I\left\{\mathcal{E}_{t}\right\},
\end{aligned}
\end{align}
where (a) is due to the fact that $ab\leq\frac{1}{2}\left(a^2+b^2\right)$.
Combining Equation (\ref{equ:D.12}) and (\ref{equ:D.13}) into (\ref{equ:D.11}), we obtain:
\begin{align}\label{equ:equ:D.14-2}
\begin{aligned}
	&\mathbb{E}\left[\left\langle 4 \mathcal{T}^{(t)}-\mathcal{T}_{\mathcal{G}}^{(t)}-\sum_{k=1}^3 \mathcal{T}_k^{(t)}, \nabla_\mathcal{T}f\left(\mathcal{T}^{(t)}\right)\right\rangle	\mid\mathcal{F}_t\right]I\left\{\mathcal{E}_{t}\right\}\\
\geq & \left(\frac{1}{2}\left\|\mathcal{T}^{(t)}-\mathcal{T}^\star\right\|_{\mathrm{F}}^2 - \frac{625}{32}c_d \lambda_{\min}^{2} J^{(t)}\right)I\left\{\mathcal{E}_{t}\right\}.
\end{aligned}
\end{align}
By Lemma E.2 in \cite{han2022optimal}, we have
\begin{align}\label{equ:J and T equ}
\begin{aligned}
 J^{(t)} &\leq 7 \left\|\mathcal{T}^{(t)}-\mathcal{T}^\star\right\|_{\mathrm{F}}^2+38 \sum_{k=1}^3 \left\|\mathbf{U}_k^{(t)}-\mathbf{U}_k^\star \mathbf{R}_k^{(t)}\right\|_{\mathrm{F}}^2\\
&\leq 480 \lambda_{\min}^{-2}\left\|\mathcal{T}^{(t)}-\mathcal{T}^\star\right\|_{\mathrm{F}}^2+80 \sum_{k=1}^3\left\|\mathbf{U}_k^{(t)\top} \mathbf{U}_k^{(t)}-\mathbf{U}_k^{\star\top} \mathbf{U}_k^{\star}\right\|_{\mathrm{F}}^2.
\end{aligned}
\end{align}
Now combining Equation (\ref{equ:C.15}) and (\ref{equ:equ:D.14-2}), we obtain that
\begin{align*}
\begin{aligned}
&\mathbb{E}\left[J_{\mathcal{G}, 1}^{(t)}+\sum_{k=1}^3 J_{k, 1}^{(t)}\mid\mathcal{F}_t\right]I\left\{\mathcal{E}_{t}\cap\mathcal{E}_t^\epsilon\right\} \\
 \geq & \left(\frac{1}{2}\left\|\mathcal{T}^{(t)}-\mathcal{T}^\star\right\|_{\mathrm{F}}^2-\frac{625}{32}c_d \lambda_{\min}^{2} J^{(t)}\right)I\left\{\mathcal{E}_{t}\right\} 
 \\& +\frac{1}{8} \sum_{k=1}^3\left( \left\|\mathbf{U}_k^{(t) \top} \mathbf{U}_k^{(t)}-\mathbf{U}_k^{\star \top} \mathbf{U}_k^{\star}\right\|_{\mathrm{F}}^2 - c_d\lambda_{\min}^2 \left\|\mathbf{U}_k^{(t)}-\mathbf{U}_k^{\star} \mathbf{R}_k^{(t)}\right\|_{\mathrm{F}}^2\right) I\left\{\mathcal{E}_{t}\right\} \\
= & \frac{1}{2}\left(\left\|\mathcal{T}^{(t)}-\mathcal{T}^\star\right\|_{\mathrm{F}}^2+\frac{1}{6} \sum_{k=1}^3\left\|\mathbf{U}_k^{(t) \top} \mathbf{U}_k^{(t)}-\mathbf{U}_k^{\star \top} \mathbf{U}_k^{\star}\right\|_{\mathrm{F}}^2\right)I\left\{\mathcal{E}_{t}\right\}  \\
&  -\left(\frac{c_d\lambda_{\min}^2}{8} \sum_{k=1}^3 \left\|\mathbf{U}_k^{(t)}-\mathbf{U}_k^{\star} \mathbf{R}_k^{(t)}\right\|_{\mathrm{F}}^2+\frac{625}{32}c_d \lambda_{\min}^{2} J^{(t)}\right) I\left\{\mathcal{E}_{t}\right\}  \\
& + \frac{1}{24} \sum_{k=1}^3\left(\left\|\mathbf{U}_k^{(t) \top} \mathbf{U}_k^{(t)}-\mathbf{U}_k^{\star \top} \mathbf{U}_k^{\star}\right\|_{\mathrm{F}}^2\right)I\left\{\mathcal{E}_{t}\right\}  \\
 \stackrel{(\ref{equ:J and T equ})}{\geq}&\left( \frac{\lambda_{\min}^{2}}{960} J^{(t)}-\frac{c_d\lambda_{\min}^2}{8}\sum_{k=1}^3\left\|\mathbf{U}_k^{(t)}-\mathbf{U}_k^{\star} \mathbf{R}_k^{(t)}\right\|_{\mathrm{F}}^2  -\frac{625}{32}c_d\lambda_{\min}^{2} J^{(t)} \right)I\left\{\mathcal{E}_{t}\right\}   \\
 & + \frac{1}{24} \sum_{k=1}^3\left(\left\|\mathbf{U}_k^{(t) \top} \mathbf{U}_k^{(t)}-\mathbf{U}_k^{\star \top} \mathbf{U}_k^{\star}\right\|_{\mathrm{F}}^2\right)I\left\{\mathcal{E}_{t}\right\}  \\
= & c_0 \lambda_{\min}^{2} J^{(t)}I\left\{\mathcal{E}_{t}\right\} + \frac{1}{24} \sum_{k=1}^3\left(\left\|\mathbf{U}_k^{(t) \top} \mathbf{U}_k^{(t)}-\mathbf{U}_k^{\star \top} \mathbf{U}_k^{\star}\right\|_{\mathrm{F}}^2\right)I\left\{\mathcal{E}_{t}\right\}.
\end{aligned}
\end{align*}
In the final step, we define  \(c_0\) to be a universal small constant such that $ c_0:=\frac{1}{960}-\frac{1}{8}c_d -\frac{625}{32}c_d >0$, and we set $c_d = \frac{1}{20000}$. 

\end{proof}

\subsection{Proof of Lemma \ref{lem:super-martingale}}\label{section:proof of super-martingale}
\begin{proof}
	Recall that we define
\begin{align*}
\mathcal{J}_t=\prod_{\tau=1}^t\left(1-\frac{\eta_\tau}{\phi}\right)^{-1}\left( J^{\prime(t)}I\left\{\mathcal{E}_{t-1}\cap\mathcal{E}_{t-1}^\epsilon\right\}-J_{M,t}\right),
\end{align*}
and we define $J_{M,t}=R \eta_t$, and $R= 21C_\epsilon\phi \sigma^2 df$ for some constant $C_\epsilon>0$ that does not depend on $t$. To prove $\mathcal{J}_t$ is a super-martingale, we need to show that $\mathbb{E}\left[\mathcal{J}_{t} \mid \mathcal{F}_{t-1}\right] \leq \mathcal{J}_{t-1}$. First, we note that
\begin{align*}
\mathbb{E}\left[\mathcal{J}_{t} \mid \mathcal{F}_{t-1}\right]=\prod_{s=1}^{t}\left(1-\frac{\eta_s}{\phi}\right)^{-1}\left(\mathbb{E}\left[J^{\prime(t)}I\left\{\mathcal{E}_{t-1}\cap\mathcal{E}_{t-1}^\epsilon\right\}\mid\mathcal{F}_{t-1}\right]-R \eta_{t}\right).
\end{align*}
Then recall that from Equation (\ref{equ:one contra of EJ}), we have
\begin{align*}
\begin{aligned}
\mathbb{E}\left[J^{\prime(t)}\mid\mathcal{F}_{t-1}\right]I\left\{\mathcal{E}_{t-1}\cap\mathcal{E}_{t-1}^\epsilon\right\}  \leq  \left(1-\frac{\eta_{t}}{\phi} \right) J^{(t-1)} I\left\{\mathcal{E}_{t-1}\right\} + 21C_\epsilon \sigma^2 df\eta_{t+1}^2 .
\end{aligned}
\end{align*}
Let us define $H = 216C_\epsilon \sigma^2 df$. Given that $\phi = \left(2c_0\lambda_{\min}^2\right)^{-1}\geq 1$, it follows from the definition of $\phi$ that $H \leq R$. 
By combining the above relationship, we have
\begin{align*}
\begin{aligned}
\mathbb{E}\left[\mathcal{J}_{t} \mid \mathcal{F}_{t-1}\right] & \leq \prod_{s=1}^{t}\left(1-\frac{\eta_s}{\phi}\right)^{-1}\left[\left(1-\frac{\eta_{t}}{\phi}\right)J^{(t-1)}I\left\{\mathcal{E}_{t-1}\right\}+H \eta_{t}^2-R \eta_{t}\right] \\
& =\prod_{s=1}^{t-1}\left(1-\frac{\eta_s}{\phi}\right)^{-1}J^{(t-1)}I\left\{\mathcal{E}_{t-1}\right\}+\prod_{s=1}^{t}\left(1-\frac{\eta_s}{\phi}\right)^{-1}\left(H \eta_{t}^2-R \eta_{t}\right).
\end{aligned}
\end{align*}
On the other hand, from the definition of $\mathcal{J}^{(t-1)}$, we have
\begin{align*}
\mathcal{J}_{t-1}=\prod_{s=1}^{t-1}\left(1-\frac{\eta_s}{\phi}\right)^{-1}J^{\prime(t-1)}I\left\{\mathcal{E}_{t-2}\right\}-\prod_{s=1}^{t-1}\left(1-\frac{\eta_s}{\phi}\right)^{-1} R \eta_{t-1}.
\end{align*}
Since $J^{(t-1)}\leq J^{\prime(t-1)}$, it only remains to show that
\begin{align*}
H \eta_t^2-R \eta_t \leq-\left(1-\frac{\eta_t}{\phi}\right) R \eta_{t-1}.
\end{align*}
Equivalently, we need to prove
\begin{align*}
H \eta_t^2+R\left(\eta_{t-1}-\eta_t\right) \leq \frac{R \eta_t \eta_{t-1}}{\phi}.
\end{align*}
We initiate our proof by demonstrating that
\begin{align*}
R\left(\eta_{t-1}-\eta_t\right) \leq \frac{R \eta_t \eta_{t-1}}{2\phi}.
\end{align*}
Recall the definition of $\eta_t=c\left(\max \left\{t^{\star}, t\right\}\right)^{-\alpha}$, we can reformulate the above inequality as follows:
\begin{align*}
\eta_{t-1}-\eta_t=c(t-1)^{-\alpha}-c t^{-\alpha} \leq \frac{\eta_t \eta_{t-1}}{2 \phi}=\frac{c^2 t^{-\alpha}(t-1)^{-\alpha}}{2 \phi}.
\end{align*}

Since
\begin{align*}
\frac{c(t-1)^{-\alpha}-c t^{-\alpha}}{c t^{-\alpha}(t-1)^{-\alpha}}=t^\alpha-(t-1)^\alpha,
\end{align*}
we only need to show
\begin{align*}
t^\alpha-(t-1)^\alpha \leq \frac{c}{2 \phi},
\end{align*}
for large enough $t$. 
For $t>\left(\frac{2 \phi \alpha}{c}\right)^{\frac{1}{1-\alpha}}+1$, we can see that
\begin{align*}
t^\alpha-(t-1)^\alpha=\alpha \int_{t-1}^t x^{\alpha-1} d x = \alpha t^{\alpha-1} \leq \frac{c}{2 \phi}.
\end{align*}
Thus, we have $\eta_{t-1}-\eta_t \leq \frac{\eta_t \eta_{t-1}}{2 \phi}$. On the other hand, given that $H \leq R$ and considering the relationship $\eta_t^2 \leq \eta_t \eta_{t-1}$, we can deduce that $H \eta_t^2 \leq R \eta_t \eta_{t-1} / 2 \phi$. Therefore, combining the above inequalities, we have shown that $R\left(\eta_{t-1}-\eta_t\right) \leq \frac{R \eta_t \eta_{t-1}}{2 \phi}$ and $H \eta_t^2 \leq R \eta_t \eta_{t-1} / 2 \phi$. We thus conclude the proof of Lemma \ref{lem:super-martingale}.
\end{proof}

\subsection{Proof of Lemma \ref{lem:bound of J1}}\label{section:proof of bound of J1}
First of all, we have
\begin{align*}
  J_1^{(t+1)}=& -2 \eta_{t+1}\sum_{k=1}^3\left\langle\mathbf{U}_k^{(t)}-\mathbf{U}_k^\star \mathbf{R}_k^{(t)}, \mathcal{M}_k\left( \nabla_\mathcal{T}f\left(\mathcal{T}^{(t)}\right)\right) \breve{\mathbf{U}}^{(t)}_k\right\rangle\\
  & -2 \eta_{t+1}\left\langle\mathcal{G}^{(t)}- \times_{k\in[3]} \mathbf{R}_k^{(t)  \top} ,  \nabla_\mathcal{T}f\left(\mathcal{T}^{(t)}\right) \times_{k\in[3]} \mathbf{U}_k^{(t) \top}\right\rangle.
\end{align*}
By utilizing Equation (\ref{equ:C.15}) and the notations defined in Equation (\ref{equ:T notation}), we can derive the following:
\begin{align}\label{equ:J1-1}
  \left| J_1^{(t+1)}\right|\leq 4\eta_{t+1}\left|\left\langle 4 \mathcal{T}^{(t)}-\mathcal{T}_{\mathcal{G}}^{(t)}-\sum_{k=1}^3 \mathcal{T}_k^{(t)}, \nabla_\mathcal{T}f\left(\mathcal{T}^{(t)}\right)\right\rangle\right|.
\end{align}
And by the definition of $\mathcal{H}_{\varepsilon}^{(t)}$ in Equation (\ref{equ:D.10}), we have 
\begin{align}\label{equ:J1-2}
\begin{aligned}
    &\left\langle 4 \mathcal{T}^{(t)}-\mathcal{T}_{\mathcal{G}}^{(t)}-\sum_{k=1}^3 \mathcal{T}_k^{(t)}, \nabla_\mathcal{T}f\left(\mathcal{T}^{(t)}\right)\right\rangle\\
=& \left\langle\mathcal{T}^{(t)}-\mathcal{T}^\star, \nabla_\mathcal{T}f\left(\mathcal{T}^{(t)}\right)-\nabla_\mathcal{T}f\left(\mathcal{T}^{\star}\right)\right\rangle  +\left\langle\mathcal{H}_{\varepsilon}^{(t)}, \nabla_\mathcal{T}f\left(\mathcal{T}^{(t)}\right)-\nabla_\mathcal{T}f\left(\mathcal{T}^{\star}\right)\right\rangle\\
&+\left\langle\mathcal{T}^{(t)}-\mathcal{T}^\star+\mathcal{H}_{\varepsilon}^{(t)}, \nabla_\mathcal{T}f\left(\mathcal{T}^{\star}\right)\right\rangle \\
=&  \left\langle\mathcal{T}^{(t)}-\mathcal{T}^\star, \left\langle\mathcal{T}^{(t)}-\mathcal{T}^\star, \mathcal{X}_{t+1}\right\rangle\mathcal{X}_{t+1}\right\rangle + \left\langle\mathcal{H}_{\varepsilon}^{(t)},\left\langle\mathcal{T}^{(t)}-\mathcal{T}^\star, \mathcal{X}_{t+1}\right\rangle\mathcal{X}_{t+1} \right\rangle\\
&- \left\langle \mathcal{T}^{(t)}-\mathcal{T}^\star+\mathcal{H}_{\varepsilon}^{(t)}, \xi_{t+1}\mathcal{X}_{t+1}\right\rangle.
\end{aligned}
\end{align}
First of all, we notice that $\left\langle 4 \mathcal{T}^{(t)}-\mathcal{T}_{\mathcal{G}}^{(t)}-\sum_{k=1}^3 \mathcal{T}_k^{(t)}, \nabla_\mathcal{T}f\left(\mathcal{T}^{(t)}\right)\right\rangle$ is a sub-exponential random variable. Thus, we will bound this term using Bernstein concentration inequality. We first note that,
\begin{align*}
  &\left\|\prod_{s=\tau+1}^t\left(1-\frac{\eta_s}{\phi}\right)\rbr{J_1^{(\tau)}-\mathbb{E}\left[J_1^{(\tau)}\mid \mathcal{F}_{\tau-1}\right]} I\left\{\mathcal{E}_{t-1}\right\}\right\|_{\psi_1}\\
  \stackrel{(a)}{\leq}& 2\left\|\prod_{s=\tau+1}^t\left(1-\frac{\eta_s}{\phi}\right)\eta_{\tau}\left\langle 4 \mathcal{T}^{(\tau-1)}-\mathcal{T}_{\mathcal{G}}^{(\tau-1)}-\sum_{k=1}^3 \mathcal{T}_k^{(\tau-1)}, \nabla_\mathcal{T}f\left(\mathcal{T}^{(\tau-1)}\right)\right\rangle I\left\{\mathcal{E}_{t-1}\right\}\right\|_{\psi_1}\\
  \stackrel{(\ref{equ:J1-2})}{\leq} & 2 \prod_{s=\tau+1}^t\left(1-\frac{\eta_s}{\phi}\right)\eta_\tau\left[ \left\|\mathcal{T}^{(\tau-1)}-\mathcal{T}^\star\right\|^2_\mathrm{F} + \left\|\mathcal{H}_{\varepsilon}^{(\tau-1)}\right\|_\mathrm{F}\left\|\mathcal{T}^{(\tau-1)}-\mathcal{T}^\star\right\|_\mathrm{F} \right.\\
  &\left. + \sigma\left(\left\|\mathcal{H}_{\varepsilon}^{(\tau-1)}\right\|_\mathrm{F}+ \left\|\mathcal{T}^{(\tau-1)}-\mathcal{T}^\star\right\|_\mathrm{F}\right) \right]  I\left\{\mathcal{E}_{t-1}\right\},
\end{align*}
where (a) is due to sub-exponential random variables $x$: $\|x-\mathbb{E} x\|_{\psi_1} \leq 2\|x\|_{\psi_1}.$ Due to Equation (\ref{equ:bound of H eps}) and Lemma E.2 in \cite{han2022optimal}, we have
\begin{align*}
\left\|\mathcal{T}^{(\tau-1)}-\mathcal{T}^\star\right\|_\mathrm{F}^2\leq 42 J^{(\tau-1)} \text{ and } \left\|\mathcal{H}_{\varepsilon}^{(\tau-1)}\right\|_\mathrm{F} \leq 6.5 J^{(\tau-1)}.
\end{align*}
Recall that by the event $\mathcal{E}_t$ in Equation (\ref{equ:def of event e}), we have
\begin{align}\label{equ:J2-4}
\sqrt{J^{(\tau-1)}}\leq C_\mathcal{E}\sigma\sqrt{df\eta_{\tau-1}}\leq C_\mathcal{E}^\prime\sigma\sqrt{df\eta_{\tau}}.
\end{align}
Therefore, we have 
\begin{align*}
  \left\|\prod_{s=\tau+1}^t\left(1-\frac{\eta_s}{\phi}\right)J_1^{(\tau)}-\mathbb{E}\left[J_1^{(\tau)}\mid \mathcal{F}_{\tau-1}\right] I\left\{\mathcal{E}_{t-1}\right\}\right\|_{\psi_1}\leq C_\mathcal{E}^\prime \sigma^2\sqrt{df}\prod_{s=\tau+1}^t\left(1-\frac{\eta_s}{\phi}\right)\sqrt{\eta_\tau^3}.
\end{align*}

Due to Lemma B.3 in \cite{han2022online}, we have
\begin{align*}
  \max_{\tau}  \left\|\prod_{s=\tau+1}^t\left(1-\frac{\eta_s}{\phi}\right) J_1^{(\tau)}-\mathbb{E}\left[J_1^{(\tau)}\mid \mathcal{F}_{\tau-1}\right] I\left\{\mathcal{E}_{t-1}\right\}\right\|_{\psi_1} \leq C_\mathcal{E}^\prime \sigma^2\sqrt{df\eta_t^3}.
\end{align*}
Then, if we apply Bernstein concentration for a sub-exponential random variable \cite[Corollary 5.17]{vershynin2011introduction},  we have
\begin{align*}
\begin{aligned}
& \mathbb{P}\left(\left|\sum_{\tau=1}^{t-1} \prod_{s=\tau+1}^t\left(1-\frac{\eta_s}{\phi}\right) J_1^{(\tau)}-\mathbb{E}\left[J_1^{(\tau)}\mid \mathcal{F}_{\tau-1}\right]\right| I\left\{\mathcal{E}_{t-1}\right\}>\varkappa\right) \\
\leq & 2 \exp \left\{-C\min\left(\frac{\varkappa^2}{C_\mathcal{E}^\prime\left( \sigma^2\sqrt{df\eta_t^3}\right)^2},\frac{\varkappa}{C_\mathcal{E}^\prime \sigma^2\sqrt{df} \sqrt{\eta_t^3}}\right)\right\},
\end{aligned}
\end{align*}
for some absolute constant $C$. Then by taking
\begin{align*}
\varkappa=C_\mathcal{E}^\prime\gamma\sigma^2\log p\sqrt{df\eta_t^3},
\end{align*}
we have
\begin{align*}
\mathbb{P}\left(\left|\sum_{\tau=1}^{t-1} \prod_{s=\tau+1}^t\left(1-\frac{\eta_s}{\phi}\right) \left(J_1^{(\tau)}-\mathbb{E}\left[J_1^{(\tau)}\mid \mathcal{F}_{\tau-1}\right]\right)\right| I\left\{\mathcal{E}_{t-1}\right\}>\varkappa\right) \leq 2 p^{-\gamma},
\end{align*}
which means
\begin{align*}
\begin{aligned}
&\left|\sum_{\tau=1}^{t-1} \prod_{s=\tau+1}^t\left(1-\frac{\eta_s}{\phi}\right) \left(J_1^{(\tau)}-\mathbb{E}\left[J_1^{(\tau)}\mid \mathcal{F}_{\tau-1}\right]\right)\right| I\left\{\mathcal{E}_{t-1}\right\}  \leq C_\mathcal{E}^\prime\gamma\sigma^2\log p\sqrt{df\eta_t^3},
\end{aligned}
\end{align*}
with probability at least $1-2p^{-\gamma}$.  We thus conclude the proof of Lemma \ref{lem:bound of J1}.

\subsection{Proof of Lemma \ref{lem:bound of J2}}\label{section:proof of bound of J2}
\paragraph{For the second term}
\begin{align*}
  J_2^{(t+1)}
  =&\eta_{t+1}^2\sum_{k=1}^3\left\|\mathcal{M}_k\left( \nabla_\mathcal{T}f\left(\mathcal{T}^{(t)}\right)\right) \breve{\mathbf{U}}^{(t)}_k\right\|_{\mathrm{F}}^2  +\eta_{t+1}^2\left\| \nabla_\mathcal{T}f\left(\mathcal{T}^{(t)}\right) \times_{k\in[3]} \mathbf{U}_k^{(t) \top} \right\|_{\mathrm{F}}^2.
\end{align*}
By Equation (\ref{equ:J2-1}), (\ref{equ:J2-2}), and (\ref{equ:J2-3}), we have
\begin{align*}
\begin{aligned}
& J_2^{(t+1)}I\left\{\mathcal{E}_{t}\right\}  
 \leq  12 \eta_{t+1}^2\left( \left(\epsilon_{1}^{(t)}\right)^2+ \left(\epsilon_{2}^{(t)}\right)^2\right) I\left\{\mathcal{E}_{t}\right\}.
\end{aligned}
\end{align*}
Under Lemma \ref{lem:def of Xi},   we have
\begin{align}
\begin{aligned}\label{equ:J2-5}
&\left|\sum_{\tau=1}^{t} \prod_{s=\tau+1}^t\left(1-\frac{\eta_s}{\phi}\right)J_2^{(\tau)}-\mathbb{E}\left[J_2^{(\tau)}\mid \mathcal{F}_{\tau-1}\right]\right|I\left\{\mathcal{E}_{t-1}\cap\mathcal{E}_{t-1}^\epsilon\right\}\\
\stackrel{(a)}{\leq} &\left|\sum_{\tau=1}^{t} \prod_{s=\tau+1}^t\left(1-\frac{\eta_s}{\phi}\right)J_2^{(\tau)}\right|I\left\{\mathcal{E}_{t-1}\cap\mathcal{E}_{t-1}^\epsilon\right\}\\
\leq & C_\epsilon \sigma^2 \df \sum_{\tau=1}^{t} \prod_{s=\tau+1}^t\left(1-\frac{\eta_s}{\phi}\right) \eta_\tau^2\\
\leq & C_\epsilon \sigma^2 \df \eta_t,
\end{aligned}
\end{align}
where (a) is due to $J_2^{(\tau)}\geq 0 $ and the last inequality is due to Lemma B.2 in \cite{han2022online}. 

\paragraph{For the third term}
 \begin{align*}
       J_3^{(t+1)}= & \eta_{t+1}^2\sum_{k=1}^3\left\langle\mathcal{M}_k\left( \nabla_\mathcal{T}f\left(\mathcal{T}^{(t)}\right)\right) \breve{\mathbf{U}}^{(t)}_k,  \mathbf{U}_k^{(t)}\left(\mathbf{U}_k^{(t) \top} \mathbf{U}_k^{(t)} - \mathbf{U}_k^{* \top} \mathbf{U}_k^{\star}\right)\right\rangle.
\end{align*}      
Note that $J_3^{(\tau)}$ is a mean zero random variable, and $X_\tau, \xi_\tau$ are i.i.d. for all $\tau$. By Equation (\ref{equ:J2-1}) and (\ref{equ:J2-2}), we have
\begin{align*}
\begin{aligned}
 J_3^{(t+1)}I\left\{\mathcal{E}_{t}\right\}
&\stackrel{(a)}{\leq }  \eta_{t+1}^2\sum_{k=1}^3\left\|\mathcal{M}_k\left( \nabla_\mathcal{T}f\left(\mathcal{T}^{(t)}\right)\right) \breve{\mathbf{U}}^{(t)}_k\right\|_\mathrm{F}\left\| \mathbf{U}_k^{(t)}\left(\mathbf{U}_k^{(t) \top} \mathbf{U}_k^{(t)} - \mathbf{U}_k^{* \top} \mathbf{U}_k^{\star}\right) \right\|_\mathrm{F}\\
 &\leq  11 \eta_{t+1}^2\left( \epsilon_{1}^{(t)}+ \epsilon_{2}^{(t)}\right)\sqrt{J^{(t)}} I\left\{\mathcal{E}_{t}\right\},
\end{aligned}
\end{align*}
where (a) follows from the Cauchy-Schwarz inequality.
Utilizing Lemma \ref{lem:def of Xi}, and following the similar approach of Equation (\ref{equ:J2-4}) and (\ref{equ:J2-5}), we derive
\begin{align*}
\begin{aligned}
&\left|\sum_{\tau=1}^{t} \prod_{s=\tau+1}^t\left(1-\frac{\eta_s}{\phi}\right)\rbr{J_3^{(\tau)}-\mathbb{E}\left[J_3^{(\tau)}\mid \mathcal{F}_{\tau-1}\right]}\right|I\left\{\mathcal{E}_{t-1}\cap\mathcal{E}_{t-1}^\epsilon\right\}\\
=&\left|\sum_{\tau=1}^{t} \prod_{s=\tau+1}^t\left(1-\frac{\eta_s}{\phi}\right)J_3^{(\tau)}\right|I\left\{\mathcal{E}_{t-1}\cap\mathcal{E}_{t-1}^\epsilon\right\}\\
\leq & C_\epsilon^\prime \sigma^2 \df \sum_{\tau=1}^{t} \prod_{s=\tau+1}^t\left(1-\frac{\eta_s}{\phi}\right) \sqrt{\eta_\tau^5}
\leq  C_\epsilon^{\prime\prime} \sigma^2 \df \sqrt{\eta_\tau^3}.
\end{aligned}
\end{align*}
We thus conclude the proof of Lemma \ref{lem:bound of J2}.

%% file: body/LemmaEntry.tex
\subsection{Proof of Lemma \ref{lem:Norm T and Norm U}}
\label{sec:proof of Lemma B3}
 Suppose $\left\| \mathcal{M}_k\left(\mathcal{T}^{(t)} - \mathcal{T}^\star\right) \right\| \leq \frac{\delta \lambda_{\text{min}}}{2}$. By \cite[Lemma 6]{zhang2020tensor}, we have
\begin{align*}
\left\| {\mathbf{U}}_{k \perp}^{(t) \top} \mathcal{M}_k\left(\mathcal{T}^{\star}\right)\right\| \leq 2 \left\| \mathcal{M}_k\left(\mathcal{T}^{(t)} - \mathcal{T}^\star\right) \right\| \leq \delta \lambda_{\min},
\end{align*}
and consequently,
\begin{align*}
\left\|\sin \Theta\left(\mathbf{U}_k^{(t)}, \mathbf{U}_k^\star\right)\right\| = \left\|{\mathbf{U}}_{k \perp}^{(t) \top} \mathbf{U}_k^\star\right\| \stackrel{(a)}{\leq} \frac{\left\|{\mathbf{U}}_{k \perp}^{(t) \top} \mathcal{P}_{\mathbf{U}_k^\star} \mathcal{M}_k\left(\mathcal{T}^\star\right)\right\|}{\sigma_{\min }\left(\mathbf{U}_k^{\star\top} \mathcal{M}_k\left(\mathcal{T}^\star\right)\right)}=\frac{\left\|{\mathbf{U}}_{k \perp}^{(t) \top} \mathcal{M}_k\left(\mathcal{T}^\star\right)\right\|}{\lambda_{\min }} \leq \delta .
\end{align*}
Here, (a) arises from $\|\mathbf{A} \mathbf{B}\| \geq\|\mathbf{A}\| \sigma_{\min }(\mathbf{B})$, and we thus conclude the proof of Lemma \ref{lem:Norm T and Norm U}.

\subsection{Proof of Lemma \ref{lem:compare z1 and z2}}\label{sec:Proof of Lemma compare z1 and z2}
\begin{proof}
 For the term $\mathbf{U}_1^{\top} \mathcal{M}_1\left(\mathcal{Z}_1^{(t)}\right) \left(\mathbf{U}_2 \otimes \mathbf{U}_3\right)=\frac{1}{t}\sum_{\tau=1}^t \xi_\tau \mathbf{U}_1^{\top} \mathcal{M}_1\left(\mathcal{X}_\tau\right) \left(\mathbf{U}_2 \otimes \mathbf{U}_3\right)$, by Assumption \ref{cond:1}, we have $$\mathbb{E}\left\|\mathbf{U}_1^{\top} \mathcal{M}_1\left(\mathcal{X}_\tau\right) \left(\mathbf{U}_2 \otimes \mathbf{U}_3\right)\right\|^2 \leq r^2, \mathbb{E}\left[\xi_t^2 \mid \mathcal{F}_{t-1}\right] \leq \sigma^2.$$ We have
\begin{align*}
\mathbb{E}\left[\left\|\mathbf{U}_1^{\top} \mathcal{M}_1\left(\mathcal{Z}^{(t)}\right) \left(\mathbf{U}_2 \otimes \mathbf{U}_3\right)\right\|^2\right] \lesssim \frac{\sigma^2 r^2}{t}
\end{align*}
where we recall that $r=\max_{k\in[3]} r_k$. Therefore, we have $\left\|\mathbf{U}_1^{\top} \mathcal{M}_1\left(\mathcal{Z}_1^{(t)}\right) \left(\mathbf{U}_2 \otimes \mathbf{U}_3\right)\right\|=\Op{\sigma \sqrt{r^2 / t}}$. For
\begin{align*}
\mathbf{U}_1^{\top} \mathcal{M}_1\left(\mathcal{Z}_2^{(t)}\right) \left(\mathbf{U}_2 \otimes \mathbf{U}_3\right) = \frac{1}{t}\sum_{\tau=1}^t \left\langle \mathcal{X}_\tau, \Delta_{\tau-1}\right\rangle \mathbf{U}_1^{\top} \mathcal{M}_1\left(\mathcal{X}_\tau\right) \left(\mathbf{U}_2 \otimes \mathbf{U}_3\right)-\mathbf{U}_1^{\top} \mathcal{M}_1\left(\Delta_{\tau-1}\right) \left(\mathbf{U}_2 \otimes \mathbf{U}_3\right)
\end{align*}
we have $\E{\left\langle \mathcal{X}_\tau, \Delta_{\tau-1}\right\rangle \mathbf{U}_1^{\top} \mathcal{M}_1\left(\mathcal{X}_\tau\right) \left(\mathbf{U}_2 \otimes \mathbf{U}_3\right)} = \mathbf{U}_1^{\top} \mathcal{M}_1\left(\Delta_{\tau-1}\right) \left(\mathbf{U}_2 \otimes \mathbf{U}_3\right)$, and 
\begin{align*}
	 \E{\nbr{\left\langle \mathcal{X}_\tau, \Delta_{\tau-1}\right\rangle \mathbf{U}_1^{\top} \mathcal{M}_1\left(\mathcal{X}_\tau\right) \left(\mathbf{U}_2 \otimes \mathbf{U}_3\right)}^2}
	 =  \E{\left\langle \mathcal{X}_\tau, \Delta_{\tau-1}\right\rangle^2\nbr{\mathbf{U}_1^{\top} \mathcal{M}_1\left(\mathcal{X}_\tau\right) \left(\mathbf{U}_2 \otimes \mathbf{U}_3\right)}^2}.
\end{align*}
Thus, we have
\begin{align*}
\mathbb{E}\left[\left\|\mathbf{U}_1^{\top} \mathcal{M}_1\left(\mathcal{Z}_2^{(t)}\right) \left(\mathbf{U}_2 \otimes \mathbf{U}_3\right)\right\|^2\right] \lesssim \frac{r^2}{t^2} \sum_{\tau=1}^t \mathbb{E}\left\|\Delta_{\tau-1}\right\|_{\mathrm{F}}^2
\end{align*}
Following the same argument in the proof of Theorem \ref{thm:converge}, we have
\begin{align}\label{equ:b1-1}
\left\|\mathbf{U}_1^{\top} \mathcal{M}_1\left(\mathcal{Z}_2^{(t)}\right) \left(\mathbf{U}_2 \otimes \mathbf{U}_3\right)\right\|=\Op{\sigma \sqrt{\frac{r^2\df}{t^{1+\alpha}}} }
\end{align}
Lastly by Assumption \ref{cond:4}, we have
\begin{align*}
\left\|\mathbf{U}_1^{\top} \mathcal{M}_1\left(\mathcal{Z}^{(t)}\right)\left(\mathbf{U}_2 \otimes \mathbf{U}_3\right)\right\| \leq \left\|\mathbf{U}_1^{\top} \mathcal{M}_1\left(\mathcal{Z}_1^{(t)}\right)\left(\mathbf{U}_2 \otimes \mathbf{U}_3\right)\right\|  + \left\|\mathbf{U}_1^{\top} \mathcal{M}_1\left(\mathcal{Z}_2^{(t)}\right)\left(\mathbf{U}_2 \otimes \mathbf{U}_3\right)\right\| =\Op{\sigma \sqrt{\frac{r^2}{t}}}.
\end{align*}
\end{proof}

\subsection{Proof of Lemma \ref{lem:bound for inference}}\label{sec:Proof of Lemma bound for inference}

Remember that for each \( k \in [3]\), \( \mathbf{U}_k^\star \) represents the left singular subspace of \( \mathcal{M}_k\left(\mathcal{T}^\star\right) \). 
From Theorem \ref{thm:converge} and  Lemma \ref{lem:Norm T and Norm U}, we can say that when \(t \geq t_0:=  C_1\left(\df \sigma^2/\lambda_{\min}^2\right)^{1/\alpha} \), with high probability, we have
\begin{align}\label{25-1}
L_t = \max _{k\in[3]}\left\|\sin \Theta\left(\widehat{\mathbf{U}}_k^{(t+1)}, \mathbf{U}_k^\star\right)\right\| \leq C_1 \frac{\sigma}{\lambda_{\min }}\sqrt{\frac{\df}{t^\alpha}} \leq \frac{1}{2} .
\end{align}
Next, we aim to prove that for $t\geq t_0$,
\begin{align}\label{29}
L_{t+1}= \Op{\frac{\sigma}{\lambda_{\min}}\sqrt{\frac{p}{t+1}}+\frac{\sigma}{\lambda_{\min}}\sqrt{\frac{pr}{t+1}} L_t}\leq \frac{1}{2}.
\end{align}
We proof Equation (\ref{29}) by induction. We first focus on the upper bound of $\left\|\sin \Theta\left(\widehat{\mathbf{U}}_1^{(t+1)}, \mathbf{U}_1^\star\right)\right\|$ when $t=t_0$. By the definition of $\mathbf{R}_k^{(t)}$ in Equation (\ref{equ:def of Rt}) and Lemma 6 in \cite{xia2022inference}, for any $t\in\mathbb{R}$ and $k\in[3]$, we have
\begin{align}\label{equ:bound of UU-R}
\left\|\widehat{\mathbf{U}}_k^{(t) \top} \mathbf{U}_k^\star-\mathbf{R}_k^{(t)}\right\| \leq \left\|\mathbf{U}_{k \perp}^{\star\top} \widehat{\mathbf{U}}_k^{(t)}\right\|^2 \leq L_t^2,
\end{align}
and 
\begin{align}\label{equ:bound of U-UR}
\begin{aligned}
\left\|\mathbf{U}_k^\star-\widehat{\mathbf{U}}_k^{(t)} \mathbf{R}_k^{(t)}\right\| & \leq\left\|\mathcal{P}_{\widehat{\mathbf{U}}_k^{(t)}}\left(\mathbf{U}_k^\star-\widehat{\mathbf{U}}_k^{(t)} \mathbf{R}_k^{(t)}\right)\right\|+\left\|\mathcal{P}_{\widehat{\mathbf{U}}_k^{(t)}}^{\perp}\left(\mathbf{U}_k^\star-\widehat{\mathbf{U}}_k^{(t)} \mathbf{R}_k^{(t)}\right)\right\| \\
& =\left\|\widehat{\mathbf{U}}_k^{(t) \top}\mathbf{U}_k^\star-\mathbf{R}_k^{(t)}\right\|+\left\|\widehat{\mathbf{U}}_{k\perp}^{(t)\top} \mathbf{U}_k^\star\right\| \leq 2L_t.
\end{aligned}
\end{align}
Recall that our model is $\widehat{\mathcal{T}}^{(t+1)}= \mathcal{T}^\star+ \mathcal{Z}^{(t+1)}$. Define the following key components in our analysis:
\begin{align*}
\begin{aligned}
& \mathbf{T}_1^{(t+1)}=\mathcal{M}_1\left(\mathcal{T}^\star  \times_2\left(\widehat{\mathbf{U}}_2^{(t)}\right)^{\top} \times_3\left(\widehat{\mathbf{U}}_3^{(t)}\right)^{\top}\right) = {\mathbf{T}}_1^\star \cdot\left(\widehat{\mathbf{U}}_2^{(t)} \otimes \widehat{\mathbf{U}}_3^{(t)}\right) \in \mathbb{R}^{p_1 \times r_2 r_3} \text {, } \\
&  \widehat{\mathbf{Z}}_1^{(t+1)}=\mathcal{M}_1\left(\mathcal{Z}  ^{(t+1)}\times_2\left(\widehat{\mathbf{U}}_2^{(t)}\right)^{\top} \times_3\left(\widehat{\mathbf{U}}_3^{(t)}\right)^{\top}\right)=\mathcal{M}_1\left(\mathcal{Z}^{(t+1)}\right) \cdot\left(\widehat{\mathbf{U}}_2^{(t)} \otimes \widehat{\mathbf{U}}_3^{(t)}\right) \in \mathbb{R}^{p_1 \times r_2 r_3}. 
\end{aligned}
\end{align*}
By definition, the left and right singular subspaces of $\mathbf{T}_1^\star$ are $\mathbf{U}_1^\star \in \mathbb{O}_{p_1, r_1}$ and $\mathbf{U}_2^\star \otimes \mathbf{U}_3^\star \in$ $\mathbb{O}_{p_2 p_3, r_2 r_3}$. Then,
\begin{align*}
\begin{aligned}
 \lambda_{\min}\left(\mathbf{T}_1^{(t+1)}\right) = &\lambda_{\min}\left( \mathbf{T}_1^\star \cdot\left(\widehat{\mathbf{U}}_2^{(t)} \otimes \widehat{\mathbf{U}}_3^{(t)}\right)\right)=\lambda_{\min}\left(\mathbf{T}_1^\star \cdot \mathcal{P}_{\mathbf{U}_2^\star \otimes \mathbf{U}_3^\star} \cdot\left(\widehat{\mathbf{U}}_2^{(t)} \otimes \widehat{\mathbf{U}}_3^{(t)}\right)\right) \\
= & \lambda_{\min}\left(\mathbf{T}_1^\star \cdot\left(\mathbf{U}_2^\star \otimes \mathbf{U}_3^\star\right) \cdot\left(\mathbf{U}_2^\star \otimes \mathbf{U}_3^\star\right)^{\top} \cdot\left(\widehat{\mathbf{U}}_2^{(t)} \otimes \widehat{\mathbf{U}}_3^{(t)}\right)\right) \\
\geq & \lambda_{\min}\left(\mathbf{T}_1^\star \cdot\left(\mathbf{U}_2^\star \otimes \mathbf{U}_3^\star\right)\right) \cdot \lambda_{\min }\left(\left(\mathbf{U}_2^\star \otimes \mathbf{U}_3^\star\right)^{\top} \cdot\left(\widehat{\mathbf{U}}_2^{(t)} \otimes \widehat{\mathbf{U}}_3^{(t)}\right)\right) \\
= & \lambda_{\min}\left(\mathbf{T}_1^\star\right) \cdot \lambda_{\min }\left(\left(\mathbf{U}_2^{\star\top} \widehat{\mathbf{U}}_2^{(t)}\right) \otimes\left(\mathbf{U}_3^{\star\top} \widehat{\mathbf{U}}_3^{(t)}\right)\right) \\
\geq & \lambda_{\min}\left(\mathbf{T}_1^\star\right) \cdot \lambda_{\min }\left(\mathbf{U}_2^{\star\top} \widehat{\mathbf{U}}_2^{(t)}\right) \cdot \lambda_{\min }\left(\mathbf{U}_3^{\star\top} \widehat{\mathbf{U}}_3^{(t)}\right) \\
\stackrel{(a)}{\geq} & \lambda_{\min} \cdot\left(1-L_t^2\right).
\end{aligned}
\end{align*}
Here, (a) is derived from the equivalent expressions for the sin-$\Theta$ distance. 

Similar to Lemma \ref{lem:compare z1 and z2} and by Lemma 10 in \cite{xia2022inference}, we can have 
\begin{align}\label{equ:bound of Z}
\begin{aligned}
		\nbr{\widehat{\mathbf{Z}}_1^{(t+1)}} = & \left\|\mathcal{M}_1\left(\mathcal{Z}^{(t+1)}\right) \cdot\left(\widehat{\mathbf{U}}_2^{(t)} \otimes \widehat{\mathbf{U}}_3^{(t)}\right)\right\| \\
	\leq & \left\|\mathcal{M}_1\left(\mathcal{Z}^{(t+1)}\right) \cdot \left( \U{2} \otimes \U{3}\right)\right\|
	 +  \left\|\mathcal{M}_1\left(\mathcal{Z}^{(t+1)}\right) \cdot\left(\U{2} \otimes \rbr{\widehat{\mathbf{U}}_3^{(t)}\fR_{3}^{(t)}-\U{3}} \right)\right\| \\
	& + \left\|\mathcal{M}_1\left(\mathcal{Z}^{(t+1)}\right) \cdot\left( \rbr{\widehat{\mathbf{U}}_2^{(t)}\fR_{2}^{(t)}-\U{2}} \otimes \U{3} \right)\right\|\\
	& +   \left\|\mathcal{M}_1\left(\mathcal{Z}^{(t+1)}\right) \cdot\left( \rbr{\widehat{\mathbf{U}}_2^{(t)}\fR_{2}^{(t)}-\U{2}} \otimes \rbr{\widehat{\mathbf{U}}_3^{(t)}\fR_{3}^{(t)}-\U{3}}\right)\right\| \\
	= & \Op{\sigma\sqrt{\frac{p}{t+1}} + \sigma\sqrt{\frac{pr}{t+1}}L_t}
\end{aligned}
\end{align}
In accordance with Wedin's $\sin \Theta$ theorem \citep{wedin1972perturbation}, we can assert that the following inequality holds:
\begin{align*}
\begin{aligned}
\left\|\sin \Theta\left(\widehat{\mathbf{U}}_1^{(t+1)}, \mathbf{U}_1^\star\right)\right\| \leq  \frac{\left\|\widehat{\mathbf{Z}}_1^{(t+1)}\right\|}{\lambda_{\min}\left(\mathbf{T}_1^{(t+1)}\right)} = & \Op{\frac{\sigma\sqrt{\frac{p}{t+1}} + \sigma\sqrt{\frac{pr}{t+1}}L_t}{\lambda_{\min}\left(1-L_t^2\right)}}.
\end{aligned}
\end{align*}
Since $L_t\leq \frac{1}{2}$, for any $k\in[3]$, we have
\begin{align*}
\begin{gathered}
\left\|\sin \Theta\left(\widehat{\mathbf{U}}_k^{(t+1)}, \mathbf{U}_k^\star\right)\right\| =\Op{\frac{\sigma}{\lambda_{\min}}\sqrt{\frac{p}{t+1}}+\frac{\sigma}{\lambda_{\min}}\sqrt{\frac{pr}{t+1}} L_t}.
\end{gathered}
\end{align*}
Finally, there exists a large constant $C_{1}>0$ such that when $t (\lambda_{\min}/\sigma )^2\geq C_1 pr$, we have
\begin{align}\label{33}
\frac{\sigma}{\lambda_{\min}}\sqrt{\frac{p}{t+1}}+ \frac{\sigma}{\lambda_{\min}}\sqrt{\frac{pr}{t+1}} L_t \leq \frac{1}{2}, \quad \frac{\sigma}{\lambda_{\min}}\sqrt{\frac{pr}{t+1}}\leq \frac{1}{2}.
\end{align}
Then we have finished the proof for (\ref{29}) for $t=t_0+1$. We can sequentially prove that (\ref{29}) for all $t \geq t_0+1$ by induction. 
\begin{align*}
\begin{aligned}
 L_{t+1} = & \Op{\frac{\sigma}{\lambda_{\min}}\sqrt{\frac{p}{t+1}} + \frac{\sigma}{\lambda_{\min}}\sqrt{\frac{pr}{t+1}} L_t} \\
=& \Op{\frac{\sigma}{\lambda_{\min}}\sqrt{\frac{p}{t+1}}+\frac{\sigma}{\lambda_{\min}}\sqrt{\frac{pr}{t+1}}\left(L_t-\frac{\sigma}{\lambda_{\min}}\sqrt{\frac{p}{t}}\right)}
, \\
=& \Op{\frac{\sigma}{\lambda_{\min}}\sqrt{\frac{p}{t+1}} + \frac{L_{t_0}}{2^{t+1-t_0}} } \\
=& \Op{\frac{\sigma}{\lambda_{\min}}\sqrt{\frac{p}{t+1}}}
\end{aligned}
\end{align*}
when $t \geq t_0 -1 +  \log _2\left(L_{t_0} \frac{\lambda_{\min }}{\sigma}\right)+\frac{1}{2} \log _2\left(\frac{t+1}{p}\right)$. Thus, we conclude the proof of Lemma \ref{lem:bound for inference}.

\subsection{Proof of Lemma \ref{lem:bound for first round}}
\label{sec:Proof of Lemma bound for first round}
By definition, 
\begin{align}\label{B.139}
\left\|\mathfrak{E}_1^{(t)}\right\| \leq\left\|\mathfrak{J}_1^{(t)}\right\|+\left\|\mathfrak{J}_2^{(t)}\right\|+\left\|\mathfrak{J}_3^{(t)}\right\|+\left\|\mathfrak{J}_4^{(t)}\right\|.
\end{align}
We first proved the upper bound for $\left\|\mathfrak{J}_1^{(t)}\right\|$. By the definition of $\left\|\mathfrak{J}_1^{(t)}\right\|$, 
\begin{align*}
\begin{aligned}
	\left\|\mathfrak{J}_1^{(t)}\right\| 
\leq\left\|\mathbf{T}_1^\star\left(\mathcal{P}_{\widehat{\mathbf{U}}_2^{(t-1)}} \otimes \mathcal{P}_{\widehat{\mathbf{U}}_3^{(t-1)}}\right) \mathbf{Z}_1^{(t)\top}\right\| 
\leq &\left\|\mathbf{T}_1^\star\right\|\left\|\left(\mathcal{P}_{\widehat{\mathbf{U}}_2^{(t-1)}} \otimes \mathcal{P}_{\widehat{\mathbf{U}}_3^{(t-1)}}\right) \mathbf{Z}_1^{(t)\top}\right\| \\
\leq & \kappa_0 \lambda_{\text {min }}\left\|\mathbf{Z}_1^{(t)} \cdot\left(\widehat{\mathbf{U}}_2^{(t-1)} \otimes \widehat{\mathbf{U}}_3^{(t-1)}\right)\right\|.
\end{aligned}
\end{align*}
By Equation \eqref{equ:bound of Z}, we have
$\left\|\mathbf{Z}_1^{(t)} \cdot\left(\widehat{\mathbf{U}}_2^{(t-1)} \otimes \widehat{\mathbf{U}}_3^{(t-1)}\right)\right\| =\Op{\sigma\sqrt{\frac{p}{t}}}$. 
Thus,
\begin{align}\label{B.145}
\left\|\mathfrak{J}_1^{(t)}\right\| = \Op{\kappa_0 \lambda_{\min }\sigma \sqrt{\frac{p}{t}}}.
\end{align}
Since $\mathfrak{J}_2=\mathfrak{J}_1^{\top}$, we also have
\begin{align}\label{B.146}
\left\|\mathfrak{J}_2^{(t)}\right\| = \Op{\kappa_0 \lambda_{\min }\sigma \sqrt{\frac{p}{t}}}.
\end{align}
For $\mathfrak{J}_3^{(t)}$, by definition,
\begin{align}\label{B.147}
\left\|\mathfrak{J}_3^{(t)}\right\|=\left\|\mathbf{Z}_1^{(t)}\left(\widehat{\mathbf{U}}_2^{(t-1)} \otimes \widehat{\mathbf{U}}_3^{(t-1)}\right)\right\|^2 = \Op{\sigma ^2 \frac{p}{t} } .
\end{align}
Considering $\mathfrak{J}_4^{(t)}$, Lemma \ref{lem:bound for inference} implies that for $k \in [3]$, we have
\begin{align}\label{B.142}
	\left\|\mathcal{P}_{\mathbf{\mathbf{U}}_k}^{\perp} \widehat{\mathbf{U}}_k^{(t-1)}\right\|=\left\|\mathbf{U}_{k \perp}^{\star\top} \widehat{\mathbf{U}}_k^{(t-1)}\right\| \leq\left\|\widehat{\mathbf{U}}_k^{(t-1)} \widehat{\mathbf{U}}_k^{(t-1) \top}-\mathcal{P}_{\mathbf{U}_k^\star}\right\| = \Op{\frac{\sigma}{ \lambda_{\min}}\sqrt{\frac{p}{t-1}}}.
\end{align}
By Equation (\ref{B.142}), 
\begin{align}\label{B.149}
\begin{aligned}
\| \mathfrak{J}_4^{(t)}\| \leq & \left\|\mathbf{T}_1^\star\left(\left(\mathcal{P}_{\widehat{\mathbf{U}}_2^{(t-1)}}-\mathcal{P}_{\mathbf{U}_2^\star}\right) \otimes \mathcal{P}_{\widehat{\mathbf{U}}_3^{(t-1)}}\right) \mathbf{T}_1^{\star\top}\right\|\\
&+\| \mathbf{T}_1^\star\left(\mathcal{P}_{\mathbf{U}_2^\star} \otimes\left(\mathcal{P}_{\widehat{\mathbf{U}}_3^{(t-1)}}-\mathcal{P}_{\mathbf{U}_3^\star}\right) \mathbf{T}_1^{\star\top} \|\right. \\
= & \left\|\mathbf{U}_1^\star \mathbf{G}_1^\star\left(\left(\mathbf{U}_2^{\star\top}\left(\mathcal{P}_{\widehat{\mathbf{U}}_2^{(t-1)}}-\mathcal{P}_{\mathbf{U}_2^\star}\right) \mathbf{U}_2^\star\right) \otimes\left(\mathbf{U}_3^{\star\top} \mathcal{P}_{\widehat{\mathbf{U}}_3^{(t-1)}} \mathbf{U}_3^\star\right)\right) \mathbf{G}_1^{\star\top} \mathbf{U}_1^{\star\top}\right\| \\
& +\left\|\mathbf{U}_1^\star \mathbf{G}_1^\star\left(\mathcal{P}_{\mathbf{U}_2^\star} \otimes\left(\mathbf{U}_3^{\star\top}\left(\mathcal{P}_{\widehat{\mathbf{U}}_3^{(t-1)}}-\mathcal{P}_{\mathbf{U}_3^\star}\right) \mathbf{U}_3^\star\right)\right) \mathbf{G}_1^{\star\top} \mathbf{U}_1^{\star\top}\right\| \\
= & \left\|\mathbf{U}_1^\star \mathbf{G}_1^\star\left(\left(\mathbf{U}_2^{\star\top} \mathcal{P}_{\widehat{\mathbf{U}}_2^{(t-1)}}^{\perp} \mathbf{U}_2^\star\right) \otimes\left(\mathbf{U}_3^{\star\top} \mathcal{P}_{\widehat{\mathbf{U}}_3^{(t-1)}} \mathbf{U}_3^\star\right)\right) \mathbf{G}_1^{\star\top} \mathbf{U}_1^{\star\top}\right\|\\
& +\left\|\mathbf{U}_1^\star \mathbf{G}_1^\star\left(\mathcal{P}_{\mathbf{U}_2^\star} \otimes\left(\mathbf{U}_3^{\star\top} \mathcal{P}_{\widehat{\mathbf{U}}_3^{(t-1)}}^{\perp} \mathbf{U}_3^\star\right)\right) \mathbf{G}_1^{\star\top} \mathbf{U}_1^{\star\top}\right\| \\
\leq & \left\|\mathbf{G}_1^\star\right\|^2\left\|\mathbf{U}_2^{\star\top} \widehat{\mathbf{U}}_{2 \perp}^{(t-1)}\right\|^2+\left\|\mathbf{G}_1^\star\right\|^2\left\|\mathbf{U}_3^{\star\top} \widehat{\mathbf{U}}_{3 \perp}^{(t-1)}\right\|^2 \\
= & \Op{\kappa_0^2 \lambda_{\min }^2\left(\frac{\sigma}{\lambda_{\min }}\right)^2\left(\sqrt{\frac{p}{t-1}}\right)^2} =  \Op{\frac{\kappa_0^2 \sigma^2 p}{t-1}}.
\end{aligned}
\end{align}
In conclusion, using Equations (\ref{B.139}), (\ref{B.145}), (\ref{B.146}), (\ref{B.147}), and (\ref{B.149}), 
we find that, $\left\|\mathfrak{E}_1^{(t)}\right\| = \Op{\kappa_0 \lambda_{\min} \sigma \sqrt{\frac{p}{t}}}$ and 
\begin{align}\label{equ:a2-1}
	\left\|\mathfrak{E}_1^{(t)} - \mathfrak{J}_1^{(t)} - \mathfrak{J}_2^{(t)}\right\| = \Op{\kappa_0^2 \sigma^2\frac{p}{t}}.
\end{align}
Under Lemma \ref{lem:bound for inference}, we can identify the existence of two (random) matrices: \( \mathbf{R}_k^{(t)} \in \mathbb{O}_{r_k} \) for $k=2,3$. These matrices satisfy the following inequalities:
\begin{align}\label{B.8-1}
 \max_{k=2,3}\left\{ \left\|\widehat{\mathbf{U}}_k^{(t-1)}-\mathbf{U}_k^\star \mathbf{R}_k^{(t-1)}\right\|\right\} = \Op{\frac{\sigma}{\lambda_{\min}}\sqrt{\frac{p}{t-1}}}.
\end{align} 
We have:
\begin{align}\label{B.9}
\begin{aligned}
& \left\|\mathfrak{J}_1^{(t)}-\mathbf{T}_1^\star\left(\mathcal{P}_{\mathbf{U}_2^\star} \otimes \mathcal{P}_{\mathbf{U}_3^\star}\right) \mathbf{Z}_1^{(t)\top}\right\|
=\left\|\mathbf{T}_1^\star\left(\mathcal{P}_{\widehat{\mathbf{U}}_2^{(t-1)}} \otimes \mathcal{P}_{\widehat{\mathbf{U}}_3^{(t-1)}}\right) \mathbf{Z}_1^{(t)\top}-\mathbf{T}_1^\star\left(\mathcal{P}_{\mathbf{U}_2^\star} \otimes \mathcal{P}_{\mathbf{U}_3^\star}\right) \mathbf{Z}_1^{(t)\top}\right\| \\
\leq & \left\|\left[\mathbf{T}_1^\star\left(\widehat{\mathbf{U}}_2^{(t-1)} \otimes \widehat{\mathbf{U}}_3^{(t-1)}\right)\right]\left[\mathbf{Z}_1^{(t)}\left(\widehat{\mathbf{U}}_2^{(t-1)} \otimes \widehat{\mathbf{U}}_3^{(t-1)}\right)\right]^{\top}\right.
\\ & \left.-\left[\mathbf{T}_1^\star\left(\left(\mathbf{U}_2^\star \mathbf{R}_2^{(t-1)}\right) \otimes\left(\mathbf{U}_3^\star \mathbf{R}_3^{(t-1)}\right)\right)\right]\left[\mathbf{Z}_1^{(t)}\left(\left(\mathbf{U}_2^\star \mathbf{R}_2^{(t-1)}\right) \otimes\left(\mathbf{U}_3^\star \mathbf{R}_3^{(t-1)}\right)\right)\right]^{\top}\right\| \\
\leq & \left\|\mathbf{Z}_1^{(t)}\left[\left(\widehat{\mathbf{U}}_2^{(t-1)} \otimes \widehat{\mathbf{U}}_3^{(t-1)}\right)-\left(\left(\mathbf{U}_2^\star \mathbf{R}_2^{(t-1)}\right) \otimes\left(\mathbf{U}_3^\star \mathbf{R}_3^{(t-1)}\right)\right)\right]\right\|\left\|\mathbf{T}_1^\star\left(\widehat{\mathbf{U}}_2^{(t-1)} \otimes \widehat{\mathbf{U}}_3^{(t-1)}\right)\right\| \\
& + \left\|\mathbf{Z}_1^{(t)}\left(\left(\mathbf{U}_2^\star \mathbf{R}_2^{(t-1)}\right) \otimes\left(\mathbf{U}_3^\star \mathbf{R}_3^{(t-1)}\right)\right)\right\| \cdot\\
&\quad\left\|\mathbf{T}_1^\star\left[\left(\widehat{\mathbf{U}}_2^{(t-1)} \otimes \widehat{\mathbf{U}}_3^{(t-1)}\right)-\left(\left(\mathbf{U}_2^\star \mathbf{R}_2^{(t-1)}\right) \otimes\left(\mathbf{U}_3^\star \mathbf{R}_3^{(t-1)}\right)\right)\right]\right\| \\
\leq & \kappa_0 \lambda_{\min}\left\|\mathbf{Z}_1^{(t)}\left[\left(\widehat{\mathbf{U}}_2^{(t-1)} \otimes \widehat{\mathbf{U}}_3^{(t-1)}\right)-\left(\left(\mathbf{U}_2^\star \mathbf{R}_2^{(t-1)}\right) \otimes\left(\mathbf{U}_3^\star \mathbf{R}_3^{(t-1)}\right)\right)\right]\right\| \\
& + \left\|\mathbf{Z}_1^{(t)}\left(\left(\mathbf{U}_2^\star \mathbf{R}_2^{(t-1)}\right) \otimes\left(\mathbf{U}_3^\star \mathbf{R}_3^{(t-1)}\right)\right)\right\|\left\|\mathbf{T}_1^\star\left[\left(\widehat{\mathbf{U}}_2^{(t-1)} \otimes \widehat{\mathbf{U}}_3^{(t-1)}\right)-\left(\left(\mathbf{U}_2^\star \mathbf{R}_2^{(t-1)}\right) \otimes\left(\mathbf{U}_3^\star \mathbf{R}_3^{(t-1)}\right)\right)\right]\right\| \\
\leq & \kappa_0 \lambda_{\min}\left(\left\|\mathbf{Z}_1^{(t)}\left(\left(\widehat{\mathbf{U}}_2^{(t-1)}-\mathbf{U}_2^\star \mathbf{R}_2^{(t-1)}\right) \otimes \widehat{\mathbf{U}}_3^{(t-1)}\right)\right\|\right.\\
&\left.+\left\|\mathbf{Z}_1^{(t)}\left(\left(\mathbf{U}_2^\star \mathbf{R}_2^{(t-1)}\right) \otimes\left(\widehat{\mathbf{U}}_3^{(t-1)}-\mathbf{U}_3^\star \mathbf{R}_3^{(t-1)}\right)\right)\right\|\right) \\
& + \left\|\mathbf{Z}_1^{(t)}\left(\left(\mathbf{U}_2^\star \mathbf{R}_2^{(t-1)}\right) \otimes\left(\mathbf{U}_3^\star \mathbf{R}_3^{(t-1)}\right)\right)\right\|\left(\left\|\mathbf{T}_1^\star\left(\left(\widehat{\mathbf{U}}_2^{(t-1)}-\mathbf{U}_2^\star \mathbf{R}_2^{(t-1)}\right) \otimes \widehat{\mathbf{U}}_3^{(t-1)}\right)\right\|\right.\\
&\left.+\left\|\mathbf{T}_1^\star\left(\left(\mathbf{U}_2^\star \mathbf{R}_2^{(t-1)}\right) \otimes\left(\widehat{\mathbf{U}}_3^{(t-1)}-\mathbf{U}_3^\star \mathbf{R}_3^{(t-1)}\right)\right)\right\|\right)\\
= & \Op{\kappa_0 \lambda_{\min } \cdot \sigma \sqrt{p/t}\left(\left\|\widehat{\mathbf{U}}_2^{(t-1)}-\mathbf{U}_2^\star \mathbf{R}_2^{(t-1)}\right\|+\left\|\widehat{\mathbf{U}}_3^{(t-1)}-\mathbf{U}_3^\star \mathbf{R}_3^{(t-1)}\right\|\right)} \\
= & \Op{\kappa_0 \sigma^2 \frac{p}{t}}.
\end{aligned}
\end{align}
Since $\mathfrak{J}_2^{(t)} = \mathbf{Z}_1^{(t)}\left(\mathcal{P}_{\widehat{\mathbf{U}}_2^{(t-1)}} \otimes \mathcal{P}_{\widehat{\mathbf{U}}_3^{(t-1)}}\right) \mathbf{T}_1^{\star\top} = \mathfrak{J}_1^{(t)\top} $, by Equation \eqref{equ:a2-1} and similar to Equation \eqref{equ:b1-1}, we have
	\begin{align*}
		\left\|\mathfrak{E}_1^{(t)} - \mathbf{T}_1^{\star}\left(\mathcal{P}_{\mathbf{U}_2^{\star}} \otimes \mathcal{P}_{\mathbf{U}_3^{\star}}\right) \cM_1^\top\rbr{\mathcal{Z}_1^{(t)}}  -\cM_1\rbr{\mathcal{Z}_1^{(t)}} \left(\mathcal{P}_{\mathbf{U}_2^{\star}} \otimes \mathcal{P}_{\mathbf{U}_3^{\star}}\right) \mathbf{T}_1^{\star \top}\right\| = \Op{\kappa_0^2\sigma^2\frac{p}{t} + \kappa_0\lambda_{\min}\sigma\sqrt{\frac{p\df}{t^{1+\alpha}}}}.
	\end{align*}
Thus, we conclude the proof of Lemma \ref{lem:bound for first round}.

\subsection{Proof of Lemma \ref{lem:bound of O1}}\label{sec:proof of lemma bound of O1}
First, we start to deal with the term $\mathcal{T}^\star 
        \times_1 \Pdu{1} 
        \times_{2}\Pdu{2}
        \times_{3}\Pu{3}$.
Since $\fI_{r_k} = \UT{k} \U{k}$ for $k \in [3]$ and $\U{k} = \U{k}\fI_{r_k} = \U{k} \UT{k} \U{k} = \Pu{k}\U{k}$, we have
 \begin{align*}
 \begin{aligned}
 & \left\langle \; \mathcal{T}^\star 
        \times_1 \Pdu{1} 
        \times_{2}\Pdu{2}
        \times_{3}\Pu{3}, \cH \right\rangle \\
 & = \left\langle \cG^\star   \times_1 \U{1}\fI_{r_1}  \times_2 \U{2}\fI_{r_2}  \times_3 \U{3}\fI_{r_3}, \cH \times_1 \Pdu{1} 
        \times_{2}\Pdu{2}
        \times_{3}\Pu{3}\right\rangle \\
 & = \left\langle \cG^\star   \times_1 \Pu{1}\U{1}  \times_2\Pu{2}\U{2}  \times_3 \Pu{3}\U{3}, \cH \times_1 \Pdu{1} 
        \times_{2}\Pdu{2}
        \times_{3}\Pu{3}\right\rangle \\
 & = \left\langle \cG^\star   \times_1 \U{1}  \times_2 \U{2}  \times_3 \U{3}, \cH \times_1 \Pu{1}\Pdu{1} 
        \times_{2} \Pu{2}\Pdu{2}
        \times_{3} \Pu{3}\right\rangle.
 \end{aligned}
 \end{align*}
Next lemma helps us to figure out $\Pu{k}\Pdu{k}$ term
\begin{lemma}\label{lem:bound of PuPdu}
Under Lemma \ref{lem:bound for inference},\ref{lem:bound for first round},\ref{lem:theorem 1 in xia2021normal}, we have
\begin{align*}
		\nbr{\UT{1}\Pdu{1} - \Lambda_{1}^{-2} \G{1} \rbr{\U{2}\otimes\U{3}} \cM_1^\top\rbr{\cZ_1^{(t)}}  \Pu{1}^\perp } = \Op{\kappa_0^2\frac{\sigma^2}{\lambda_{\min}^2}\sqrt{\frac{p^2}{t^{1+\alpha}}} } .
	\end{align*}
\end{lemma}
Proof in Section \ref{sec:proof of lemma bound of PuPdu}.

By Lemma \ref{lem:bound of PuPdu}, we can have
\begin{align*}
	& \nbr{\cH \times_1 \Pu{1}\Pdu{1} \times_2 \Pu{2}\Pdu{2} \times_3 \Pu{3}}\\
	= & \nbr{\cH \times_1 \Lambda_{1}^{-2} \G{1} \rbr{\U{3}\otimes\U{2}} \cM_1^\top\rbr{\cZ_1^{(t)}}  \Pu{1}^\perp \times_2 \Lambda_{2}^{-2} \G{2} \rbr{\U{1}\otimes\U{3}} \cM_2^\top\rbr{\cZ_1^{(t)}}  \Pu{2}^\perp \times_3 \Pu{3}  }\\
	& + \Op{\nbr{\cH\times  \Pu{3}} \rbr{\kappa_0^2\frac{\sigma^2}{\lambda_{\min}^2}\sqrt{\frac{p^2}{t^{1+\alpha}}}}^2 }\\
    = & \nbr{\rbr{\Pu{3}\otimes\Lambda_{2}^{-2} \G{2} \rbr{\U{1}\otimes\U{3}} \cM_2^\top\rbr{\cZ_1^{(t)}}  \Pu{2}^\perp } \fH_1^\top  \Pu{1}^\perp \cM_1 \rbr{\cZ_1^{(t)}} 
    \rbr{\U{3}\otimes\U{2}}^\top \rbr{\G{1}}^\top  \Lambda_{1}^{-2} }\\
    & + \Op{\nbr{\cH\times  \Pu{3}} \rbr{\kappa_0^2\frac{\sigma^2}{\lambda_{\min}^2}\sqrt{\frac{p^2}{t^{1+\alpha}}}}^2}\\
     = & \Op{\rbr{\frac{\sigma}{\lambda_{\min}}}^2\fbr{\rbr{\Pu{3}\otimes \Pu{2}^\perp }\fH^\top  \Pu{1}^\perp}\sqrt{\frac{\log p}{t^2}}+ \nbr{\cH\times  \Pu{3}} \rbr{\kappa_0^2\frac{\sigma^2}{\lambda_{\min}^2}\sqrt{\frac{p^2}{t^{1+\alpha}}}}^2}
\end{align*}
We need to consider
\begin{align*}
\rbr{\U{1}\otimes\U{3}} \cM_2^\top\rbr{\cZ_1^{(t)}} \underbrace{\rbr{\Pu{3}\otimes \Pu{2}^\perp }\fH_1^\top  \Pu{1}^\perp}_{\text{fixed}} \cM_1 \rbr{\cZ_1^{(t)}} \rbr{\U{3}\otimes\U{2}}^\top.
\end{align*}
First, $\rbr{\U{1}\otimes\U{3}} \cM_2^\top\rbr{\cZ_1^{(t)}}\Pu{2}^\perp $ is uncorrelated with $\Pu{1}^\perp \cM_1 \rbr{\cZ_1^{(t)}} \rbr{\U{3}\otimes\U{2}}^\top$. Then by Hanson-Wright inequality for the sub-exponential random variables in \cite[Proposition 2.1]{gotze2021concentration}, we have 
\begin{align*}
	\nbr{\rbr{\U{1}\otimes\U{3}} \cM_2^\top\rbr{\cZ_1^{(t)}} \rbr{\Pu{3}\otimes \Pu{2}^\perp }\fH_1^\top  \Pu{1}^\perp \cM_1 \rbr{\cZ_1^{(t)}} \rbr{\U{3}\otimes\U{2}}^\top} = \\\Op{\sigma^2\fbr{\rbr{\Pu{3}\otimes \Pu{2}^\perp }\fH_1^\top  \Pu{1}^\perp}\sqrt{\log p/t^2}}.
\end{align*}
\xin{I’m afraid this isn’t correct. See the Gaussian case in Equation \eqref{B.24-1}.}
 Thus, we have
\begin{align*}
	&\abr{\left\langle \; \mathcal{T}^\star 
        \times_1 \Pdu{1} 
        \times_{2}\Pdu{2}
        \times_{3}\Pu{3}, \cH \right\rangle}
         \\
        =& \Op{\sqrt{r}\kappa_0\lambda_{\min} \cdot \sbr{\fbr{\rbr{\Pu{3}\otimes \Pu{2}^\perp }\fH_1^\top  \Pu{1}^\perp}\frac{\sigma^2}{\lambda_{\min}^{2}} \sqrt{\frac{\log p}{t^2}} + \fbr{\cH\times  \Pu{3}} \rbr{\kappa_0^2\frac{\sigma^2}{\lambda_{\min}^2}\sqrt{\frac{p^2}{t^{1+\alpha}}}}^2}}\\
        =& \Op{\kappa_0 \frac{\sigma^2}{\lambda_{\min}^{}} \sqrt{\frac{r\log p}{t^2}}\fbr{\cH\times  \Pu{3}}}
\end{align*}
when $t^\alpha \rbr{\lambda_{\min}/\sigma}^2\geq Cp^2.$

A similar term for the term $\mathcal{T}^\star 
        \times_1 \Pdu{1} 
        \times_{2}\Pdu{2}
        \times_{3}\Pdu{3}$ with three difference will be
\begin{align*}
	& \left\| \cH \times_1 \Pu{1}\Pdu{1} \times_2 \Pu{2}\Pdu{2} \times_3 \Pu{3} \Pdu{3}\right\|\\
	= & \nbr{\cH \times_1 \Lambda_{1}^{-2} \G{1} \rbr{\U{3}\otimes\U{2}} \cM_1^\top\rbr{\cZ_1^{(t)}}  \Pu{1}^\perp 
	\times_2 \Lambda_{2}^{-2} \G{2} \rbr{\U{1}\otimes\U{3}} \cM_2^\top\rbr{\cZ_1^{(t)}}  \Pu{2}^\perp
	 \times_3 \Lambda_{3}^{-2} \G{3} \rbr{\U{2}\otimes\U{1}} \cM_3^\top\rbr{\cZ_1^{(t)}}  \Pu{3}^\perp  }
\end{align*}
Using the Hanson-Wright inequality for the sub-exponential random variables in \cite[Theorem 1.6]{gotze2021concentration}, we need to consider
\begin{align*}
	& \sbr{\rbr{\U{2}\otimes\U{1}} \cM_3^\top\rbr{\cZ_1^{(t)}}  \Pu{3}^\perp \otimes \rbr{\U{1}\otimes\U{3}} \cM_2^\top\rbr{\cZ_1^{(t)}}  \Pu{2}^\perp } \fH_1^\top  \Pu{1}^\perp \cM_1 \rbr{\cZ_1^{(t)}} \rbr{\U{3}\otimes\U{2}}^\top \\
	 = & \sbr{\rbr{\U{2}\otimes\U{1}} \cM_3^\top\rbr{\cZ_1^{(t)}}\otimes \rbr{\U{1}\otimes\U{3}} \cM_2^\top\rbr{\cZ_1^{(t)}}} \underbrace{\rbr{\Pu{3}^\perp \otimes  \Pu{2}^\perp } \fH_1^\top  \Pu{1}^\perp}_{\text{fixed}} \cM_1 \rbr{\cZ_1^{(t)}} \rbr{\U{3}\otimes\U{2}}^\top\\
	  = & \Op{\fbr{\rbr{\Pu{3}^\perp \otimes  \Pu{2}^\perp } \fH_1^\top  \Pu{1}^\perp}\sigma^3\sqrt{\frac{\log p}{t^3}}}
\end{align*}

\begin{align*}
	&\abr{\left\langle \; \mathcal{T}^\star 
        \times_1 \Pdu{1} 
        \times_{2}\Pdu{2}
        \times_{3}\Pdu{3}, \cH \right\rangle}\\
         = &  \Op{\sqrt{r}\kappa_0\lambda_{\min} \cdot \sbr{\fbr{\rbr{\Pu{3}^\perp\otimes \Pu{2}^\perp }\fH_1^\top  \Pu{1}^\perp}\frac{\sigma^3}{\lambda_{\min}^{3}} \sqrt{\frac{\log p}{t^3}} + \fbr{\cH} \rbr{\kappa_0^2\frac{\sigma^2}{\lambda_{\min}^2}\sqrt{\frac{p^2}{t^{1+\alpha}}}}^3}}\\
         =& \Op{\kappa_0 \frac{\sigma^3}{\lambda_{\min}^{2}} \sqrt{\frac{r\log p}{t^3}}\fbr{\cH }}. 
\end{align*}
when $t^\alpha \rbr{\lambda_{\min}/\sigma}^2\geq Cp^2.$ 
\paragraph{Z part} 
This part is repeated using the following bound. 
\begin{align*}
	\sup _{\substack{\fU_j \in \mathbb{R}^{p_j \times 2 r_j,\left\|\fU_j\right\|=1} \\ j=1,2,3}}\left\|\fU_1^{\top} \widehat{Z}_1\left(\fU_3 \otimes \fU_2\right)\right\|_{\mathrm{F}}=\Op{\sigma\sqrt{\frac{pr}{t}}}
\end{align*}
First, we have the following decomposition. For the first term on the right-hand side of the equality above, using the Cauchy-Schwarz inequality and by Lemma \ref{lem:compare z1 and z2}, we have

\begin{align}\label{equ:bound of ZU1}
\begin{aligned}
		&\left|\left\langle \mathcal{Z}^{(t)} \times_1 \left(\mathcal{P}_{\widehat{\mathbf{U}}_1^{(t)}} - \mathcal{P}_{\mathbf{U}_1^\star}\right) \times_2\left(\mathcal{P}_{\widehat{\mathbf{U}}_2^{(t)}}-\mathcal{P}_{\mathbf{U}_2^\star}\right) \times_3\left(\mathcal{P}_{\widehat{\mathbf{U}}_3^{(t)}}-\mathcal{P}_{\mathbf{U}_3^\star}\right), \mathcal{H}\right\rangle\right|\\
  		= &\left|\left\langle \left(\mathcal{P}_{\widehat{\mathbf{U}}_1^{(t)}}-\mathcal{P}_{\mathbf{U}_1^\star}\right)  \mathcal{M}_1\left(\mathcal{Z}^{(t)}\right)\left( \left(\mathcal{P}_{\widehat{\mathbf{U}}_2^{(t)}}-\mathcal{P}_{\mathbf{U}_2^\star}\right) \otimes\left(\mathcal{P}_{\widehat{\mathbf{U}}_3^{(t)}}-\mathcal{P}_{\mathbf{U}_3^\star}\right)\right), \mathbf{H}_1\right\rangle\right|\\
	\leq & \sup _{\substack{\fW_k \in \mathbb{R}^{p_k \times 2 r_k,\left\|\fW_k\right\|=1} \\ k\in[3]}}\left\|\fW_1^{\top} \cM_1\rbr{\mathcal{Z}^{(t)}}\left(\fW_3 \otimes \fW_2\right)\right\|_{\mathrm{F}} L_t^3\left\|\mathcal{H}\right\|_\mathrm{F} \\
	= & \Op{\sigma L_t^3 \sqrt{\frac{pr}{t}}\left\|\mathcal{H}\right\|_\mathrm{F}}.
\end{aligned}
\end{align}
Similarly, for the second and third terms, we derive
\begin{align}\label{equ:bound of ZU2}
\begin{aligned}
	 & \left|\sum_{k=1}^3\left\langle{\mathcal{Z}}^{(t)} \times_k\left(\mathcal{P}_{\widehat{\mathbf{U}}_1^{(t)}}-\mathcal{P}_{\mathbf{U}_k^\star}\right) \times_{k+1}\left(\mathcal{P}_{\widehat{\mathbf{U}}_{k+1}^{(t)}}-\mathcal{P}_{\mathbf{U}_{k+1}^\star}\right) \times_{k+2} \mathcal{P}_{\mathbf{U}_{k+2}^\star}, \mathcal{H}\right\rangle \right|\\
  	\stackrel{(a)}{\leq} & \sum_{k=1}^3\left|\left\langle{\mathcal{Z}}^{(t)} \times_k\left(\mathcal{P}_{\widehat{\mathbf{U}}_k^{(t)}}-\mathcal{P}_{\mathbf{U}_k^\star}\right) \times_{k+1}\left(\mathcal{P}_{\widehat{\mathbf{U}}_{k+1}^{(t)}}-\mathcal{P}_{\mathbf{U}_{k+1}^\star}\right) \times_{k+2}\mathbf{U}_{k+2}^\star, \mathcal{H} \times_{k+2}\mathbf{U}_{k+2} ^{\star\top}  \right\rangle \right|\\
  =  & \Op{\sigma  L_t^2 \sqrt{\frac{pr}{t}} \sum_{k=1}^3 \left\|\cH\times_k \mathbf{U}_{k} ^{\star}  \right\|_\mathrm{F}},
\end{aligned}
\end{align}
and
\begin{align}\label{equ:bound of ZU3}
\begin{aligned}
	 & \left|\sum_{k=1}^3\left\langle{\mathcal{Z}}^{(t)} \times{ }_k\left(\widehat{\mathbf{U}}_k ^{(t)}\widehat{\mathbf{U}}_k^{(t)\top}-\mathcal{P}_{\mathbf{U}_k^\star}\right) \times_{k+1}\mathcal{P}_{\mathbf{U}_{k+1}^\star} \times_{k+2} \mathcal{P}_{\mathbf{U}_{k+2}^\star}, \mathcal{H}\right\rangle\right| \\
  	\stackrel{(b)}{\leq}&  \sum_{k=1}^3 \left|\left\langle \left(\widehat{\mathbf{U}}_k ^{(t)}\widehat{\mathbf{U}}_k^{(t)\top}-\mathcal{P}_{\mathbf{U}_k^\star}\right)  \mathcal{M}_1\left(\mathcal{Z}^{(t)}\right) \left(\mathbf{U}_{k+1} ^{\star} \otimes \mathbf{U}_{k+2} ^{\star}\right)  ,  \mathbf{H}_k\left( \mathbf{U}_{k+1} ^{\star} \otimes\mathbf{U}_{k+2} ^{\star}\right) \right\rangle\right| \\
  = & \Op{\sigma  L_t \sqrt{\frac{pr}{t}} \sum_{k=1}^3\left\|\mathbf{H}_k\left( \mathbf{U}_{k+1} ^{\star} \otimes\mathbf{U}_{k+2} ^{\star}\right)  \right\|_\mathrm{F}} .
\end{aligned}
\end{align}
Here, (a) and (b) arise from the basic property of inner product of a tensor $\left\langle\mathcal{X}, \mathcal{Y} \times{ }_n \mathbf{A}\right\rangle=\left\langle\mathcal{X} \times{ }_n \mathbf{A}^{\top}, \mathcal{Y}\right\rangle$.

\subsection{Proof of Lemma \ref{lem:bound of PuPdu}}\label{sec:proof of lemma bound of PuPdu}
First, 
  \begin{align*}
\UT{1}\Pdu{1}
         \stackrel{(a)}{=} &  \UT{1} \sum_{n \geq 1} \mathcal{S}_{\mathbf{G}_1, n}\left(\mathfrak{E}_1^{(t)}\right) 
       =  \UT{1} \sbr{ \mathcal{S}_{\mathbf{G}_1, 1}\left(\mathfrak{E}_1^{(t)}\right) + \sum_{n \geq 2} \mathcal{S}_{\mathbf{G}_1, n}\left(\mathfrak{E}_1^{(t)}\right)}, 
  \end{align*}
where (a) is due to the Lemma \ref{lem:theorem 1 in xia2021normal} and Equation (\ref{B.3}).
We first consider the first part
\begin{align*}
	\UT{1} \mathcal{S}_{\mathbf{G}_1, 1}\left(\mathfrak{E}_1^{(t)}\right) = \UT{1} \sbr{ \mathfrak{P}_1^{-1} \mathfrak{E}_1^{(t)} \mathfrak{P}_1^{\perp}+\mathfrak{P}_1^{\perp} \mathfrak{E}_1^{(t)} \mathfrak{P}_1^{-1} } =  \UT{1} \mathfrak{P}_1^{-1} \mathfrak{E}_1^{(t)} \mathfrak{P}_1^{\perp}.
\end{align*}
And since the definition of $\mathfrak{E}_1^{(t)}$ in Equation (\ref{equ:def of E1}), we have $\mathfrak{J}_2^{(t)}\Pu{1}^\perp = \mathfrak{J}_4^{(t)}\Pu{1}^\perp = 0$. By Lemma \ref{lem:bound for inference}, we have
	\begin{align*}
		\nbr{\UT{1} \mathcal{S}_{\mathbf{G}_1, 1}\left(\mathfrak{E}_1^{(t)}\right) - \Lambda_{1}^{-2} \G{1} \rbr{\U{2}\otimes\U{3}} \cM_1^\top\rbr{\cZ_1^{(t)}}  \Pu{1}^\perp } =  \Op{\kappa_0^2\sigma^2\frac{p}{t} + \kappa_0\lambda_{\min}\sigma\sqrt{\frac{p\df}{t^{1+\alpha}}}}.
	\end{align*}
	Note that  under the Lemma \ref{lem:bound for first round}, we have
\begin{align*}
\begin{aligned}
& \left\|\mathcal{S}_{\mathbf{G}_1, n}\left(\mathfrak{E}_1^{(t)} \right)\right\| 
\leq & \binom{2n}{n} \left(\frac{\left\|\mathfrak{E}_1^{(t)} \right\|}{\lambda_{\text{min }}^{2}} \right)^n \stackrel{(a)}{\leq} \left(\frac{4\left\|\mathfrak{E}_1^{(t)} \right\|}{\lambda_{\text{min }}^2}\right)^n,
\end{aligned}
\end{align*}
and 
\begin{align}\label{equ:bound of Sg gep 2}
\left\|\sum_{n\geq 2}\mathcal{S}_{\mathbf{G}_1, n}\left(\mathfrak{E}_1^{(t)} \right)\right\| 
\leq \sum_{n \geq 2} \left(\frac{4\left\|\mathfrak{E}_1^{(t)} \right\|}{\lambda_{\min }^2}\right)^n 
= \Op{\kappa_0^2 \frac{\sigma^2p}{\lambda_{\min }^2t}}.
\end{align}
Here, (a) arises from $\binom{2n}{n} \leq \frac{4^n}{\sqrt{\pi n}}\left(1-\frac{1}{9 n}\right)$.
	Combined with Equation (\ref{equ:bound of Sg gep 2}) for the second term, we have
	\begin{align*}
		\nbr{\UT{1}\Pdu{1} - \Lambda_{1}^{-2} \G{1} \rbr{\U{2}\otimes\U{3}} \cM_1^\top\rbr{\cZ_1^{(t)}}  \Pu{1}^\perp } =  \Op{\kappa_0^2\frac{\sigma^2}{\lambda_{\min}^2}\sqrt{\frac{p^2}{t^{1+\alpha}}} }.
	\end{align*}

\subsection{Proof of Lemma \ref{lem:bound of O3}}\label{sec:proof of lemma bound of O3}

We apply the first-order expansion to \(\left(\mathcal{P}_{\widehat{\mathbf{U}}_1^{(t)}}-\mathcal{P}_{\mathbf{U}_1^\star}\right)\) in Equation (\ref{B.3}), and we observe:
\begin{align}\label{equ:m2}
\begin{aligned}
		&\left\langle\mathcal{T}^\star \times_1\left(\mathcal{P}_{\widehat{\mathbf{U}}_1^{(t)}}-\mathcal{P}_{\mathbf{U}_1^\star}\right) \times_2 \mathcal{P}_{\mathbf{U}_2^\star} \times_3\mathcal{P}_{\mathbf{U}_3^\star}, \mathcal{H}\right\rangle \\
	=& \left\langle\mathcal{T}^\star \times_1 \left[ \mathcal{S}_{\mathbf{G}_1, 1}\left(\mathfrak{E}_1^{(t)}\right) + \sum_{n \geq 2} \mathcal{S}_{\mathbf{G}_1, n}\left(\mathfrak{E}_1^{(t)}\right)\right]\times_2 \mathcal{P}_{\mathbf{U}_2^\star} \times_3\mathcal{P}_{\mathbf{U}_3^\star}, \mathcal{H}\right\rangle .
\end{aligned}
\end{align}
For the second term, using the Cauchy-Schwarz inequality and Equation (\ref{equ:bound of Sg gep 2}),  
we obtain:
\begin{align*}
\begin{aligned}
 \left\langle\mathcal{T}^\star \times_1 \sum_{n \geq 2} \mathcal{S}_{\mathbf{G}_1, n}\left(\mathfrak{E}_1^{(t)}\right), \mathcal{H} \times_2 \mathcal{P}_{\mathbf{U}_2^\star} \times_3\mathcal{P}_{\mathbf{U}_3^\star}\right\rangle
        \leq & \sqrt{r}\left\|\mathbf{T}_1^\star\right\|\left\|\sum_{n \geq 2} \mathcal{S}_{\mathbf{G}_1, n}\left(\mathfrak{E}_1^{(t)}\right)\right\|\left\| \mathbf{H}_1\left(\mathbf{U}_3^\star  \otimes \mathbf{U}_2^\star\right)\right\|_\mathrm{F}\\
  	= & \Op{\sqrt{r}\kappa_0 \lambda_{\min}\kappa_0^2 \frac{\sigma^2p}{\lambda_{\min }^2t}\left\|\mathbf{H}_1\left( \mathbf{U}_3^\star \otimes \mathbf{U}_2^\star \right)\right\|_\mathrm{F}} .
  	\end{aligned}
\end{align*}
Next, we will proceed to further expand the first term on the right-hand side of Equation (\ref{equ:m2}). Since $\mathcal{T}^\star = \mathcal{G}^\star\times_1\mathbf{U}_1^\star\times_2\mathbf{U}_2^\star
\times_3\mathbf{U}_3^\star$, we have  $\mathfrak{P}_1^{\perp}\mathbf{U}_1^\star = 0$ and $\mathfrak{P}_1^{\perp}\mathbf{T}_1^\star = 0$. Hence, 
\begin{align*}
	\mathcal{S}_{\mathbf{G}_1, 1}\left(\mathfrak{E}_1^{(t)}\right)\U{1} = & \mathfrak{P}_1^{-1} \mathfrak{E}_1^{(t)} \mathfrak{P}_1^{\perp}\mathbf{U}_1^\star + \mathfrak{P}_1^{\perp} \mathfrak{E}_1^{(t)} \mathfrak{P}_1^{-1}\U{1}  = \mathfrak{P}_1^{\perp} \mathfrak{E}_1^{(t)} \mathfrak{P}_1^{-1}\U{1} \\
 = & \mathfrak{P}_1^{\perp}\left(\mathfrak{J}_1^{(t)}+\mathfrak{J}_2^{(t)}+\mathfrak{J}_3^{(t)}+\mathfrak{J}_4^{(t)}\right)\mathfrak{P}_1^{-1}\U{1} 
	=	\mathfrak{P}_1^{\perp}\left(\mathfrak{J}_2^{(t)}+\mathfrak{J}_3^{(t)}\right)\mathfrak{P}_1^{-1}\U{1}.
\end{align*}
Substituting the above equation into the first term on the right-hand side of Equation (\ref{equ:m2}), we get:
\begin{align*}
	&\mathcal{T}^\star \times_1\mathcal{S}_{\mathbf{G}_1, 1}\left(\mathfrak{E}_1^{(t)}\right) \times_2 \mathcal{P}_{\mathbf{U}_2^\star} \times_3\mathcal{P}_{\mathbf{U}_3^\star}\\
	  =& \mathcal{T}^\star \times_1 \mathfrak{P}_1^{\perp}\left(\mathfrak{J}_2^{(t)}+\mathfrak{J}_3^{(t)}\right)\mathfrak{P}_1^{-1}  \times_2 \mathcal{P}_{\mathbf{U}_2^\star} \times_3\mathcal{P}_{\mathbf{U}_3^\star} \\
	=& \mathcal{T}^\star \times_1 \mathfrak{P}_1^{\perp}\mathfrak{J}_2^{(t)}\mathfrak{P}_1^{-1}  \times_2 \mathcal{P}_{\mathbf{U}_2^\star} \times_3\mathcal{P}_{\mathbf{U}_3^\star} + \mathcal{T}^\star \times_1 \mathfrak{P}_1^{\perp}\mathfrak{J}_3^{(t)}\mathfrak{P}_1^{-1}  \times_2 \mathcal{P}_{\mathbf{U}_2^\star} \times_3\mathcal{P}_{\mathbf{U}_3^\star}.
\end{align*}
By Lemma \ref{lem:bound for first round} and Cauchy-Schwarz inequality, 
we have
  \begin{align*}
  \left\langle\mathcal{T}^\star \times_1 \mathfrak{P}_1^{\perp}\mathfrak{J}_3^{(t)}\mathfrak{P}_1^{-1} , \mathcal{H} \times_2 \mathcal{P}_{\mathbf{U}_2^\star} \times_3\mathcal{P}_{\mathbf{U}_3^\star}\right\rangle
   = & \left\langle \mathfrak{P}_1^{\perp}\mathfrak{J}_3^{(t)}\mathfrak{P}_1^{-1} \mathbf{T}_1^\star, \mathbf{H}_1 
   \left(\mathcal{P}_{\mathbf{U}_3^\star} \otimes\mathcal{P}_{\mathbf{U}_2^\star}\right)\right\rangle\\
   \leq & \sqrt{r}\left\|\mathbf{T}_1^\star\right\|\left\|\mathfrak{P}_1^{\perp}\mathfrak{J}_3^{(t)}\mathfrak{P}_1^{-1}\right\|
   \left\|\mathbf{H}_1 
   \left(\mathbf{U}_3^\star \otimes\mathbf{U}_2^\star \right)\right\|_\mathrm{F}\\
    = & \Op{\kappa_0 \frac{\sigma^2}{\lambda_{\min}} \sqrt{\frac{p^2r}{t^2}} \left\|\mathbf{H}_1 
   \left(\mathbf{U}_3^\star \otimes\mathbf{U}_2^\star \right)\right\|_\mathrm{F}},
  \end{align*}
and
\begin{align*}
\begin{aligned}
\left|\left\langle\mathcal{T}^\star \times_1 \mathfrak{P}_1^{\perp}\left[\mathfrak{J}_2^{(t)}-\mathbf{Z}_1^{(t)}\left(\mathcal{P}_{\mathbf{U}_3^\star} \otimes \mathcal{P}_{\mathbf{U}_2^\star}\right) \mathbf{T}_1^{\top}  \right]\mathfrak{P}_1^{-1}  \times_2 \mathcal{P}_{\mathbf{U}_2^\star} \times_3\mathcal{P}_{\mathbf{U}_3^\star}, \mathcal{H}\right\rangle\right|
		= \Op{\kappa_0^2 \frac{\sigma^2}{\lambda_{\min}}\sqrt{\frac{p^2r}{t^2}}\left\|\mathbf{H}_1 
   \left(\mathbf{U}_3^\star \otimes\mathbf{U}_2^\star \right)\right\|_\mathrm{F}}.
\end{aligned}
\end{align*}
By combining the above inequality with Equation (\ref{equ:m2}), we can derive:
\begin{align}\label{equ:m4}
\begin{aligned}
        & \left|\left\langle\mathcal{T}^\star \times_1\left(\mathcal{P}_{\widehat{\mathbf{U}}_1^{(t)}}-\mathcal{P}_{\mathbf{U}_1^\star}\right) \times_2 \mathcal{P}_{\mathbf{U}_2^\star} \times_3\mathcal{P}_{\mathbf{U}_3^\star}, \mathcal{H}\right\rangle\right| \\
    =& \left|\left\langle\mathcal{T}^\star \times_1 \mathfrak{P}_1^{\perp}\mathbf{Z}_1^{(t)}\left(\mathcal{P}_{\mathbf{U}_3^\star} \otimes \mathcal{P}_{\mathbf{U}_2^\star}\right) \mathbf{T}_1^{\top} \mathfrak{P}_1^{-1}  \times_2 \mathcal{P}_{\mathbf{U}_2^\star} \times_3\mathcal{P}_{\mathbf{U}_3^\star}, \mathcal{H}\right\rangle\right|\\
    & + \Op{\kappa_0^3 \frac{\sigma^2}{\lambda_{\min }}\sqrt{\frac{p^2r}{t^2}} \left\|\mathbf{H}_1 
   \left(\mathbf{U}_3^\star \otimes\mathbf{U}_2^\star \right)\right\|_\mathrm{F}}.
\end{aligned}
\end{align}
Regarding the first term on the right-hand side of Equation (\ref{equ:m4}), by applying Equation (\ref{B.2}), we find the following:
\begin{align*}
	 \mathfrak{P}_1^{\perp}\mathbf{Z}_1^{(t)}\left(\mathcal{P}_{\mathbf{U}_3^\star} \otimes \mathcal{P}_{\mathbf{U}_2^\star}\right) \mathbf{T}_1^{\top} \mathfrak{P}_1^{-1} 
	= & \mathcal{P}_{\mathbf{U}_1^\star}^{\perp}\mathbf{Z}_1^{(t)}\left(\mathcal{P}_{\mathbf{U}_3^\star} \otimes \mathcal{P}_{\mathbf{U}_2^\star}\right)\left(\mathbf{U}_3^\star\otimes\mathbf{U}_2^\star\right)\mathbf{G}_1^{\star\top}\mathbf{U}_1^{\star\top}\mathbf{U}_1^\star\mathbf{\Lambda}_1^{-2}\mathbf{U}_1^{\star\top}\\
	= & \mathcal{P}_{\mathbf{U}_1^\star}^{\perp}\mathbf{Z}_1^{(t)}\left(\mathbf{U}_3^\star\otimes\mathbf{U}_2^\star\right)\mathbf{G}_1^{\star\top}\mathbf{\Lambda}_1^{-2}\mathbf{U}_1^{\star\top}.
\end{align*}
Therefore, the first term in the right of Equation (\ref{equ:m4}) can be written as: 
\begin{align*}
	&\left\langle\mathcal{T}^\star \times_1 \mathfrak{P}_1^{\perp}\mathbf{Z}_1^{(t)}\left(\mathcal{P}_{\mathbf{U}_3^\star} \otimes \mathcal{P}_{\mathbf{U}_2^\star}\right) \mathbf{T}_1^{\star\top} \mathfrak{P}_1^{-1}  \times_2 \mathcal{P}_{\mathbf{U}_2^\star} \times_3\mathcal{P}_{\mathbf{U}_3^\star}, \mathcal{H}\right\rangle\\
	= &  \left\langle\mathcal{T}^\star \times_1 \mathcal{P}_{\mathbf{U}_1^\star}^{\perp}\mathbf{Z}_1^{(t)}\left(\mathbf{U}_3^\star\otimes\mathbf{U}_2^\star\right)\mathbf{G}_1^{\star\top}\mathbf{\Lambda}_1^{-2}\mathbf{U}_1^{\star\top}  \times_2 \mathcal{P}_{\mathbf{U}_2^\star} \times_3\mathcal{P}_{\mathbf{U}_3^\star}, \mathcal{H}\right\rangle\\
	\stackrel{(a)}{=} &  \left\langle\mathcal{G}^\star \times_1 \mathcal{P}_{\mathbf{U}_1^\star}^{\perp}\mathbf{Z}_1^{(t)}\left(\mathbf{U}_3^\star\otimes\mathbf{U}_2^\star\right)\mathbf{G}_1^{\star\top}\mathbf{\Lambda}_1^{-2} \times_2 \mathbf{U}_2^\star \times_3 \mathbf{U}_3^\star, \mathcal{H}\right\rangle\\
		= &  \left\langle \mathcal{P}_{\mathbf{U}_1^\star}^{\perp}\mathbf{Z}_1^{(t)}\left(\mathbf{U}_3^\star\otimes\mathbf{U}_2^\star\right)\mathbf{G}_1^{\star\top}\mathbf{\Lambda}_1^{-2} \mathbf{G}_1^\star \left(\mathbf{U}_3^\star \otimes \mathbf{U}_2^\star\right)^\top, \mathbf{H}_1 \right\rangle\\
		= &  \left\langle \mathbf{Z}_1^{(t)}, \mathcal{P}_{\mathbf{U}_1^\star}^{\perp}\mathbf{H}_1 \left(\mathbf{U}_3^\star\otimes\mathbf{U}_2^\star\right)\mathbf{G}_1^{\star\top}\mathbf{\Lambda}_1^{-2} \mathbf{G}_1^\star \left(\mathbf{U}_3^\star \otimes \mathbf{U}_2^\star\right)^\top\right\rangle .
\end{align*}
Here, (a) arises from $\mathcal{T}^\star = \mathcal{G}^\star \times_1\mathbf{U}_1^\star \times_2\mathbf{U}_2^\star \times_3\mathbf{U}_3^\star$.
Due to from Equation (\ref{B.2}), there exists matrix $\mathbf{V}_k^\star\in\mathbb{O}_{r_{-k}, r_k}$, such that  $\mathbf{G}_k^{\star} = \boldsymbol{\Lambda}_k^\star \mathbf{V}_k^{\star \top}$.  
Thus we have 
\begin{align*}
    \mathbf{G}_1^{\star\top}\left(\mathbf{\Lambda}_1^{\star}\right)^{-2} \mathbf{G}_1^\star = \mathbf{V}_1^\star \boldsymbol{\Lambda}_1^\star\left(\mathbf{\Lambda}_1^{\star}\right)^{-2}  \boldsymbol{\Lambda}_1^{\star} \mathbf{V}_1^{\star\top} = \mathbf{V}_1^\star\mathbf{V}_1^{\star\top} 
   = \mathbf{G}_k^{\star\dagger}\boldsymbol{\Lambda}_k\boldsymbol{\Lambda}_k^{-1}\mathbf{G}_k^{\star} = \mathbf{G}_k^{\star\dagger}\mathbf{G}_k^{\star}.
\end{align*}
Thus we have 
\begin{align*}
\mathcal{P}_{\mathbf{U}_1^\star}^{\perp}\mathbf{H}_1\left(\mathbf{U}_3^\star\otimes\mathbf{U}_2^\star\right)\mathbf{G}_1^{\star\top}\mathbf{\Lambda}_1^{-2} \mathbf{G}_1^\star \left(\mathbf{U}_3^\star \otimes \mathbf{U}_2^\star\right)^\top = 
\mathcal{P}_{\mathbf{U}_1^\star}^{\perp} \mathbf{H}_1\left(\mathbf{U}_3^\star\otimes\mathbf{U}_2^\star\right) \mathbf{V}_1^\star\mathbf{V}_1^{\star\top}\left(\mathbf{U}_3^\star \otimes \mathbf{U}_2^\star\right)^\top
\end{align*}
We use notation $\mathcal{P}_{\left(\mathbf{U}_{k+2}^\star\otimes\mathbf{U}_{k+1}^\star\right)  \mathbf{V}_k^\star} := \left(\mathbf{U}_{k+2}^\star\otimes\mathbf{U}_{k+1}^\star\right)  \mathbf{V}_k^\star\mathbf{V}_k^{\star\top} \left(\mathbf{U}_{k+1}^\star \otimes \mathbf{U}_{k+2}^\star\right)^\top$. Put the above equality into previous Equation, we can have 
\begin{align*}
    & \left\langle \mathbf{Z}_1^{(t)}, \mathcal{P}_{\mathbf{U}_1^\star}^{\perp}\mathbf{H}_1 \left(\mathbf{U}_3^\star\otimes\mathbf{U}_2^\star\right)\mathbf{G}_1^{\star\top}\mathbf{\Lambda}_1^{-2} \mathbf{G}_1^\star \left(\mathbf{U}_3^\star \otimes \mathbf{U}_2^\star\right)^\top\right\rangle  \\
    = & \left\langle \mathbf{Z}_1^{(t)}, \mathcal{P}_{\mathbf{U}_1^\star}^{\perp}\mathbf{H}_1 \left(\mathbf{U}_3^\star\otimes\mathbf{U}_2^\star\right)  \mathbf{V}_1^\star\mathbf{V}_1^{\star\top} \left(\mathbf{U}_3^\star \otimes \mathbf{U}_2^\star\right)^\top\right\rangle\\
    = & \left\langle \mathbf{Z}_1^{(t)}, \mathcal{P}_{\mathbf{U}_1^\star}^{\perp}\mathbf{H}_1 \mathcal{P}_{\left(\mathbf{U}_3^\star\otimes\mathbf{U}_2^\star\right)  \mathbf{V}_1^\star} \right\rangle.
\end{align*}

Substituting the above result into Equation (\ref{equ:m4}), 
we obtain:
\begin{align*}
\begin{aligned}
        &\left\langle\mathcal{T}^\star \times_1\left(\mathcal{P}_{\widehat{\mathbf{U}}_1^{(t)}}-\mathcal{P}_{\mathbf{U}_1^\star}\right) \times_2 \mathcal{P}_{\mathbf{U}_2^\star} \times_3\mathcal{P}_{\mathbf{U}_3^\star}, \mathcal{H}\right\rangle \\
    =& \left\langle \mathbf{Z}_1^{(t)}, \mathcal{P}_{\mathbf{U}_1^\star}^{\perp}\mathbf{H}_1 \mathcal{P}_{\left(\mathbf{U}_3^\star\otimes\mathbf{U}_2^\star\right)  \mathbf{V}_1^\star} \right\rangle  + \Op{\kappa_0^3 \frac{\sigma^2}{\lambda_{\min }}\sqrt{\frac{p^2r}{t^2}} \left\|\mathbf{H}_1 
   \left(\mathbf{U}_3^\star \otimes\mathbf{U}_2^\star \right)\right\|_\mathrm{F}}.
\end{aligned}
\end{align*}
Thus, we have proofed Lemma \ref{lem:bound of O3}.

\subsection{Proof of Lemma \ref{lem:clt}} \label{sec:proof of lemma clt}
We first observe that from Equation (\ref{equ:t=t+z}), we can express ${\mathcal{Z}}^{(t)}$ as the sum of ${\mathcal{Z}_1}^{(t)}$ and ${\mathcal{Z}_2}^{(t)}$. Consequently, we can rewrite the primary term into the following two components:
\begin{align}\label{equ:CLT-1}
\left\langle {\mathcal{Z}_1}^{(t)} 
        \bigtimes_{k\in[3]} \mathcal{P}_{\mathbf{U}_k^\star}, \mathcal{H} \right\rangle 
        + \sum_{k=1}^3   \left\langle \mathcal{M}_k\left({\mathcal{Z}_1}^{(t)}\right) , \mathcal{P}_{\mathbf{U}_k^\star}^{\perp}\mathbf{H}_k \mathcal{P}_{\left(\mathbf{U}_{k+1}^\star\otimes\mathbf{U}_{k+2}^\star\right)  \mathbf{V}_k^\star} \right\rangle ,
\end{align}
and
\begin{align}\label{equ:CLT-2}
\left\langle {\mathcal{Z}_2}^{(t)} 
        \bigtimes_{k\in[3]} \mathcal{P}_{\mathbf{U}_k^\star}, \mathcal{H} \right\rangle 
        + \sum_{k=1}^3   \left\langle \mathcal{M}_k\left({\mathcal{Z}_2}^{(t)}\right) , \mathcal{P}_{\mathbf{U}_k^\star}^{\perp}\mathbf{H}_k \mathcal{P}_{\left(\mathbf{U}_{k+1}^\star\otimes\mathbf{U}_{k+2}^\star\right)  \mathbf{V}_k^\star} \right\rangle .
\end{align}
We then need to prove that Equation (\ref{equ:CLT-1}) is asymptotic normal while Equation (\ref{equ:CLT-2}) has a smaller order than the variance of the first term and thus converges to zero.

First, we can see that 
\begin{align*}
    & \left\langle {\mathcal{Z}_1}^{(t)} 
       \bigtimes_{k\in[3]} \mathcal{P}_{\mathbf{U}_k^\star}, \mathcal{H} \right\rangle 
        + \sum_{k=1}^3   \left\langle \mathcal{M}_k\left({\mathcal{Z}_1}^{(t)}\right) , \mathcal{P}_{\mathbf{U}_k^\star}^{\perp}\mathbf{H}_k \mathcal{P}_{\left(\mathbf{U}_{k+1}^\star\otimes\mathbf{U}_{k+2}^\star\right)  \mathbf{V}_k^\star} \right\rangle  \\
    = & \frac{1}{t} \sum_{\tau=1}^t  \xi_\tau \left\langle \mathcal{X}_\tau,\mathcal{H} \bigtimes_{k\in[3]} \mathcal{P}_{\mathbf{U}_k^\star}  \right\rangle
    + \frac{1}{t} \sum_{\tau=1}^t \sum_{k=1}^3  \xi_\tau \left\langle \mathcal{M}_k\left(\mathcal{X}_\tau\right), \mathcal{P}_{\mathbf{U}_k^\star}^{\perp}\mathbf{H}_k \mathcal{P}_{\left(\mathbf{U}_{k+1}^\star\otimes\mathbf{U}_{k+2}^\star\right)  \mathbf{V}_k^\star}  \right\rangle \\
      = & \frac{1}{t} \sum_{\tau=1}^t  \xi_\tau \left( \left\langle \mathcal{X}_\tau,\mathcal{H}  \bigtimes_{k\in[3]} \mathcal{P}_{\mathbf{U}_k^\star}  \right\rangle
    + \sum_{k=1}^3  \left\langle \mathcal{M}_k\left(\mathcal{X}_\tau\right), \mathcal{P}_{\mathbf{U}_k^\star}^{\perp}\mathbf{H}_k \mathcal{P}_{\left(\mathbf{U}_{k+1}^\star\otimes\mathbf{U}_{k+2}^\star\right)  \mathbf{V}_k^\star}  \right\rangle \right) .
\end{align*}
By the distribution of $\xi$ and $\mathcal{X}$ in Assumption \ref{cond:1}, we can get that 
\begin{align*}
	\mathbb{E}\left[\left\langle {\mathcal{Z}_1}^{(t)} 
        \bigtimes_{k\in[3]} \mathcal{P}_{\mathbf{U}_k^\star}, \mathcal{H} \right\rangle 
        + \sum_{k=1}^3   \left\langle \mathcal{M}_k\left({\mathcal{Z}_1}^{(t)}\right) , \mathcal{P}_{\mathbf{U}_k^\star}^{\perp}\mathbf{H}_k \mathcal{P}_{\left(\mathbf{U}_{k+1}^\star\otimes\mathbf{U}_{k+2}^\star\right)  \mathbf{V}_k^\star} \right\rangle \right] = 0.
\end{align*}
and
\begin{align*}
	& \text{Var}\left[ \left\langle \mathcal{X}_\tau,\mathcal{H} \bigtimes_{k\in[3]} \mathcal{P}_{\mathbf{U}_k^\star}  \right\rangle
    + \sum_{k=1}^3  \left\langle \mathcal{M}_k\left(\mathcal{X}_\tau\right), \mathcal{P}_{\mathbf{U}_k^\star}^{\perp}\mathbf{H}_k \mathcal{P}_{\left(\mathbf{U}_{k+1}^\star\otimes\mathbf{U}_{k+2}^\star\right)  \mathbf{V}_k^\star}  \right\rangle  \right]\\
    = & \text{Var}\left[ \left\langle \mathcal{X}_\tau,\mathcal{H} \bigtimes_{k\in[3]} \mathcal{P}_{\mathbf{U}_k^\star}  \right\rangle\right]
         +  \sum_{k=1}^3 \text{Var}\left[ \left\langle \mathcal{M}_k\left(\mathcal{X}_\tau\right), \mathcal{P}_{\mathbf{U}_k^\star}^{\perp}\mathbf{H}_k \mathcal{P}_{\left(\mathbf{U}_{k+1}^\star\otimes\mathbf{U}_{k+2}^\star\right)  \mathbf{V}_k^\star}  \right\rangle \right] \\
        & + \sum_{k=1}^3 \mathbb{E}\left[\left\langle \mathcal{X}_\tau,\mathcal{H} \bigtimes_{k\in[3]} \mathcal{P}_{\mathbf{U}_k^\star}  \right\rangle \left\langle \mathcal{M}_k\left(\mathcal{X}_\tau\right), \mathcal{P}_{\mathbf{U}_k^\star}^{\perp}\mathbf{H}_k \mathcal{P}_{\left(\mathbf{U}_{k+1}^\star\otimes\mathbf{U}_{k+2}^\star\right)  \mathbf{V}_k^\star}  \right\rangle   \right]\\
          & +  \sum_{j,k\in [3], j\neq k} \mathbb{E}\left[\left\langle \mathcal{M}_j\left(\mathcal{X}_\tau\right), \mathcal{P}_{\mathbf{U}_j^\star}^{\perp}\mathbf{H}_j \mathcal{P}_{\left(\mathbf{U}_{j+1}^\star\otimes\mathbf{U}_{j+2}^\star\right)  \mathbf{V}_j^\star}  \right\rangle  \left\langle \mathcal{M}_k\left(\mathcal{X}_\tau\right), \mathcal{P}_{\mathbf{U}_k^\star}^{\perp}\mathbf{H}_k \mathcal{P}_{\left(\mathbf{U}_{k+1}^\star\otimes\mathbf{U}_{k+2}^\star\right)  \mathbf{V}_k^\star}  \right\rangle  \right].
\end{align*}
We find that 
\begin{align*}
	\text{Var}\left[ \left\langle \mathcal{X}_\tau,\mathcal{H} \bigtimes_{k\in[3]} \mathcal{P}_{\mathbf{U}_k^\star}  \right\rangle\right] = \left\|\mathcal{H} \bigtimes_{k\in[3]} \mathbf{U}_k^\star  \right\|_{\mathrm{F}}^2.
\end{align*}
and
\begin{align*}
	\sum_{k=1}^3 \text{Var}\left[ \left\langle \mathcal{M}_k\left(\mathcal{X}_\tau\right), \mathcal{P}_{\mathbf{U}_k^\star}^{\perp}\mathbf{H}_k \mathcal{P}_{\left(\mathbf{U}_{k+1}^\star\otimes\mathbf{U}_{k+2}^\star\right)  \mathbf{V}_k^\star}  \right\rangle \right] = \sum_{k=1}^3 \left\| \mathcal{P}_{\mathbf{U}_k^\star}^{\perp}\mathbf{H}_k \mathcal{P}_{\left(\mathbf{U}_{k+1}^\star\otimes\mathbf{U}_{k+2}^\star\right)  \mathbf{V}_k^\star}  \right\|_\mathrm{F} ^2. 
\end{align*}
Since $\mathcal{P}_{{\mathbf{U}}_k^\star}^{\perp} \mathcal{P}_{{\mathbf{U}}_k^\star} = 0$, thus for the intersection terms we have:
\begin{align*}
	\mathbb{E}\left[\left\langle \mathcal{X}_\tau,\mathcal{H} \bigtimes_{k\in[3]} \mathcal{P}_{\mathbf{U}_k^\star}  \right\rangle \left\langle \mathcal{M}_k\left(\mathcal{X}_\tau\right), \mathcal{P}_{\mathbf{U}_k^\star}^{\perp}\mathbf{H}_k \mathcal{P}_{\left(\mathbf{U}_{k+1}^\star\otimes\mathbf{U}_{k+2}^\star\right)  \mathbf{V}_k^\star}  \right\rangle   \right] = 0,
\end{align*}
and
\begin{align*}
	\sum_{j,k=[3], j\neq k} \mathbb{E}\left[\left\langle \mathcal{M}_j\left(\mathcal{X}_\tau\right), \mathcal{P}_{\mathbf{U}_j^\star}^{\perp}\mathbf{H}_j \mathcal{P}_{\left(\mathbf{U}_{j+1}^\star\otimes\mathbf{U}_{j+2}^\star\right)  \mathbf{V}_j^\star}  \right\rangle  \left\langle \mathcal{M}_k\left(\mathcal{X}_\tau\right), \mathcal{P}_{\mathbf{U}_k^\star}^{\perp}\mathbf{H}_k \mathcal{P}_{\left(\mathbf{U}_{k+1}^\star\otimes\mathbf{U}_{k+2}^\star\right)  \mathbf{V}_k^\star}  \right\rangle  \right] = 0.
\end{align*}
Thus, we have
\begin{align*}
& \text{Var}\left[ \left\langle \mathcal{X}_\tau,\mathcal{H} \bigtimes_{k\in[3]} \mathcal{P}_{\mathbf{U}_k^\star}  \right\rangle
    + \sum_{k=1}^3  \left\langle \mathcal{M}_k\left(\mathcal{X}_\tau\right), \mathcal{P}_{\mathbf{U}_k^\star}^{\perp}\mathbf{H}_k \mathcal{P}_{\left(\mathbf{U}_{k+1}^\star\otimes\mathbf{U}_{k+2}^\star\right)  \mathbf{V}_k^\star}  \right\rangle  \right]\\
    & = \left\|\mathcal{H} \bigtimes_{k\in[3]} \mathbf{U}_k^\star  \right\|_{\mathrm{F}}^2 +  \sum_{k=1}^3 \left\| \mathcal{P}_{\mathbf{U}_k^\star}^{\perp}\mathbf{H}_k \mathcal{P}_{\left(\mathbf{U}_{k+1}^\star\otimes\mathbf{U}_{k+2}^\star\right)  \mathbf{V}_k^\star}  \right\|_\mathrm{F} ^2 .
\end{align*}
Since the distribution of $\xi$ and $\mathcal{X}$ are independent, we have
\begin{align*}
	& \text{Var}\left[\left\langle {\mathcal{Z}_1}^{(t)} 
        \bigtimes_{k\in[3]} \mathcal{P}_{\mathbf{U}_k^\star}, \mathcal{H} \right\rangle 
        + \sum_{k=1}^3   \left\langle \mathcal{M}_k\left({\mathcal{Z}_1}^{(t)}\right) , \mathcal{P}_{\mathbf{U}_k^\star}^{\perp}\mathbf{H}_k \mathcal{P}_{\left(\mathbf{U}_{k+1}^\star\otimes\mathbf{U}_{k+2}^\star\right)  \mathbf{V}_k^\star} \right\rangle \right] \\
        & = \frac{1}{t} \sigma^2\left(\left\|\mathcal{H} \bigtimes_{k\in[3]} \mathbf{U}_k^\star  \right\|_{\mathrm{F}}^2 +  \sum_{k=1}^3 \left\| \mathcal{P}_{\mathbf{U}_k^\star}^{\perp}\mathbf{H}_k \mathcal{P}_{\left(\mathbf{U}_{k+1}^\star\otimes\mathbf{U}_{k+2}^\star\right)  \mathbf{V}_k^\star}  \right\|_\mathrm{F} ^2\right).
\end{align*}
We define:
\begin{align}\label{equ:def of S}
S_\mathcal{H}^2 = \left\|\mathcal{H} \bigtimes_{k\in[3]} \mathbf{U}_k^\star \right\|_{\mathrm{F}}^2 
+  \sum_{k=1}^3 \left\| \mathcal{P}_{\mathbf{U}_k^\star}^{\perp}\mathbf{H}_k \mathcal{P}_{\left(\mathbf{U}_{k+1}^\star\otimes\mathbf{U}_{k+2}^\star\right)  \mathbf{V}_k^\star}  \right\|_\mathrm{F} ^2 . 
\end{align}
Drawing from the Central Limit Theorem as presented in \cite[Theorem 5]{ferguson2017course}, we deduce:
\begin{align}\label{equ:clt-3}
\frac{\sqrt{t}}{\sigma S_\mathcal{H}}  \left( \left\langle {\mathcal{Z}_1}^{(t)} 
        \bigtimes_{k\in[3]} \mathcal{P}_{\mathbf{U}_k^\star}, \mathcal{H} \right\rangle 
        + \sum_{k=1}^3   \left\langle \mathcal{M}_k\left({\mathcal{Z}_1}^{(t)}\right) , \mathcal{P}_{\mathbf{U}_k^\star}^{\perp}\mathbf{H}_k \mathcal{P}_{\left(\mathbf{U}_{k+1}^\star\otimes\mathbf{U}_{k+2}^\star\right)  \mathbf{V}_k^\star} \right\rangle  \right)\stackrel{d}{\longrightarrow} \mathcal{N}(0,1).
\end{align}
Next, we focus on the term in Equation \eqref{equ:CLT-2}.

By Lemma \ref{lem:compare z1 and z2}, we have
\begin{align*}
& \left|  \left\langle {\mathcal{Z}_2}^{(t)} 
        \bigtimes_{k\in[3]} \mathcal{P}_{\mathbf{U}_k^\star}, \mathcal{H} \right\rangle 
        + \sum_{k=1}^3   \left\langle \mathcal{M}_k\left({\mathcal{Z}_2}^{(t)}\right) , \mathcal{P}_{\mathbf{U}_k^\star}^{\perp}\mathbf{H}_k \mathcal{P}_{\left(\mathbf{U}_{k+1}^\star\otimes\mathbf{U}_{k+2}^\star\right)  \mathbf{V}_k^\star} \right\rangle \right|  \\
\leq &  \left|  \left\langle {\mathcal{Z}_2}^{(t)}, \mathcal{H} 
        \bigtimes_{k\in[3]} \mathcal{P}_{\mathbf{U}_k^\star} \right\rangle \right| +  \sum_{k=1}^3  \left|  \left\langle \mathcal{M}_k\left({\mathcal{Z}_2}^{(t)}\right) , \mathcal{P}_{\mathbf{U}_k^\star}^{\perp}\mathbf{H}_k \mathcal{P}_{\left(\mathbf{U}_{k+1}^\star\otimes\mathbf{U}_{k+2}^\star\right)  \mathbf{V}_k^\star} \right\rangle \right|\\
        = & \Op{\sigma \sqrt{\frac{\df^2}{t^{1+\alpha} } } 
        \left( \left\|\mathcal{H} \bigtimes_{k\in[3]} \mathbf{U}_k^\star  \right\|_{\mathrm{F}} + \sum_{k=1}^3 \left\| \mathcal{P}_{\mathbf{U}_k^\star}^{\perp}\mathbf{H}_k \mathcal{P}_{\left(\mathbf{U}_{k+1}^\star\otimes\mathbf{U}_{k+2}^\star\right)  \mathbf{V}_k^\star}  \right\|_\mathrm{F}\right)},
\end{align*}
We only need 
\begin{align*}
 & \frac{\sqrt{t}}{\sigma S_\mathcal{H}}  \left( \left\langle {\mathcal{Z}_2}^{(t)} 
        \bigtimes_{k\in[3]} \mathcal{P}_{\mathbf{U}_k^\star}, \mathcal{H} \right\rangle 
        + \sum_{k=1}^3   \left\langle \mathcal{M}_k\left({\mathcal{Z}_2}^{(t)}\right) , \mathcal{P}_{\mathbf{U}_k^\star}^{\perp}\mathbf{H}_k \mathcal{P}_{\left(\mathbf{U}_{k+1}^\star\otimes\mathbf{U}_{k+2}^\star\right)  \mathbf{V}_k^\star} \right\rangle  \right)  \\
         = & \Op{ \frac{1}{S_\mathcal{H}} \sqrt{\frac{\df^2 }{t^\alpha}} 
        \left( \left\|\mathcal{H} \bigtimes_{k\in[3]} \mathbf{U}_k^\star  \right\|_{\mathrm{F}} + \sum_{k=1}^3 \left\| \mathcal{P}_{\mathbf{U}_k^\star}^{\perp}\mathbf{H}_k \mathcal{P}_{\left(\mathbf{U}_{k+1}^\star\otimes\mathbf{U}_{k+2}^\star\right)  \mathbf{V}_k^\star}  \right\|_\mathrm{F}\right)} \\
         = & \Op{\sqrt{\frac{\df^2}{t^\alpha}}} = o_\bP(1).
\end{align*}
Together with Equation (\ref{equ:clt-3}) and the fact above, if we apply Slutsky's theorem, we conclude the proof of the Lemma \ref{lem:clt}.

%% file: body/LemmaComponent.tex
\subsection{Proof of Lemma \ref{lem:bound of S2}}\label{sec:Proof of Lemma S2}
\label{sec:Proof of Lemmas of Theorem U normality}
	Recall that 
\begin{align}\label{B.7-3}
	-\left\langle\mathcal{S}_{\mathbf{G}_1, 2}\left(\mathfrak{E}_1\right), \mathbf{U}_1^\star \mathbf{U}_1^{\star\top}\right\rangle=\operatorname{tr}\left(\mathfrak{P}_1^{-1} \mathfrak{E}_1^{(t)} \mathfrak{P}_1^{\perp} \mathfrak{E}_1^{(t)} \mathfrak{P}_1^{-1}\right).
\end{align}
Since $\mathbf{T}_1^{\star\top} \mathfrak{P}_1^{\perp}=0$ and $\mathfrak{P}_1^{\perp} \mathbf{T}_1^\star=0$, we write
\begin{align}\label{B.13}
\begin{aligned}
\operatorname{tr}\left(\mathfrak{P}_1^{-1} \mathfrak{E}_1^{(t)} \mathfrak{P}_1^{\perp} \mathfrak{E}_1^{(t)} \mathfrak{P}_1^{-1}\right)=&\operatorname{tr}\left(\mathfrak{P}_1^{-1}\left(\mathfrak{J}_1^{(t)}+\mathfrak{J}_3^{(t)} \right) \mathfrak{P}_1^{\perp}\left(\mathfrak{J}_2^{(t)}+\mathfrak{J}_3^{(t)} \right) \mathfrak{P}_1^{-1}\right) \\
= & \operatorname{tr}\left(\mathfrak{P}_1^{-1} \mathfrak{J}_1^{(t)} \mathfrak{P}_1^{\perp} \mathfrak{J}_2^{(t)} \mathfrak{P}_1^{-1}\right)+\operatorname{tr}\left(\mathfrak{P}_1^{-1} \mathfrak{J}_1^{(t)} \mathfrak{P}_1^{\perp} \mathfrak{J}_3^{(t)} \mathfrak{P}_1^{-1}\right)
\\&\quad+\operatorname{tr}\left(\mathfrak{P}_1^{-1} \mathfrak{J}_3^{(t)} \mathfrak{P}_1^{\perp} \mathfrak{J}_2^{(t)} \mathfrak{P}_1^{-1}\right)+\operatorname{tr}\left(\mathfrak{P}_1^{-1} \mathfrak{J}_3^{(t)} \mathfrak{P}_1^{\perp} \mathfrak{J}_3^{(t)} \mathfrak{P}_1^{-1}\right)
\\=&: \mathrm{I}+\mathrm{II}+\mathrm{III}+\mathrm{IV}.
\end{aligned}
\end{align}
By Lemma \ref{lem:bound for first round}, we have
 \begin{align}\label{B.14}
|\mathrm{IV}| \stackrel{(a)}{\leq} r_1\left\|\mathfrak{P}_1^{-1}\right\|^2\left\|\mathfrak{J}_3\right\|^2 = \Op{r_1 \frac{p^2\sigma^4}{t^2\lambda_{\min}^4}}.
\end{align}
Here, (a) is due to $\operatorname{tr}\left(\mathbf{A}\right)\leq \operatorname{rank}\left( \mathbf{A}\right)\cdot\left\|\mathbf{A}\right\|$.
We use the following lemma to  measure the quantities of $\mathrm{II}$ and $\mathrm{III}$:
\begin{lemma}\label{lem:lemma 3}
Under the assumption of Theorem \ref{thm:U normality}, we have
\begin{align}\label{B.16}
\begin{aligned}
		&\left|\mathrm{I}-\operatorname{tr}\left(\mathbf{\Lambda}_1^{-4} \mathbf{G}_1^\star \left(\mathbf{U}_2^{\star\top} \otimes \mathbf{U}_3^{\star\top}\right) \mathbf{Z}_1^{(t)\top} \mathbf{U}_{1 \perp}^\star \mathbf{U}_{1 \perp}^{\star\top} \mathbf{Z}_1^{(t)}\left(\mathbf{U}_2^\star \otimes \mathbf{U}_3^\star\right) \mathbf{G}_1^{\star\top}\right)\right|\\
	= &\Op{\frac{r_1 \kappa_0^2\sigma^4p^2}{t^2 \lambda_{\min }^4}+ \frac{r^{1 / 2}\sigma^3 p }{t^{3/2} \lambda_{\min}^3}},
\end{aligned}
\end{align} 
and
\begin{align}\label{B.15-1}
	|\mathrm{II}|=|\mathrm{III}| = \Op{\frac{r^2p\sigma^3}{t^{3/2}\lambda_{\min}^{3}}+ \frac{r \kappa_0p^2\sigma^4}{t^{2}\lambda_{\min }^4}}.
\end{align}
\end{lemma}
Proof in Section \ref{sec:Proof of Lemma 3}. Combining Equation (\ref{B.13}), (\ref{B.14}), (\ref{B.15-1}) and (\ref{B.16}), we have
\begin{align*}
\begin{aligned}
& \left|\left\langle\mathcal{S}_{\mathbf{G}_1, 2}\left(\mathfrak{E}_1\right), \mathbf{U}_1^\star \mathbf{U}_1^{\star\top}\right\rangle +\operatorname{tr}\left(\mathbf{\Lambda}_1^{-4} \mathbf{G}_1^\star \left(\mathbf{U}_2^{\star\top} \otimes \mathbf{U}_3^{\star\top}\right) \mathbf{Z}_1^{(t)\top} \mathbf{U}_{1 \perp}^\star \mathbf{U}_{1 \perp}^{\star\top} \mathbf{Z}_1^{(t)}\left(\mathbf{U}_2^\star \otimes \mathbf{U}_3^\star\right) \mathbf{G}_1^{\star\top}\right)\right| \\
= & \left|\operatorname{tr}\left(\mathfrak{P}_1^{-1} \mathfrak{E}_1^{(t)} \mathfrak{P}_1^{\perp} \mathfrak{E}_1^{(t)} \mathfrak{P}_1^{-1}\right)-\operatorname{tr}\left(\mathbf{\Lambda}_1^{-4} \mathbf{G}_1^\star \left(\mathbf{U}_2^{\star\top} \otimes \mathbf{U}_3^{\star\top}\right) \mathbf{Z}_1^{(t)\top} \mathbf{U}_{1 \perp}^\star \mathbf{U}_{1 \perp}^{\star\top} \mathbf{Z}_1^{(t)}\left(\mathbf{U}_2^\star \otimes \mathbf{U}_3^\star\right) \mathbf{G}_1^{\star\top}\right)\right| \\
 = & \Op{\frac{r^2p\sigma^3}{t^{3/2}\lambda_{\min}^{3}}
+ \frac{r_1 \kappa_0^2\sigma^4p^2}{t^2 \lambda_{\min }^4}}.
\end{aligned}
\end{align*}

For the second term \eqref{equ:lemma d1-2} in the Lemma \ref{lem:bound of S2}, 
given that $\mathbf{T}_1^{\star\top} \mathfrak{P}_1^{\perp}=0$, $\mathfrak{P}_1^{\perp} \mathbf{T}_1^\star=0$, and $\mathfrak{E}_1^{(t)}=\mathfrak{J}_1^{(t)}+\mathfrak{J}_2^{(t)}+\mathfrak{J}_3^{(t)}+\mathfrak{J}_4^{(t)}$, we can exploit these properties to simplify $\left\langle\mathcal{S}_{\mathbf{G}_1, 3}\left(\mathfrak{E}_1\right), \mathbf{U}_1^\star \mathbf{U}_1^{\star\top}\right\rangle$ as follows:
\begin{align}\label{B.5}
\begin{aligned}
&\left\langle\mathcal{S}_{\mathbf{G}_1, 3}\left(\mathfrak{E}_1\right), \mathbf{U}_1^\star \mathbf{U}_1^{\star\top}\right\rangle\\
= & -2 \operatorname{tr}\left(\mathfrak{P}_1^{-1} \mathfrak{E}_1^{(t)} \mathfrak{P}_1^{\perp} \mathfrak{E}_1^{(t)} \mathfrak{P}_1^{\perp} \mathfrak{E}_1^{(t)} \mathfrak{P}_1^{-2}\right)+2 \operatorname{tr}\left(\mathfrak{P}_1^{-1} \mathfrak{E}_1^{(t)} \mathfrak{P}_1^{\perp} \mathfrak{E}_1^{(t)} \mathfrak{P}_1^{-1} \mathfrak{E}_1^{(t)} \mathfrak{P}_1^{-1}\right) \\
= & -2 \operatorname{tr}\left(\mathfrak{P}_1^{-1}\left(\mathfrak{J}_1^{(t)}+\mathfrak{J}_3^{(t)} \right) \mathfrak{P}_1^{\perp} \mathfrak{J}_3^{(t)} \mathfrak{P}_1^{\perp}\left(\mathfrak{J}_2^{(t)}+\mathfrak{J}_3^{(t)} \right) \mathfrak{P}_1^{-2}\right) \\
& +2 \operatorname{tr}\left(\mathfrak{P}_1^{-1}\left(\mathfrak{J}_1^{(t)}+\mathfrak{J}_3^{(t)} \right) \mathfrak{P}_1^{\perp}\left(\mathfrak{J}_2^{(t)}+\mathfrak{J}_3^{(t)} \right) \mathfrak{P}_1^{-1}\left(\mathfrak{J}_1^{(t)}+\mathfrak{J}_2^{(t)}+\mathfrak{J}_3^{(t)}+\mathfrak{J}_4^{(t)}\right) \mathfrak{P}_1^{-1}\right)\\
= & -2 \operatorname{tr}\left(\mathfrak{P}_1^{-1}\left(\mathfrak{J}_1^{(t)}+\mathfrak{J}_3^{(t)} \right) \mathfrak{P}_1^{\perp} \mathfrak{J}_3^{(t)} \mathfrak{P}_1^{\perp}\left(\mathfrak{J}_2^{(t)}+\mathfrak{J}_3^{(t)} \right) \mathfrak{P}_1^{-2}\right) \\
& +2 \operatorname{tr}\left(\mathfrak{P}_1^{-1} \mathfrak{J}_1^{(t)} \mathfrak{P}_1^{\perp} \mathfrak{J}_2^{(t)} \mathfrak{P}_1^{-1}\left(\mathfrak{J}_1^{(t)}+\mathfrak{J}_2^{(t)}\right) \mathfrak{P}_1^{-1}\right)\\
& +2 \operatorname{tr}\left(\mathfrak{P}_1^{-1} \mathfrak{J}_1^{(t)} \mathfrak{P}_1^{\perp} \mathfrak{J}_2^{(t)} \mathfrak{P}_1^{-1}\left(\mathfrak{J}_3^{(t)}+\mathfrak{J}_4^{(t)}\right) \mathfrak{P}_1^{-1}\right) \\
& +2 \operatorname{tr}\left(\mathfrak{P}_1^{-1} \mathfrak{J}_3^{(t)} \mathfrak{P}_1^{\perp}\left(\mathfrak{J}_2^{(t)}+\mathfrak{J}_3^{(t)} \right) \mathfrak{P}_1^{-1}\left(\mathfrak{J}_1^{(t)}+\mathfrak{J}_2^{(t)}+\mathfrak{J}_3^{(t)}+\mathfrak{J}_4^{(t)}\right) \mathfrak{P}_1^{-1}\right) \\
& +2 \operatorname{tr}\left(\mathfrak{P}_1^{-1} \mathfrak{J}_1^{(t)} \mathfrak{P}_1^{\perp} \mathfrak{J}_3^{(t)} \mathfrak{P}_1^{-1}\left(\mathfrak{J}_1^{(t)}+\mathfrak{J}_2^{(t)}+\mathfrak{J}_3^{(t)}+\mathfrak{J}_4^{(t)}\right) \mathfrak{P}_1^{-1}\right).
\end{aligned}
\end{align}
The second term on the right-hand side of the above equation dominates as the leading term. Let $\mathfrak{M}$ represent the negligible terms:
\begin{align*}
\begin{aligned}
\mathfrak{M}= & \left\langle\mathcal{S}_{\mathbf{G}_1, 3}\left(\mathfrak{E}_1\right), \mathbf{U}_1^\star \mathbf{U}_1^{\star\top}\right\rangle - 2 \operatorname{tr}\left(\mathfrak{P}_1^{-1} \mathfrak{J}_1^{(t)} \mathfrak{P}_1^{\perp} \mathfrak{J}_2^{(t)} \mathfrak{P}_1^{-1}\left(\mathfrak{J}_1^{(t)}+\mathfrak{J}_2^{(t)}\right) \mathfrak{P}_1^{-1}\right) \\
= & -2 \operatorname{tr}\left(\mathfrak{P}_1^{-1}\left(\mathfrak{J}_1^{(t)}+\mathfrak{J}_3^{(t)} \right) \mathfrak{P}_1^{\perp} \mathfrak{J}_3^{(t)} \mathfrak{P}_1^{\perp}\left(\mathfrak{J}_2^{(t)}+\mathfrak{J}_3^{(t)} \right) \mathfrak{P}_1^{-2}\right)\\
& +2 \operatorname{tr}\left(\mathfrak{P}_1^{-1} \mathfrak{J}_1^{(t)} \mathfrak{P}_1^{\perp} \mathfrak{J}_2^{(t)} \mathfrak{P}_1^{-1}\left(\mathfrak{J}_3^{(t)}+\mathfrak{J}_4^{(t)}\right) \mathfrak{P}_1^{-1}\right) \\
& +2 \operatorname{tr}\left(\mathfrak{P}_1^{-1} \mathfrak{J}_3^{(t)} \mathfrak{P}_1^{\perp}\left(\mathfrak{J}_2^{(t)}+\mathfrak{J}_3^{(t)} \right) \mathfrak{P}_1^{-1}\left(\mathfrak{J}_1^{(t)}+\mathfrak{J}_2^{(t)}+\mathfrak{J}_3^{(t)}+\mathfrak{J}_4^{(t)}\right) \mathfrak{P}_1^{-1}\right) \\
& +2 \operatorname{tr}\left(\mathfrak{P}_1^{-1} \mathfrak{J}_1^{(t)} \mathfrak{P}_1^{\perp} \mathfrak{J}_3^{(t)} \mathfrak{P}_1^{-1}\left(\mathfrak{J}_1^{(t)}+\mathfrak{J}_2^{(t)}+\mathfrak{J}_3^{(t)}+\mathfrak{J}_4^{(t)}\right) \mathfrak{P}_1^{-1}\right).
\end{aligned}
\end{align*}
By Lemma \ref{lem:bound for first round}, we have,
\begin{align}\label{B.6}
\mathfrak{M} =\Op{r_1 \frac{\kappa_0^2 \sigma^2 p/\left(t-1\right) \cdot\left(\kappa_0 \sigma \lambda_{\min } \sqrt{p/t} \right)\cdot\left(\kappa_0 \sigma \lambda_{\min }\sqrt{p/t}\right)}{\lambda_{\min}^6}} = \Op{r_1  \kappa_0^4 \frac{\sigma^4 p^2}{\lambda_{\min}^4 t^2}}.
\end{align}
Therefore, by Equation (\ref{B.5}) and (\ref{B.6}), we conclude that,   
\begin{align}\label{B.7}
\begin{aligned}
	\left\langle\mathcal{S}_{\mathbf{G}_1, 3}\left(\mathfrak{E}_1\right), \mathbf{U}_1^\star \mathbf{U}_1^{\star\top}\right\rangle =  &2 \operatorname{tr}\left(\mathfrak{P}_1^{-1} \mathfrak{J}_1^{(t)} \mathfrak{P}_1^{\perp} \mathfrak{J}_2^{(t)} \mathfrak{P}_1^{-1}\left(\mathfrak{J}_1^{(t)}+\mathfrak{J}_2^{(t)}\right) \mathfrak{P}_1^{-1}\right) \\
 &+\Op{r_1 \kappa_0^4 \frac{\sigma^4p^2}{\lambda_{\min }^4t^2}} .
\end{aligned}
\end{align}

We begin with considering $\operatorname{tr}\left(\mathfrak{P}_1^{-1} \mathfrak{J}_1^{(t)} \mathfrak{P}_1^{\perp} \mathfrak{J}_2^{(t)} \mathfrak{P}_1^{-1}\left(\mathfrak{J}_1^{(t)}+\mathfrak{J}_2^{(t)}\right) \mathfrak{P}_1^{-1}\right)$. Clearly,
\begin{align}\label{B.7-1}
\begin{aligned}
\left|\operatorname{tr}\left(\mathfrak{P}_1^{-1} \mathfrak{J}_1^{(t)} \mathfrak{P}_1^{\perp} \mathfrak{J}_2^{(t)} \mathfrak{P}_1^{-1}\left(\mathfrak{J}_1^{(t)}+\mathfrak{J}_2^{(t)}\right) \mathfrak{P}_1^{-1}\right)\right| 
&\leq  \left|\operatorname{tr}\left(\mathfrak{P}_1^{-1} \mathfrak{J}_1^{(t)} \mathfrak{P}_1^{\perp} \mathfrak{J}_2^{(t)} \mathfrak{P}_1^{-1} \mathfrak{J}_1^{(t)} \mathfrak{P}_1^{-1}\right)\right|\\
 &+\left|\operatorname{tr}\left(\mathfrak{P}_1^{-1} \mathfrak{J}_1^{(t)} \mathfrak{P}_1^{\perp} \mathfrak{J}_2^{(t)} \mathfrak{P}_1^{-1} \mathfrak{J}_2^{(t)} \mathfrak{P}_1^{-1}\right)\right| .
\end{aligned}
\end{align}
It suffices to bound $\left|\operatorname{tr}\left(\mathfrak{P}_1^{-1} \mathfrak{J}_1^{(t)} \mathfrak{P}_1^{\perp} \mathfrak{J}_2^{(t)} \mathfrak{P}_1^{-1} \mathfrak{J}_1^{(t)} \mathfrak{P}_1^{-1}\right)\right|$ and $\left|\operatorname{tr}\left(\mathfrak{P}_1^{-1} \mathfrak{J}_1^{(t)} \mathfrak{P}_1^{\perp} \mathfrak{J}_2^{(t)} \mathfrak{P}_1^{-1} \mathfrak{J}_2^{(t)} \mathfrak{P}_1^{-1}\right)\right|$, respectively. 

\begin{lemma}\label{lem:lemma 2}
Under the assumption of Theorem \ref{thm:U normality}, we have:
\begin{align*}
	\left|\operatorname{tr}\left(\mathfrak{P}_1^{-1} \mathfrak{J}_1^{(t)} \mathfrak{P}_1^{\perp} \mathfrak{J}_2^{(t)} \mathfrak{P}_1^{-1} \mathfrak{J}_1^{(t)} \mathfrak{P}_1^{-1}\right)\right|=\Op{\frac{ \sigma^3r^2 p}{t^{3/2}\lambda_{\min}^{3}} 
	+ \frac{r_1\kappa_0^3 \sigma^4p^2}{t^{2}\lambda_{\min}^{4}}}\\
		\left|\operatorname{tr}\left(\mathfrak{P}_1^{-1} \mathfrak{J}_1^{(t)} \mathfrak{P}_1^{\perp} \mathfrak{J}_2^{(t)} \mathfrak{P}_1^{-1} \mathfrak{J}_2^{(t)} \mathfrak{P}_1^{-1}\right)\right|=\Op{\frac{ \sigma^3r^2 p}{t^{3/2}\lambda_{\min}^{3}} 
	+ \frac{r_1\kappa_0^3 \sigma^4p^2}{t^{2}\lambda_{\min}^{4}}}.
\end{align*}
\end{lemma}
Proof in Section \ref{sec:Proof of Lemma 2}. Utilizing Lemma \ref{lem:lemma 2} and referencing Equations (\ref{B.7}) and (\ref{B.7-1}), we deduce that:
  \begin{align}
\begin{aligned}
	\left|\left\langle\mathcal{S}_{\mathbf{G}_1, 3}\left(\mathfrak{E}_1^{(t)} \right), \mathbf{U}_1^\star \mathbf{U}_1^{\star\top}\right\rangle \right|=\Op{ \frac{ \sigma^3r^2 p}{t^{3/2}\lambda_{\min}^{3}}
	+ \frac{r_1\kappa_0^4 \sigma^4p^2}{t^2\lambda_{\min}^{4}}}
	.
\end{aligned}
\end{align}
Thus, we conclude the proof of Lemma \ref{lem:bound of S2}.

\subsection{Proof of Lemma \ref{lem:lemma 3}}\label{sec:Proof of Lemma 3}

\begin{proof}	
Similar to Equations (\ref{B.8-1}) and (\ref{B.9}), we first have:
\begin{align*}
    &\left\|\mathfrak{J}_3^{(t)}-\mathbf{Z}_1^{(t)}\left(\mathcal{P}_{\mathbf{U}_2^\star} \otimes \mathcal{P}_{\mathbf{U}_3^\star}\right) \mathbf{Z}_1^{(t)\top}\right\|\\
=&\left\|\mathbf{Z}_1^{(t)}\left(\mathcal{P}_{\widehat{\mathbf{U}}_2^{(t-1)}} \otimes \mathcal{P}_{\widehat{\mathbf{U}}_3^{(t-1)}}\right) \mathbf{Z}_1^{(t)\top}-\mathbf{Z}_1^{(t)}\left(\mathcal{P}_{\mathbf{U}_2^\star} \otimes \mathcal{P}_{\mathbf{U}_3^\star}\right) \mathbf{Z}_1^{(t)\top}\right\| \\
 \leq&\left\|\left[\mathbf{Z}_1^{(t)}\left(\widehat{\mathbf{U}}_2^{(t-1)} \otimes \widehat{\mathbf{U}}_3^{(t-1)}\right)\right]\left[\mathbf{Z}_1^{(t)}\left(\widehat{\mathbf{U}}_2^{(t-1)} \otimes \widehat{\mathbf{U}}_3^{(t-1)}\right)\right]^{\top}\right.
\\&\left.-\left[\mathbf{Z}_1^{(t)}\left(\left(\mathbf{U}_2^\star \mathbf{R}_2^{(t-1)}\right) \otimes\left(\mathbf{U}_3^\star \mathbf{R}_3^{(t-1)}\right)\right)\right]\left[\mathbf{Z}_1^{(t)}\left(\left(\mathbf{U}_2^\star \mathbf{R}_2^{(t-1)}\right) \otimes\left(\mathbf{U}_3^\star \mathbf{R}_3^{(t-1)}\right)\right)\right]^{\top}\right\| \\
 \leq&\left\|\mathbf{Z}_1^{(t)}\left[\left(\widehat{\mathbf{U}}_2^{(t-1)} \otimes \widehat{\mathbf{U}}_3^{(t-1)}\right)-\left(\left(\mathbf{U}_2^\star \mathbf{R}_2^{(t-1)}\right) \otimes\left(\mathbf{U}_3^\star \mathbf{R}_3^{(t-1)}\right)\right)\right]\right\|\left\|\mathbf{Z}_1^{(t)}\left(\widehat{\mathbf{U}}_2^{(t-1)} \otimes \widehat{\mathbf{U}}_3^{(t-1)}\right)\right\| \\
&+\left\|\mathbf{Z}_1^{(t)}\left(\left(\mathbf{U}_2^\star \mathbf{R}_2^{(t-1)}\right) \otimes\left(\mathbf{U}_3^\star \mathbf{R}_3^{(t-1)}\right)\right)\right\|\cdot\\
&\quad\left\|\mathbf{Z}_1^{(t)}\left[\left(\widehat{\mathbf{U}}_2^{(t-1)} \otimes \widehat{\mathbf{U}}_3^{(t-1)}\right)-\left(\left(\mathbf{U}_2^\star \mathbf{R}_2^{(t-1)}\right) \otimes\left(\mathbf{U}_3^\star \mathbf{R}_3^{(t-1)}\right)\right)\right]\right\|. 
\end{align*}
Similar to Lemma \ref{lem:compare z1 and z2}, we conclude:
\begin{align}\label{B.15}
\begin{aligned}
&\left\|\mathfrak{J}_3^{(t)}-\mathbf{Z}_1^{(t)}\left(\mathcal{P}_{\mathbf{U}_2^\star} \otimes \mathcal{P}_{\mathbf{U}_3^\star}\right) \mathbf{Z}_1^{(t)\top}\right\|\\
=  &\Op{ \sigma\sqrt{p/t}\left\|\mathbf{Z}_1^{(t)}\left[\left(\widehat{\mathbf{U}}_2^{(t-1)} \otimes \widehat{\mathbf{U}}_3^{(t-1)}\right)-\left(\left(\mathbf{U}_2^\star \mathbf{R}_2^{(t-1)}\right) \otimes\left(\mathbf{U}_3^\star \mathbf{R}_3^{(t-1)}\right)\right)\right]\right\|} \\
= & \Op{\sigma\sqrt{p/t}\left(\left\|\mathbf{Z}_1^{(t)}\left(\left(\widehat{\mathbf{U}}_2^{(t-1)}-\mathbf{U}_2^\star \mathbf{R}_2^{(t-1)}\right) \otimes \widehat{\mathbf{U}}_3^{(t-1)}\right)\right\| + \left\|\mathbf{Z}_1^{(t)}\left(\left(\mathbf{U}_2^\star \mathbf{R}_2^{(t-1)}\right) \otimes\left(\widehat{\mathbf{U}}_3^{(t-1)}-\mathbf{U}_3^\star \mathbf{R}_3^{(t-1)}\right)\right)\right\|\right)} \\
= & \Op{\sigma\sqrt{p/t} \cdot  \sigma\sqrt{p/t}\left(\left\|\widehat{\mathbf{U}}_2^{(t-1)}-\mathbf{U}_2^\star \mathbf{R}_2^{(t-1)}\right\|+\left\|\widehat{\mathbf{U}}_3^{(t-1)}-\mathbf{U}_3^\star \mathbf{R}_3^{(t-1)}\right\|\right)} \\
= & \Op{ \frac{\sigma^3p^{3 / 2}}{t(t-1)^{1/2}\lambda_{\min }}} 
= \Op{\frac{\sigma^3p^{3 / 2}}{t^{3/2}\lambda_{\min }}}.
\end{aligned}
\end{align}
Because $\mathfrak{J}_1^{(t)}$ equals $\left(\mathfrak{J}_2^{(t)}\right)^\top$, we can find that
\begin{align*}
    & |\mathrm{II}|=|\mathrm{III}| \\
= & \left|\operatorname{tr}\left(\mathbf{U}_1^\star \mathbf{\Lambda}_1^{-2} \mathbf{U}_1^{\star\top} \mathfrak{J}_1^{(t)} \mathbf{U}_{1 \perp}^\star \mathbf{U}_{1 \perp}^{\star\top} \mathfrak{J}_3^{(t)} \mathbf{U}_1^\star \mathbf{\Lambda}_1^{-2} \mathbf{U}_1^{\star\top}\right)\right| \\
\leq & \left|\operatorname{tr}\left(\mathbf{U}_1^\star \mathbf{\Lambda}_1^{-2} \mathbf{U}_1^{\star\top} \mathbf{T}_1^\star\left(\mathcal{P}_{\mathbf{U}_2^\star} \otimes \mathcal{P}_{\mathbf{U}_3^\star}\right) \mathbf{Z}_1^{(t)\top} \mathbf{U}_{1 \perp}^\star \mathbf{U}_{1 \perp}^{\star\top} \mathbf{Z}_1^{(t)}\left(\mathcal{P}_{\mathbf{U}_2^\star} \otimes \mathcal{P}_{\mathbf{U}_3^\star}\right) \mathbf{Z}_1^{(t)\top} \mathbf{U}_1^\star \mathbf{\Lambda}_1^{-2} \mathbf{U}_1^{\star\top}\right)\right| \\
& +\left|\operatorname{tr}\left(\mathbf{U}_1^\star \mathbf{\Lambda}_1^{-2} \mathbf{U}_1^{\star\top}\left(\mathfrak{J}_1^{(t)}-\mathbf{T}_1^\star\left(\mathcal{P}_{\mathbf{U}_2^\star} \otimes \mathcal{P}_{\mathbf{U}_3^\star}\right) \mathbf{Z}_1^{(t)\top}\right) \mathbf{U}_{1 \perp}^\star \mathbf{U}_{1 \perp}^{\star\top} \mathbf{Z}_1^{(t)}\left(\mathcal{P}_{\mathbf{U}_2^\star} \otimes \mathcal{P}_{\mathbf{U}_3^\star}\right) \mathbf{Z}_1^{(t)\top} \mathbf{U}_1^\star \mathbf{\Lambda}_1^{-2} \mathbf{U}_1^{\star\top}\right)\right| \\
& +\left|\operatorname{tr}\left(\mathbf{U}_1^\star \mathbf{\Lambda}_1^{-2} \mathbf{U}_1^{\star\top} \mathfrak{J}_1^{(t)} \mathbf{U}_{1 \perp}^\star \mathbf{U}_{1 \perp}^{\star\top}\left(\mathfrak{J}_3^{(t)}-\mathbf{Z}_1^{(t)}\left(\mathcal{P}_{\mathbf{U}_2^\star} \otimes \mathcal{P}_{\mathbf{U}_3^\star}\right) \mathbf{Z}_1^{(t)\top}\right) \mathbf{U}_1^\star \mathbf{\Lambda}_1^{-2} \mathbf{U}_1^{\star\top}\right)\right|\\
 \leq&\left|\operatorname{tr}\left(\mathbf{U}_1^\star \mathbf{\Lambda}_1^{-2} \mathbf{U}_1^{\star\top} \mathbf{T}_1^\star\left(\mathcal{P}_{\mathbf{U}_2^\star} \otimes \mathcal{P}_{\mathbf{U}_3^\star}\right) \mathbf{Z}_1^{(t)\top} \mathbf{U}_{1 \perp}^\star \mathbf{U}_{1 \perp}^{\star\top} \mathbf{Z}_1^{(t)}\left(\mathcal{P}_{\mathbf{U}_2^\star} \otimes \mathcal{P}_{\mathbf{U}_3^\star}\right) \mathbf{Z}_1^{(t)\top} \mathbf{U}_1^\star \mathbf{\Lambda}_1^{-2} \mathbf{U}_1^{\star\top}\right)\right| \\
 & +r \frac{\left\|\mathfrak{J}_1^{(t)}-\mathbf{T}_1^\star\left(\mathcal{P}_{\mathbf{U}_2^\star} \otimes \mathcal{P}_{\mathbf{U}_3^\star}\right) \mathbf{Z}_1^{(t)\top}\right\|\left\|\mathbf{Z}_1^{(t)}\left(\mathcal{P}_{\mathbf{U}_2^\star} \otimes \mathcal{P}_{\mathbf{U}_3^\star}\right) \mathbf{Z}_1^{(t)\top}\right\|}{\lambda_{\min }^4} \\
& +r \frac{\left\|\mathbf{T}_1^\star\left(\mathcal{P}_{\widehat{\mathbf{U}}_2} \otimes \mathcal{P}_{\widehat{\mathbf{U}}_3}\right) \mathbf{Z}_1^{(t)\top}\right\|\left\|\mathfrak{J}_3^{(t)}-\mathbf{Z}_1^{(t)}\left(\mathcal{P}_{\mathbf{U}_2^\star} \otimes \mathcal{P}_{\mathbf{U}_3^\star}\right) \mathbf{Z}_1^{(t)\top}\right\|}{\lambda_{\min }^4}.
\end{align*}
 For the sake of clarity, we introduce the following notations:
 \begin{align}\label{equ:def of W}
\begin{aligned}
		&\mathbf{W}_1=\mathbf{U}_1^{\star\top} \mathbf{Z}_1^{(t)}\left(\mathbf{U}_2^\star \otimes \mathbf{U}_3^\star\right) \in \mathbb{R}^{r_1 \times\left(r_2 r_3\right)}, \\
	&\mathbf{W}_2= \mathbf{U}_{1 \perp}^{\star\top} \mathbf{Z}_1^{(t)}\left(\mathbf{U}_2^\star \otimes \mathbf{U}_3^\star\right) \in \mathbb{R}^{\left(p_1-r_1\right) \times\left(r_2 r_3\right)}.
\end{aligned}
\end{align}
By merging Equations (\ref{B.9}) and (\ref{B.15}), we can conclude:
\begin{align*}
\begin{aligned}
& |\mathrm{II}|=|\mathrm{III}| \\
 = &\left|\operatorname{tr}\left(\mathbf{U}_1^\star \mathbf{\Lambda}_1^{-2} \mathbf{U}_1^{\star\top} \mathbf{T}_1^\star\left(\mathcal{P}_{\mathbf{U}_2^\star} \otimes \mathcal{P}_{\mathbf{U}_3^\star}\right) \mathbf{Z}_1^{(t)\top} \mathbf{U}_{1 \perp}^\star \mathbf{U}_{1 \perp}^{\star\top} \mathbf{Z}_1^{(t)}\left(\mathcal{P}_{\mathbf{U}_2^\star} \otimes \mathcal{P}_{\mathbf{U}_3^\star}\right) \mathbf{Z}_1^{(t)\top} \mathbf{U}_1^\star \mathbf{\Lambda}_1^{-2} \mathbf{U}_1^{\star\top}\right)\right|\\
&+ \Op{\kappa_0 r \frac{p^2\sigma^4}{t^{2}\lambda_{\min }^4}} \\
 = &\left|\operatorname{tr}\left(\mathbf{U}_1^\star \mathbf{\Lambda}_1^{-2} \mathbf{G}_1^\star \mathbf{W}_2^{\top} \mathbf{W}_2 \mathbf{W}_1^{\top} \mathbf{\Lambda}_1^{-2} \mathbf{U}_1^{\star\top}\right)\right| + 
 \Op{ \kappa_0 r \frac{p^2\sigma^4}{t^{2}\lambda_{\min }^4}}\\
= &\left|\operatorname{tr}\left(\mathbf{\Lambda}_1^{-4} \mathbf{G}_1^\star \mathbf{W}_2^{\top} \mathbf{W}_2 \mathbf{W}_1^{\top}\right)\right| +
\Op{\kappa_0 r \frac{p^2\sigma^4}{t^{2}\lambda_{\min }^4}},
\end{aligned}
\end{align*}
where $\mathbf{W}_1$ and $\mathbf{W}_2$ are defined in Equation (\ref{equ:def of W}).
From Equations (\ref{B.2}), (\ref{equ:W1}) and (\ref{equ:W2}), we deduce, we have:
\begin{align*}
	\left|\operatorname{tr}\left(\mathbf{\Lambda}_1^{-4} \mathbf{G}_1^\star \mathbf{W}_2^{\top} \mathbf{W}_2 \mathbf{W}_1^{\top}\right)\right| = \Op{\frac{r^2p\sigma^3}{t^{3/2}\lambda_{\min}^{3}}}.
	\end{align*}
Hence, we have
\begin{align*}
	|\mathrm{II}|=|\mathrm{III}| =  \Op{\frac{r^2p\sigma^3}{t^{3/2}\lambda_{\min}^{3}}
	+ \frac{r \kappa_0p^2\sigma^4}{t^{2}\lambda_{\min }^4}}.
\end{align*}
For $\mathrm{I}=\operatorname{tr}\left(\mathfrak{P}_1^{-1} \mathfrak{J}_1^{(t)} \mathfrak{P}_1^{\perp} \mathfrak{J}_2^{(t)} \mathfrak{P}_1^{-1}\right)$, we have
\begin{align}\label{equ:I-1}
\begin{aligned}
&\left|\mathrm{I}-\operatorname{tr}\left(\mathbf{\Lambda}_1^{-4} \mathbf{G}_1^\star \left(\mathbf{U}_2^{\star\top} \otimes \mathbf{U}_3^{\star\top}\right) \mathbf{Z}_1^{(t)\top} \mathbf{U}_{1 \perp}^\star \mathbf{U}_{1 \perp}^{\star\top} \mathbf{Z}_1^{(t)}\left(\mathbf{U}_2^\star \otimes \mathbf{U}_3^\star\right) \mathbf{G}_1^{\star\top}\right)\right|\\
=&\left|\operatorname{tr}\left(\mathbf{U}_1^\star \mathbf{\Lambda}_1^{-2} \mathbf{U}_1^{\star\top} \mathfrak{J}_1^{(t)} \mathbf{U}_{1 \perp}^\star \mathbf{U}_{1 \perp}^{\star\top} \mathfrak{J}_1^{\top} \mathbf{U}_1^\star \mathbf{\Lambda}_1^{-2} \mathbf{U}_1^{\star\top}\right)-\right.\\
&\left.\quad\operatorname{tr}\left(\mathbf{\Lambda}_1^{-4} \mathbf{G}_1^\star \left(\mathbf{U}_2^{\star\top} \otimes \mathbf{U}_3^{\star\top}\right) \mathbf{Z}_1^{(t)\top} \mathbf{U}_{1 \perp}^\star \mathbf{U}_{1 \perp}^{\star\top} \mathbf{Z}_1^{(t)}\left(\mathbf{U}_2^\star \otimes \mathbf{U}_3^\star\right) \mathbf{G}_1^{\star\top}\right)\right| \\
=&\left|\operatorname{tr}\left(\mathbf{U}_1^\star \mathbf{\Lambda}_1^{-2} \mathbf{U}_1^{\star\top} \mathbf{T}_1^{\star}\left(\mathcal{P}_{\widehat{\mathbf{U}}_2^{(t-1)}} \otimes \mathcal{P}_{\widehat{\mathbf{U}}_3^{(t-1)}}\right) \mathbf{Z}_1^{(t) \top} \mathbf{U}_{1 \perp}^\star \mathbf{U}_{1 \perp}^{\star\top} 
 \mathbf{Z}_1^{(t)}\left(\mathcal{P}_{\widehat{\mathbf{U}}_2^{(t-1)}} \otimes \mathcal{P}_{\widehat{\mathbf{U}}_3^{(t-1)}}\right)\mathbf{T}_1^{\star\top}
\mathbf{U}_1^\star \mathbf{\Lambda}_1^{-2} \mathbf{U}_1^{\star\top}\right)-\right.\\
&\left.\quad\operatorname{tr}\left(\mathbf{\Lambda}_1^{-4} \mathbf{G}_1^\star \left(\mathbf{U}_2^{\star\top} \otimes \mathbf{U}_3^{\star\top}\right) \mathbf{Z}_1^{(t)\top} \mathbf{U}_{1 \perp}^\star \mathbf{U}_{1 \perp}^{\star\top} \mathbf{Z}_1^{(t)}\left(\mathbf{U}_2^\star \otimes \mathbf{U}_3^\star\right) \mathbf{G}_1^{\star\top}\right)\right| \\
 =&\left| \operatorname{tr}\left(\mathbf{\Lambda}_1^{-4} \mathbf{G}_1^\star\left(\left(\mathbf{U}_2^{\star\top} \mathcal{P}_{\widehat{\mathbf{U}}_2^{(t-1)}}\right) \otimes\left(\mathbf{U}_3^{\star\top} \mathcal{P}_{\widehat{\mathbf{U}}_3^{(t-1)}}\right)\right) \mathbf{Z}_1^{(t)\top} \mathbf{U}_{1 \perp}^\star\mathbf{U}_{1 \perp}^{\star\top} \mathbf{Z}_1^{(t)}\left(\left(\mathcal{P}_{\widehat{\mathbf{U}}_2^{(t-1)}} \mathbf{U}_2^\star\right) \otimes\left(\mathcal{P}_{\widehat{\mathbf{U}}_3^{(t-1)}} \mathbf{U}_3^\star\right)\right) \mathbf{G}_1^{\star\top}\right)\right.\\
&-\left.\operatorname{tr}\left(\mathbf{\Lambda}_1^{-4} \mathbf{G}_1^\star \left(\mathbf{U}_2^{\star\top} \otimes \mathbf{U}_3^{\star\top}\right) \mathbf{Z}_1^{(t)\top} \mathbf{U}_{1 \perp}^\star \mathbf{U}_{1 \perp}^{\star\top} \mathbf{Z}_1^{(t)}\left(\mathbf{U}_2^\star \otimes \mathbf{U}_3^\star\right) \mathbf{G}_1^{\star\top}\right) \right|.
\end{aligned}
\end{align}
In dealing with the term $\mathbf{U}_2^{\star\top} \mathcal{P}_{\widehat{\mathbf{U}}_2^{(t-1)}}$ within Equation (\ref{equ:I-1}), we exploit the property of this term to provide a more refined approximation. By Algorithm \ref{alg:Single-step Tensor Linear Form Estimator Update}, $ \widehat{\mathbf{U}}_2^{(t-1)} \widehat{\mathbf{U}}_2^{(t-1) \top}$ is the spectral projector for the top- $r_2$ eigenvectors of
\begin{align}\label{equ:decomp of U2}
\begin{aligned}
& \mathcal{M}_2(\widehat{\mathcal{T}}^{(t-1)})\left(\mathcal{P}_{\widehat{\mathbf{U}}_1^{(t-2)}} \otimes \mathcal{P}_{\widehat{\mathbf{U}}_3^{(t-2)}}\right) \mathcal{M}_2(\widehat{\mathcal{T}})^{(t-1)\top}\\
= &  \mathbf{U}_2^\star \mathbf{G}_2^\star \mathbf{G}_2^{\star\top} \mathbf{U}_2^{\star\top}
-\mathbf{U}_2^\star \mathbf{G}_2^\star\left(\mathbf{U}_1^{\star\top} \mathcal{P}_{\widehat{\mathbf{U}}_1^{(t-2)}}^{\perp} \mathbf{U}_1^\star \otimes \mathbf{U}_3^{\star\top} \mathcal{P}_{\widehat{\mathbf{U}}_3^{(t-2)}} \mathbf{U}_3^\star\right) \mathbf{G}_2^{\star\top} \mathbf{U}_2^{\star\top} \\
 & - \mathbf{U}_2^\star \mathbf{G}_2^\star\left(\mathbf{I}_{r_1} \otimes \mathbf{U}_3^{\star\top} \mathcal{P}_{\widehat{\mathbf{U}}_3^{(t-2)}}^{\perp} \mathbf{U}_3^\star\right) \mathbf{G}_2^{\star\top} \mathbf{U}_2^{\star\top}+\mathbf{T}_2^\star\left(\mathcal{P}_{\widehat{\mathbf{U}}_1^{(t-2)}} \otimes \mathcal{P}_{\widehat{\mathbf{U}}_3^{(t-2)}}\right) \mathbf{Z}_2^{(t-1)\top} \\
 & + \mathbf{Z}_2^{(t-1)}\left(\mathcal{P}_{\widehat{\mathbf{U}}_1^{(t-2)}} \otimes \mathcal{P}_{\widehat{\mathbf{U}}_3^{(t-2)}}\right) \mathbf{T}_2^{\star\top}+\mathbf{Z}_2^{(t-1)}\left(\mathcal{P}_{\widehat{\mathbf{U}}_1^{(t-2)}} \otimes \mathcal{P}_{\widehat{\mathbf{U}}_3^{(t-2)}}\right) \mathbf{Z}_2^{(t-1)\top} \\
= &  \mathbf{U}_2^\star \mathbf{G}_2^\star \mathbf{G}_2^{\star\top} \mathbf{U}_2^{\star\top}+{\mathfrak{E}}_2^{(t-1)} .
\end{aligned}
\end{align}
Similarly, we can define ${\mathfrak{E}}_3^{(t-1)}$.
Recall that $\mathfrak{P}_k^n=\mathbf{U}_k^\star \mathbf{\Lambda}_k^{-2 n} \mathbf{U}_k^{\star\top}$ for positive integer $n$, and $\mathfrak{P}_k^0:=\mathfrak{P}_k^{\perp}:=\mathcal{P}_{\mathbf{U}_k^\star}^{\perp}$ for $k\in[3]$. By Lemma \ref{lem:bound for first round} and Lemma \ref{lem:theorem 1 in xia2021normal}, 
\begin{align*}
\left\|{\mathfrak{E}}_k^{(t-1)}\right\| =\Op{\kappa_0  \lambda_{\min}\sigma\sqrt{\frac{p}{t-1}}},
\end{align*}
and
\begin{align*}
\widehat{\mathbf{U}}_k^{(t-1)} \widehat{\mathbf{U}}_k^{(t-1) \top}-\mathbf{U}_k^\star \mathbf{U}_k^{\star\top}=\sum_{n \geq 1} \mathcal{S}_{\mathbf{G}_k, n}\left(\mathfrak{E}_k^{(t-1)}\right),
\end{align*}
where for positive integer $n$,
\begin{align*}
\mathcal{S}_{\mathbf{G}_k, n}\left(\mathfrak{E}_k^{(t-1)}\right)=\sum_{s_1+\cdots+s_{n+1}=n}(-1)^{1+\tau(\mathbf{s})} \cdot \mathfrak{P}_k^{-s_1} \mathfrak{E}_k^{(t-1)} \mathfrak{P}_k^{-s_2} \mathfrak{E}_k^{(t-1)} \mathfrak{P}_k^{-s_3} \cdots \mathfrak{P}_k^{-s_k} \mathfrak{E}_k^{(t-1)} \mathfrak{P}_k^{-s_{n+1}}.
\end{align*}
For $n \geq 2$, similarly to Equation (\ref{equ:bound of Sg gep 2}), we have
\begin{align*}
\left\|\mathcal{S}_{\mathbf{G}_k, n}\left(\mathfrak{E}_k^{(t-1)}\right)\right\| \leq\left(\frac{4\left\|\mathfrak{E}_k^{(t-1)}\right\|}{\lambda_{\min }^2}\right)^n.
\end{align*}
Then 
\begin{align*}
\left\|\sum_{n \geq 2} \mathcal{S}_{\mathbf{G}_n, n}\left(\mathfrak{E}_k^{(t-1)}\right)\right\| \leq \sum_{n \geq 2}\left(\frac{4\left\|\mathfrak{E}_k^{(t-1)}\right\|}{\lambda_{\min}^2}\right)^n =\Op{\frac{\kappa_0^2 \sigma^2 p}{(t-1)\lambda_{\min }^2}}.
\end{align*}
Note that since $\mathbf{U}_k^\star\mathfrak{P}_k^{\perp}= 0$, we have
\begin{align*}
\mathcal{P}_{\mathbf{U}_k^\star} \mathcal{S}_{\mathbf{G}_k, 1}\left(\mathfrak{E}_k^{(t-1)}\right)=\mathcal{P}_{\mathbf{U}_k^\star}\left(\mathfrak{P}_k^{-1} \mathfrak{E}_k^{(t-1)} \mathfrak{P}_k^{\perp}+\mathfrak{P}_k^{\perp} \mathfrak{E}_k^{(t-1)} \mathfrak{P}_k^{-1}\right)=\mathbf{U}_k^\star \mathbf{\Lambda}_k^{-2} \mathbf{U}_k^{\star\top} \mathfrak{E}_k^{(t-1)} \mathcal{P}_{\mathbf{U}_k^\star}^{\perp}.
\end{align*}
Building upon the preceding equation, for the term $\mathbf{U}_2^{\star\top} \mathcal{P}_{\widehat{\mathbf{U}}_2^{(t-1)}}$ in Equation (\ref{equ:I-1}), and for $k = 2,3$, we derive its first-order approximation:
\begin{align}\label{B.21}
\begin{aligned}
	&\left\|\mathbf{U}_k^{\star\top} \mathcal{P}_{\widehat{\mathbf{U}}_k^{(t)}}-\mathbf{U}_k^{\star\top}-\mathbf{\Lambda}_k^{-2} \mathbf{U}_k^{\star\top} \mathfrak{E}_k^{(t-1)} \mathcal{P}_{\mathbf{U}_k^\star}^{\perp}\right\|\\
 = &\left\|\mathcal{P}_{\mathbf{U}_k^\star} \mathcal{P}_{\widehat{\mathbf{U}}_k^{(t)}}-\mathcal{P}_{\mathbf{U}_k^\star}-\mathbf{U}_k^\star \mathbf{\Lambda}_k^{-2} \mathbf{U}_k^{\star\top} \mathfrak{E}_k^{(t-1)} \mathcal{P}_{\mathbf{U}_k^\star}^{\perp}\right\| =\Op{\frac{\kappa_0^2 \sigma^2 p}{(t-1)\lambda_{\min }^2}
 } = \Op{\frac{\kappa_0^2 \sigma^2 p}{t\lambda_{\min }^2}}.
\end{aligned}
\end{align}

By Equation (\ref{B.2}), (\ref{equ:W2}), and (\ref{B.21}), we have
\begin{align}\label{B.23}
\begin{aligned}
& \left|\operatorname{tr}\left(\mathbf{\Lambda}_1^{-4} \mathbf{G}_1^\star\left(\left(\mathbf{U}_2^{\star\top} \mathcal{P}_{\widehat{\mathbf{U}}_2^{(t-1)}}\right) \otimes\left(\mathbf{U}_3^{\star\top} \mathcal{P}_{\widehat{\mathbf{U}}_3^{(t-1)}}\right)\right) \mathbf{Z}_1^{(t)\top} \mathcal{P}_{\mathbf{U}_1^\star}^{\perp} \mathbf{Z}_1^{(t)}\left(\left(\mathcal{P}_{\widehat{\mathbf{U}}_2^{(t-1)}} \mathbf{U}_2^\star\right) \otimes\left(\mathcal{P}_{\widehat{\mathbf{U}}_3^{(t-1)}} \mathbf{U}_3^\star\right)\right) \mathbf{G}_1^{\star\top}\right)\right. \\
 &\left.-\operatorname{tr}\left(\mathbf{\Lambda}_1^{-4} \mathbf{G}_1^\star \left(\mathbf{U}_2^{\star\top} \otimes \mathbf{U}_3^{\star\top}\right) \mathbf{Z}_1^{(t)\top} \mathbf{U}_{1 \perp}^\star \mathbf{U}_{1 \perp}^{\star\top} \mathbf{Z}_1^{(t)}\left(\mathbf{U}_2^\star \otimes \mathbf{U}_3^\star\right) \mathbf{G}_1^{\star\top}\right) \right| \\
 \leq & 2\left|\operatorname{tr}\left(\mathbf{\Lambda}_1^{-4} \mathbf{G}_1^\star\left(\left(\mathbf{\Lambda}_2^{-2} \mathbf{U}_2^{\star\top} {\mathfrak{E}}_2^{(t-1)} \mathcal{P}_{\mathbf{U}_2^\star}^{\perp}\right) \otimes \mathbf{U}_3^{\star\top}\right) \mathbf{Z}_1^{(t)\top} \mathbf{U}_{1 \perp}^\star \mathbf{U}_{1 \perp}^{\star\top} \mathbf{Z}_1^{(t)}\left(\mathbf{U}_2^\star \otimes \mathbf{U}_3^\star\right) \mathbf{G}_1^{\star\top}\right)\right| \\
&+2\left|\operatorname{tr}\left(\mathbf{\Lambda}_1^{-4} \mathbf{G}_1^\star\left(\mathbf{U}_2^{\star\top} \otimes\left(\mathbf{\Lambda}_3^{-2} \mathbf{U}_3^{\star\top} {\mathfrak{E}}_3^{(t-1)} \mathcal{P}_{\mathbf{U}_3^\star}^{\perp}\right)\right) \mathbf{Z}_1^{(t)\top} \mathbf{U}_{1 \perp}^\star \mathbf{U}_{1 \perp}^{\star\top} \mathbf{Z}_1^{(t)}\left(\mathbf{U}_2^\star \otimes \mathbf{U}_3^\star\right) \mathbf{G}_1^{\star\top}\right)\right| \\
& + \Op{ r_1 \max _{k=2,3}\left(\left\|\mathbf{\Lambda}_k^{-2} \mathbf{U}_k^{\star\top} \mathfrak{E}_k^{(t-1)} \mathcal{P}_{\mathbf{U}_k^\star}^{\perp}\right\|\right)^2 \cdot\left(\sigma\sqrt{\frac{p}{t}}\right)^2 \cdot \lambda_{\min }^{-2}}
\\
&+\Op{  r_1 \frac{\kappa_0^2 \sigma^2 p}{t\lambda_{\min }^2} \cdot \left(\sigma\sqrt{\frac{p}{t}}\right)^2 \cdot \lambda_{\min }^{-2}} \\
& +\Op{ r_1 \frac{\kappa_0^2 \sigma^2 p}{t\lambda_{\min }^2} \cdot \max _{k=2,3}\left\|\mathbf{\Lambda}_k^{-2} \mathbf{U}_k^{\star\top} \mathfrak{E}_k^{(t-1)} \mathcal{P}_{\mathbf{U}_k^\star}^{\perp}\right\| \cdot\left(\sigma\sqrt{\frac{p}{t}}\right)^2 \cdot \lambda_{\min }^{-2}}
\\
&+ \Op{  r_1\left(\frac{\kappa_0^2 \sigma^2 p}{t\lambda_{\min }^2}\right)^2 \cdot\left(\sigma\sqrt{\frac{p}{t}}\right)^2\cdot \lambda_{\min }^{-2}} \\
 \leq & 2 \left|\operatorname{tr}\left(\mathbf{\Lambda}_1^{-4} \mathbf{G}_1^\star\left(\left(\mathbf{\Lambda}_2^{-2} \mathbf{U}_2^{\star\top} {\mathfrak{E}}_2^{(t-1)} \mathcal{P}_{\mathbf{U}_2^\star}^{\perp}\right) \otimes \mathbf{U}_3^{\star\top}\right) \mathbf{Z}_1^{(t)\top} \mathbf{U}_{1 \perp}^\star \mathbf{U}_{1 \perp}^{\star\top} \mathbf{Z}_1^{(t)}\left(\mathbf{U}_2^\star \otimes \mathbf{U}_3^\star\right) \mathbf{G}_1^{\star\top}\right)\right|\\
& + 2 \left|\operatorname{tr}\left(\mathbf{\Lambda}_1^{-4} \mathbf{G}_1^\star\left(\mathbf{U}_2^{\star\top} \otimes\left(\mathbf{\Lambda}_3^{-2} \mathbf{U}_3^{\star\top} {\mathfrak{E}}_3^{(t-1)} \mathcal{P}_{\mathbf{U}_3^\star}^{\perp}\right)\right) \mathbf{Z}_1^{(t)\top} \mathbf{U}_{1 \perp}^\star \mathbf{U}_{1 \perp}^{\star\top} \mathbf{Z}_1^{(t)}\left(\mathbf{U}_2^\star \otimes \mathbf{U}_3^\star\right) \mathbf{G}_1^{\star\top}\right)\right|\\
& + \Op{ \frac{ r_1 \kappa_0^2\sigma^4p^2}{t^2\lambda_{\min }^4}}.
\end{aligned}
\end{align}
For the first term on the right-hand of the above inequality, by the definition of ${\mathfrak{E}}_2^{(t-1)}$ and recall that $\mathbf{T}_2^{\star\top} \mathcal{P}_{\mathbf{U}_2^\star}^{\perp}=0$ and $\mathbf{U}_2^{\star\top} \mathcal{P}_{\mathbf{U}_2^\star}^{\perp}=0$ , we have
\begin{align*}
    {\mathfrak{E}}_2^{(t-1)} \mathcal{P}_{\mathbf{U}_2^\star}^{\perp} = & -\mathbf{U}_2^\star \mathbf{G}_2^\star\left(\mathbf{U}_1^{\star\top} \mathcal{P}_{\widehat{\mathbf{U}}_1^{(t-2)}}^{\perp} \mathbf{U}_1^\star \otimes \mathbf{U}_3^{\star\top} \mathcal{P}_{\widehat{\mathbf{U}}_3^{(t-2)}} \mathbf{U}_3^\star\right) \mathbf{G}_2^{\star\top} \mathbf{U}_2^{\star\top}\mathcal{P}_{\mathbf{U}_2^\star}^{\perp} \\
 & - \mathbf{U}_2^\star \mathbf{G}_2^\star\left(\mathbf{I}_{r_1} \otimes \mathbf{U}_3^{\star\top} \mathcal{P}_{\widehat{\mathbf{U}}_3^{(t-2)}}^{\perp} \mathbf{U}_3^\star\right) \mathbf{G}_2^{\star\top} \mathbf{U}_2^{\star\top}\mathcal{P}_{\mathbf{U}_2^\star}^{\perp}
 +\mathbf{T}_2^\star\left(\mathcal{P}_{\widehat{\mathbf{U}}_1^{(t-2)}} \otimes \mathcal{P}_{\widehat{\mathbf{U}}_3^{(t-2)}}\right) \mathbf{Z}_2^{(t-1)\top} \mathcal{P}_{\mathbf{U}_2^\star}^{\perp} \\
 & + \mathbf{Z}_2^{(t-1)}\left(\mathcal{P}_{\widehat{\mathbf{U}}_1^{(t-2)}} \otimes \mathcal{P}_{\widehat{\mathbf{U}}_3^{(t-2)}}\right) \mathbf{T}_2^{\star\top}\mathcal{P}_{\mathbf{U}_2^\star}^{\perp}
 +\mathbf{Z}_2^{(t-1)}\left(\mathcal{P}_{\widehat{\mathbf{U}}_1^{(t-2)}} \otimes \mathcal{P}_{\widehat{\mathbf{U}}_3^{(t-2)}}\right) \mathbf{Z}_2^{(t-1)\top} \mathcal{P}_{\mathbf{U}_2^\star}^{\perp}\\
 = & \mathbf{T}_2^\star\left(\mathcal{P}_{\widehat{\mathbf{U}}_1^{(t-2)}} \otimes \mathcal{P}_{\widehat{\mathbf{U}}_3^{(t-2)}}\right) \mathbf{Z}_2^{(t-1)\top} \mathcal{P}_{\mathbf{U}_2^\star}^{\perp} +\mathbf{Z}_2^{(t-1)}\left(\mathcal{P}_{\widehat{\mathbf{U}}_1^{(t-2)}} \otimes \mathcal{P}_{\widehat{\mathbf{U}}_3^{(t-2)}}\right) \mathbf{Z}_2^{(t-1)\top} \mathcal{P}_{\mathbf{U}_2^\star}^{\perp}.
\end{align*}
Equation (\ref{B.2}), (\ref{equ:W2}) and (\ref{equ:decomp of U2}) imply that
\begin{align*}
\begin{aligned}
	&\left|\operatorname{tr}\left(\mathbf{\Lambda}_1^{-4} \mathbf{G}_1^\star\left(\left(\mathbf{\Lambda}_2^{-2} \mathbf{U}_2^{\star\top} {\mathfrak{E}}_2^{(t-1)} \mathcal{P}_{\mathbf{U}_2^\star}^{\perp}\right) \otimes \mathbf{U}_3^{\star\top}\right) \mathbf{Z}_1^{(t)\top} \mathbf{U}_{1 \perp}^\star \mathbf{U}_{1 \perp}^{\star\top} \mathbf{Z}_1^{(t)}\left(\mathbf{U}_2^\star \otimes \mathbf{U}_3^\star\right) \mathbf{G}_1^{\star\top}\right)\right| \\ 
	 \leq & \left|\operatorname{tr}\left(\mathbf{\Lambda}_1^{-4} \mathbf{G}_1^\star\left(\left(\mathbf{\Lambda}_2^{-2} \mathbf{U}_2^{\star\top} \mathbf{T}_2^\star\left(\mathcal{P}_{\widehat{\mathbf{U}}_1^{(t-2)}} \otimes \mathcal{P}_{\widehat{\mathbf{U}}_3^{(t-2)}}\right) \mathbf{Z}_2^{(t-1)\top} \mathcal{P}_{\mathbf{U}_2^\star}^{\perp}\right) \otimes \mathbf{U}_3^{\star\top}\right)\cdot\right.\right.\\
  &\left.\left.\qquad\mathbf{Z}_1^{(t)\top} \mathbf{U}_{1 \perp}^\star \mathbf{U}_{1 \perp}^{\star\top} \mathbf{Z}_1^{(t)}\left(\mathbf{U}_2^\star \otimes \mathbf{U}_3^\star\right) \mathbf{G}_1^{\star\top}\right)\right| 
	 \\ & + \left|\operatorname{tr}\left(\mathbf{\Lambda}_1^{-4} \mathbf{G}_1^\star\left(\left(\mathbf{\Lambda}_2^{-2} \mathbf{U}_2^{\star\top} \mathbf{Z}_2^{(t-1)}\left(\mathcal{P}_{\widehat{\mathbf{U}}_1^{(t-2)}} \otimes \mathcal{P}_{\widehat{\mathbf{U}}_3^{(t-2)}}\right) \mathbf{Z}_2^{(t-1)\top} \mathcal{P}_{\mathbf{U}_2^\star}^{\perp}\right) \otimes \mathbf{U}_3^{\star\top}\right)\cdot \right.\right.\\
  &\left.\left.\qquad\mathbf{Z}_1^{(t)\top} \mathbf{U}_{1 \perp}^\star \mathbf{U}_{1 \perp}^{\star\top} \mathbf{Z}_1^{(t)}\left(\mathbf{U}_2^\star \otimes \mathbf{U}_3^\star\right) \mathbf{G}_1^{\star\top}\right)\right| 
	\\  \leq &\left|\operatorname{tr}\left(\mathbf{\Lambda}_1^{-4} \mathbf{G}_1^\star\left(\left(\mathbf{\Lambda}_2^{-2} \mathbf{U}_2^{\star\top} \mathbf{T}_2^\star\left(\mathcal{P}_{\widehat{\mathbf{U}}_1^{(t-2)}} \otimes \mathcal{P}_{\widehat{\mathbf{U}}_3^{(t-2)}}\right) \mathbf{Z}_2^{(t-1)\top} \mathcal{P}_{\mathbf{U}_2^\star}^{\perp}\right) \otimes \mathbf{U}_3^{\star\top}\right) \cdot\right.\right.\\
  &\left.\left.\qquad\mathbf{Z}_1^{(t)\top} \mathbf{U}_{1 \perp}^\star \mathbf{U}_{1 \perp}^{\star\top} \mathbf{Z}_1^{(t)}\left(\mathbf{U}_2^\star \otimes \mathbf{U}_3^\star\right) \mathbf{G}_1^{\star\top}\right)\right| 
	+ \Op{ r_1 \lambda_{\min }^{-2} \cdot \sigma^2\frac{p}{t-1} \cdot \lambda_{\min }^{-2} \cdot \left(\sigma\sqrt{\frac{p}{t}} \right)^2}.
\end{aligned}
\end{align*}
We further measure the first term on the right hand of the above inequality:  
\begin{align*}
\begin{aligned}
 	&\left|\operatorname{tr}\left(\mathbf{\Lambda}_1^{-4} \mathbf{G}_1^\star\left(\left(\mathbf{\Lambda}_2^{-2} \mathbf{U}_2^{\star\top} {\mathfrak{E}}_2^{(t-1)} \mathcal{P}_{\mathbf{U}_2^\star}^{\perp}\right) \otimes \mathbf{U}_3^{\star\top}\right) \mathbf{Z}_1^{(t)\top} \mathbf{U}_{1 \perp}^\star \mathbf{U}_{1 \perp}^{\star\top} \mathbf{Z}_1^{(t)}\left(\mathbf{U}_2^\star \otimes \mathbf{U}_3^\star\right) \mathbf{G}_1^{\star\top}\right)\right| \\ 
	 \leq&\left|\operatorname{tr}\left(\mathbf{\Lambda}_1^{-4} \mathbf{G}_1^\star\left(\left(\mathbf{\Lambda}_2^{-2} \mathbf{U}_2^{\star\top} \mathbf{T}_2^\star\left(\mathcal{P}_{\mathbf{U}_1^\star} \otimes \mathcal{P}_{\mathbf{U}_3^\star}\right) \mathbf{Z}_2^{(t-1)\top} \mathcal{P}_{\mathbf{U}_2^\star}^{\perp}\right) \otimes \mathbf{U}_3^{\star\top}\right) \cdot\right.\right.\\
  &\left.\left.\qquad \mathbf{Z}_1^{(t)\top} \mathbf{U}_{1 \perp}^\star \mathbf{U}_{1 \perp}^{\star\top} \mathbf{Z}_1^{(t)}\left(\mathbf{U}_2^\star \otimes \mathbf{U}_3^\star\right) \mathbf{G}_1^{\star\top}\right)\right| \\ &+\left|\operatorname{tr}\left(\mathbf{\Lambda}_1^{-4} \mathbf{G}_1^\star\left(\left(\mathbf{\Lambda}_2^{-2} \mathbf{U}_2^{\star\top} \mathbf{T}_2^\star\left(\left(\mathcal{P}_{\widehat{\mathbf{U}}_1^{(t-2)}}-\mathcal{P}_{\mathbf{U}_1^\star}\right) \otimes \mathcal{P}_{\mathbf{U}_3^\star}\right) \mathbf{Z}_2^{(t-1)\top} \mathcal{P}_{\mathbf{U}_2^\star}^{\perp}\right) \otimes \mathbf{U}_3^{\star\top}\right)  \cdot\right.\right.\\
  &\left.\left.\qquad\mathbf{Z}_1^{(t)\top} \mathbf{U}_{1 \perp}^\star \mathbf{U}_{1 \perp}^{\star\top} \mathbf{Z}_1^{(t)}\left(\mathbf{U}_2^\star \otimes \mathbf{U}_3^\star\right) \mathbf{G}_1^{\star\top}\right)\right| 
	\\ &+\left|\operatorname{tr}\left(\mathbf{\Lambda}_1^{-4} \mathbf{G}_1^\star\left(\left(\mathbf{\Lambda}_2^{-2} \mathbf{U}_2^{\star\top} \mathbf{T}_2^\star\left(\mathcal{P}_{\widehat{\mathbf{U}}_1^{(t-2)}} \otimes\left(\mathcal{P}_{\widehat{\mathbf{U}}_3^{(t-2)}} - \mathcal{P}_{\mathbf{U}_3^\star}\right)\right) \mathbf{Z}_2^{(t-1)\top} \mathcal{P}_{\mathbf{U}_2^\star}^{\perp}\right) \otimes \mathbf{U}_3^{\star\top}\right) \cdot\right.\right.\\
  &\left.\left.\qquad \mathbf{Z}_1^{(t)\top} \mathbf{U}_{1 \perp}^\star \mathbf{U}_{1 \perp}^{\star\top} \mathbf{Z}_1^{(t)}\left(\mathbf{U}_2^\star \otimes \mathbf{U}_3^\star\right) \mathbf{G}_1^{\star\top}\right)\right| 
	+ \Op{ r_1 \frac{\sigma^4p^2}{t^2 \lambda_{\min }^4}}.
\end{aligned}
\end{align*}
Based on Lemma \ref{lem:bound for inference}, Lemma \ref{lem:bound for first round}, and Equation (\ref{equ:W2}), we find that
\begin{align}\label{B.24-1}
\begin{aligned}
 	&\left|\operatorname{tr}\left(\mathbf{\Lambda}_1^{-4} \mathbf{G}_1^\star\left(\left(\mathbf{\Lambda}_2^{-2} \mathbf{U}_2^{\star\top} {\mathfrak{E}}_2^{(t-1)} \mathcal{P}_{\mathbf{U}_2^\star}^{\perp}\right) \otimes \mathbf{U}_3^{\star\top}\right) \mathbf{Z}_1^{(t)\top} \mathbf{U}_{1 \perp}^\star \mathbf{U}_{1 \perp}^{\star\top} \mathbf{Z}_1^{(t)}\left(\mathbf{U}_2^\star \otimes \mathbf{U}_3^\star\right) \mathbf{G}_1^{\star\top}\right)\right| \\ 
\leq & \left|\operatorname{tr}\left(\mathbf{\Lambda}_1^{-4} \mathbf{G}_1^\star\left(\left(\mathbf{\Lambda}_2^{-2} \mathbf{G}_2^\star\left(\mathbf{U}_1^{\star\top} \otimes \mathbf{U}_3^{\star\top}\right) \mathbf{Z}_2^{(t-1)\top} \mathcal{P}_{\mathbf{U}_2^\star}^{\perp}\right) \otimes \mathbf{U}_3^{\star\top}\right) \mathbf{Z}_1^{(t)\top} \mathbf{U}_{1 \perp}^\star \mathbf{U}_{1 \perp}^{\star\top} \mathbf{Z}_1^{(t)}\left(\mathbf{U}_2^\star \otimes \mathbf{U}_3^\star\right) \mathbf{G}_1^{\star\top}\right)\right| 
	\\ & + \Op{ r_1 \lambda_{\min }^{-2} \cdot \lambda_{\min }^{-2} \cdot\kappa_0\frac{\sigma}{\lambda_{\min }} \sqrt{\frac{p }{t-1}}\cdot\kappa_0\lambda_{\min}\sigma \sqrt{\frac{p}{t-1}} \cdot\left(\sigma \sqrt{\frac{p}{t}} \right)^2
	+ r_1 \frac{\sigma^4p^2}{t^2 \lambda_{\min }^4}}\\
\leq & \left|\operatorname{tr}\left(\mathbf{\Lambda}_1^{-4} \mathbf{G}_1^\star\left(\left(\mathbf{\Lambda}_2^{-2} \mathbf{G}_2^\star\left(\mathbf{U}_1^{\star\top} \otimes \mathbf{U}_3^{\star\top}\right) \mathbf{Z}_2^{(t-1)\top} \mathcal{P}_{\mathbf{U}_2^\star}^{\perp}\right) \otimes \mathbf{U}_3^{\star\top}\right) \mathbf{Z}_1^{(t)\top} \mathbf{U}_{1 \perp}^\star \mathbf{U}_{1 \perp}^{\star\top} \mathbf{Z}_1^{(t)}\left(\mathbf{U}_2^\star \otimes \mathbf{U}_3^\star\right) \mathbf{G}_1^{\star\top}\right)\right| \\
 &+ \Op{ r_1 \frac{\kappa_0^2\sigma^4p^2}{t^2 \lambda_{\min }^4}}.
\end{aligned}
\end{align}
Now, we define three random variables to simplify the proof:
\begin{align*}
  & \overline{\mathbf{V}}_i=\mathbf{U}_{1 \perp}^{\star\top} \mathcal{M}_1\left(\mathcal{X}_i\right)\left( \mathbf{U}_{2 \perp}^\star \otimes \mathbf{U}_3^\star\right) \in \mathbb{R}^{\left(p_1-r_1\right) \times\left(\left(p_2-r_2\right) r_3\right)}, \\
 & \mathbf{V}_i=\mathbf{U}_{1 \perp}^{\star\top} \mathcal{M}_1\left(\mathcal{X}_i\right)\left(\mathbf{U}_2^\star \otimes \mathbf{U}_3^\star\right) \in\mathbb{R}^{\left(p_1-r_1\right) \times r_2 r_3},\\
 & \widetilde{\mathbf{V}}_i = \mathbf{U}_{2 \perp}^{\star\top} \mathcal{M}_2\left(\mathcal{X}_i\right)\left(\mathbf{U}_1^\star \otimes \mathbf{U}_3^\star\right) \in \mathbb{R}^{\left(p_2-r_2\right) \times r_1 r_3}.
\end{align*}
Since 
 \begin{align*}
\mathcal{X}_i \times_1\left[
\begin{array}{ll}\mathbf{U}_1^\star & \mathbf{U}_{1\perp}^\star\end{array}\right]  \times_2 \left[
\begin{array}{ll}\mathbf{U}_2^\star & \mathbf{U}_{2\perp}^\star\end{array}\right] \times_3\left[
\begin{array}{ll}\mathbf{U}_3^\star & \mathbf{U}_{3\perp}^\star\end{array}\right] \stackrel{\text { i.i.d. }}{\sim} \mathcal{N}(0,1),
 \end{align*}
 we know that  
 \begin{align}\label{equ:def of V}
 	\overline{\mathbf{V}}_i \stackrel{i . i . d .}{\sim} \mathcal{N}(0,1), \quad
 	 \mathbf{V}_i \stackrel{\text { i.i.d. }}{\sim} \mathcal{N}(0,1), \quad
 	\widetilde{\mathbf{V}}_i \stackrel{\text { i.i.d. }}{\sim} \mathcal{N}(0,1),
 \end{align}
 and $\overline{\mathbf{V}}_i,  \mathbf{V}_i$ and $\widetilde{\mathbf{V}}_i$ are independent.  Returning to Equation (\ref{B.24-1}), focusing on the initial term in the right-hand side of the equation, utilizing the previously introduced notation, we obtain:
 \begin{align*}
 & \left|\operatorname{tr}\left(\mathbf{\Lambda}_1^{-4} \mathbf{G}_1^\star\left(\left(\mathbf{\Lambda}_2^{-2} \mathbf{G}_2^\star\left(\mathbf{U}_1^{\star\top} \otimes \mathbf{U}_3^{\star\top}\right) \mathbf{Z}_2^{(t-1)\top} \mathcal{P}_{\mathbf{U}_2^\star}^{\perp}\right) \otimes \mathbf{U}_3^{\star\top}\right) \mathbf{Z}_1^{(t)\top} \mathbf{U}_{1 \perp}^\star \mathbf{U}_{1 \perp}^{\star\top} \mathbf{Z}_1^{(t)}\left(\mathbf{U}_2^\star \otimes \mathbf{U}_3^\star\right) \mathbf{G}_1^{\star\top}\right)\right| \\
&  =\left|\operatorname{tr}\left(\mathbf{\Lambda}_1^{-4} \mathbf{G}_1^\star\left(\left(\mathbf{\Lambda}_2^{-2} \mathbf{G}_2^\star\left(\frac{1}{t-1} \sum_{i=1}^{t-1} \xi_i \widetilde{\mathbf{V}}_i\right)^{\top} \right) \otimes \mathbf{I}_{r_3}\right) \left(\frac{1}{t} \sum_{i=1}^t \xi_i \overline{\mathbf{V}}_i\right)^{\top} \left(\frac{1}{t} \sum_{i=1}^t \xi_i  \mathbf{V}_i \right) \mathbf{G}_1^{\star\top}\right)\right|\\
&  =\left|\operatorname{tr}\left(\left(\frac{1}{t} \sum_{i=1}^t \xi_i \mathbf{V}_i\right) \mathbf{G}_1^{\star\top}\mathbf{\Lambda}_1^{-4} \mathbf{G}_1^\star\left(\left(\mathbf{\Lambda}_2^{-2} \mathbf{G}_2^\star\left(\frac{1}{t-1} \sum_{i=1}^{t-1} \xi_i \widetilde{\mathbf{V}}_i\right)^{\top} \right) \otimes \mathbf{I}_{r_3}\right) \left(\frac{1}{t} \sum_{i=1}^t \xi_i \overline{\mathbf{V}}_i\right)^{\top} \right)\right|\\
& = \left|\left\langle\left(\frac{1}{t} \sum_{i=1}^t \xi_i \mathbf{V}_i\right) \mathbf{G}_1^{\star\top}\mathbf{\Lambda}_1^{-4} \mathbf{G}_1^\star\left(\left(\mathbf{\Lambda}_2^{-2} \mathbf{G}_2^\star\left(\frac{1}{t-1} \sum_{i=1}^{t-1} \xi_i \widetilde{\mathbf{V}}_i\right)^{\top} \right) \otimes \mathbf{I}_{r_3}\right) ,\frac{1}{t} \sum_{i=1}^t \xi_i \overline{\mathbf{V}}_i\right\rangle\right|.
\end{align*}
Therefore, 
\begin{align*}
\begin{aligned}
&\left.\left\langle\left(\frac{1}{t} \sum_{i=1}^t \xi_i \mathbf{V}_i\right) \mathbf{G}_1^{\star\top}\mathbf{\Lambda}_1^{-4} \mathbf{G}_1^\star\left(\left(\mathbf{\Lambda}_2^{-2} \mathbf{G}_2^\star\left(\frac{1}{t-1} \sum_{i=1}^{t-1} \xi_i \widetilde{\mathbf{V}}_i\right)^{\top} \right) \otimes \mathbf{I}_{r_3}\right) ,\frac{1}{t} \sum_{i=1}^t \xi_i \overline{\mathbf{V}}_i\right\rangle\right|\left\{ \mathbf{V}_i, \widetilde{\mathbf{V}}_i, \xi_i\right\}_{i=1}^t \\
& \sim \mathcal{N}\left(0,\left\|\left(\frac{1}{t} \sum_{i=1}^t \xi_i \mathbf{V}_i\right) \mathbf{G}_1^{\star\top}\mathbf{\Lambda}_1^{-4} \mathbf{G}_1^\star\left(\left(\mathbf{\Lambda}_2^{-2} \mathbf{G}_2^\star\left(\frac{1}{t-1} \sum_{i=1}^{t-1} \xi_i \widetilde{\mathbf{V}}_i\right)^{\top} \right) \otimes \mathbf{I}_{r_3}\right)\right\|_{\mathrm{F}}^2\cdot \frac{\sum_{i=1}^t\xi_i^2}{t^2}\right) .
\end{aligned}
\end{align*}
Note that, by Equation (\ref{B.2}), we have
\begin{align*}
\begin{aligned}
&\left\|\left(\frac{1}{t} \sum_{i=1}^t \xi_i \mathbf{V}_i\right) \mathbf{G}_1^{\star\top}\mathbf{\Lambda}_1^{-4} \mathbf{G}_1^\star\left(\left(\mathbf{\Lambda}_2^{-2} \mathbf{G}_2^\star\left(\frac{1}{t-1} \sum_{i=1}^{t-1} \xi_i \widetilde{\mathbf{V}}_i\right)^{\top} \right) \otimes \mathbf{I}_{r_3}\right)\right\|_{\mathrm{F}} \\
& \leq \sqrt{r}\left\|\left(\frac{1}{t} \sum_{i=1}^t \xi_i \mathbf{V}_i\right) \mathbf{G}_1^{\star\top}\mathbf{\Lambda}_1^{-4} \mathbf{G}_1^\star\right\|\left\|\mathbf{\Lambda}_2^{-2} \mathbf{G}_2^\star\left(\frac{1}{t-1} \sum_{i=1}^{t-1} \xi_i \widetilde{\mathbf{V}}_i\right)^{\top}\right\| \\
& =\Op{ \sqrt{r}\sigma\sqrt{\frac{p}{t}} \lambda_{\min}^{-2}\cdot \sigma\sqrt{\frac{p}{t-1}} \lambda_{\min}^{-1}} \\
& =  \Op{ \frac{\sigma^2p r^{1 / 2}}{\sqrt{t(t-1)} \lambda_{\min}^3}} = \Op{ \frac{\sigma^2p r^{1 / 2}}{t \lambda_{\min}^3}},
\end{aligned}
\end{align*}
and under Lemma \ref{lem:Op bound}, we have
\begin{align*}
\sum_{i=1}^t\xi_i^2  = \Op{ t\sigma^2},
\end{align*}
and as a result, we have
\begin{align}\label{B.25}
\begin{aligned}
&\left|\operatorname{tr}\left(\mathbf{\Lambda}_1^{-4} \mathbf{G}_1^\star\left(\left(\mathbf{\Lambda}_2^{-2} \mathbf{G}_2^\star\left(\mathbf{U}_1^{\star\top} \otimes \mathbf{U}_3^{\star\top}\right) \mathbf{Z}_2^{(t-1)\top} \mathcal{P}_{\mathbf{U}_2^\star}^{\perp}\right) \otimes \mathbf{U}_3^{\star\top}\right) \mathbf{Z}_1^{(t)\top} \mathbf{U}_{1 \perp}^\star \mathbf{U}_{1 \perp}^{\star\top} \mathbf{Z}_1^{(t)}\left(\mathbf{U}_2^\star \otimes \mathbf{U}_3^\star\right) \mathbf{G}_1^{\star\top}\right)\right|\\
& = \Op{ \frac{r^{1 / 2}\sigma^3 p}{t^{3/2} \lambda_{\min}^3}}.
\end{aligned}
\end{align}
By Equation (\ref{B.24-1}) and (\ref{B.25}),  
\begin{align*}
&\left|\operatorname{tr}\left(\mathbf{\Lambda}_1^{-4} \mathbf{G}_1^\star\left(\left(\mathbf{\Lambda}_2^{-2} \mathbf{U}_2^{\star\top} {\mathfrak{E}}_2^{(t-1)} \mathcal{P}_{\mathbf{U}_2^\star}^{\perp}\right) \otimes \mathbf{U}_3^{\star\top}\right) \mathbf{Z}_1^{(t)\top} \mathbf{U}_{1 \perp}^\star \mathbf{U}_{1 \perp}^{\star\top} \mathbf{Z}_1^{(t)}\left(\mathbf{U}_2^\star \otimes \mathbf{U}_3^\star\right) \mathbf{G}_1^{\star\top}\right)\right|\\
& =\Op{r_1 \kappa_0^2 \frac{\sigma^4p^2}{t^2 \lambda_{\min }^4}+ \frac{r^{1 / 2}\sigma^3 p }{t^{3/2} \lambda_{\min}^3}}.
\end{align*}
Similarly, 
\begin{align*}
&\left|\operatorname{tr}\left(\mathbf{\Lambda}_1^{-4} \mathbf{G}_1^\star\left(\mathbf{U}_2^{\star\top} \otimes\left(\mathbf{\Lambda}_3^{-2} \mathbf{U}_3^{\star\top} {\mathfrak{E}}_3^{(t-1)} \mathcal{P}_{\mathbf{U}_3^\star}^{\perp}\right)\right) \mathbf{Z}_1^{(t)\top} \mathbf{U}_{1 \perp}^\star \mathbf{U}_{1 \perp}^{\star\top} \mathbf{Z}_1^{(t)}\left(\mathbf{U}_2^\star \otimes \mathbf{U}_3^\star\right) \mathbf{G}_1^{\star\top}\right)\right|\\
&=\Op{r_1 \kappa_0^2 \frac{\sigma^4p^2}{t^2 \lambda_{\min }^4}+ \frac{r^{1 / 2}\sigma^3 p}{t^{3/2} \lambda_{\min}^3}}.
\end{align*}
By Equation (\ref{B.23}) and above two equations, \begin{align*}
	&\left|\mathrm{I}-\operatorname{tr}\left(\mathbf{\Lambda}_1^{-4} \mathbf{G}_1^\star \left(\mathbf{U}_2^{\star\top} \otimes \mathbf{U}_3^{\star\top}\right) \mathbf{Z}_1^{(t)\top} \mathbf{U}_{1 \perp}^\star \mathbf{U}_{1 \perp}^{\star\top} \mathbf{Z}_1^{(t)}\left(\mathbf{U}_2^\star \otimes \mathbf{U}_3^\star\right) \mathbf{G}_1^{\star\top}\right)\right|\\
	& =\Op{r_1 \kappa_0^2 \frac{\sigma^4p^2}{t^2 \lambda_{\min }^4}+ \frac{r^{1 / 2}\sigma^3 p}{t^{3/2} \lambda_{\min}^3}}.
\end{align*}
Thus, we conclude the proof of  Lemma \ref{lem:lemma 3}.
\end{proof}

\subsection{Proof of Lemma \ref{lem:lemma 2}}\label{sec:Proof of Lemma 2}
By Lemma \ref{lem:bound for first round}, we have
\begin{align}\label{B.9-1}
\begin{aligned}
&\left|\operatorname{tr}\left(\mathfrak{P}_1^{-1} \mathfrak{J}_1^{(t)} \mathfrak{P}_1^{\perp} \mathfrak{J}_2^{(t)} \mathfrak{P}_1^{-1} \mathfrak{J}_1^{(t)} \mathfrak{P}_1^{-1}\right)\right| \\
 \leq&\left|\operatorname{tr}\left(\mathfrak{P}_1^{-1} \mathbf{T}_1^\star\left(\mathcal{P}_{\mathbf{U}_2^\star} \otimes \mathcal{P}_{\mathbf{U}_3^\star}\right) \mathbf{Z}_1^{(t)\top} \mathfrak{P}_1^{\perp} \mathbf{Z}_1^{(t)}\left(\mathcal{P}_{\mathbf{U}_2^\star} \otimes \mathcal{P}_{\mathbf{U}_3^\star}\right) \mathbf{T}_1^{\star\top} \mathfrak{P}_1^{-1} \mathbf{T}_1^\star\left(\mathcal{P}_{\mathbf{U}_2^\star} \otimes \mathcal{P}_{\mathbf{U}_3^\star}\right) \mathbf{Z}_1^{(t)\top} \mathfrak{P}_1^{-1}\right)\right| \\
&+\left|\operatorname{tr}\left(\mathfrak{P}_1^{-1}\left(\mathfrak{J}_1^{(t)}-\mathbf{T}_1^\star\left(\mathcal{P}_{\mathbf{U}_2^\star} \otimes \mathcal{P}_{\mathbf{U}_3^\star}\right) \mathbf{Z}_1^{(t)\top}\right) \mathfrak{P}_1^{\perp} \cdot\right.\right.\\
&\left.\left.\quad\mathbf{Z}_1^{(t)}\left(\mathcal{P}_{\mathbf{U}_2^\star} \otimes \mathcal{P}_{\mathbf{U}_3^\star}\right) \mathbf{T}_1^{\star\top} \mathfrak{P}_1^{-1} \mathbf{T}_1^\star\left(\mathcal{P}_{\mathbf{U}_2^\star} \otimes \mathcal{P}_{\mathbf{U}_3^\star}\right) \mathbf{Z}_1^{(t)\top} \mathfrak{P}_1^{-1}\right)\right| \\
&+\left|\operatorname{tr}\left(\mathfrak{P}_1^{-1} \mathfrak{J}_1^{(t)} \mathfrak{P}_1^{\perp}\left(\mathfrak{J}_1^{(t)}-\mathbf{T}_1^\star\left(\mathcal{P}_{\mathbf{U}_2^\star} \otimes \mathcal{P}_{\mathbf{U}_3^\star}\right) \mathbf{Z}_1^{(t)\top}\right)^{\top} \mathfrak{P}_1^{-1} \mathbf{T}_1^\star\left(\mathcal{P}_{\mathbf{U}_2^\star} \otimes \mathcal{P}_{\mathbf{U}_3^\star}\right) \mathbf{Z}_1^{(t)\top} \mathfrak{P}_1^{-1}\right)\right| \\
&+\left|\operatorname{tr}\left(\mathfrak{P}_1^{-1} \mathfrak{J}_1^{(t)} \mathfrak{P}_1^{\perp} \mathfrak{J}_2^{(t)} \mathfrak{P}_1^{-1}\left(\mathfrak{J}_1^{(t)}-\mathbf{T}_1^\star\left(\mathcal{P}_{\mathbf{U}_2^\star} \otimes \mathcal{P}_{\mathbf{U}_3^\star}\right) \mathbf{Z}_1^{(t)\top}\right) \mathfrak{P}_1^{-1}\right)\right| \\
 \leq&\left|\operatorname{tr}\left(\mathfrak{P}_1^{-1} \mathbf{T}_1^\star\left(\mathcal{P}_{\mathbf{U}_2^\star} \otimes \mathcal{P}_{\mathbf{U}_3^\star}\right) \mathbf{Z}_1^{(t)\top} \mathfrak{P}_1^{\perp} \mathbf{Z}_1^{(t)}\left(\mathcal{P}_{\mathbf{U}_2^\star} \otimes \mathcal{P}_{\mathbf{U}_3^\star}\right) \mathbf{T}_1^{\star\top} \mathfrak{P}_1^{-1} \mathbf{T}_1^\star\left(\mathcal{P}_{\mathbf{U}_2^\star} \otimes \mathcal{P}_{\mathbf{U}_3^\star}\right) \mathbf{Z}_1^{(t)\top} \mathfrak{P}_1^{-1}\right)\right| \\
&+\Op{r_1 \frac{\kappa_0 \lambda_{\min}\sigma \sqrt{p/t}  \cdot \kappa_0 \lambda_{\min}\sigma \sqrt{p/t} \cdot \kappa_0 \sigma^2\left(p/t\right)}{\lambda_{\min}^6} }\\
=& \left|\operatorname{tr}\left(\mathbf{U}_1^\star \mathbf{\Lambda}_1^{-2} \mathbf{U}_1^{\star\top} \mathbf{T}_1^\star\left(\mathcal{P}_{\mathbf{U}_2^\star} \otimes \mathcal{P}_{\mathbf{U}_3^\star}\right) \mathbf{Z}_1^{(t)\top} \mathbf{U}_{1 \perp}^\star \mathbf{U}_{1 \perp}^{\star\top} \mathbf{Z}_1^{(t)}\left(\mathcal{P}_{\mathbf{U}_2^\star} \otimes \mathcal{P}_{\mathbf{U}_3^\star}\right) \mathbf{T}_1^{\star\top}\cdot\right.\right.\\
&\left.\left.\qquad\mathbf{U}_1^\star \mathbf{\Lambda}_1^{-2} \mathbf{U}_1^{\star\top} \mathbf{T}_1^\star\left(\mathcal{P}_{\mathbf{U}_2^\star} \otimes \mathcal{P}_{\mathbf{U}_3^\star}\right) \mathbf{Z}_1^{(t)\top} \mathbf{U}_1^\star \mathbf{\Lambda}_1^{-2} \mathbf{U}_1^{\star\top}\right) \right|  +\Op{r_1\kappa_0^3 \sigma^4\lambda_{\min}^{-4}p^2 t^{-2}}.
\end{aligned}
\end{align}
By the notation defined in Equation (\ref{equ:def of W}), we have
\begin{align*}
    &\left|\operatorname{tr}\left(\mathbf{U}_1^\star \mathbf{\Lambda}_1^{-2} \mathbf{U}_1^{\star\top} \mathbf{T}_1^\star\left(\mathcal{P}_{\mathbf{U}_2^\star} \otimes \mathcal{P}_{\mathbf{U}_3^\star}\right) \mathbf{Z}_1^{(t)\top} \mathbf{U}_{1 \perp}^\star \mathbf{U}_{1 \perp}^{\star\top} \mathbf{Z}_1^{(t)}\left(\mathcal{P}_{\mathbf{U}_2^\star} \otimes \mathcal{P}_{\mathbf{U}_3^\star}\right) \mathbf{T}_1^{\star\top}\cdot\right.\right.\\
&\left.\left.\qquad\mathbf{U}_1^\star \mathbf{\Lambda}_1^{-2} \mathbf{U}_1^{\star\top} \mathbf{T}_1^\star\left(\mathcal{P}_{\mathbf{U}_2^\star} \otimes \mathcal{P}_{\mathbf{U}_3^\star}\right) \mathbf{Z}_1^{(t)\top} \mathbf{U}_1^\star \mathbf{\Lambda}_1^{-2} \mathbf{U}_1^{\star\top}\right) \right|\\
=&\left|\operatorname{tr}\left(\mathbf{U}_1^\star \mathbf{\Lambda}_1^{-2} \mathbf{G}_1^\star \mathbf{W}_2^{\top} \mathbf{W}_2 \mathbf{G}_1^{\star\top} \mathbf{\Lambda}_1^{-2} \mathbf{G}_1^\star \mathbf{W}_1^{\top} \mathbf{\Lambda}_1^{-2} \mathbf{U}_1^{\star\top}\right)\right|\\
=&\left|\operatorname{tr}\left(\mathbf{\Lambda}_1^{-4} \mathbf{G}_1^\star \mathbf{W}_2^{\top} \mathbf{W}_2 \mathbf{G}_1^{\star\top} \mathbf{\Lambda}_1^{-2} \mathbf{G}_1^\star \mathbf{W}_1^{\top}\right)\right|.
\end{align*}
In a manner analogous to Lemma \ref{lem:compare z1 and z2}, we derive:
\begin{align}\label{equ:W1}
\left\|\mathbf{W}_1\right\|=\left\|\mathbf{U}_1^{\star\top} \mathbf{Z}_1^{(t)}\left(\mathbf{U}_2^\star \otimes \mathbf{U}_3^\star\right) \right\| = \Op{ \sigma\sqrt{\frac{r^2}{t}}},
\end{align}
and 
\begin{align}\label{equ:W2}
\left\|\mathbf{W}_2\right\|=\left\|\mathbf{U}_{1 \perp}^{\star\top} \mathbf{Z}_1^{(t)}\left(\mathbf{U}_2^\star \otimes \mathbf{U}_3^\star\right)\right\| = \Op{ \sigma\sqrt{\frac{p}{t}}}.
\end{align}
From Equations (\ref{B.2}), (\ref{equ:W1}) and (\ref{equ:W2}), we deduce that
\begin{align}\label{B.9-2}
\begin{aligned}
		\left|\operatorname{tr}\left(\mathbf{\Lambda}_1^{-4} \mathbf{G}_1^\star \mathbf{W}_2^{\top} \mathbf{W}_2 \mathbf{G}_1^{\star\top} \mathbf{\Lambda}_1^{-2} \mathbf{G}_1^\star \mathbf{W}_1^{\top}\right)\right| 
  &\leq r \left\|\mathbf{\Lambda}_1^{-4} \mathbf{G}_1^\star  \right\|  \left\| \mathbf{W}_2\right\|^2  \left\| \mathbf{G}_1^{\star\top} \mathbf{\Lambda}_1^{-2} \mathbf{G}_1^\star \right\| \left\| \mathbf{W}_1\right\|\\ 
  & =\Op{ r \sigma^3\lambda_{\min}^{-3} \frac{p}{t}\sqrt{\frac{r^2}{t}}} \\
	& = \Op{\frac{ \sigma^3r^2 p}{t^{3/2}\lambda_{\min}^{3}} }.
\end{aligned}
\end{align}
Combining  the findings from Equations (\ref{B.9-1}), we arrive at the following conclusion:
\begin{align*}
	\left|\operatorname{tr}\left(\mathfrak{P}_1^{-1} \mathfrak{J}_1^{(t)} \mathfrak{P}_1^{\perp} \mathfrak{J}_2^{(t)} \mathfrak{P}_1^{-1} \mathfrak{J}_1^{(t)} \mathfrak{P}_1^{-1}\right)\right| = \Op{\frac{ \sigma^3r^2 p}{t^{3/2}\lambda_{\min}^{3}} 
	+ \frac{r_1\kappa_0^3 \sigma^4p^2}{t^{2}\lambda_{\min}^{4}}}.
\end{align*}
Similarly, we have:
\begin{align*}
	\left|\operatorname{tr}\left(\mathfrak{P}_1^{-1} \mathfrak{J}_1^{(t)} \mathfrak{P}_1^{\perp} \mathfrak{J}_2^{(t)} \mathfrak{P}_1^{-1} \mathfrak{J}_2^{(t)} \mathfrak{P}_1^{-1}\right)\right| = \Op{ \frac{ \sigma^3r^2 p}{t^{3/2}\lambda_{\min}^{3}} 
	+ \frac{r_1\kappa_0^3 \sigma^4p^2}{t^{2}\lambda_{\min}^{4}}}.
\end{align*}
Thus, we conclude the proof of  Lemma \ref{lem:lemma 2}.